\pdfoutput=1 
\documentclass[acmsmall,screen,nonacm]{acmart}

\newif\ifdraftediting   
\newif\ifextendedbuild  
\newif\ifshowlinenumbers 
\newif\ifreportmode     
\extendedbuildtrue

\usepackage{graphicx}
\usepackage{subcaption}
\graphicspath{{artifacts/figures/}}
\usepackage{booktabs}
\usepackage{tabularx}
\usepackage{listings}
\usepackage{listingsutf8}
\usepackage{tikz}
\usetikzlibrary{arrows.meta, calc}
\usepackage{microtype}

\makeatletter
\newcommand{\refappendixsloppy}{%
  \rightskip=0pt plus 2em%
  \@rightskip=0pt plus 2em\relax}
\makeatother

\usepackage{longtable}
\usepackage{array}

\definecolor{algofill}{HTML}{A6CEE3}  
\definecolor{trainfill}{HTML}{B2DF8A} 
\definecolor{benchfill}{HTML}{FDBF6F} 
\definecolor{classfill}{HTML}{DDEEFF} 
\definecolor{extfill}{HTML}{E8E8E8}   

\definecolor{instancefill}{HTML}{FFF4CC}

\lstdefinelanguage{sparql}{
  morekeywords={SELECT,WHERE,FILTER,PREFIX,ASK,CONSTRUCT,DESCRIBE,
    INSERT,DELETE,OPTIONAL,UNION,BIND,AS,ORDER,BY,LIMIT,GROUP,HAVING,
    COUNT,AVG,MIN,MAX,SUM,VALUES,a},
  sensitive=true,
  morecomment=[l]{\#},
  morestring=[b]",
}

\lstdefinelanguage{turtle}{
  morekeywords={@prefix,a},
  sensitive=true,
  morecomment=[l]{\#},
  morestring=[b]",
}

\newcommand{\DanielHernandez}[1]{}
\newcommand{\JongHyunJung}[1]{}
\newcommand{\YujiIkeda}[1]{}
\newcommand{\YongliangOu}[1]{}
\newcommand{\PranavKumar}[1]{}
\newcommand{\TomSchaechtel}[1]{}
\newcommand{\WenchuanLiu}[1]{}
\newcommand{\XinLi}[1]{}
\newcommand{\XiZhang}[1]{}
\newcommand{\XiangXu}[1]{}
\newcommand{\LifangZhu}[1]{}
\newcommand{\FritzKoermann}[1]{}
\newcommand{\SteffenStaab}[1]{}
\newcommand{\BlazejGrabowski}[1]{}

\newcommand{\MLIPsOntology}{MLIPs ontology}
\newcommand{\MLIPsKG}{MLIPs knowledge graph}

\newcommand{\mlipsns}{https://w3id.org/mlips/}
\newcommand{\mlipsprefix}{\texttt{mlips:}}

\newcommand{\paperref}[1]{\texttt{#1}}

\newcommand{\extref}[1]{%
  \ifextendedbuild
    Appendix~\ref{#1}%
  \else
    the supplemental material%
  \fi
}
\newcommand{\extrefshort}[1]{%
  \ifextendedbuild
    \S\ref{#1}%
  \else
    supplemental material%
  \fi
}

\newcommand{\bodyref}[2]{%
  \ifreportmode
    #2%
  \else
    \ref{#1}%
  \fi
}

\newcommand{\dlConcept}[1]{\ensuremath{\mathsf{#1}}}
\newcommand{\dlRole}[1]{\ensuremath{\mathsf{#1}}}

\newcommand{\sqsub}{\sqsubseteq}
\newcommand{\existsR}[2]{\exists\, \mathsf{#1}.\mathsf{#2}}

\newcommand{\exactR}[3]{(\mathord{=}#1\;\mathsf{#2}.\mathsf{#3})}

\newcommand{\MLIPMethodC}{\dlConcept{MLIPMethod}}
\newcommand{\FunctionalFormC}{\dlConcept{FunctionalForm}}
\newcommand{\LossFunctionC}{\dlConcept{LossFunction}}
\newcommand{\HyperparameterC}{\dlConcept{Hyperparameter}}
\newcommand{\HyperparameterSettingC}{\dlConcept{HyperparameterSetting}}
\newcommand{\ImplementationC}{\dlConcept{Implementation}}
\newcommand{\SimulationTypeC}{\dlConcept{SimulationType}}
\newcommand{\LibraryC}{\dlConcept{Library}}
\newcommand{\MLIPRunC}{\dlConcept{MLIPRun}}
\newcommand{\TrainedModelC}{\dlConcept{TrainedModel}}
\newcommand{\AtomicEnvironmentDescriptorC}{\dlConcept{AtomicEnvironmentDescriptor}}
\newcommand{\hasDescriptorR}{\dlRole{hasDescriptor}}
\newcommand{\TrainingDatasetC}{\dlConcept{TrainingDataset}}
\newcommand{\DFTCalculationC}{\dlConcept{DFTCalculation}}
\newcommand{\DFTSettingsC}{\dlConcept{DFTSettings}}
\newcommand{\ReferenceCalculationC}{\dlConcept{ReferenceCalculation}}
\newcommand{\ReferenceSettingsC}{\dlConcept{ReferenceSettings}}
\newcommand{\WaveFunctionCalculationC}{\dlConcept{WaveFunctionCalculation}}
\newcommand{\WaveFunctionSettingsC}{\dlConcept{WaveFunctionSettings}}
\newcommand{\hasReferenceCalculationR}{\dlRole{hasReferenceCalculation}}
\newcommand{\hasReferenceSettingsR}{\dlRole{hasReferenceSettings}}

\newcommand{\wfMethodR}{\dlRole{wfMethod}}
\newcommand{\basisSetR}{\dlRole{basisSet}}
\newcommand{\frozenCoreR}{\dlRole{frozenCore}}
\newcommand{\XCFunctionalC}{\dlConcept{XCFunctional}}
\newcommand{\PseudopotentialTypeC}{\dlConcept{PseudopotentialType}}
\newcommand{\WfMethodC}{\dlConcept{WfMethod}}
\newcommand{\DftBasisSetC}{\dlConcept{DftBasisSet}}

\newcommand{\candidateForVocabularyR}{\dlRole{candidateForVocabulary}}
\newcommand{\MaterialSystemC}{\dlConcept{MaterialSystem}}
\newcommand{\AtomicConfigurationC}{\dlConcept{AtomicConfiguration}}
\newcommand{\DatasetProvenanceC}{\dlConcept{DatasetProvenance}}
\newcommand{\SamplingStrategyC}{\dlConcept{SamplingStrategy}}
\newcommand{\samplingStrategyR}{\dlRole{samplingStrategy}}
\newcommand{\CoveredPropertyC}{\dlConcept{CoveredProperty}}
\newcommand{\BenchmarkStudyC}{\dlConcept{BenchmarkStudy}}
\newcommand{\BenchmarkResultC}{\dlConcept{BenchmarkResult}}
\newcommand{\AccuracyMetricC}{\dlConcept{AccuracyMetric}}
\newcommand{\MetricTypeC}{\dlConcept{MetricType}}
\newcommand{\MetricPropertyC}{\dlConcept{MetricProperty}}

\newcommand{\hasHyperparameterR}{\dlRole{hasHyperparameter}}
\newcommand{\hasImplementationR}{\dlRole{hasImplementation}}
\newcommand{\supportsSimulationR}{\dlRole{supportsSimulation}}
\newcommand{\implementedInR}{\dlRole{implementedIn}}
\newcommand{\forHyperparameterR}{\dlRole{forHyperparameter}}
\newcommand{\appliesMethodR}{\dlRole{appliesMethod}}
\newcommand{\hasFunctionalFormR}{\dlRole{hasFunctionalForm}}
\newcommand{\hasLossFunctionR}{\dlRole{hasLossFunction}}
\newcommand{\hasTrainingAlgorithmR}{\dlRole{hasTrainingAlgorithm}}
\newcommand{\hasTrainingRunR}{\dlRole{hasTrainingRun}}
\newcommand{\producesR}{\dlRole{produces}}

\newcommand{\runsOnR}{\dlRole{runsOn}}
\newcommand{\trainedWithR}{\dlRole{trainedWith}}
\newcommand{\trainedOnR}{\dlRole{trainedOn}}
\newcommand{\trainedUsingR}{\dlRole{trainedUsing}}
\newcommand{\hasHyperparameterSettingR}{\dlRole{hasHyperparameterSetting}}
\newcommand{\evaluatesModelR}{\dlRole{evaluatesModel}}
\newcommand{\hasResultR}{\dlRole{hasResult}}
\newcommand{\hasAccuracyMetricR}{\dlRole{hasAccuracyMetric}}
\newcommand{\targetMaterialR}{\dlRole{targetMaterial}}

\newcommand{\reportedInR}{\dlRole{reportedIn}}
\newcommand{\metricTypeR}{\dlRole{metricType}}
\newcommand{\metricValueR}{\dlRole{metricValue}}
\newcommand{\metricPropertyR}{\dlRole{metricProperty}}
\newcommand{\coversMaterialR}{\dlRole{coversMaterial}}
\newcommand{\coversPropertyR}{\dlRole{coversProperty}}
\newcommand{\hasConfigurationR}{\dlRole{hasConfiguration}}
\newcommand{\datasetProvenanceR}{\dlRole{datasetProvenance}}
\newcommand{\hasDFTSettingsR}{\dlRole{hasDFTSettings}}
\newcommand{\hasDFTCalculationR}{\dlRole{hasDFTCalculation}}
\newcommand{\cutoffRadiusR}{\dlRole{cutoffRadius}}
\newcommand{\numLayersR}{\dlRole{numLayers}}

\newcommand{\xcFunctionalR}{\dlRole{xcFunctional}}
\newcommand{\kPointMeshR}{\dlRole{kPointMesh}}
\newcommand{\energyCutoffR}{\dlRole{energyCutoff}}
\newcommand{\pseudopotentialTypeR}{\dlRole{pseudopotentialType}}
\newcommand{\numConfigurationsR}{\dlRole{numConfigurations}}
\newcommand{\trainingDurationR}{\dlRole{trainingDuration}}
\newcommand{\gpuHoursR}{\dlRole{gpuHours}}
\newcommand{\trainingHardwareR}{\dlRole{trainingHardware}}
\newcommand{\peakMemoryR}{\dlRole{peakMemory}}
\newcommand{\inferenceTimePerAtomR}{\dlRole{inferenceTimePerAtom}}
\newcommand{\inferenceHardwareR}{\dlRole{inferenceHardware}}

\newcommand{\mlsAlgorithmC}{\dlConcept{mls{:}Algorithm}}
\newcommand{\mlsHyperParameterC}{\dlConcept{mls{:}HyperParameter}}
\newcommand{\mlsImplementationC}{\dlConcept{mls{:}Implementation}}
\newcommand{\mlsRunC}{\dlConcept{mls{:}Run}}
\newcommand{\mlsModelC}{\dlConcept{mls{:}Model}}
\newcommand{\mlsDatasetC}{\dlConcept{mls{:}Dataset}}
\newcommand{\provActivityC}{\dlConcept{prov{:}Activity}}
\newcommand{\provEntityC}{\dlConcept{prov{:}Entity}}
\newcommand{\mdoCalculationC}{\dlConcept{mdo{:}Calculation}}
\newcommand{\cmsoCrystallineMaterialC}{\dlConcept{cmso{:}CrystallineMaterial}}
\newcommand{\cmsoAtomicStructureC}{\dlConcept{cmso{:}AtomicStructure}}
\newcommand{\schemaScholarlyArticleC}{\dlConcept{schema{:}ScholarlyArticle}}

\newcommand{\mlsExecutesR}{\dlRole{mls{:}executes}}
\newcommand{\mlsHasOutputR}{\dlRole{mls{:}hasOutput}}
\newcommand{\mlsHasInputR}{\dlRole{mls{:}hasInput}}
\newcommand{\provWasDerivedFromR}{\dlRole{prov{:}wasDerivedFrom}}

\newcommand{\neigh}[1]{\mathcal{N}_{#1}}            
\newcommand{\Rcut}{R_{\mathrm{cut}}}                 
\newcommand{\Rmin}{R_{\mathrm{min}}}                 
\newcommand{\levmax}{\operatorname{lev}_{\mathrm{max}}}  
\newcommand{\nRadialBasis}{N_Q}                      
\newcommand{\fitParams}{\boldsymbol{\xi}}            
\newcommand{\localEnergy}{V}                         
\newcommand{\basisFunc}{B}                           
\newcommand{\radialBasis}{Q}                         
\newcommand{\totalEnergy}{E^{\mathrm{mtp}}}          
\newcommand{\relPos}{\mathbf{r}}                     
\newcommand{\atomType}{z}                            
\newcommand{\config}{\mathrm{cfg}}                   

\newcounter{axiom}
\newcommand{\axiomtag}{\refstepcounter{axiom}\tag{A\theaxiom}\label{ax:\theaxiom}}

\usepackage{orcidlink}
\makeatletter
\def\orcid#1{\unskip\ignorespaces%
  \IfBeginWith{#1}{http}{%
    \expandafter\gdef\csname
        typeset@author\the\num@authors\endcsname##1{%
          \href{#1}{##1}}}{%
    \expandafter\gdef\csname
        typeset@author\the\num@authors\endcsname##1{%
          \href{https://orcid.org/#1}{##1}\,\orcidlink{#1}}}}
\makeatother

\title{An Ontology for Machine Learning Interatomic Potentials}

\author{Daniel Hern\'andez}
\orcid{0000-0002-7896-0875}
\email{daniel.hernandez@ki.uni-stuttgart.de}
\affiliation{%
  \institution{Institute for Artificial Intelligence, University of Stuttgart}
  \city{Stuttgart}
  \country{Germany}}

\author{Jong Hyun Jung}
\orcid{0000-0002-2409-975X}
\email{jong-hyun.jung@imw.uni-stuttgart.de}
\affiliation{%
  \institution{Institute for Materials Science, University of Stuttgart}
  \city{Stuttgart}
  \country{Germany}}

\author{Yuji Ikeda}
\orcid{0000-0001-9176-3270}
\email{yuji.ikeda@imw.uni-stuttgart.de}
\affiliation{%
  \institution{Institute for Materials Science, University of Stuttgart}
  \city{Stuttgart}
  \country{Germany}}

\author{Yongliang Ou}
\orcid{0000-0003-1486-1197}
\email{yongliang.ou@imw.uni-stuttgart.de}
\affiliation{%
  \institution{Institute for Materials Science, University of Stuttgart}
  \city{Stuttgart}
  \country{Germany}}

\author{Pranav Kumar}
\orcid{0000-0002-3661-5870}
\email{pranav.kumar@imw.uni-stuttgart.de}
\affiliation{%
  \institution{Institute for Materials Science, University of Stuttgart}
  \city{Stuttgart}
  \country{Germany}}

\author{Tom Sch\"achtel}
\orcid{0009-0001-4894-0195}
\affiliation{%
  \institution{Institute for Materials Science, University of Stuttgart}
  \city{Stuttgart}
  \country{Germany}}

\author{Wenchuan Liu}
\orcid{0009-0009-9043-2491}
\affiliation{%
  \institution{Institute for Materials Science, University of Stuttgart}
  \city{Stuttgart}
  \country{Germany}}

\author{Xin Li}
\orcid{0009-0003-3860-8388}
\affiliation{%
  \institution{Institute for Materials Science, University of Stuttgart}
  \city{Stuttgart}
  \country{Germany}}

\author{Xi Zhang}
\orcid{0000-0003-0319-8203}
\email{xi.zhang@imw.uni-stuttgart.de}
\affiliation{%
  \institution{Institute for Materials Science, University of Stuttgart}
  \city{Stuttgart}
  \country{Germany}}

\author{Xiang Xu}
\orcid{0000-0002-4399-0580}
\affiliation{%
  \institution{Institute for Materials Science, University of Stuttgart}
  \city{Stuttgart}
  \country{Germany}}

\author{Lifang Zhu}
\orcid{0000-0003-2740-8172}
\email{lifang.zhu@imw.uni-stuttgart.de}
\affiliation{%
  \institution{Institute for Materials Science, University of Stuttgart}
  \city{Stuttgart}
  \country{Germany}}

\author{Fritz K\"ormann}
\orcid{0000-0003-3050-6291}
\email{fritz.koermann@rub.de}
\affiliation{%
  \institution{Interdisciplinary Centre for Advanced Materials Simulation (ICAMS), Ruhr-Universit\"at Bochum}
  \city{Bochum}
  \country{Germany}}
\affiliation{%
  \institution{Department for Computational Materials Design, Max-Planck-Institut for Sustainable Materials}
  \city{D\"usseldorf}
  \country{Germany}}

\author{Steffen Staab}
\orcid{0000-0002-0780-4154}
\email{steffen.staab@ki.uni-stuttgart.de}
\affiliation{%
  \institution{Institute for Artificial Intelligence, University of Stuttgart}
  \city{Stuttgart}
  \country{Germany}}

\author{Blazej Grabowski}
\orcid{0000-0003-4281-5665}
\email{blazej.grabowski@imw.uni-stuttgart.de}
\affiliation{%
  \institution{Institute for Materials Science, University of Stuttgart}
  \city{Stuttgart}
  \country{Germany}}

\renewcommand{\shortauthors}{Hern\'andez et al.}

\begin{abstract}
Machine learning interatomic potentials (MLIPs) approximate
quantum-mechanical energies and forces---conventionally computed by
density functional theory (DFT) or wave-function methods---at a
fraction of the cost. The field encompasses a growing ecosystem of
algorithms, training datasets, hyperparameters, and target
materials, yet the metadata needed to systematically compare,
reproduce, and build upon MLIP studies remains scattered across
papers, scripts, and ad-hoc file formats. We present the
\MLIPsOntology{}, an OWL\,2~DL ontology that captures the concepts
needed to describe MLIP methods, their hyperparameters, training
datasets with DFT provenance, and published benchmarks. The
ontology is organized into three modules---Method, Training Data,
and Benchmark---and connects existing ontologies in materials
science (MDO, CMSO/ASMO) and machine learning (ML-Schema),
complementing dataset-side schemas such as Croissant. It declares
27 formal axioms enforcing data completeness and consistency,
including property chains that link trained models to their
methods and training data. We demonstrate the ontology through a
running example based on Moment Tensor Potentials and evaluate it
through competency-question execution on a 20-paper seeded
knowledge graph, OWL reasoning, and comparison with existing
ontologies.
\end{abstract}


\begin{document}

\maketitle

\section{Introduction}
\label{sec:introduction}
The MLIP community is sizable and growing: VASP, LAMMPS, and
DeePMD-kit drew 13{,}200, 4{,}190, and 1{,}510 citations in
2025~\cite{talirz21}, and MLIP-related publications rose from
$\sim$50/year in 2015 to over 1{,}500 in
2024~\cite{kocer2022nnp,deringer2021gpr}.
Machine learning interatomic potentials
(MLIPs)~\cite{behler2007hdnnp,bartok2010gap} predict atomic
energies and forces at near the accuracy of their
quantum-mechanical reference method---conventionally density
functional theory (DFT)~\cite{kohn1965dft}, increasingly also
wave-function methods---at a fraction of its cost, enabling
atomistic simulations at scale.
MLIP families have proliferated over
the past decade---HDNNP, GAP, ACE, MACE, NequIP, M3GNet, MTP, and
others~\cite{drautz2019ace,batatia2022mace,batzner2022nequip,chen2022m3gnet,shapeev2016mtp}---each
with its own architecture, hyperparameter space, and application
domain.

This diversity makes routine tasks---selecting an algorithm for a
given material, building a training set, comparing results across
papers---hard~\cite{grabowski2024metalearnerc}. Study metadata (algorithm, training set, material, DFT settings,
accuracy) is scattered across papers, supplements, scripts, and
library-specific formats. Existing ontologies do not close the
gap: ML-Schema~\cite{publio2018mlschema}, DMOP~\cite{keet2015dmop},
MEX~\cite{esteves2015mex}, and CMO~\cite{foltin2024cmo} model
generic ML workflows; EMMO~\cite{horsch2020emmo},
MDO~\cite{li2024mdo}, PMDco~\cite{bayerlein2024pmdco}, and
CMSO/ASMO~\cite{menon2024atomrdf} model the physical
domain---neither captures the MLIP-specific concepts:
physics-aware hyperparameters, training-dataset construction
strategies, algorithm-aligned DFT reference settings, and
benchmark accuracy across method/material pairs.

We present the \emph{\MLIPsOntology{}}, an OWL\,2~DL
ontology~\cite{owl2overview} organized into three modules
(\emph{Method}, \emph{Training Data}, \emph{Benchmark}) that
connects existing schemas rather than displacing them, reusing
established vocabularies wherever an existing class or property
already names the concept at hand: 22
formal alignment axioms plus 14
\texttt{rdfs:subClassOf}/\texttt{subPropertyOf} edges thread
its terms through ML-Schema, PROV-O, MDO, CMSO/ASMO,
schema.org, and QUDT~(\S\ref{sec:alignment}); 27 additional
axioms enforce the MLIP-specific structure that no single
existing ontology supplies, and SHACL shapes validate
instance data. Its conceptual contributions are three
MLIP-specific design decisions:
hyperparameters are first-class type-level entities with
units, defaults, and ranges (not literal values on runs); a
DFT / wave-function reference-calculation hierarchy uses
\texttt{rdfs:subPropertyOf} so legacy DFT data validates
against family-agnostic super-properties; and an
\texttt{mlips:metaSort} annotation per class carries the
OntoClean classification. The schema is exercised on a
20-paper knowledge graph (all nine competency
questions return non-empty rows), populated via an agentic-AI
extraction protocol whose output was returned to the domain
experts for review---an extensional validation cycle that
extends LOT's intensional CQ-agreement loop
(\S\ref{sec:requirements})---and is served at
\url{https://w3id.org/mlips} with content negotiation.
For consumers, cross-paper questions---which method reaches a
target accuracy on a material class, which training data exists
with compatible DFT settings---reduce to single SPARQL queries
instead of a manual literature survey, for human researchers and
autonomous agents alike; for producers, an encoded
study is one that those selection queries can find, and the
encoding protocol doubles as a reporting checklist that surfaces
unreported metadata before reviewers do.
A concrete usage plan sustains the resource beyond this paper
(\S\ref{sec:sustainability}): the META-LEARN project
(2026--2030) commits the two Stuttgart institutes to
maintenance and feeds the knowledge graph from its own MLIP
studies; practitioners extend the controlled vocabulary through
the candidate-vocabulary workflow without ontology-engineering
skills; and we are in contact with the maintainers of
complementary materials-science vocabularies toward federated
semantic services, with our service contributing the MLIP
metadata.


\section{Running Example: Moment Tensor Potentials for TiCr$_2$ Laves Phases}
\label{sec:running-example}

We introduce the \MLIPsOntology{} (presented in
Section~\ref{sec:ontology}) through a concrete published study: training a
Moment Tensor Potential (MTP)~\cite{shapeev2016mtp,novikov2021mlip} for
hydrogen absorption in the TiCr$_2$--H system by Kumar et
al.~\cite{kumar2025ticr2h}. TiCr$_2$ is an \emph{intermetallic
compound} which absorbes hydrogen ($H$). It crystallizes
in two close-packed forms (a cubic one and a hexagonal one), and Kumar
et al.\ train one MTP per form over the hydrogen-content range
$0 < x \le 6$ for TiCr$_2$H$_x$ (the stoichiometric ratio of H atoms
per TiCr$_2$ formula unit). We use the cubic form (called \emph{C15}
in the materials literature) as the running example. The example
illustrates how the ontology captures the full lifecycle of an MLIP
study---from the underlying physics, through the method's
hyperparameters and training data, to the evaluation of the trained
model.

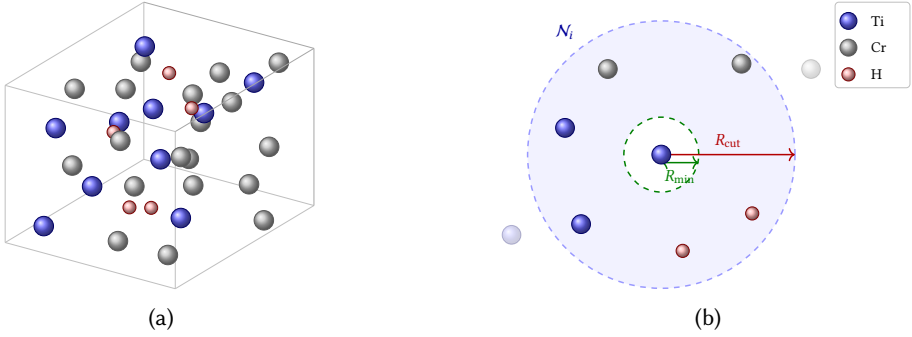
\begin{figure}[t]
\centering
\newlength{\simfigh}\setlength{\simfigh}{3.8cm}
\begin{subfigure}[t]{0.48\textwidth}
  \centering
  \resizebox{!}{\simfigh}{\begin{tikzpicture}[
    x={(0.92cm,-0.25cm)},
    y={(0.75cm,0.45cm)},
    z={(0cm,0.85cm)},
    ti/.style={circle, ball color=blue!60, draw=blue!40!black, minimum size=12pt, inner sep=0pt},
    cr/.style={circle, ball color=gray!50, draw=gray!70!black, minimum size=12pt, inner sep=0pt},
    h/.style={circle, ball color=red!30, draw=red!50!black, minimum size=8pt, inner sep=0pt},
]

\def\L{4}

\draw[gray!50, thin] (0,0,0) -- (\L,0,0);
\draw[gray!50, thin] (0,0,0) -- (0,\L,0);
\draw[gray!50, thin] (0,0,0) -- (0,0,\L);
\draw[gray!50, thin] (\L,0,0) -- (\L,\L,0);
\draw[gray!50, thin] (\L,0,0) -- (\L,0,\L);
\draw[gray!50, thin] (0,\L,0) -- (\L,\L,0);
\draw[gray!50, thin] (0,\L,0) -- (0,\L,\L);
\draw[gray!50, thin] (0,0,\L) -- (\L,0,\L);
\draw[gray!50, thin] (0,0,\L) -- (0,\L,\L);

\node[ti] at (0.5,0.5,0.3) {};
\node[cr] at (2.0,0.8,0.2) {};
\node[cr] at (3.5,0.4,0.5) {};
\node[cr] at (1.0,2.5,0.4) {};
\node[ti] at (2.5,2.0,0.3) {};
\node[cr] at (3.8,2.8,0.2) {};
\node[ti] at (0.8,3.5,0.5) {};
\node[cr] at (2.8,3.6,0.4) {};
\node[h]  at (1.7,1.5,0.6) {};

\node[cr] at (0.6,1.2,1.5) {};
\node[ti] at (1.8,0.3,1.8) {};
\node[cr] at (3.2,1.5,1.6) {};
\node[ti] at (0.4,2.8,1.7) {};
\node[cr] at (2.2,2.6,1.4) {};
\node[cr] at (3.6,3.2,1.8) {};
\node[cr] at (1.5,3.8,1.5) {};
\node[h]  at (2.7,0.9,1.2) {};
\node[h]  at (1.0,1.9,2.1) {};

\node[ti] at (0.7,0.6,2.8) {};
\node[cr] at (2.3,0.5,3.0) {};
\node[cr] at (3.4,0.9,2.7) {};
\node[cr] at (1.2,2.0,3.2) {};
\node[ti] at (2.8,2.3,2.9) {};
\node[cr] at (0.5,3.3,3.0) {};
\node[ti] at (3.0,3.5,3.1) {};
\node[cr] at (1.8,3.2,2.6) {};
\node[h]  at (3.0,1.7,3.4) {};

\node[cr] at (0.9,0.9,3.7) {};
\node[ti] at (2.5,1.2,3.5) {};
\node[cr] at (3.7,2.0,3.6) {};
\node[ti] at (1.0,2.8,3.8) {};
\node[cr] at (2.6,3.0,3.5) {};
\node[cr] at (3.5,3.6,3.7) {};
\node[h]  at (1.9,2.4,3.6) {};

\draw[gray!70, thin] (\L,\L,0) -- (\L,\L,\L);
\draw[gray!70, thin] (\L,0,\L) -- (\L,\L,\L);
\draw[gray!70, thin] (0,\L,\L) -- (\L,\L,\L);

\end{tikzpicture}}
  \caption{}
  \label{fig:simulation-cell-3d}
\end{subfigure}
\hfill
\begin{subfigure}[t]{0.48\textwidth}
  \centering
  \resizebox{!}{\simfigh}{\begin{tikzpicture}[
    x=1cm, y=1cm,
    ti/.style={circle, ball color=blue!60, draw=blue!40!black, minimum size=10pt, inner sep=0pt},
    cr/.style={circle, ball color=gray!50, draw=gray!70!black, minimum size=10pt, inner sep=0pt},
    h/.style={circle, ball color=red!30, draw=red!50!black, minimum size=7pt, inner sep=0pt},
    outside/.style={opacity=0.25},
]

\draw[dashed, thick, blue!40, fill=blue!5] (0,0) circle (2.5cm);

\draw[dashed, thick, green!50!black, fill=white] (0,0) circle (0.7cm);

\draw[<->, thick, red!70!black] (0,0) -- node[above, font=\small, red!70!black] {$\Rcut$} (2.5,0);
\draw[<->, thick, green!50!black] (0,-0.15) -- node[below, font=\small, green!50!black] {$\Rmin$} (0.7,-0.15);

\node[ti] (center) at (0,0) {};

\node[cr] (j1) at (-1.0, 1.6) {};
\node[ti] (j2) at (-1.8, 0.5) {};
\node[cr] (j3) at (1.5, 1.7) {};
\node[ti] (j4) at (-1.5, -1.3) {};
\node[h]  (j5) at (1.7, -1.1) {};
\node[h]  (j6) at (0.4, -1.8) {};

\node[cr, outside] (j7) at (2.8, 1.6) {};
\node[ti, outside] (j8) at (-2.8, -1.5) {};

\node[font=\small, blue!60!black] at (-1.8, 2.3) {$\neigh{i}$};

\draw[gray!50, thin, rounded corners=2pt] (3.25, 1.25) rectangle (4.65, 2.85);
\node[ti, minimum size=9pt] at (3.5,2.5) {};
\node[font=\footnotesize, anchor=west] at (3.8,2.5) {Ti};
\node[cr, minimum size=9pt] at (3.5,2.0) {};
\node[font=\footnotesize, anchor=west] at (3.8,2.0) {Cr};
\node[h, minimum size=7pt] at (3.5,1.5) {};
\node[font=\footnotesize, anchor=west] at (3.8,1.5) {H};

\end{tikzpicture}}
  \caption{}
  \label{fig:simulation-cell-cutoff}
\end{subfigure}
\caption{(a)~A TiCr$_2$H$_x$ simulation cell. (b)~Neighborhood
  set $\neigh{i}$ of atom $i$ (calligraphic $\mathcal{N}$, not
  nickel): atoms within $\Rcut$ contribute to
  $\localEnergy(\neigh{i})$; greyed atoms (outside $\Rcut$) are
  excluded; $\Rmin$ bounds the radial domain.}
\label{fig:simulation-cell}
\end{figure}

A \emph{simulation cell} is a finite region of space with
periodic boundary conditions containing $n$ atoms at positions
$\relPos_1, \ldots, \relPos_n$ with atomic types
$\atomType_1, \ldots, \atomType_n$
($\atomType_i \in \{\textrm{Ti},\textrm{Cr},\textrm{H}\}$ for our
TiCr$_2$H$_x$ running example). An interatomic potential predicts the total energy
$E$ and forces $\mathbf{F}_i = -\nabla_{\relPos_i} E$ acting on
each atom given only the atomic positions and types
(Fig.~\ref{fig:simulation-cell-3d}).
In quantum mechanics, the total energy can be computed accurately (within
the approximations of density functional theory) by solving the
Kohn--Sham equations~\cite{kohn1965dft}. However, this is
computationally expensive---scaling as $O(n^3)$ with the number of
atoms---limiting DFT to systems of a few hundred atoms. Machine learning
interatomic potentials approximate the DFT energy surface at a
fraction of the cost---scaling linearly in $n$ for strictly local
short-ranged potentials---enabling simulations of millions of atoms.

\begin{table}[t]
\centering
\caption{Namespace prefixes used by the \MLIPsOntology{}.}
\label{tab:prefixes}
\small
{\renewcommand{\texttt}[1]{{\scriptsize\ttfamily #1}}%
\begin{tabular*}{\textwidth}{@{}l@{\hspace{4pt}}l@{\extracolsep{\fill}}l@{\hspace{4pt}}l@{}}
\toprule
Prefix & Namespace & Prefix & Namespace \\
\midrule
\texttt{schema:} & \texttt{https://schema.org/}        & \texttt{entity:}   & \texttt{https://w3id.org/mlips/entity/} \\
\texttt{mlips:}  & \texttt{\mlipsns}                   & \texttt{wd:}       & \texttt{http://www.wikidata.org/entity/} \\
\texttt{mls:}    & \texttt{http://www.w3.org/ns/mls\#} & \texttt{mdo-calc:} & \texttt{https://w3id.org/mdo/calculation/} \\
\texttt{prov:}   & \texttt{http://www.w3.org/ns/prov\#} & \texttt{cmso:}    & \texttt{https://purls.helmholtz-metadaten.de/cmso/} \\
\texttt{mdo:}    & \texttt{https://w3id.org/mdo/core/} & \texttt{asmo:}     & \texttt{https://purls.helmholtz-metadaten.de/asmo/} \\
\texttt{qudt:}   & \texttt{http://qudt.org/schema/qudt/} & & \\
\bottomrule
\end{tabular*}}
\end{table}

\paragraph{The MTP method.}
In the MTP framework~\cite{shapeev2016mtp}, the total energy of
an atomic configuration $\config$ is decomposed into local
contributions
$\totalEnergy(\config) = \sum_{i=1}^{n} \localEnergy(\neigh{i})$,
where $\neigh{i}$ is the \emph{neighborhood} of atom~$i$, defined
as the set of all atoms within the cutoff radius $\Rcut$. The
neighborhood consists of the atomic type $\atomType_i$ of the
central atom, and for each neighbor $j$ within $\Rcut$: the
atomic type $\atomType_j$ and the relative position vector
$\relPos_{ij} = \relPos_j - \relPos_i$. Atoms beyond $\Rcut$ do
not contribute to $\localEnergy(\neigh{i})$---this
\emph{locality assumption} keeps the energy computation
tractable. The local energy
$\localEnergy(\neigh{i}) = \sum_{\alpha} \xi_\alpha\,
\basisFunc_\alpha(\neigh{i})$ is a linear combination of basis
functions, where the $\basisFunc_\alpha$ are \emph{moment tensor
descriptors}---invariant functions of the neighborhood that
capture the local atomic environment---and
$\fitParams = \{\xi_\alpha\}$ are the \emph{parameters to be
learned} by fitting to DFT reference data. The basis functions
$\basisFunc_\alpha$ are contractions of moment tensors
$M_{\mu,\nu}$ involving radial functions
$\radialBasis_\mu(|\relPos_{ij}|,z_i,z_j)$ that vanish smoothly
at $\Rcut$ and are evaluated above a minimum interatomic distance
$\Rmin$ (a numerical lower bound; physical interactions are not
truncated). MTP thus has four hyperparameters:
two geometric (the cutoff radius $\Rcut$ and minimum interatomic
distance $\Rmin$, illustrated in
Fig.~\ref{fig:simulation-cell-cutoff}) and two algorithmic that
control basis size (the number of radial basis functions
$\nRadialBasis$ and the maximum polynomial level $\levmax$ of
the moment-tensor descriptors, both systematically increasable
for higher accuracy).

The MLIP workflow---and the ontology---separate three levels of
parameterization: \emph{hyperparameter definitions} (a
parameter's name, data type, valid range, default; declared once
in the controlled vocabulary or paper-locally with a
\texttt{candidateForVocabulary} marker), \emph{hyperparameter
settings} (concrete values chosen for a specific MLIP run, e.g.,
$\Rcut = 5.0$~\AA, $\levmax = 16$), and the \emph{learned
parameters} $\fitParams$ that define the trained model.

To enable queries such as ``which physically motivated
hyperparameters does this method accept?'', we classify
hyperparameters into three non-disjoint subclasses of
$\HyperparameterC$: physical (parameters set by reference to a
physical scale, such as an interaction range---strictly
speaking, every hyperparameter is a parameter of the
mathematical approximation rather than of the physics, and this
subclass marks those whose values domain scientists choose on
physical grounds), architectural (model-architecture choice), and
training (training-procedure parameter). The non-disjointness is
a modelling choice informed by domain expertise: $\Rcut$ and
$\Rmin$ are physically motivated (they tie to interaction-range and
short-distance scales), the optimizer's learning rate is purely
training, but $\nRadialBasis$ and $\levmax$ are both
architectural \emph{and} physically motivated---they control the model size
and also determine how finely the radial and angular expansions
are resolved. A strict
partition would misrepresent how materials scientists think about
these parameters.

\paragraph{Encoding: run, settings, model.}
The example is expository, not the design method: the schema's
decisions were elicited from the competency questions of
Section~\ref{sec:requirements}, their rationale is given in
Section~\ref{sec:ontology} and \extref{sec:appendix-run}, and
their validation is extensional---competency-question execution
over the 20-paper corpus of Section~\ref{sec:evaluation}, of
which this study is one entry. The listings below show what the
finished vocabulary looks like in use.
Throughout the section, prefix
\texttt{entity:} is bound to
\texttt{https://w3id.org/mlips/entity/}.
An $\MLIPRunC$ applies a method to a training dataset under
concrete hyperparameter settings---each an instance of
$\HyperparameterSettingC$ linking to a definition---and produces
the trained model. A $\TrainedModelC$ holds the fitted
coefficients $\fitParams$:
\begin{lstlisting}[language=turtle,caption={MLIP run, a hyperparameter setting, and the trained model.},label=lst:trained-model]
entity:run-mtp-TiCr2H-c15 a mlips:MLIPRun ;
    mlips:appliesMethod entity:MTP ;
    mlips:runsOn entity:ds-TiCr2H-c15 ;
    mlips:produces entity:model-mtp-TiCr2H-c15 ;
    mlips:hasHyperparameterSetting entity:setting-rcut .
entity:setting-rcut a mlips:HyperparameterSetting ;
    mlips:forHyperparameter mlips:cutoffRadius ;
    mlips:settingValue "5.0"^^xsd:double ;
    mlips:hasUnit unit:ANGSTROM .
entity:model-mtp-TiCr2H-c15 a mlips:TrainedModel ;
    rdfs:label "C15-MTP for TiCr2-H (lev16, rcut5)" .
\end{lstlisting}

\noindent
Hyperparameter \emph{definitions} are attached to the method,
whereas the \emph{settings} of
Listing~\ref{lst:trained-model} belong to the run. The
method also names its three components (functional form, loss
function, training algorithm), supported simulation types, and
implementations (Listing~\ref{lst:algorithm}):
\begin{lstlisting}[language=turtle,caption={MTP method: functional form, loss function, training algorithm, hyperparameter definitions, and implementation.},label=lst:algorithm]
entity:MTP a mlips:MLIPMethod ;
    rdfs:label "Moment Tensor Potential" ;
    mlips:hasFunctionalForm entity:mtp-functional-form ;
    mlips:hasLossFunction entity:mtp-loss ;
    mlips:hasTrainingAlgorithm entity:bfgs ;
    mlips:hasHyperparameter mlips:cutoffRadius, mlips:minInteratomicDistance,
        entity:hp-nq, mlips:momentLevel ;
    mlips:hasImplementation entity:mlip-package-v2 ;
    mlips:supportsSimulation entity:sim-md, entity:sim-geopt .
entity:mtp-functional-form a mlips:FunctionalForm ; rdfs:label "MTP moment-tensor polynomial" .
entity:mtp-loss a mlips:LossFunction ; rdfs:label "Weighted MSE (energy 1, force 0.01, stress*V 0.001)" .
entity:bfgs a mls:Algorithm ; rdfs:label "BFGS" .
entity:hp-nq a mlips:Hyperparameter ;
    mlips:hyperparameterName "N_Q" ;
    mlips:hyperparameterDatatype xsd:integer ;
    mlips:candidateForVocabulary mlips:Hyperparameter .
entity:mlip-package-v2 a mlips:Implementation ;
    mlips:implementedIn mlips:MLIP ; mlips:version "2" .
entity:sim-md a mlips:SimulationType ; rdfs:label "molecular dynamics" .
entity:sim-geopt a mlips:SimulationType ; rdfs:label "geometry optimization" .
\end{lstlisting}

\paragraph{Training data.}
Ground truth comes from DFT calculations of total energy, forces,
and (optionally) stresses for each configuration; configurations
are drawn from MD trajectories, random perturbations, or active
learning~\cite{novikov2021mlip}. The initial low-hydrogen
enumeration in Kumar et al.~\cite{kumar2025ticr2h} yields 1{,}019
configurations on $2{\times}2{\times}2$ C15 supercells, computed
with PBE/PAW and a 400~eV plane-wave cutoff in VASP:

\begin{lstlisting}[language=turtle,caption={Training dataset and DFT settings.},label=lst:training-data]
entity:ds-TiCr2H-c15 a mlips:TrainingDataset ;
    rdfs:label "TiCr2-H DFT dataset (C15, step 1)" ;
    mlips:coversMaterial entity:mat-TiCr2 ;
    mlips:coversProperty mlips:Energy, mlips:Forces, mlips:Stresses ;
    mlips:datasetProvenance mlips:Published ;
    mlips:numConfigurations "1019"^^xsd:integer ;
    mlips:hasDFTCalculation entity:dft-TiCr2H .
entity:mat-TiCr2 a mlips:MaterialSystem ;
    mlips:chemicalFormula "TiCr2" ;
    mlips:materialClass "Laves phase, C15 cubic (Fd-3m)" .
entity:dft-TiCr2H a mlips:DFTCalculation ;
    mlips:hasDFTSettings entity:dft-settings-TiCr2H .
entity:dft-settings-TiCr2H a mlips:DFTSettings ;
    mlips:usedDFTCode mlips:VASP ;
    mlips:xcFunctional mlips:PBE ;
    mlips:pseudopotentialType mlips:PAW ;
    mlips:energyCutoff "400"^^xsd:double ;
    mlips:kPointMesh "4x4x4 Gamma-centered" .
\end{lstlisting}

\paragraph{Evaluation.}
Each evaluation produces a $\BenchmarkResultC$ within a
$\BenchmarkStudyC$:
\begin{lstlisting}[language=turtle,caption={Benchmark study and evaluation result.},label=lst:benchmark]
entity:study-kumar2025 a mlips:BenchmarkStudy ;
    rdfs:label "Kumar et al. (2025) - MTP for TiCr2-H Laves phases" ;
    mlips:hasResult entity:result-01 .
entity:result-01 a mlips:BenchmarkResult ;
    mlips:evaluatesModel entity:model-mtp-TiCr2H-c15 ;
    mlips:targetMaterial entity:mat-TiCr2 ;
    mlips:hasAccuracyMetric entity:metric-rmse-energy .
entity:metric-rmse-energy a mlips:AccuracyMetric ;
    mlips:metricType mlips:RMSE ;
    mlips:metricProperty mlips:EnergyProperty ;
    mlips:metricValue "3.17"^^xsd:double ;
    mlips:hasUnit mlips:MilliEV-PER-ATOM .
\end{lstlisting}

\section{Related Work}
\label{sec:related-work}

We organize the related work into three streams: ontologies for
machine learning, ontologies for materials science, and
cross-domain metadata templates and
FAIR~\cite{wilkinson2016fair,Wilkinson2019appendum}-data consortia.
The MLIPs Ontology is a domain-specific schema; for cross-domain
terminology covering scientific-workflow systems generally we
refer to Suter et al.~\cite{suter2026workflows}.
Section~\ref{sec:eval-comparison} returns to the ontologies
surveyed here: Table~\ref{tab:comparison} assesses, term by
term, which of them can answer the competency questions of
Section~\ref{sec:requirements}.

\paragraph{Ontologies for Machine Learning.}
ML-Schema~\cite{publio2018mlschema}, from a W3C community group, provides
top-level classes for ML algorithms, datasets, and experiments.
DMOP~\cite{keet2015dmop} supports ontology-driven algorithm selection and
hyperparameter optimization for data mining workflows. The MEX
vocabulary~\cite{esteves2015mex} provides a lightweight RDF vocabulary for ML
experiment metadata, built on PROV-O. The Common Metadata Ontology
(CMO)~\cite{foltin2024cmo} integrates pipeline metadata from multiple ML
platforms. These efforts cover generic ML workflows but miss MLIP-specific
concepts: algorithm families with physics-specific
hyperparameters (cutoff radii, angular/radial basis functions),
training-set construction strategies, and DFT reference settings.

\paragraph{Ontologies for Materials Science.}
The European Materials Modelling Ontology (EMMO) provides a
top-level ontology grounded in mereocausality
theory~\cite{horsch2020emmo}. The Materials Design Ontology
(MDO)~\cite{li2024mdo} is an OWL\,2~DL ontology with four modules
(Core, Structure, Calculation, Provenance); it has been used for
SPARQL-based integrated querying over the Materials
Project~\cite{jain2013mp} and OQMD~\cite{saal2013oqmd}.
PMDco~\cite{bayerlein2024pmdco} provides a mid-level ontology
aligned with BFO. CMSO and ASMO~\cite{menon2024atomrdf} describe
atomic structures and simulation methods, and a recent survey by
Norouzi et al.~\cite{norouzi2024survey} analyzes 60~ontologies in
the field. These ontologies cover the physical domain but miss
MLIP-training concepts: per-algorithm hyperparameters,
training-dataset construction and DFT settings, and benchmark
comparisons across algorithms, materials, and properties.
Concurrent work by Ravari et
al.~\cite{ravari2026reproduction} pursues LLM-assisted extraction
of DFT workflows aligned to CMSO/ASMO via atomRDF, addressing
workflow reproducibility rather than benchmark-comparison
knowledge graphs.

\paragraph{Metadata schemas and platforms in computational materials science.}
Major computational-materials-science platforms define rich
metadata schemas for simulation data, often in platform-specific
formats but increasingly with Semantic Web surfaces.
NOMAD~\cite{draxl2023nomad}'s NOMAD Metainfo normalizes outputs
from diverse simulation codes and is progressively aligning with
EMMO. The Materials Project~\cite{jain2013mp},
AFLOW~\cite{curtarolo2012aflow}, and OQMD~\cite{saal2013oqmd}
expose data through custom REST APIs with their own JSON
schemas; OPTIMADE~\cite{andersen2021optimade} standardizes
cross-database queries via a JSON:API spec.
AiiDA~\cite{huber2020aiida} automatically tracks full data
provenance as a directed acyclic graph and, with Materials
Cloud~\cite{talirz2020materialscloud}, provides
data-dissemination infrastructure.
Kadi4Mat~\cite{brandt2021kadi4mat} is a virtual research
environment for FAIR data management with hierarchical
collections, access control, and metadata records (e.g., the
file repository in Circular Factory
CRC~1574~\cite{thapa2025inf}); recent versions add RDF export
and SPARQL endpoints. Recent work by
Azoc{\'a}r Guzm{\'a}n et al.~\cite{azocarguzman2026workflows}
demonstrates FAIR-compliant RDF workflows using CMSO, ASMO, and
PROV-O within pyiron, and a follow-on infrastructure paper from
the same group~\cite{guzman2026kg} generalizes the approach to
a knowledge graph of $\sim$750\,K triples across $\sim$8\,K
samples covering grain-boundary, cross-dataset, and thermodynamic
queries beyond pyiron-specific workflows.
These schemas, platforms, and workflow graphs focus on DFT
calculations, material structures, and simulation provenance,
but lack descriptions of specific MLIP algorithms and
workflows---hyperparameters, training-set construction, and
cross-paper benchmarks.
Concretely, a user of these platforms today cannot ask: which
trained models were fitted to PBE reference data with a
5\,\AA{} cutoff, and how do they rank on phonon
benchmarks?---the platforms store MLIP runs as opaque files, so
selection and comparison queries over method, training-data,
and benchmark metadata have nothing to bind to.

The \MLIPsOntology{} offers these platforms an MLIP-specific
shared vocabulary they can adopt to expose their metadata with
explicit, machine-interpretable semantics.

\paragraph{Cross-domain metadata templates and FAIR-data consortia.}
Four further streams target ML-wide metadata templates or
coordinate FAIR-data efforts at consortium scale, each
complementary to the \MLIPsOntology{}.
\emph{Datasheets for Datasets}~\cite{gebru2021datasheets} and
\emph{Model Cards}~\cite{mitchell2019modelcards} are narrative
templates for ML datasets and models, now de-facto practice in
mainstream ML publishing.
\emph{Croissant}~\cite{akhtar2024croissant} formalizes the
dataset side as a JSON-LD schema, now used by the Hugging Face
Hub, Kaggle, and OpenML; Hugging Face's own model-card
metadata~\cite{huggingface_metadata} is the YAML-frontmatter
counterpart for trained-model descriptions. The \MLIPsOntology{}
serves the algorithm- and benchmark-side counterpart of this stack:
where Croissant and Datasheets describe \emph{what is in a dataset},
our $\TrainingDatasetC{}$ captures \emph{how the dataset was
generated} via DFT or wavefunction reference calculations, and our
$\BenchmarkResultC{}$ structures the cross-paper accuracy
comparisons that Model Cards leave to free-text.
At cross-sector level, the \emph{DCAT Application Profile}
(DCAT-AP)~\cite{dcatap} is the European standard for data-portal
metadata, implemented by \texttt{data.europa.eu} and many
national portals; the \MLIPsOntology{} extends DCAT-AP at the
algorithm-and-benchmark level: an MLIP-related catalog entry
is a DCAT-AP \texttt{Dataset} whose distribution carries our
\texttt{mlips:} predicates in addition to DCAT-AP's generic ones.
Adjacent EMMO-aligned domain ontologies form a third stream:
\emph{CHAMEO}~\cite{chameo} formalizes characterization-method
metadata across analytical chemistry, and
\emph{BattINFO}~\cite{battinfo} extends EMMO to battery
interfaces and electrochemistry. The \MLIPsOntology{} aligns
with the same EMMO-rooted ecosystem (\S\ref{sec:eval-comparison});
where CHAMEO and BattINFO model the underlying chemistry, our
schema models the ML-driven \emph{interpolation} of that
chemistry.
At consortium scale, the German
\emph{NFDI-MatWerk}~\cite{nfdi_matwerk} consortium operates an
ontology registry that materials-science vocabularies (CMSO,
ASMO, BattINFO, MDO) are progressively adopting; sister consortia
(NFDI4Chem for chemistry, NFDI4Ing for engineering) use the same
Linked Open Vocabularies pattern and are candidate alignment
targets.


\section{Requirements}
\label{sec:requirements}

Requirements were gathered through iterative discussions with
domain experts (materials scientists and MLIP developers) in the
META-LEARN project~\cite{grabowski2024metalearnerc}. We followed
the LOT (Linked Open Terms)
methodology~\cite{povedavillalon2022lot}, whose requirements
phase yielded the following competency questions:

\begin{enumerate}
\item[CQ1.] Which MLIP algorithms exist, and what hyperparameters does each
  accept (with types, ranges, and default values)?
\item[CQ2.] Which libraries implement a given algorithm, and what versions
  are available?
\item[CQ3.] For a given material system (a chemical composition
  together with its phase or crystal structure), which training
  datasets are available,
  and what DFT code and settings were used to generate them?
\item[CQ4.] What is the provenance of a training dataset (published source,
  in-house calculation, augmented from existing data)?
\item[CQ5.] How many atomic configurations does a training dataset contain,
  what properties are covered (energies, forces, stresses), and along which
  sampling strategy were the configurations drawn (e.g., chemical, vibrational,
  or a combination)?
\item[CQ6.] Which published studies have benchmarked a given algorithm on a
  given material property, and what accuracy metrics were reported?
\item[CQ7.] For a given material and target property, how do different
  algorithm and hyperparameter combinations rank by accuracy?
\item[CQ8.] Which simulation types (molecular dynamics, Monte Carlo, geometry
  optimization, phonon calculations, thermodynamic integration) have been
  performed with a given MLIP?
\item[CQ9.] For a given target accuracy, which algorithms and trained
  models are most efficient---both in asymptotic complexity, known
  from the method's design, and in measured training and inference
  cost, as reported by studies?
\end{enumerate}

We extended LOT's intensional CQ-agreement loop with an
extensional validation cycle: the seeded knowledge graph
(\S\ref{sec:eval-kg}), extracted from twenty MLIP papers via an
agentic-AI protocol
(\extrefshort{sec:appendix-example-protocol}), was returned to
domain experts for review. Reviewing encoded data rather than
the schema in isolation surfaced gaps and ambiguities most
efficiently; round-trip validation per paper catches schema
drift across iterations.


\section{The MLIPs Ontology}
\label{sec:ontology}

\begin{figure}[t]
\centering
\resizebox{\textwidth}{!}{
%
\begin{tikzpicture}[
    x=1cm, y=1cm,
    every node/.style={font=\footnotesize},
    methodnode/.style={draw, rounded corners=2pt, fill=algofill,
        minimum height=0.55cm, inner sep=3pt, align=center,
        font=\footnotesize, text height=1.6ex, text depth=0.4ex},
    trainnode/.style={draw, rounded corners=2pt, fill=trainfill,
        minimum height=0.55cm, inner sep=3pt, align=center,
        font=\footnotesize, text height=1.6ex, text depth=0.4ex},
    benchnode/.style={draw, rounded corners=2pt, fill=benchfill,
        minimum height=0.55cm, inner sep=3pt, align=center,
        font=\footnotesize, text height=1.6ex, text depth=0.4ex},
    centerwidth/.style={minimum width=2.5cm},
    edge/.style={->, >=stealth, thin},
    crossedge/.style={->, >=stealth, thin, dashed},
    shortcut/.style={->, >=stealth, thin},
    edgelabel/.style={font=\scriptsize, fill=white, inner sep=0.5pt,
        text height=1.4ex, text depth=0.3ex},
]
  \node[methodnode, centerwidth] (Implementation)  at (5, 7.0) {\ImplementationC};
  \node[methodnode, centerwidth] (MLIPMethod)      at (5, 5.7) {\MLIPMethodC};
  \node[methodnode, centerwidth] (MLIPRun)         at (5, 4.4) {\MLIPRunC};
  \node[methodnode, centerwidth] (TrainedModel)    at (5, 3.1) {\TrainedModelC};
  \node[trainnode,  centerwidth] (MaterialSystem)  at (5, 1.8) {\MaterialSystemC};
  \node[trainnode,  centerwidth] (TrainingDataset) at (5, 0.5) {\TrainingDatasetC};
  \node[methodnode, anchor=west] (Library)              at (-1.1, 7.0) {\LibraryC};
  \node[methodnode, anchor=west] (SimulationType)       at (-1.1, 5.7) {\SimulationTypeC};
  \node[trainnode,  anchor=west] (AtomicConfiguration)  at (-1.1, 1.8) {\AtomicConfigurationC};
  \node[methodnode, minimum width=3.2cm] (HyperparameterSetting) at (11.25, 3.1) {\HyperparameterSettingC};
  \node[benchnode,  anchor=east, minimum width=2.5cm] (BenchmarkStudy)  at (16.5, 3.1) {\BenchmarkStudyC};
  \node[benchnode,  anchor=east, minimum width=2.5cm] (BenchmarkResult) at (16.5, 1.8) {\BenchmarkResultC};
  \node[benchnode,  anchor=east, minimum width=2.5cm] (AccuracyMetric)  at (16.5, 0.5) {\AccuracyMetricC};
  \node[methodnode, minimum width=3.2cm] (FunctionalForm) at (11.25, 7.0) {\FunctionalFormC};
  \node[methodnode, minimum width=3.2cm] (LossFunction)   at (11.25, 5.7) {\LossFunctionC};
  \node[methodnode, minimum width=3.2cm] (Hyperparameter) at (11.25, 4.4) {\HyperparameterC};
  \node[trainnode]  (DatasetProvenance) at (3,    -0.8) {\DatasetProvenanceC};
  \node[trainnode]  (CoveredProperty)   at (7,    -0.8) {\CoveredPropertyC};
  \node[benchnode]  (MetricType)        at (12.5, -0.8) {\MetricTypeC};
  \node[benchnode, anchor=east]  (MetricProperty)    at (16.5, -0.8) {\MetricPropertyC};

  \node[trainnode] (DFTSettings)
    at ($(CoveredProperty.east)!0.5!(MetricType.west)$) {\DFTSettingsC};
  \node[trainnode] (DFTCalculation)
    at ($(CoveredProperty.east)!0.5!(MetricType.west) + (0, 1.3)$) {\DFTCalculationC};

  \draw[edge] (MLIPMethod) -- node[edgelabel] {\hasFunctionalFormR} (FunctionalForm.west);
  \draw[edge] (MLIPMethod) -- node[edgelabel] {\hasLossFunctionR} (LossFunction.west);
  \draw[edge] (MLIPMethod) -- node[edgelabel] {\hasHyperparameterR} (Hyperparameter.west);
  \draw[edge] (MLIPMethod) -- node[edgelabel] {\hasImplementationR} (Implementation);
  \draw[edge] (MLIPMethod) -- node[edgelabel] {\supportsSimulationR} (SimulationType);
  \draw[edge] (Implementation) -- node[edgelabel] {\implementedInR} (Library);
  \draw[edge] (HyperparameterSetting) -- node[edgelabel] {\forHyperparameterR} (Hyperparameter);
  \draw[edge] (MLIPRun) -- node[edgelabel] {\appliesMethodR} (MLIPMethod);
  \draw[edge] (MLIPRun) -- node[edgelabel] {\producesR} (TrainedModel.north);
  \draw[edge] (TrainingDataset) -- node[edgelabel] {\hasDFTCalculationR} (DFTCalculation);
  \draw[edge] (TrainingDataset) -- node[edgelabel] {\coversMaterialR} (MaterialSystem);
  \draw[edge] ($(TrainingDataset.north west)!2/3!(TrainingDataset.south west)$)
    to[out=180, in=270] node[edgelabel, pos=0.6] {\hasConfigurationR} (AtomicConfiguration.south);
  \draw[edge] (DFTCalculation) -- node[edgelabel] {\hasDFTSettingsR} (DFTSettings);
  \draw[edge] (BenchmarkStudy) -- node[edgelabel] {\hasResultR} (BenchmarkResult);
  \draw[edge] (BenchmarkResult) -- node[edgelabel] {\hasAccuracyMetricR} (AccuracyMetric);

  \draw[edge] (TrainingDataset) -- node[edgelabel] {\datasetProvenanceR} (DatasetProvenance);
  \draw[edge] (TrainingDataset) -- node[edgelabel] {\coversPropertyR} (CoveredProperty);
  \draw[edge] (AccuracyMetric) -- node[edgelabel] {\metricTypeR} (MetricType);
  \draw[edge] (AccuracyMetric) -- node[edgelabel] {\metricPropertyR} (MetricProperty);

  \coordinate (runsOnEnd) at ($(TrainingDataset.north west)!1/3!(TrainingDataset.south west)$);
  \coordinate (corridor)  at ($(AtomicConfiguration.east)!0.5!(MaterialSystem.west)$);
  \path let \p1=(corridor), \p2=(MLIPRun.west), \p3=(runsOnEnd) in
    coordinate (runsOnTopCorner) at (\x1, \y2)
    coordinate (runsOnBotCorner) at (\x1, \y3);
  \draw[crossedge, rounded corners=4pt]
    (MLIPRun.west) -- (runsOnTopCorner) --
    node[edgelabel] {\runsOnR}
    (runsOnBotCorner) -- (runsOnEnd);

  \draw[crossedge]
    ($(BenchmarkResult.north west)!1/4!(BenchmarkResult.south west)$)
    to[out=180, in=0]
    node[edgelabel, pos=0.5] {\evaluatesModelR}
    ($(TrainedModel.north east)!3/4!(TrainedModel.south east)$);
  \draw[crossedge] (BenchmarkResult) -- node[edgelabel] {\targetMaterialR} (MaterialSystem);

  \draw[edge] (MLIPRun.east) to[out=0, in=180]
    node[edgelabel, pos=0.5] {\hasHyperparameterSettingR}
    ($(HyperparameterSetting.north west)!1/4!(HyperparameterSetting.south west)$);
  \draw[shortcut] (TrainedModel.east) to
    node[edgelabel, pos=0.55] {\trainedUsingR}
    (HyperparameterSetting.west);

  \draw[shortcut, bend left=70, looseness=2.5]
    (TrainedModel.west) to node[edgelabel, pos=0.5] {\trainedWithR}
    ($(MLIPMethod.north west)!3/4!(MLIPMethod.south west)$);
\end{tikzpicture}}
\caption{Overview of the \MLIPsOntology{}: three modules and
  their core classes and relationships. Node fill color encodes
  module membership (blue: Method; green: Training Data;
  orange: Benchmark); cross-module relationships are drawn as
  dashed arrows.}
\label{fig:ontology-overview}
\end{figure}
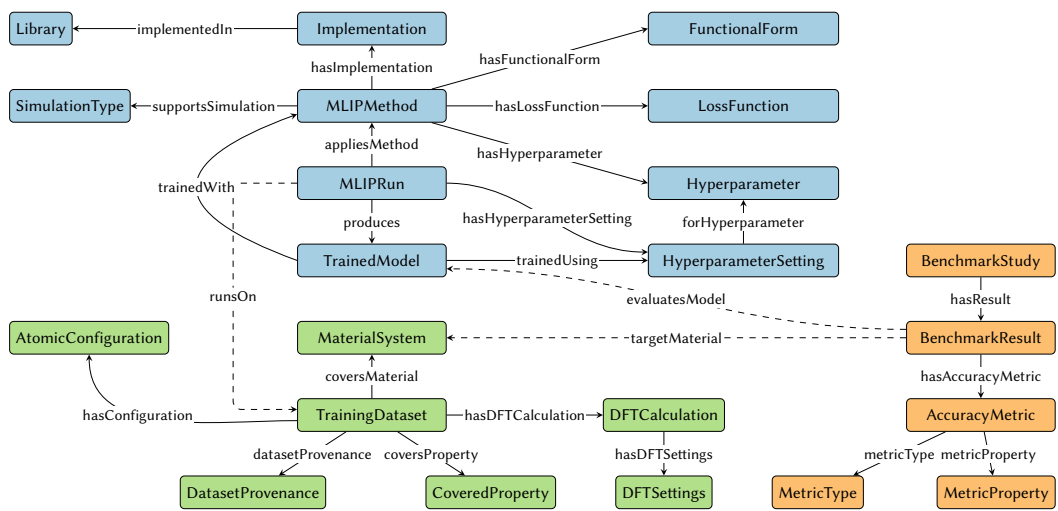

The \MLIPsOntology{} is an OWL\,2~DL ontology organized into three
modules that address the competency questions of
Section~\ref{sec:requirements}; Figure~\ref{fig:ontology-overview}
provides an overview. Beyond answering the competency questions,
the schema reflects four engineering choices of our own---adopted
as Semantic Web good practice rather than elicited from the domain
experts: alignment with existing ontologies
(\S\ref{sec:alignment}), SHACL validation (\S\ref{sec:axioms},
\S\ref{sec:sustainability}), Linked-Data integration via Wikidata
(\S\ref{sec:sustainability}), and extensibility for new algorithm
families and hyperparameters (candidate-vocabulary workflow,
\S\ref{sec:sustainability}). The ontology uses the namespace
\texttt{\mlipsns} (prefix
\mlipsprefix{}); Table~\ref{tab:prefixes} lists the external prefixes used.

\subsection{Ontology Modules}
\label{sec:modules}

\paragraph{Method module.}
\label{sec:algorithm-module}
{\sloppy
$\MLIPMethodC$ is the named recipe (e.g.,
MACE, ACE, HDNNP) for turning an atomic configuration into an
energy. It bundles a parametrized functional form
($\FunctionalFormC$, via $\hasFunctionalFormR$), a loss function
($\LossFunctionC$, via $\hasLossFunctionR$), a training algorithm
(an $\mlsAlgorithmC$ via $\hasTrainingAlgorithmR$, e.g., Adam,
L-BFGS---the sole alignment point with ML-Schema), and an atomic
environment descriptor ($\AtomicEnvironmentDescriptorC$) capturing
the local-neighborhood representation each method uses (moment
tensors for MTP, SOAP for GAP, symmetry functions for HDNNP,
equivariant message passing for MACE/NequIP, atomic cluster
expansion for ACE). Each method declares the $\HyperparameterC$
instances it accepts via $\hasHyperparameterR$, names its
$\ImplementationC$s in specific $\LibraryC$s and versions via
$\hasImplementationR$, and marks the kinds of $\SimulationTypeC$
it supports via $\supportsSimulationR$. An $\MLIPRunC$
$\appliesMethodR$ a method, $\runsOnR$ a $\TrainingDatasetC$, and
$\producesR$ a $\TrainedModelC$, with an optional
$\hasTrainingRunR$ to an underlying $\mlsRunC$. An $\MLIPRunC$
denotes one concrete execution: training is generally
non-deterministic (initialization, data shuffling, parallel
reduction order), so repeated runs with identical settings are
distinct individuals that may produce distinct
$\TrainedModelC$s. $\TrainedModelC$
carries shortcut roles $\trainedWithR$ and $\trainedOnR$ (property
chains through the run). Hyperparameter values are recorded as
domain-specific datatype properties (e.g., $\cutoffRadiusR$,
$\numLayersR$) with QUDT-annotated
units~\cite{hodgson2014qudt}. Efficiency metadata is split:
prior-knowledge complexity (e.g., $\dlRole{trainingComplexity}$,
$\dlRole{inferenceComplexity}$, $\dlRole{supportsGPU}$) on
$\MLIPMethodC$; measured cost (e.g., wall-clock, peak memory)
on $\MLIPRunC$; measured inference cost on the corresponding
$\BenchmarkResultC$. The module thus addresses CQ1 (methods and
their hyperparameters), CQ2 (implementations and versions), CQ8
(supported simulation types), and CQ9 (efficiency).\par}

\paragraph{Training Data module.}
\label{sec:training-data-module}
{\sloppy
A
$\TrainingDatasetC$ holds $\AtomicConfigurationC$ instances over
one or more $\MaterialSystemC$ instances; $\numConfigurationsR$
records the dataset size, and $\coversPropertyR$ marks the
physical properties included (energy, forces, stresses, virials).
Reference data comes from a $\ReferenceCalculationC$, with two
concrete subclasses ($\DFTCalculationC$ and
$\WaveFunctionCalculationC$\footnote{We include wave-function
methods (CCSD(T), MP2, CASPT2) because some MLIPs are now trained
on CCSD(T) reference data.}) each governed by its own settings
type ($\DFTSettingsC$, $\WaveFunctionSettingsC$); method-specific
data properties ($\xcFunctionalR$, $\kPointMeshR$ for DFT;
$\wfMethodR$, $\basisSetR$ for wave-function methods) live on
each subtype. The DFT-specific predicates $\hasDFTCalculationR$,
$\hasDFTSettingsR$, and $\dlRole{usedDFTCode}$ are declared
\texttt{rdfs:subPropertyOf} their family-agnostic counterparts
$\hasReferenceCalculationR$, $\hasReferenceSettingsR$, and
$\dlRole{usedReferenceCode}$, so legacy data validates and
supertype queries pick up DFT-only data via sub-property
entailment. $\datasetProvenanceR$ classifies the training data as
$\dlConcept{Published}$, $\dlConcept{InHouse}$, or
$\dlConcept{Augmented}$ (with PROV-O's $\provWasDerivedFromR$
tracking source datasets in the latter case); the optional
$\samplingStrategyR$ records how configurations were drawn
(e.g., chemical, vibrational, active-learning). The module thus
addresses CQ3 (datasets and their reference calculations), CQ4
(provenance), and CQ5 (size, covered properties, and sampling
strategy).\par}

\paragraph{Benchmark module.}
\label{sec:benchmark-module}
A $\BenchmarkStudyC$
contains one or more $\BenchmarkResultC$ instances, each capturing
a single evaluation of a $\TrainedModelC$ on a material system,
linked via $\evaluatesModelR$ and $\targetMaterialR$. Since a
trained model already encodes the method, training data, and
hyperparameter settings, a result need only reference the model
and the test conditions. Results report $\AccuracyMetricC$
instances via $\hasAccuracyMetricR$, each characterized by a
$\MetricTypeC$ (RMSE, MAE, R$^2$, via $\metricTypeR$), a
$\MetricPropertyC$ (energy, force, stress, via $\metricPropertyR$),
and a numeric value with unit (via $\metricValueR$);
$\reportedInR$ links a result to its source publication. The
module thus addresses CQ6 and CQ7; the
``target property'' of CQ7 corresponds to $\MetricPropertyC$,
and is distinct from the dataset-side $\CoveredPropertyC$ class;
applied target properties (e.g., hydrogen-adsorption energy) are
operationalized through energy or force metrics on specific
configurations, not as a first-class concept.

\subsection{Axioms, alignment, and meta-categorization}
\label{sec:axioms}

The schema's formal axioms fall into three categories.
\emph{Alignment axioms} (22 labelled (L1)--(L22), see below)
connect \texttt{mlips:} terms to ML-Schema, PROV-O, MDO, CMSO,
schema.org, and QUDT. \emph{Meta-categorization axioms} (see
below) are existential constraints inferred from each class's
OntoClean classification (every reified-relation class is
required to carry the entity it reifies). \emph{Other axioms}---existential restrictions, cardinality
constraints, inverse-existence axioms, property chains, and
domain restrictions---enforce data completeness and consistency
at the schema level. The full catalog is published alongside the
ontology at \url{\mlipsns}.

\paragraph{Alignment.}
\label{sec:alignment}
{\sloppy
The \MLIPsOntology{} reuses established vocabularies wherever an
existing class or property already names the concept at hand:
ML-Schema for the ML workflow, PROV-O for activity-based
provenance, MDO and CMSO for materials science, schema.org for
bibliographic metadata, and QUDT for units. Figure~\ref{fig:alignments}
shows the resulting alignment graph; each arc carries the labels
of the formal alignment axioms (L1)--(L22) it represents, with
line width proportional to the number of axioms collapsed onto
it. The full axiom catalog is provided in
\extref{tab:alignment-axioms}.
\begin{figure}[t]
\centering
\resizebox{0.7\textwidth}{!}{
%
\begin{tikzpicture}[
    x=1cm, y=1cm,
    every node/.style={font=\small},
    nodemod/.style={draw, circle, fill=classfill,
        minimum size=1.5cm, inner sep=0pt, align=center,
        font=\fontsize{8}{9.6}\selectfont},
    nodeext/.style={draw, dashed, circle, fill=extfill,
        minimum size=1.5cm, inner sep=0pt, align=center,
        font=\fontsize{8}{9.6}\selectfont},
    arc/.style={-},
    edgelabel/.style={font=\scriptsize, fill=white, inner sep=1pt},
]


  \node[nodemod] (method) at (0, 1.25)       {Method};
  \node[nodemod] (train)  at (4.72, -0.625)  {Training\\Data};
  \node[nodemod] (bench)  at (-2.17, -0.625) {Benchmark};

  \node[nodeext] (mls)    at (3.45, 1.6)   {MLS};
  \node[nodeext] (prov)   at (0, 3.35)     {PROV-O};
  \node[nodeext] (mdo)    at (5.75, 2.25)  {MDO};
  \node[nodeext] (cmso)   at (7.3, -0.6)   {CMSO};
  \node[nodeext] (schema) at (-5, -0.6)    {schema.org};
  \node[nodeext] (qudt)   at (1, -0.5)     {QUDT};

  \draw[arc, line width=3.5pt] (method) to
    node[edgelabel, pos=0.5, align=center] {L1, L12, L13 \\ L6, L17 \\ L7, L8} (mls);

  \draw[arc, line width=0.5pt] (method) to node[edgelabel, pos=0.5] {L3} (prov);
  \draw[arc, line width=0.5pt, bend right, looseness=1.25] (method) to
    node[edgelabel, pos=0.5] {L9} (qudt);

  \draw[arc, line width=0.5pt] (train) to node[edgelabel, pos=0.5] {L2} (mls);
  \draw[arc, line width=0.5pt, bend right=45, looseness=1.5] (train) to
    node[edgelabel, pos=0.5] {L4} (prov);
  \draw[arc, line width=0.5pt, bend right=15] (train) to
    node[edgelabel, pos=0.5] {L14} (mdo);
  \draw[arc, line width=1.0pt] (train) to
    node[edgelabel, pos=0.5, align=center] {L15 \\ L16} (cmso);

  \draw[arc, line width=0.5pt, bend left=60, looseness=1.25] (bench) to
    node[edgelabel, pos=0.5] {L5} (prov);
  \draw[arc, line width=0.5pt] (bench) to node[edgelabel, pos=0.5] {L18} (schema);

  \draw[arc, line width=1.0pt] (method) to
    node[edgelabel, pos=0.5, align=center] {L10 \\ L11} (train);
  \draw[arc, line width=1.0pt] (bench)  to
    node[edgelabel, pos=0.5, align=center] {L19 \\ L20} (method);
  \draw[arc, line width=1.0pt, bend right] (bench)  to
    node[edgelabel, pos=0.5, align=center] {L21 \\ L22} (train);

\end{tikzpicture}}
\caption{Alignment graph for the \MLIPsOntology{}. Solid blue
  circles are MLIPs ontology modules; dashed grey circles are
  external ontologies. Each arc carries the labels of the
  alignment axioms it represents (cf.\ (L1)--(L22)); arc
  thickness is proportional to the number of axioms collapsed
  onto it.}
\label{fig:alignments}
\end{figure}
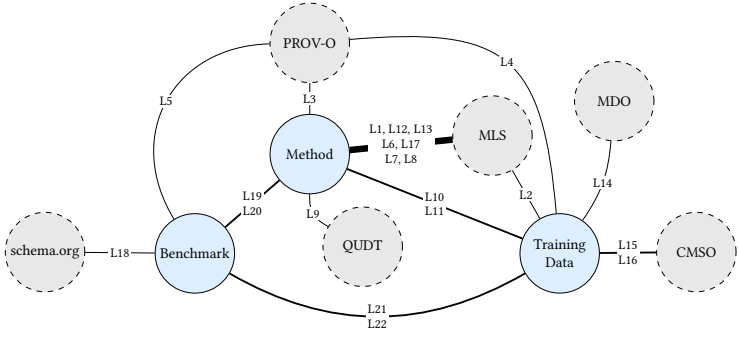
\ifextendedbuild

\begin{table}[t]
\centering
\caption{Catalogue of the 22 alignment axioms of the \MLIPsOntology{}
  (cf.\ Fig.~\ref{fig:alignments}).}
\label{tab:alignment-axioms}
\resizebox{\textwidth}{!}{$
\begin{array}{r@{\;\;}l @{\hspace{1.5em}} r@{\;\;}l}
  \text{(L1)}  & \TrainedModelC \sqsub \mlsModelC
    & \text{(L12)} & \HyperparameterC \sqsub \mlsHyperParameterC \\
  \text{(L2)}  & \TrainingDatasetC \sqsub \mlsDatasetC
    & \text{(L13)} & \ImplementationC \sqsub \mlsImplementationC \\
  \text{(L3)}  & \MLIPRunC \sqsub \provActivityC
    & \text{(L14)} & \DFTCalculationC \sqsub \mdoCalculationC \\
  \text{(L4)}  & \TrainingDatasetC \sqsub \provEntityC
    & \text{(L15)} & \MaterialSystemC \sqsub \cmsoCrystallineMaterialC \\
  \text{(L5)}  & \BenchmarkStudyC \sqsub \provActivityC
    & \text{(L16)} & \AtomicConfigurationC \sqsub \cmsoAtomicStructureC \\
  \text{(L6)}  & \MLIPMethodC \sqcap \mlsAlgorithmC \sqsub \bot
    & \text{(L17)} & \MLIPMethodC \sqsub \existsR{hasTrainingAlgorithm}{\mlsAlgorithmC} \\
  \text{(L7)}  & \exists\, \dlRole{hasTrainingRun}.\top \sqsub \MLIPRunC
    & \text{(L18)} & \top \sqsub \forall\, \reportedInR.\schemaScholarlyArticleC \\
  \text{(L8)}  & \top \sqsub \forall\, \dlRole{hasTrainingRun}.\mlsRunC
    & \text{(L19)} & \exists\, \evaluatesModelR.\top \sqsub \BenchmarkResultC \\
  \text{(L9)}  & \top \sqsub \forall\, \dlRole{hasUnit}.\dlConcept{qudt{:}Unit}
    & \text{(L20)} & \top \sqsub \forall\, \evaluatesModelR.\TrainedModelC \\
  \text{(L10)} & \exists\, \runsOnR.\top \sqsub \MLIPRunC
    & \text{(L21)} & \exists\, \targetMaterialR.\top \sqsub \BenchmarkResultC \\
  \text{(L11)} & \top \sqsub \forall\, \runsOnR.\TrainingDatasetC
    & \text{(L22)} & \top \sqsub \forall\, \targetMaterialR.\MaterialSystemC \\
\end{array}$}
\end{table}
\fi
The single salient design decision---the disjointness in
(L6) between $\MLIPMethodC$ and $\mlsAlgorithmC$, with
$\hasTrainingAlgorithmR$ (L17) as the sole connecting
role---keeps the domain-level recipe distinct from the generic
ML training algorithm it delegates to; the decision is
discussed in detail in \extref{sec:appendix-method}.
Beyond the 22 cataloged axioms, the schema carries 14 additional
\texttt{rdfs:subClassOf} / \texttt{rdfs:subPropertyOf} edges from
\texttt{mlips:} terms to their closest external parents (e.g.,
\texttt{mlips:Library} $\sqsubseteq$ \texttt{schema:SoftwareApplication},
\texttt{mlips:HyperparameterSetting} $\sqsubseteq$
\texttt{mls:HyperParameterSetting},
\texttt{mlips:metricValue} $\sqsubseteq$ \texttt{qudt:value}),
making the alignment queryable rather than narrative-only.
Enumerative classes (e.g., \texttt{XCFunctional},
\texttt{MetricType}), descriptor classes, and bespoke constructs
(e.g., \texttt{FunctionalForm}, \texttt{LossFunction},
\texttt{SamplingStrategy}) lack a clear external parent and
remain unaligned.%
\ifextendedbuild{} The full edge list:\else{} The full edge list
is provided in the supplemental material.\fi
\par}

\ifextendedbuild
\begin{itemize}
\item \texttt{mlips:Library} $\sqsubseteq$
  \texttt{schema:SoftwareApplication};
\item \texttt{mlips:Implementation} $\sqsubseteq$
  \texttt{schema:SoftwareSourceCode} (in addition to L13);
\item \texttt{mlips:BenchmarkResult} $\sqsubseteq$ \texttt{prov:Entity};
\item \texttt{mlips:AccuracyMetric} $\sqsubseteq$
  \texttt{qudt:QuantityValue};
\item \texttt{mlips:MaterialSystem} $\sqsubseteq$ \texttt{mdo:Material}
  (in addition to L15);
\item \texttt{mlips:ReferenceCalculation} $\sqsubseteq$
  \texttt{prov:Activity} (and inherited by \texttt{DFTCalculation}
  and \texttt{WaveFunctionCalculation});
\item \texttt{mlips:TrainedModel} $\sqsubseteq$ \texttt{prov:Entity}
  (in addition to L1);
\item \texttt{mlips:HyperparameterSetting} $\sqsubseteq$
  \texttt{mls:HyperParameterSetting};
\item \texttt{mlips:hasHyperparameter} $\sqsubseteq$
  \texttt{mls:hasHyperParameter};
\item \texttt{mlips:hasUnit} $\sqsubseteq$ \texttt{qudt:hasUnit};
\item \texttt{mlips:metricValue} $\sqsubseteq$ \texttt{qudt:value};
\item \texttt{mlips:produces} $\sqsubseteq$ \texttt{prov:generated};
\item \texttt{mlips:runsOn} $\sqsubseteq$ \texttt{prov:used};
\item \texttt{mlips:hasResult} $\sqsubseteq$ \texttt{prov:generated}.
\end{itemize}
A SPARQL query over the merged ontology returns the full set of
external-vocabulary parents directly.
\fi

\paragraph{Meta-categorization.}
\label{sec:metasort}
We classify every \texttt{mlips:} class by its OntoClean
meta-properties~\cite{guarino2009ontoclean} into one of four
named categories (\emph{Sortal}, \emph{Subordinate Sortal},
\emph{Role}, \emph{Reified Relation}) attached as an
\texttt{mlips:metaSort} annotation. We define the four values
locally rather than importing OntoClean's OWL serialization,
which currently does not resolve and lacks a first-class entry
for reified $n$-ary relations (the most common non-sortal kind
here). One consequence is formal: each reified-relation class
carries an existential axiom requiring its carrier (e.g.,
$\HyperparameterSettingC \sqsubseteq
\existsR{isSettingOf}{MLIPRun}$); the full per-class
classification, meta-property analysis, and the CONSTRUCT rules
for computed labels and inverse triples are in
\extref{sec:appendix-metasort} and, machine-readably, in the
ontology source and its DaRUS deposit~\cite{darus2026mlips}.


\section{Evaluation}
\label{sec:evaluation}

We evaluate the \MLIPsOntology{} along multiple dimensions: a seeded
knowledge graph of published MLIP studies, competency question coverage
via SPARQL queries against that graph, consistency verification via
OWL reasoning, automated pitfall detection, and comparison with
existing ontologies.

\subsection{Seeded Knowledge Graph}
\label{sec:eval-kg}

To exercise the ontology against real data, we encoded 20
published MLIP studies into a seeded knowledge graph. The
full catalog (one subsection per paper, organized around a
12-question extraction protocol with a round-trip check) is in
\extref{sec:appendix-paper-catalogue}, and the complete
triple-level encoding of the running example (Kumar et
al.~\cite{kumar2025ticr2h}), of which
Section~\ref{sec:running-example} shows the core listings, is in
\extref{sec:appendix-worked-examples}; both are also archived on
DaRUS~\cite{darus2026mlips}. The graph contains
3{,}561 triples across the 20 paper-graph instances and 376 more
in the controlled-vocabulary file \texttt{mlips-vocab.ttl};
together with the 1{,}098-triple T-Box, this is a 5{,}027-triple
graph.

\paragraph{Diversity.}
The corpus deliberately covers multiple axes of variation
(Table~\ref{tab:kg-diversity}). Method families span the chronological
arc of the field (HDNNP from 2007 through MACE-MP-0 in 2024) and the
architectural spectrum (descriptor + regressor in GAP, MTP, ACE;
deep-network in HDNNP, ANI, SchNet, DeepMD; equivariant message passing
in MACE, NequIP, Allegro; universal/foundation models in M3GNet,
CHGNet, MACE-MP-0). Reference data is predominantly DFT, but one paper
(\texttt{smith2019ccx}, the ANI-1ccx transfer-learning study) trains
to CCSD(T) data and exercises the
$\WaveFunctionCalculationC$/$\WaveFunctionSettingsC$ branch added in
this work. Materials cover elemental crystals (Si, Cu, Fe, W),
intermetallics (TiAl, Ti$_3$Al, Ni--Al, Ti--Al--V), Laves phases
(TiCr$_2$), magnetic oxides (MnO), high-entropy alloys (TaVCrW),
hydrogen-bonded liquids (water and ice polymorphs), organic molecules,
and universal coverage of the Materials Project (MACE-MP-0, CHGNet,
M3GNet).

\begin{table}[t]
\centering
\caption{Diversity axes of the seeded knowledge graph.}
\label{tab:kg-diversity}
\begin{tabularx}{\textwidth}{@{}l@{\hspace{8pt}}X@{}}
\toprule
\textbf{Axis} & \textbf{Coverage} \\
\midrule
Papers & 20 \\
Method families & HDNNP, GAP, MTP, ANI, ACE, SchNet, MACE/NequIP/Allegro, M3GNet/CHGNet, DeepMD \\
Reference method & DFT (19), wave-function (1) \\
XC functionals (DFT) & PBE (12), PBE0 (2), PW91 (2), HSE06 (1), $\omega$B97X (2), LDA (1); 19/20 studies report an XC functional; the counts sum to 20 because lysogorskiy2021ace reports two (PBE for Cu, PW91 for Si) \\
Pseudopotential / basis & PAW (8), ultrasoft (2), norm-conserving (1), all-electron with $\dlConcept{DftBasisSet}$ (3); 6 studies do not report a basis specification \\
Sampling strategies & vibrational/AIMD, chemical-space, active-learning, random-perturbation, surface enumeration, defect sampling \\
\bottomrule
\end{tabularx}
\end{table}

\paragraph{Controlled vocabulary and extension.}
The vocabulary file declares named individuals (e.g.,
$\dlConcept{mlips{:}PBE}$, $\dlConcept{mlips{:}PAW}$,
$\dlConcept{mlips{:}DLPNO\_CCSDT}$) for the four open
extension-point classes ($\XCFunctionalC$,
$\PseudopotentialTypeC$, $\WfMethodC$, $\DftBasisSetC$). When an
encoded paper reports a value not yet in the vocabulary, the
encoder mints a paper-local IRI tagged
$\candidateForVocabularyR$; the corpus currently has two such
candidates (Gaussian basis sets 6-31G(d) and 6-31G$^*$), both
flagged for promotion review.

\paragraph{Reportable-vs-reported gaps.}
A first cross-corpus pass exposes the metadata gap motivating
the ontology: 18/20 studies report energy errors and 15/20
report force errors, but only 9/20 report any training- or
inference-cost metadata, and 0/20 report peak
memory.\ifextendedbuild{} Per-predicate: training hardware 6,
inference time per atom 5, inference hardware 4, training
duration 2, GPU hours 1.\fi{} The ontology can express each of
these per paper; the literature reports them only
intermittently. Section~\ref{sec:eval-cq} formalizes this
through competency-question SPARQL queries.

\subsection{Competency Question Execution}
\label{sec:eval-cq}

All nine competency questions (Section~\ref{sec:requirements})
are expressible as SPARQL queries over the ontology, with
several spanning all three modules (e.g.\ joining a benchmark
result through its trained model to the underlying method and
training dataset). The full query templates are in
\extref{sec:appendix-sparql}.
\begin{table}[t]
\centering
\caption{Competency-question execution on the seeded corpus
  (5{,}027 triples). Row counts reflect the current corpus; the
  times were measured on the 0.1.0 submission corpus with the
  \texttt{sparql} Rust CLI on a single machine, no reasoner;
  queries in \extref{sec:appendix-sparql}.}
\label{tab:cq-results}
\small
\begin{tabularx}{\textwidth}{@{}lrrX@{}}
\toprule
CQ  & Rows & Time (ms) & Note \\
\midrule
CQ1 & 81  & 117 & Hyperparameter metadata
                 (\texttt{hyperparameterName}, datatype) populated
                 across 13 canonical \texttt{Hyperparameter} individuals
                 in \texttt{mlips-vocab.ttl} (\texttt{cutoffRadius},
                 \texttt{numLayers}, \texttt{learningRate}, \dots);
                 paper-local survivors carry the same metadata plus a
                 \texttt{candidateForVocabulary} marker for a future
                 curation pass. \\
CQ2 & 20  & 33 & One implementation per paper. \\
CQ3 & 21  & 33 & One DFT or wavefunction-method settings block per
                 dataset; 21 of the 22 datasets carry the full
                 settings tuple. \\
CQ4 & 22  & 31 & Provenance recorded for every dataset (22 datasets
                 across the 20 papers; two papers split their
                 data). \\
CQ5 & 135 & 26 & Property/sampling combinations across all datasets. \\
CQ6 & 46  & 37 & 46 benchmark results across the 20 studies. \\
CQ7 & 27  & 76 & RMSE-ranked rows; ordering reflects the cross-paper accuracy comparison. \\
CQ8 & 38  & 74 & \texttt{supportsSimulation} triples are not asserted
                 by the source literature (a metadata gap surfaced by
                 the seeded corpus); we curated them manually for the
                 17 studies whose simulation types are stated in
                 prose (6 simulation-type individuals). \\
CQ9 & 3   & 41 & Only 5 of 20 papers report
                 \texttt{inferenceTimePerAtom} (cf.\ the metadata gaps
                 in \S\ref{sec:eval-kg}); the threshold
                 $\mathrm{RMSE}<2$ further trims the result set. \\
\bottomrule
\end{tabularx}
\end{table}
All nine queries return non-empty results. CQ8 initially
returned zero rows, reflecting the source literature's habit of
stating supported simulation types only in prose---a metadata
gap in the field, not in the schema; curating those prose
statements into triples for the 17 studies that make them (38
rows over 6 simulation types) validated the
\texttt{supportsSimulation} branch against real data.

\ifextendedbuild
\paragraph{CQ3: Training datasets with DFT provenance.}
For a given material system, which training datasets are available and
what DFT settings were used? This query traverses the Training Data
module, joining datasets with their DFT calculations and settings---a
level of detail not available from any existing ontology:
\begin{lstlisting}[language=sparql]
SELECT ?ds ?xc ?cutoff ?pseudo WHERE {
  ?ds a mlips:TrainingDataset ;
      mlips:coversMaterial ?mat .
  ?mat mlips:chemicalFormula "TiCr2" .
  ?ds mlips:hasDFTCalculation ?calc .
  ?calc mlips:hasDFTSettings ?settings .
  ?settings mlips:xcFunctional ?xc ;
            mlips:energyCutoff ?cutoff ;
            mlips:pseudopotentialType ?pseudo .
}
\end{lstlisting}

\paragraph{CQ6: Published benchmarks for an algorithm and material.}
Which studies have benchmarked a given algorithm on a given material, and
what accuracy metrics were reported? This is the query that motivates the
ontology: it traverses all three modules, linking a benchmark study to the
trained model's algorithm and the target material system:
\begin{lstlisting}[language=sparql]
SELECT ?study ?metric_type ?prop ?value WHERE {
  ?study a mlips:BenchmarkStudy ;
         mlips:hasResult ?result .
  ?result mlips:evaluatesModel ?model ;
          mlips:targetMaterial ?mat ;
          mlips:hasAccuracyMetric ?metric .
  ?run a mlips:MLIPRun ;
       mlips:produces ?model ;
       mlips:appliesMethod ?algo .
  ?algo rdfs:label "MTP" .
  ?mat mlips:chemicalFormula "TiCr2" .
  ?metric mlips:metricType ?metric_type ;
          mlips:metricProperty ?prop ;
          mlips:metricValue ?value .
}
\end{lstlisting}

\paragraph{CQ7: Ranking algorithm--hyperparameter combinations.}
For a given material and target property, how do different
algorithm and hyperparameter combinations rank by accuracy? This
analytical query ranks models by accuracy, enabling systematic
comparison across the literature:
\begin{lstlisting}[language=sparql]
SELECT ?algo ?value WHERE {
  ?result mlips:evaluatesModel ?model ;
          mlips:targetMaterial ?mat ;
          mlips:hasAccuracyMetric ?metric .
  ?run a mlips:MLIPRun ;
       mlips:produces ?model ;
       mlips:appliesMethod ?algo .
  ?mat mlips:chemicalFormula "TiCr2" .
  ?metric mlips:metricType mlips:RMSE ;
          mlips:metricProperty mlips:EnergyProperty ;
          mlips:metricValue ?value .
} ORDER BY ?value LIMIT 10
\end{lstlisting}
\fi

\subsection{Consistency and Pitfall Checks}
\label{sec:eval-consistency}
\label{sec:eval-oops}

The primary consistency check is OWL reasoning.
HermiT~\cite{glimm2014hermit} via
ROBOT~\cite{jackson2019robot}, cross-checked with ELK on the
\textit{EL} fragment, confirms that the ontology is
consistent (no contradictions among axioms (A1)--(A27)),
every class is satisfiable, and the inferred class hierarchy
contains no unexpected subsumptions. HermiT completes in
$\approx$\,4\,s; the post-reasoning ontology has 669 axioms.

{\sloppy
As a supplementary check we ran the
OOPS!~\cite{poveda2014oops} open-source CLI (reproducible
via \texttt{make oops}), which implements seven pitfall
checkers (P02--P08): P02, P03, P05, P07 return zero issues,
P06 errors on a known tool-side bug, and P04 (unconnected
elements) and P08 (missing annotations) each trigger one
finding at \emph{minor} importance. Both findings trace to
external alignment-target classes (e.g.\
\texttt{mdo:Material}, \texttt{mls:HyperParameter},
\texttt{schema:Person}, \texttt{qudt:QuantityValue},
\texttt{cmso:AtomicStructure}, \texttt{prov:Entity}) that
\texttt{mlips:} terms reference as \texttt{rdfs:subClassOf}
parents; clearing them would require importing the external
ontologies (deliberately avoided to keep the theory in
OWL\,2~DL and the reasoner load tractable) or redefining
external classes locally. Four further OOPS! checks on the
ontology header (P38--P41: missing \texttt{owl:Ontology},
partial descriptions, namespace hijacking, missing licence)
are preempted by the schema's complete header
annotations\footnote{Full header: \texttt{owl:\{versionIRI,
versionInfo\}} and \texttt{dcterms:\{title, description,
creator, contributor, license, issued, modified\}}.}
including a CC~BY~4.0 licence.\par}

\ifextendedbuild
\begin{table}[h]
\centering
\caption{OOPS! pitfall results on \texttt{mlips.owl}, run locally
  via the open-source CLI (v0.3.0-SNAPSHOT). All flagged pitfalls
  are at \emph{minor} importance. The full pitfall catalog
  (\mbox{$\sim$40} pitfalls) is currently only available via the
  public web UI; the full report is deferred to a future
  revision.}
\label{tab:oops-results}
\small
\begin{tabular*}{\textwidth}{@{\extracolsep{\fill}}lll@{}}
\toprule
ID  & Pitfall                                       & Triggered \\
\midrule
P02 & Creating synonyms as classes                  & 0 \\
P03 & Creating ``is'' relationships                 & 0 \\
P04 & Creating unconnected ontology elements        & 1 (minor) \\
P05 & Defining wrong inverse relationships          & 0 \\
P06 & Including cycles in the hierarchy             & --- (tool bug) \\
P07 & Merging different concepts in the same class  & 0 \\
P08 & Missing annotations (label \& comment)        & 1 (minor) \\
\bottomrule
\end{tabular*}
\end{table}
\fi

\subsection{Comparison with Existing Ontologies}
\label{sec:eval-comparison}

Table~\ref{tab:comparison} compares the \MLIPsOntology{} with
existing ontologies from the ML and materials-science domains
(the \emph{comparators}).
For each comparator we walked the TBox and identified the terms
a SPARQL query would need to answer each of the nine competency
questions of Section~\ref{sec:requirements}. A cell is marked
\emph{answerable} ($\bullet$) when the comparator declares all
the required terms; \emph{partially answerable} ($\circ$) when
superclass reasoning approximates the question (e.g., a
comparator's general $\textsf{Calculation}$ class subsumes our
\DFTCalculationC{} but does not declare DFT-specific settings
such as energy cutoff or pseudopotential type as data
properties); and \emph{not answerable} (---) when the required
terms are absent from the comparator's vocabulary. The
classification is TBox-level; the assessment is independent of
each comparator's currently-deployed instance data. The full
per-cell mapping is archived at
\paperref{dataset/artifacts/kg/comparison/} so disagreement on
any cell can be tracked to a specific term.
\begin{table}[t]
\centering
\caption{Competency question coverage across comparator vocabularies
  (MLS = ML-Schema, Cr = Croissant). $\bullet$ = answerable,
  $\circ$ = partially, --- = not answerable.}
\label{tab:comparison}
\begin{tabular*}{\textwidth}{@{}l@{\extracolsep{\fill}}ccccccc@{}}
\toprule
 & MLS & MDO & \shortstack{CMSO\\ASMO} & EMMO & PMDco & Cr & \textbf{MLIPs} \\
\midrule
CQ1: Algo.\ hyperparams     & $\circ$ & ---     & ---     & ---     & ---     & ---     & $\bullet$ \\
CQ2: Implementations         & $\circ$ & ---     & ---     & ---     & ---     & ---     & $\bullet$ \\
CQ3: Training data + DFT     & ---     & $\circ$ & $\circ$ & $\circ$ & $\circ$ & $\circ$ & $\bullet$ \\
CQ4: Dataset provenance       & ---     & ---     & ---     & ---     & ---     & $\circ$ & $\bullet$ \\
CQ5: Dataset size + coverage  & ---     & ---     & ---     & ---     & ---     & $\circ$ & $\bullet$ \\
CQ6: Published benchmarks     & ---     & ---     & ---     & ---     & ---     & ---     & $\bullet$ \\
CQ7: Accuracy ranking         & ---     & ---     & ---     & ---     & ---     & ---     & $\bullet$ \\
CQ8: Simulation types         & ---     & ---     & $\circ$ & $\circ$ & ---     & ---     & $\bullet$ \\
\bottomrule
\end{tabular*}
\end{table}
{\sloppy
ML-Schema partially covers CQ1--CQ2 through $\mlsAlgorithmC$ and
$\mlsImplementationC$ but lacks physics-specific hyperparameters
and units; MDO and CMSO/ASMO partially cover CQ3 but do not model
MLIP-specific training-data provenance or DFT settings at the
required granularity. No existing ontology covers CQ4--CQ7---the
benchmark--method--dataset triangle central to the
\MLIPsOntology{}. The complementary scope is captured by
linking rather than duplication: ML-Schema's $\mlsRunC$
provides richer training-run metadata reachable via the
optional $\hasTrainingRunR$ link, and CMSO/ASMO, MDO, EMMO,
and PMDco provide detailed crystal-structure descriptions
linked via $\MaterialSystemC$.\par}

\paragraph{Croissant.}
Croissant~\cite{akhtar2024croissant}, the dataset-side counterpart
positioned in \S\ref{sec:related-work}, exhibits the mirror-image
profile. It partially covers the dataset-centric CQ3--CQ5---dataset
description, provenance via schema.org $\textsf{creator}$/$\textsf{citeAs}$,
and field-level property coverage via $\textsf{RecordSet}$/$\textsf{Field}$
with their $\textsf{dataType}$s---but declares none of the method-,
benchmark-, simulation-, or efficiency-specific terms (CQ1--CQ2,
CQ6--CQ9), and so answers no competency question fully. Its partials
are of a different kind from the comparators above: not superclass
subsumption, but generic dataset metadata lacking the MLIP-specific
specializations (provenance categories, sampling strategies,
configuration counts). Croissant describes the training set as a
distributable artifact; the \MLIPsOntology{} describes how that set
was produced and how the resulting model performs---confirming, at
the level of the competency questions, the complementarity stated
in \S\ref{sec:related-work}.

\paragraph{Union of comparators.}
The per-ontology cells could overstate the contribution if
the comparators combined (via their shared alignment to
PROV-O, schema.org, QUDT) close the gap. The union fully
answers CQ2 (ML-Schema $\mlsImplementationC$ with schema.org
versioning) and CQ3 (MDO declares DFT settings as data
properties); for CQ1 and CQ4--CQ9 the union remains $\circ$
or --- because the MLIP-specific specializations
(physics-specific hyperparameter individuals, dataset
provenance categories, sampling strategies, the benchmark
chain, efficiency metadata) are absent from every
comparator. The contribution concentrates in those seven CQs.


\section{Sustainability and Maintenance}
\label{sec:sustainability}


The MLIPs Ontology is developed within the META-LEARN
project~\cite{grabowski2024metalearnerc}, an ERC Advanced Grant
(2026--2030); it is an authoring artefact of the Stuttgart
Institute for Artificial Intelligence and the Institute for
Materials Science, which jointly commit to maintaining it beyond
the project funding period. The project's own MLIP studies
publish their metadata into the \MLIPsKG{} as they appear, so
the graph grows with the group's research output independently
of external adoption.

\paragraph{Versioning.}
The ontology follows Semantic
Versioning~\cite{prestonwerner2013semver}: each release has a
distinct version IRI under the \mlipsprefix{} namespace, with
major increments for breaking changes, minor for
backwards-compatible extensions, and patch for annotation-only
fixes (current release \texttt{0.1.0}). Each release also
deposits a snapshot to DaRUS~\cite{darus2024} for archival;
DaRUS uses sequential dataset versioning (V1, V2, \ldots) rather
than SemVer, with the mapping (SemVer MAJOR/MINOR $\to$ DaRUS
major; PATCH $\to$ DaRUS minor) kept in a \texttt{CHANGELOG.md}.

\paragraph{Governance and editorship.}
{\sloppy
The Institute for Artificial Intelligence contributes
Semantic-Web and ontology-engineering expertise plus the
engineering capacity to maintain the persistent IRI service and
the dataset CI infrastructure; the Institute for Materials
Science provides the materials-science domain expertise that
sustains the schema's MLIP-specific concepts. The editors and
contributors can annotate the ontology with comments, and have
weekly meetings to discuss the comments. Three editors
have special dedication to the project: Hern\'andez, Jung,
and Grabowski.\par}

\paragraph{Hosting and persistent identification.}
The persistent IRI \texttt{https://w3id.org/mlips}, registered
with the W3C Permanent Identifier Community Group, decouples the
ontology from any specific host using a redirection. The redirect
target serves the ontology with HTTP content negotiation
(e.g., XHTML+RDFa, Turtle, RDF/XML); institutional hosting is being
provisioned at the University of Stuttgart, with the maintainer's
domain as fallback. Instance IRIs under
\texttt{https://w3id.org/mlips/entity/} are reserved for
instance dereferencing; a content-negotiating service that
renders subject pages with forward and reverse properties is
planned alongside the institutional hosting.

\paragraph{Continuous validation.}
Every commit to the dataset repository runs three checks:
Turtle parsing with \texttt{rapper}; a SPARQL-based round-trip
check~(\extrefshort{sec:appendix-example-protocol}) that extracts
each paper's canonical triples via 11 \texttt{CONSTRUCT} queries
and verifies bit-equivalence with the source (catching schema
drift the moment a term is renamed); and OWL reasoning
consistency via ROBOT with HermiT and
ELK~(\S\ref{sec:eval-consistency}). A fourth project-wide check
runs the full competency-question battery
(\S\ref{sec:eval-cq}) on the merged corpus, surfacing schema
drift and encoding-coverage regressions.

\paragraph{Community engagement and adoption.}
{\sloppy
Two pathways drive adoption.
\emph{Producers}: MLIP frameworks (e.g., MACE, NequIP,
DeePMD-kit) already serialize the run configurations and
trained-model metadata that the schema captures, so a parser
layer---analogous to NOMAD's~\cite{draxl2023nomad}
normalizers for DFT codes---can emit MLIPs-Ontology triples
at training time, making the ontology self-populating; the
12-question protocol of
\extref{sec:appendix-example-protocol} covers \emph{legacy}
studies that predate such emitters. The motivation to populate
is direct: encoded studies are the ones selection queries
return---visibility for the method and its benchmark
results---and the twelve-question protocol doubles as a
completeness checklist that surfaces unreported metadata
(\S\ref{sec:eval-kg}) before a paper ships. \emph{Consumers}:
cross-paper questions reduce to single SPARQL queries
against the merged graph (e.g., ``which MLIPs support phonon
calculations on intermetallics with
$\mathrm{RMSE}<5$\,meV/atom?'').
The schema's \texttt{mlips:sameAsWikidata} links are already
populated for the major DFT and wave-function codes and for
18 of 21 encoded \texttt{MaterialSystem} instances; we will
contribute the schema to the NFDI-MatWerk ontology registry
(where CMSO and ASMO are indexed) and file interoperability
proposals with NOMAD and Materials
Cloud~\cite{talirz2020materialscloud} so that MLIP entries
can be exported in the \texttt{mlips:} vocabulary.
The contribution infrastructure is in place: the development
repository is public with an issue tracker, contribution
guidelines, and a canonical citation file, and the ontology has
been submitted to the LOV registry. Beyond registries, we are
in contact with the maintainers of complementary
materials-science vocabularies about federating semantic
services for materials science, with our service contributing
the MLIP metadata.\par}


\section{Conclusion}
\label{sec:conclusion}

We have presented the \MLIPsOntology{}, an OWL\,2~DL ontology
whose three modules---Method, Training Data, and Benchmark---fill
a gap left by existing ML and materials-science ontologies: the
metadata that connects an MLIP method to its training set, DFT
reference settings, trained model, and benchmark results. The
schema declares 27 formal axioms and aligns with ML-Schema, MDO,
CMSO/ASMO, PROV-O, schema.org, and QUDT; SHACL shapes enforce
these constraints at the data level. All competency questions
reduce to SPARQL queries over the seeded knowledge graph, and the
ontology is consistent under HermiT and ELK.

\paragraph{Outlook: AI-driven research.}
A machine-readable substrate of methods, training data, and
benchmarks is a prerequisite for AI-driven research in the MLIP
field. The agentic extraction protocol that seeded the knowledge
graph already demonstrates machine-in-the-loop population; the
same structure lets autonomous agents retrieve prior training
runs for a target material property, identify suitable training
sets and hyperparameters, reproduce a study from its recorded
settings, and plan extensions grounded in what has been
tried---speeding up bespoke MLIP development and the fine-tuning
of foundation models by humans and agents alike. We see the
ontology as a step toward MLIP studies whose results ship as
queryable data alongside the paper, supporting reproducibility
of and extension from previous work.

\paragraph{Resource Availability Statement.}
The \MLIPsOntology{} (OWL file, SHACL shapes, controlled
vocabulary, and full axiom catalog with examples) is available
at \url{https://w3id.org/mlips} under CC BY 4.0. The 20-paper
seeded knowledge graph is archived on DaRUS (DOI
10.18419/DARUS-5948~\cite{darus2026mlips}) for long-term
availability; its extraction protocol, paper catalogue, and
worked example are also included as appendices of this
preprint.
\paragraph{Acknowledgements.}
This work is funded by the European Research Council (ERC) under the European
Union's Horizon Europe Research and Innovation Programme, grant agreement
No.\ 101200433, project META-LEARN.
\paragraph{Declaration of use of Generative AI.}
The authors used Claude (Anthropic) for drafting and iterating on sections of
this manuscript. All content was reviewed, edited, and validated by the
authors.

\bibliographystyle{ACM-Reference-Format}
\bibliography{references}

\appendix
{\refappendixsloppy
\section{Class Reference}
\label{sec:class-reference}
\JongHyunJung{Could we have a table of reference for this supplementary info?}

This appendix describes all 36 classes defined in the \MLIPsOntology{}.

\subsection{\dlConcept{AccuracyMetric}}
\label{sec:term-AccuracyMetric}

\textbf{Label:} Accuracy Metric.
An accuracy measure reported in a benchmark
(e.g., RMSE of energy, MAE of forces).

\paragraph{Superclasses.} $\dlConcept{QuantityValue}$.

\paragraph{Outgoing properties.} $\dlRole{isMetricOf}$, $\dlRole{metricProperty}$, $\dlRole{metricType}$, $\dlRole{metricValue}$.

\paragraph{Incoming properties.} $\dlRole{hasAccuracyMetric}$.

\paragraph{Example.}
An RMSE of energy metric reporting 1.2~meV/atom for the MACE Ti-Al evaluation.
\begin{lstlisting}[language=turtle]
ex:metric-rmse-energy a mlips:AccuracyMetric ;
    mlips:metricType mlips:RMSE ;
    mlips:metricProperty mlips:Energy ;
    mlips:metricValue "1.2"^^xsd:float ;
    qudt:hasUnit mlips:MilliEV-PER-ATOM .
\end{lstlisting}
}

\paragraph{Related axioms.} \eqref{ax:24}, \eqref{ax:25}, \eqref{ax:27}.

\subsection{\dlConcept{ArchitecturalHyperparameter}}
\label{sec:term-ArchitecturalHyperparameter}

\textbf{Label:} Architectural Hyperparameter.
A hyperparameter that specifies the model
architecture. Not disjoint with PhysicalHyperparameter.

\paragraph{Superclasses.} $\dlConcept{Hyperparameter}$.

\paragraph{Example.}
The number of message-passing layers in MACE is a purely
architectural hyperparameter.
\begin{lstlisting}[language=turtle]
ex:hp-num-layers a mlips:Hyperparameter,
        mlips:ArchitecturalHyperparameter ;
    mlips:hyperparameterName "num_layers" ;
    mlips:hyperparameterDatatype xsd:integer ;
    mlips:defaultValue "2"^^xsd:integer .
\end{lstlisting}
}

\subsection{\dlConcept{AtomicConfiguration}}
\label{sec:term-AtomicConfiguration}

\textbf{Label:} Atomic Configuration.
A single atomic configuration (positions, cell,
properties) in a training dataset.

\paragraph{Superclasses.} $\dlConcept{AtomicStructure}$.

\paragraph{Incoming properties.} $\dlRole{hasConfiguration}$.

\paragraph{Example.}
A single atomic configuration belonging to the TiCr$_2$-H training dataset.
\begin{lstlisting}[language=turtle]
ex:config-0042 a mlips:AtomicConfiguration ;
    rdfs:label "TiCr2-H config #42" .

ex:ds-TiCr2H mlips:hasConfiguration ex:config-0042 .
\end{lstlisting}
}

\subsection{\dlConcept{AtomicEnvironmentDescriptor}}
\label{sec:term-AtomicEnvironmentDescriptor}

\textbf{Label:} Atomic Environment Descriptor.
A mathematical representation of the local atomic environment used by an MLIP method.

\paragraph{Incoming properties.} $\dlRole{hasDescriptor}$.

\paragraph{Example.}
The equivariant message-passing descriptor used by MACE.
\begin{lstlisting}[language=turtle]
mlips:EquivariantMessagePassingDescriptor a mlips:AtomicEnvironmentDescriptor ;
    rdfs:label "Equivariant message-passing descriptor"@en ;
    rdfs:comment "Descriptor family used by MACE, NequIP, Allegro."@en .
\end{lstlisting}
}

\subsection{\dlConcept{BenchmarkResult}}
\label{sec:term-BenchmarkResult}

\textbf{Label:} Benchmark Result.
A single evaluation result: an MLIP method applied to a
material system with specific hyperparameters, reporting accuracy metrics.

\paragraph{Superclasses.} $\dlConcept{Entity}$.

\paragraph{Outgoing properties.} $\dlRole{evaluatesModel}$, $\dlRole{hasAccuracyMetric}$, $\dlRole{isResultOf}$, $\dlRole{reportedIn}$, $\dlRole{targetMaterial}$, $\dlRole{usesAlgorithm}$, $\dlRole{usesTrainingData}$, $\dlRole{inferenceHardware}$, $\dlRole{inferenceTimePerAtom}$.

\paragraph{Incoming properties.} $\dlRole{hasResult}$, $\dlRole{isEvaluatedIn}$, $\dlRole{isMetricOf}$.

\paragraph{Example.}
A benchmark result evaluating the C15-MTP model on the TiCr$_2$ material system.
\begin{lstlisting}[language=turtle]
ex:result-mtp-ticr2h-c15 a mlips:BenchmarkResult ;
    mlips:evaluatesModel ex:model-mtp-TiCr2H ;
    mlips:targetMaterial ex:TiCr2 ;
    mlips:hasAccuracyMetric ex:metric-rmse-energy ;
    mlips:reportedIn ex:kumar2025 .
\end{lstlisting}
}

\paragraph{Related axioms.} \eqref{ax:21}, \eqref{ax:22}, \eqref{ax:23}, \eqref{ax:26}.

\subsection{\dlConcept{BenchmarkStudy}}
\label{sec:term-BenchmarkStudy}

\textbf{Label:} Benchmark Study.
A published study reporting MLIP evaluation results.

\paragraph{Superclasses.} $\dlConcept{Activity}$.

\paragraph{Outgoing properties.} $\dlRole{hasResult}$.

\paragraph{Incoming properties.} $\dlRole{isResultOf}$.

\paragraph{Example.}
A benchmark study by Kumar et al.\ (2025)~\cite{kumar2025ticr2h} containing
evaluation results for MTP on TiCr$_2$-H Laves phases.
\begin{lstlisting}[language=turtle]
ex:kumar2025 a mlips:BenchmarkStudy ;
    rdfs:label "Kumar et al. 2025" ;
    mlips:hasResult ex:result-mtp-ticr2h-c15 .
\end{lstlisting}
}

\paragraph{Related axioms.} \eqref{ax:20}.

\subsection{\dlConcept{CoveredProperty}}
\label{sec:term-CoveredProperty}

\textbf{Label:} Covered Property.
A physical property covered by the training dataset
(energy, forces, stresses, virials).

\paragraph{Incoming properties.} $\dlRole{coversProperty}$.

\paragraph{Example.}
A training dataset declaring that it covers energy, forces, and stresses using the predefined individuals.
\begin{lstlisting}[language=turtle]
ex:ds-TiCr2H mlips:coversProperty
    mlips:Energy ,
    mlips:Forces ,
    mlips:Stresses .
\end{lstlisting}
}

\subsection{\dlConcept{DFTCalculation}}
\label{sec:term-DFTCalculation}

\textbf{Label:} DFT Calculation.
A density functional theory calculation that produced
reference data for training.

\paragraph{Superclasses.} $\dlConcept{Calculation}$, $\dlConcept{ReferenceCalculation}$.

\paragraph{Outgoing properties.} $\dlRole{hasDFTSettings}$.

\paragraph{Incoming properties.} $\dlRole{hasDFTCalculation}$.

\paragraph{Example.}
A PBE density functional theory calculation that produced reference data for the Ti-Al dataset.
\begin{lstlisting}[language=turtle]
ex:dft-tial a mlips:DFTCalculation ;
    rdfs:label "Ti-Al PBE calculation" ;
    mlips:hasDFTSettings ex:dft-settings-pbe .
\end{lstlisting}
}

\paragraph{Related axioms.} \eqref{ax:19}.

\subsection{\dlConcept{DFTSettings}}
\label{sec:term-DFTSettings}

\textbf{Label:} DFT Settings.
Settings of a DFT calculation: exchange-correlation
functional, k-point mesh, energy cutoff, pseudopotential type.

\paragraph{Superclasses.} $\dlConcept{ReferenceSettings}$.

\paragraph{Outgoing properties.} $\dlRole{dftBasisSet}$, $\dlRole{pseudopotentialType}$, $\dlRole{usedDFTCode}$, $\dlRole{xcFunctional}$, $\dlRole{energyCutoff}$, $\dlRole{kPointMesh}$.

\paragraph{Incoming properties.} $\dlRole{hasDFTSettings}$.

\paragraph{Example.}
DFT settings specifying the PBE functional, PAW pseudopotentials, and VASP as the calculation code.
\begin{lstlisting}[language=turtle]
ex:dft-settings-pbe a mlips:DFTSettings ;
    mlips:xcFunctional "PBE" ;
    mlips:pseudopotentialType "PAW" ;
    mlips:energyCutoff 520.0 ;
    mlips:usedDFTCode ex:vasp .
\end{lstlisting}
}

\subsection{\dlConcept{DatasetProvenance}}
\label{sec:term-DatasetProvenance}

\textbf{Label:} Dataset Provenance.
The provenance type of a training dataset.

\paragraph{Incoming properties.} $\dlRole{datasetProvenance}$.

\paragraph{Example.}
A training dataset using the \texttt{Published} provenance individual to indicate publicly available data.
\begin{lstlisting}[language=turtle]
ex:ds-TiCr2H mlips:datasetProvenance
    mlips:Published .
\end{lstlisting}
}

\subsection{\dlConcept{DftBasisSet}}
\label{sec:term-DftBasisSet}

\textbf{Label:} DFT Basis Set.
The basis set used in a DFT calculation,
when the code is all-electron or otherwise uses a localised basis (e.g.,
FHI-aims NAO families, LAPW, Gaussian-type orbitals such as cc-pVDZ
or 6-31G). Range of mlips:dftBasisSet on DFTSettings. Plane-wave
codes specify the basis through mlips:energyCutoff instead and leave
this slot empty. Concrete basis sets are modelled as named individuals;
an open extension point.

\paragraph{Incoming properties.} $\dlRole{dftBasisSet}$.

\paragraph{Example.}
A paper-local candidate-vocabulary instance for a Gaussian basis set
that is not (yet) in the controlled vocabulary, plus the
\texttt{candidateForVocabulary} flag that signals it for promotion review.
\begin{lstlisting}[language=turtle]
ex:basis-6-31g-d-smith2017 a mlips:DftBasisSet ;
    rdfs:label "6-31G(d)" ;
    rdfs:comment "Pople-style split-valence double-zeta Gaussian basis with d polarisation." ;
    mlips:candidateForVocabulary mlips:DftBasisSet .
\end{lstlisting}
}

\subsection{\dlConcept{FunctionalForm}}
\label{sec:term-FunctionalForm}

\textbf{Label:} Functional Form.
A parametrised mapping from an atomic configuration to an energy, specified independently of parameter values.

\paragraph{Incoming properties.} $\dlRole{hasFunctionalForm}$.

\paragraph{Example.}
The MTP functional form: a polynomial in moment-tensor contractions evaluated over the local neighborhood.
\begin{lstlisting}[language=turtle]
ex:mtp-functional-form a mlips:FunctionalForm ;
    rdfs:label "MTP moment-tensor polynomial" .
\end{lstlisting}
}

\subsection{\dlConcept{Hyperparameter}}
\label{sec:term-Hyperparameter}

\textbf{Label:} Hyperparameter.
A hyperparameter of an MLIP method, with its name,
data type, range, and default value.

\paragraph{Superclasses.} $\dlConcept{HyperParameter}$.

\paragraph{Outgoing properties.} $\dlRole{isHyperparameterOf}$, $\dlRole{defaultValue}$, $\dlRole{hyperparameterDatatype}$, $\dlRole{hyperparameterName}$, $\dlRole{maxValue}$, $\dlRole{minValue}$.

\paragraph{Incoming properties.} $\dlRole{forHyperparameter}$, $\dlRole{hasHyperparameter}$.

\paragraph{Example.}
The cutoff radius hyperparameter with its name, data type, range, and default value.
\begin{lstlisting}[language=turtle]
ex:hp-cutoff-radius a mlips:Hyperparameter ;
    mlips:hyperparameterName "cutoff_radius" ;
    mlips:hyperparameterDatatype xsd:float ;
    mlips:minValue "3.0"^^xsd:float ;
    mlips:maxValue "8.0"^^xsd:float ;
    mlips:defaultValue "5.0"^^xsd:float .
\end{lstlisting}
}

\paragraph{Related axioms.} \eqref{ax:8}.

\subsection{\dlConcept{HyperparameterSetting}}
\label{sec:term-HyperparameterSetting}

\textbf{Label:} Hyperparameter Setting.
A concrete value assigned to a hyperparameter in a
specific training run or configuration.

\paragraph{Superclasses.} $\dlConcept{HyperParameterSetting}$.

\paragraph{Outgoing properties.} $\dlRole{forHyperparameter}$, $\dlRole{isSettingOf}$, $\dlRole{cutoffRadius}$, $\dlRole{learningRate}$, $\dlRole{numAngularBasis}$, $\dlRole{numLayers}$, $\dlRole{numRadialBasis}$, $\dlRole{settingValue}$.

\paragraph{Incoming properties.} $\dlRole{hasHyperparameterSetting}$, $\dlRole{trainedUsing}$.

\paragraph{Example.}
A concrete setting of the cutoff radius to 5.0~\AA{} for a specific training configuration.
\begin{lstlisting}[language=turtle]
ex:setting-cutoff a mlips:HyperparameterSetting ;
    mlips:forHyperparameter ex:hp-cutoff-radius ;
    mlips:settingValue "5.0"^^xsd:float ;
    mlips:cutoffRadius 5.0 ;
    qudt:hasUnit unit:ANGSTROM .
\end{lstlisting}
}

\subsection{\dlConcept{Implementation}}
\label{sec:term-Implementation}

\textbf{Label:} Implementation.
A software implementation of an MLIP method in a
specific library and version.

\paragraph{Superclasses.} $\dlConcept{SoftwareSourceCode}$.

\paragraph{Outgoing properties.} $\dlRole{implementedIn}$, $\dlRole{isImplementationOf}$, $\dlRole{version}$.

\paragraph{Incoming properties.} $\dlRole{hasImplementation}$.

\paragraph{Example.}
Version 0.3 of the MACE algorithm implemented in the MACE library.
\begin{lstlisting}[language=turtle]
ex:mace-v03 a mlips:Implementation ;
    mlips:implementedIn ex:mace-lib ;
    mlips:version "0.3" .
\end{lstlisting}
}

\paragraph{Related axioms.} \eqref{ax:3}, \eqref{ax:9}.

\subsection{\dlConcept{Library}}
\label{sec:term-Library}

\textbf{Label:} Library.
A software library or tool used in MLIP research
(e.g., ASE, LAMMPS, VASP, GPAW).

\paragraph{Superclasses.} $\dlConcept{SoftwareApplication}$.

\paragraph{Incoming properties.} $\dlRole{implementedIn}$, $\dlRole{usedDFTCode}$, $\dlRole{usedReferenceCode}$.

\paragraph{Example.}
The MACE software library used for training equivariant interatomic potentials.
\begin{lstlisting}[language=turtle]
ex:mace-lib a mlips:Library ;
    rdfs:label "MACE" ;
    schema:url <https://github.com/ACEsuit/mace> .
\end{lstlisting}
}

\subsection{\dlConcept{LossFunction}}
\label{sec:term-LossFunction}

\textbf{Label:} Loss Function.
A scalar objective whose minimisation fits the parameters of the functional form.

\paragraph{Incoming properties.} $\dlRole{hasLossFunction}$.

\paragraph{Example.}
A weighted sum of mean-squared errors on energies, forces, and stresses.
\begin{lstlisting}[language=turtle]
ex:weighted-energy-force-stress-loss a mlips:LossFunction ;
    rdfs:label "Weighted MSE on energies, forces, stresses" .
\end{lstlisting}
}

\subsection{\dlConcept{MLIPMethod}}
\label{sec:term-MLIPMethod}

\textbf{Label:} MLIP Method.
A machine learning interatomic potential method family
(e.g., MACE, ACE, HDNNP, NequIP, M3GNet, MTP). Specifies a functional form,
a loss function, and a training algorithm. Disjoint from mls:Algorithm.

{\raggedright
\paragraph{Outgoing properties.} $\dlRole{hasDescriptor}$, $\dlRole{hasFunctionalForm}$, $\dlRole{hasHyperparameter}$, $\dlRole{hasImplementation}$, $\dlRole{hasLossFunction}$, $\dlRole{hasTrainingAlgorithm}$, $\dlRole{supportsSimulation}$, $\dlRole{inferenceComplexity}$, $\dlRole{supportsGPU}$, $\dlRole{supportsParallelization}$, $\dlRole{trainingComplexity}$.
\par}

\paragraph{Incoming properties.} $\dlRole{appliesMethod}$, $\dlRole{isHyperparameterOf}$, $\dlRole{isImplementationOf}$, $\dlRole{trainedWith}$, $\dlRole{usesAlgorithm}$.

\paragraph{Example.}
The MACE equivariant message-passing method with its components, hyperparameters, and supported simulations.
\begin{lstlisting}[language=turtle]
ex:MACE a mlips:MLIPMethod ;
    rdfs:label "MACE" ;
    mlips:hasFunctionalForm ex:mace-neural-net ;
    mlips:hasLossFunction ex:weighted-energy-force-loss ;
    mlips:hasTrainingAlgorithm ex:adam ;
    mlips:hasHyperparameter ex:hp-cutoff-radius ;
    mlips:hasImplementation ex:mace-v03 ;
    mlips:supportsSimulation ex:molecular-dynamics .

ex:adam a mls:Algorithm ;
    rdfs:label "Adam" .
\end{lstlisting}
}

\subsection{\dlConcept{MLIPRun}}
\label{sec:term-MLIPRun}

\textbf{Label:} MLIP Run.
A concrete fitting activity for an MLIP method: it applies an MLIP method to a training dataset and produces a trained model.

\paragraph{Superclasses.} $\dlConcept{Activity}$.

\paragraph{Outgoing properties.} $\dlRole{appliesMethod}$, $\dlRole{hasHyperparameterSetting}$, $\dlRole{hasTrainingRun}$, $\dlRole{produces}$, $\dlRole{runsOn}$, $\dlRole{gpuHours}$, $\dlRole{peakMemory}$, $\dlRole{trainingDuration}$, $\dlRole{trainingHardware}$.

\paragraph{Incoming properties.} $\dlRole{isProducedBy}$, $\dlRole{isSettingOf}$, $\dlRole{wasRunBy}$.

\paragraph{Example.}
An MLIP run that applies the MTP method on a TiCr$_2$-H dataset and produces a trained model.
\begin{lstlisting}[language=turtle]
ex:run-mtp-TiCr2H a mlips:MLIPRun ;
    mlips:appliesMethod ex:MTP ;
    mlips:runsOn ex:ds-TiCr2H ;
    mlips:produces ex:model-mtp-TiCr2H .
\end{lstlisting}
}

\subsection{\dlConcept{MaterialSystem}}
\label{sec:term-MaterialSystem}

\textbf{Label:} Material System.
A material system studied in a dataset: element, alloy,
compound, or complex microstructure.

\paragraph{Superclasses.} $\dlConcept{CrystallineMaterial}$, $\dlConcept{Material}$.

\paragraph{Outgoing properties.} $\dlRole{chemicalFormula}$, $\dlRole{materialClass}$, $\dlRole{microstructuralFeature}$.

\paragraph{Incoming properties.} $\dlRole{coversMaterial}$, $\dlRole{targetMaterial}$.

\paragraph{Example.}
A Ti-Al binary alloy system with its chemical formula and material class.
\begin{lstlisting}[language=turtle]
ex:TiAl a mlips:MaterialSystem ;
    mlips:chemicalFormula "TiAl" ;
    mlips:materialClass "binary alloy" ;
    mlips:sameAsWikidata wd:Q2549811 .
\end{lstlisting}
}

\subsection{\dlConcept{MetaSort}}
\label{sec:term-MetaSort}

\textbf{Label:} Meta-Sort.
An OntoClean-style meta-category used to
classify classes of the ontology.

\paragraph{Example.}
\InputIfFileExists{artifacts/examples/classes/MetaSort.tex}{}{\textit{TODO: add example.}}

\subsection{\dlConcept{MetricProperty}}
\label{sec:term-MetricProperty}

\textbf{Label:} Metric Property.
The physical property being measured (energy, force, stress).

\paragraph{Incoming properties.} $\dlRole{metricProperty}$.

\paragraph{Example.}
An accuracy metric using the predefined Energy metric property individual.
\begin{lstlisting}[language=turtle]
ex:metric-rmse-energy
    mlips:metricProperty mlips:Energy .
\end{lstlisting}
}

\subsection{\dlConcept{MetricType}}
\label{sec:term-MetricType}

\textbf{Label:} Metric Type.
Type of accuracy metric (RMSE, MAE, R², etc.).

\paragraph{Incoming properties.} $\dlRole{metricType}$.

\paragraph{Example.}
An accuracy metric using the predefined RMSE metric type individual.
\begin{lstlisting}[language=turtle]
ex:metric-rmse-energy
    mlips:metricType mlips:RMSE .
\end{lstlisting}
}

\subsection{\dlConcept{PhysicalHyperparameter}}
\label{sec:term-PhysicalHyperparameter}

\textbf{Label:} Physical Hyperparameter.
A hyperparameter that has a direct physical
meaning. Not disjoint with ArchitecturalHyperparameter.

\paragraph{Superclasses.} $\dlConcept{Hyperparameter}$.

\paragraph{Example.}
The cutoff radius is a physical hyperparameter: its value bounds the
physical interaction range of the potential.
\begin{lstlisting}[language=turtle]
ex:hp-cutoff-radius a mlips:Hyperparameter,
        mlips:PhysicalHyperparameter ;
    mlips:hyperparameterName "R_cut" ;
    mlips:hyperparameterDatatype xsd:double ;
    mlips:defaultValue "5.0"^^xsd:double .
\end{lstlisting}
}

\subsection{\dlConcept{PseudopotentialType}}
\label{sec:term-PseudopotentialType}

\textbf{Label:} Pseudopotential Type.
A family of pseudopotentials used in a
plane-wave DFT calculation (e.g., PAW, ultrasoft, norm-conserving).
Range of mlips:pseudopotentialType on DFTSettings. All-electron codes
leave this slot empty. Concrete pseudopotential families are modelled
as named individuals; an open extension point.

\paragraph{Incoming properties.} $\dlRole{pseudopotentialType}$.

\paragraph{Example.}
\texttt{mlips:PAW} from the controlled vocabulary.
\begin{lstlisting}[language=turtle]
mlips:PAW a mlips:PseudopotentialType ;
    rdfs:label "PAW" ;
    rdfs:comment "Projector augmented-wave method." .
\end{lstlisting}
}

\subsection{\dlConcept{ReferenceCalculation}}
\label{sec:term-ReferenceCalculation}

\textbf{Label:} Reference Calculation.
A first-principles calculation that produced
reference energies, forces, or stresses for MLIP training. Generalisation of
DFTCalculation that also covers wave-function methods (e.g., CCSD(T), MP2,
CASPT2) and other ab initio approaches.

\paragraph{Superclasses.} $\dlConcept{Activity}$, $\dlConcept{Calculation}$.

\paragraph{Outgoing properties.} $\dlRole{hasReferenceSettings}$.

\paragraph{Incoming properties.} $\dlRole{hasReferenceCalculation}$.

\paragraph{Example.}
A reference calculation can be either a DFT calculation or a wave-function calculation; the supertype is what is required by the dataset.
\begin{lstlisting}[language=turtle]
ex:ds-water mlips:hasReferenceCalculation ex:ccsdt-water .
ex:ccsdt-water a mlips:WaveFunctionCalculation .  # subclass of ReferenceCalculation
\end{lstlisting}
}

\subsection{\dlConcept{ReferenceSettings}}
\label{sec:term-ReferenceSettings}

\textbf{Label:} Reference Settings.
Settings of a reference calculation, abstracted
across method families. Concrete subclasses (DFTSettings, WaveFunctionSettings)
carry the method-specific parameters; the supertype carries cross-family slots
such as the software code used.

\paragraph{Outgoing properties.} $\dlRole{usedReferenceCode}$.

\paragraph{Incoming properties.} $\dlRole{hasReferenceSettings}$.

\paragraph{Example.}
Reference settings are abstracted across method families; concrete subclasses carry the specific parameters.
\begin{lstlisting}[language=turtle]
ex:ccsdt-water mlips:hasReferenceSettings ex:ccsdt-settings .
ex:ccsdt-settings a mlips:WaveFunctionSettings ;  # subclass of ReferenceSettings
    mlips:usedReferenceCode ex:lib-pyscf .
\end{lstlisting}
}

\subsection{\dlConcept{SamplingStrategy}}
\label{sec:term-SamplingStrategy}

\textbf{Label:} Sampling Strategy.
A strategy used to sample the atomic
configurations in a training dataset (e.g., chemical sampling across
compositions, vibrational sampling across thermal snapshots, or a
combination). Concrete sampling strategies are modeled as instances and
will be developed in future work; this class is left open as an
extension point.

\paragraph{Incoming properties.} $\dlRole{samplingStrategy}$.

\paragraph{Example.}
An ad-hoc \texttt{ChemicalSampling} individual declared inline.
\begin{lstlisting}[language=turtle]
ex:ChemicalSampling a mlips:SamplingStrategy ;
    rdfs:label "Chemical sampling" ;
    rdfs:comment "Configurations sampled across compositions." .
\end{lstlisting}
}

\subsection{\dlConcept{SimulationType}}
\label{sec:term-SimulationType}

\textbf{Label:} Simulation Type.
A type of atomistic simulation: molecular dynamics,
Monte Carlo, geometry optimization, phonon calculations,
thermodynamic integration, etc.

\paragraph{Incoming properties.} $\dlRole{supportsSimulation}$.

\paragraph{Example.}
Molecular dynamics as a simulation type supported by an MLIP algorithm.
\begin{lstlisting}[language=turtle]
ex:molecular-dynamics a mlips:SimulationType ;
    rdfs:label "Molecular Dynamics" ;
    mlips:sameAsWikidata wd:Q901663 .
\end{lstlisting}
}

\subsection{\dlConcept{TrainedModel}}
\label{sec:term-TrainedModel}

\textbf{Label:} Trained Model.
The concrete artifact produced by fitting an MLIP method
on a specific dataset with specific hyperparameter settings---the learned
parameters (weights) that can be used for prediction.

\paragraph{Superclasses.} $\dlConcept{Model}$, $\dlConcept{Entity}$.

\paragraph{Outgoing properties.} $\dlRole{isEvaluatedIn}$, $\dlRole{isProducedBy}$, $\dlRole{trainedOn}$, $\dlRole{trainedUsing}$, $\dlRole{trainedWith}$.

\paragraph{Incoming properties.} $\dlRole{evaluatesModel}$, $\dlRole{produces}$.

\paragraph{Example.}
A trained C15-MTP for TiCr$_2$-H, linking back to method and training data.
\begin{lstlisting}[language=turtle]
ex:model-mtp-TiCr2H a mlips:TrainedModel ;
    rdfs:label "C15-MTP for TiCr2-H" ;
    mlips:trainedWith ex:MTP ;
    mlips:trainedOn ex:ds-TiCr2H .
\end{lstlisting}
}

\paragraph{Related axioms.} \eqref{ax:7}.

\subsection{\dlConcept{TrainingDataset}}
\label{sec:term-TrainingDataset}

\textbf{Label:} Training Dataset.
A dataset of atomic configurations used for MLIP
training, with provenance, size, property coverage, and DFT settings.

\paragraph{Superclasses.} $\dlConcept{Dataset}$, $\dlConcept{Entity}$.

\paragraph{Outgoing properties.} $\dlRole{coversMaterial}$, $\dlRole{coversProperty}$, $\dlRole{datasetProvenance}$, $\dlRole{hasConfiguration}$, $\dlRole{hasDFTCalculation}$, $\dlRole{hasReferenceCalculation}$, $\dlRole{samplingStrategy}$, $\dlRole{wasRunBy}$, $\dlRole{numConfigurations}$.

\paragraph{Incoming properties.} $\dlRole{runsOn}$, $\dlRole{trainedOn}$, $\dlRole{usesTrainingData}$.

\paragraph{Example.}
A published Ti-Al DFT dataset with 15\,000 configurations covering energy, forces, and stresses.
\begin{lstlisting}[language=turtle]
ex:tial-dataset a mlips:TrainingDataset ;
    mlips:numConfigurations 15000 ;
    mlips:coversMaterial ex:TiAl ;
    mlips:coversProperty mlips:Energy ;
    mlips:datasetProvenance mlips:Published .
\end{lstlisting}
}

\paragraph{Related axioms.} \eqref{ax:15}, \eqref{ax:16}, \eqref{ax:17}, \eqref{ax:18}.

\subsection{\dlConcept{TrainingHyperparameter}}
\label{sec:term-TrainingHyperparameter}

\textbf{Label:} Training Hyperparameter.
A hyperparameter of the training procedure.

\paragraph{Superclasses.} $\dlConcept{Hyperparameter}$.

\paragraph{Example.}
The optimiser learning rate is a training hyperparameter — it
governs how the model is fitted, not what the model is.
\begin{lstlisting}[language=turtle]
ex:hp-learning-rate a mlips:Hyperparameter,
        mlips:TrainingHyperparameter ;
    mlips:hyperparameterName "learning_rate" ;
    mlips:hyperparameterDatatype xsd:double ;
    mlips:defaultValue "0.001"^^xsd:double .
\end{lstlisting}
}

\subsection{\dlConcept{WaveFunctionCalculation}}
\label{sec:term-WaveFunctionCalculation}

\textbf{Label:} Wave Function Calculation.
A wave-function-based reference calculation
(e.g., Hartree-Fock, MP2, CCSD, CCSD(T), CASPT2). An alternative to
DFTCalculation as a source of reference data, typically more accurate but
more expensive.

\paragraph{Superclasses.} $\dlConcept{ReferenceCalculation}$.

\paragraph{Outgoing properties.} $\dlRole{hasWaveFunctionSettings}$.

\paragraph{Example.}
A CCSD(T) calculation used to generate reference data for an MLIP study.
\begin{lstlisting}[language=turtle]
ex:ccsdt-water a mlips:WaveFunctionCalculation ;
    mlips:hasWaveFunctionSettings ex:ccsdt-settings .
\end{lstlisting}
}

\subsection{\dlConcept{WaveFunctionSettings}}
\label{sec:term-WaveFunctionSettings}

\textbf{Label:} Wave Function Settings.
Settings of a wave-function-based reference
calculation: the method (e.g., CCSD(T), MP2), the basis set (e.g.,
cc-pVTZ, aug-cc-pVDZ), and frozen-core treatment.

\paragraph{Superclasses.} $\dlConcept{ReferenceSettings}$.

\paragraph{Outgoing properties.} $\dlRole{wfMethod}$, $\dlRole{basisSet}$, $\dlRole{frozenCore}$.

\paragraph{Incoming properties.} $\dlRole{hasWaveFunctionSettings}$.

\paragraph{Example.}
CCSD(T) with the cc-pVTZ basis, frozen-core approximation, run in PySCF.
\begin{lstlisting}[language=turtle]
ex:ccsdt-settings a mlips:WaveFunctionSettings ;
    mlips:wfMethod "CCSD(T)" ;
    mlips:basisSet "cc-pVTZ" ;
    mlips:frozenCore true ;
    mlips:usedReferenceCode ex:lib-pyscf .
\end{lstlisting}
}

\subsection{\dlConcept{WfMethod}}
\label{sec:term-WfMethod}

\textbf{Label:} Wave Function Method.
A wave-function correlation method used to
produce reference data (e.g., HF, MP2, CCSD, CCSD(T), DLPNO-CCSD(T),
CASPT2, NEVPT2). Range of mlips:wfMethod on WaveFunctionSettings.
Concrete methods are modelled as named individuals; an open
extension point.

\paragraph{Incoming properties.} $\dlRole{wfMethod}$.

\paragraph{Example.}
\texttt{mlips:DLPNO\_CCSDT} from the controlled vocabulary, used by smith2019ccx.
\begin{lstlisting}[language=turtle]
mlips:DLPNO_CCSDT a mlips:WfMethod ;
    rdfs:label "DLPNO-CCSD(T)" ;
    rdfs:comment "Domain-based local pair natural orbital approximation to CCSD(T)." .
\end{lstlisting}
}

\subsection{\dlConcept{XCFunctional}}
\label{sec:term-XCFunctional}

\textbf{Label:} Exchange-Correlation Functional.
A density-functional approximation to the
exchange-correlation energy used in a DFT calculation (e.g., LDA, PBE,
PBE0, HSE06, SCAN, BLYP, omegaB97X). Range of mlips:xcFunctional on
DFTSettings. Concrete functionals are modelled as named individuals;
an open extension point.

\paragraph{Incoming properties.} $\dlRole{xcFunctional}$.

\paragraph{Example.}
\texttt{mlips:PBE} from the controlled vocabulary, declared in
\path{artifacts/kg/mlips-vocab.ttl}.
\begin{lstlisting}[language=turtle]
mlips:PBE a mlips:XCFunctional ;
    rdfs:label "PBE" ;
    rdfs:comment "Perdew-Burke-Ernzerhof; GGA family." .
\end{lstlisting}
}

\section{Object Property Reference}
\label{sec:object-property-reference}

This appendix describes all 53 object properties defined in the \MLIPsOntology{}.

\subsection{\dlRole{appliesMethod}}
\label{sec:term-appliesMethod}

\textbf{Label:} applies method.
Links an MLIP run to the MLIP method it fits.

\textbf{Domain:} $\dlConcept{MLIPRun}$.
\textbf{Range:} $\dlConcept{MLIPMethod}$.

\paragraph{Example.}
An MLIP run applies the MTP method.
\begin{lstlisting}[language=turtle]
ex:run-mtp-TiCr2H mlips:appliesMethod ex:MTP .
\end{lstlisting}
}

\subsection{\dlRole{candidateForVocabulary}}
\label{sec:term-candidateForVocabulary}

\textbf{Label:} candidate for vocabulary.
Annotates a paper-local instance with the
controlled-vocabulary class to which it should be considered for
promotion. Used when an encoded paper introduces an entity that is not
yet a named individual in mlips-vocab.ttl. The deduplication audit
collects these candidates; a curator promotes, merges, or leaves them
paper-local.

\paragraph{Example.}
A paper-local instance flagged for vocabulary review.
\begin{lstlisting}[language=turtle]
ex:basis-6-31g-d-smith2017 mlips:candidateForVocabulary mlips:DftBasisSet .
\end{lstlisting}
}

\subsection{\dlRole{coversMaterial}}
\label{sec:term-coversMaterial}

\textbf{Label:} covers material.

\textbf{Domain:} $\dlConcept{TrainingDataset}$.
\textbf{Range:} $\dlConcept{MaterialSystem}$.

\paragraph{Example.}
A training dataset covers the TiCr$_2$-H material system.
\begin{lstlisting}[language=turtle]
ex:ds-TiCr2H mlips:coversMaterial ex:TiCr2 .
\end{lstlisting}
}

\paragraph{Related axioms.} \eqref{ax:15}.

\subsection{\dlRole{coversProperty}}
\label{sec:term-coversProperty}

\textbf{Label:} covers property.

\textbf{Domain:} $\dlConcept{TrainingDataset}$.
\textbf{Range:} $\dlConcept{CoveredProperty}$.

\paragraph{Example.}
A training dataset covers total energy as a reference property.
\begin{lstlisting}[language=turtle]
ex:ds-TiAl mlips:coversProperty mlips:Energy .
\end{lstlisting}
}

\paragraph{Related axioms.} \eqref{ax:16}.

\subsection{\dlRole{datasetProvenance}}
\label{sec:term-datasetProvenance}

\textbf{Label:} dataset provenance.

\textbf{Domain:} $\dlConcept{TrainingDataset}$.
\textbf{Range:} $\dlConcept{DatasetProvenance}$.

\paragraph{Example.}
A training dataset records its provenance as published data.
\begin{lstlisting}[language=turtle]
ex:ds-TiCr2H mlips:datasetProvenance mlips:Published .
\end{lstlisting}
}

\paragraph{Related axioms.} \eqref{ax:17}.

\subsection{\dlRole{dftBasisSet}}
\label{sec:term-dftBasisSet}

\textbf{Label:} DFT basis set.
The basis set used by an all-electron or
localised-basis DFT calculation (FHI-aims NAO, LAPW, Gaussian-type
orbitals such as cc-pVDZ or 6-31G). Plane-wave codes specify the basis
through mlips:energyCutoff and leave this slot empty.

\textbf{Domain:} $\dlConcept{DFTSettings}$.
\textbf{Range:} $\dlConcept{DftBasisSet}$.

\paragraph{Example.}
An all-electron DFT settings node references its NAO basis set.
\begin{lstlisting}[language=turtle]
ex:dft-settings-fhi-aims mlips:dftBasisSet mlips:NAOIntermediate .
\end{lstlisting}
}

\subsection{\dlRole{evaluatesModel}}
\label{sec:term-evaluatesModel}

\textbf{Label:} evaluates model.
Links a benchmark result to the trained model being evaluated.

\textbf{Domain:} $\dlConcept{BenchmarkResult}$.
\textbf{Range:} $\dlConcept{TrainedModel}$.

\paragraph{Example.}
A benchmark result evaluates the trained MTP model for TiCr$_2$-H.
\begin{lstlisting}[language=turtle]
ex:result-01 mlips:evaluatesModel ex:model-mtp-TiCr2H .
\end{lstlisting}
}

\paragraph{Related axioms.} \eqref{ax:21}.

\subsection{\dlRole{forHyperparameter}}
\label{sec:term-forHyperparameter}

\textbf{Label:} for hyperparameter.
Links a setting to the hyperparameter it configures.

\textbf{Domain:} $\dlConcept{HyperparameterSetting}$.
\textbf{Range:} $\dlConcept{Hyperparameter}$.

\paragraph{Example.}
A hyperparameter setting references the cutoff radius hyperparameter it configures.
\begin{lstlisting}[language=turtle]
ex:setting-cutoff-5A mlips:forHyperparameter
    ex:cutoff-radius .
\end{lstlisting}
}

\subsection{\dlRole{hasAccuracyMetric}}
\label{sec:term-hasAccuracyMetric}

\textbf{Label:} has accuracy metric.

\textbf{Domain:} $\dlConcept{BenchmarkResult}$.
\textbf{Range:} $\dlConcept{AccuracyMetric}$.

\paragraph{Example.}
A benchmark result includes an RMSE accuracy metric for energy.
\begin{lstlisting}[language=turtle]
ex:result-01 mlips:hasAccuracyMetric ex:metric-RMSE-E .
\end{lstlisting}
}

\paragraph{Related axioms.} \eqref{ax:23}, \eqref{ax:27}.

\subsection{\dlRole{hasConfiguration}}
\label{sec:term-hasConfiguration}

\textbf{Label:} has configuration.

\textbf{Domain:} $\dlConcept{TrainingDataset}$.
\textbf{Range:} $\dlConcept{AtomicConfiguration}$.

\paragraph{Example.}
A training dataset contains an atomic configuration.
\begin{lstlisting}[language=turtle]
ex:ds-TiAl mlips:hasConfiguration ex:config-0001 .
\end{lstlisting}
}

\paragraph{Related axioms.} \eqref{ax:18}.

\subsection{\dlRole{hasDFTCalculation}}
\label{sec:term-hasDFTCalculation}

\textbf{Label:} has DFT calculation.

\textbf{Domain:} $\dlConcept{TrainingDataset}$.
\textbf{Range:} $\dlConcept{DFTCalculation}$.

\paragraph{Example.}
A training dataset includes a DFT calculation that produced its reference data.
\begin{lstlisting}[language=turtle]
ex:ds-TiAl mlips:hasDFTCalculation ex:dft-calc-TiAl .
\end{lstlisting}
}

\subsection{\dlRole{hasDFTSettings}}
\label{sec:term-hasDFTSettings}

\textbf{Label:} has DFT settings.

\textbf{Domain:} $\dlConcept{DFTCalculation}$.
\textbf{Range:} $\dlConcept{DFTSettings}$.

\paragraph{Example.}
A DFT calculation has associated computational settings.
\begin{lstlisting}[language=turtle]
ex:dft-calc-TiAl mlips:hasDFTSettings ex:dft-PBE-PAW .
\end{lstlisting}
}

\paragraph{Related axioms.} \eqref{ax:19}.

\subsection{\dlRole{hasDescriptor}}
\label{sec:term-hasDescriptor}

\textbf{Label:} has descriptor.
Links an MLIP method to the atomic environment descriptor it uses.

\textbf{Domain:} $\dlConcept{MLIPMethod}$.
\textbf{Range:} $\dlConcept{AtomicEnvironmentDescriptor}$.

\paragraph{Example.}
MACE uses an equivariant message-passing descriptor.
\begin{lstlisting}[language=turtle]
ex:MACE mlips:hasDescriptor mlips:EquivariantMessagePassingDescriptor .
\end{lstlisting}
}

\subsection{\dlRole{hasFunctionalForm}}
\label{sec:term-hasFunctionalForm}

\textbf{Label:} has functional form.
Links an MLIP method to its parametrised functional form.

\textbf{Domain:} $\dlConcept{MLIPMethod}$.
\textbf{Range:} $\dlConcept{FunctionalForm}$.

\paragraph{Example.}
The MTP method has a moment-tensor polynomial functional form.
\begin{lstlisting}[language=turtle]
ex:MTP mlips:hasFunctionalForm ex:mtp-functional-form .
\end{lstlisting}
}

\subsection{\dlRole{hasHyperparameter}}
\label{sec:term-hasHyperparameter}

\textbf{Label:} has hyperparameter.
Links an MLIP method to a hyperparameter it accepts.

\textbf{Domain:} $\dlConcept{MLIPMethod}$.
\textbf{Range:} $\dlConcept{Hyperparameter}$.

\paragraph{Example.}
The MACE algorithm accepts a cutoff radius hyperparameter.
\begin{lstlisting}[language=turtle]
ex:MACE mlips:hasHyperparameter ex:cutoff-radius .
\end{lstlisting}
}

\paragraph{Related axioms.} \eqref{ax:1}, \eqref{ax:8}, \eqref{ax:12}.

\subsection{\dlRole{hasHyperparameterSetting}}
\label{sec:term-hasHyperparameterSetting}

\textbf{Label:} has hyperparameter setting.
Links an MLIP run to a concrete hyperparameter setting used in that run.

\textbf{Domain:} $\dlConcept{MLIPRun}$.
\textbf{Range:} $\dlConcept{HyperparameterSetting}$.

\paragraph{Example.}
A benchmark result records the hyperparameter setting used during evaluation.
\begin{lstlisting}[language=turtle]
ex:result-01 mlips:hasHyperparameterSetting
    ex:setting-cutoff-5A .
\end{lstlisting}
}

\subsection{\dlRole{hasImplementation}}
\label{sec:term-hasImplementation}

\textbf{Label:} has implementation.
Links an MLIP method to a software implementation.

\textbf{Domain:} $\dlConcept{MLIPMethod}$.
\textbf{Range:} $\dlConcept{Implementation}$.

\paragraph{Example.}
The MACE algorithm has a version 0.3 implementation.
\begin{lstlisting}[language=turtle]
ex:MACE mlips:hasImplementation ex:MACE-v03 .
\end{lstlisting}
}

\paragraph{Related axioms.} \eqref{ax:9}, \eqref{ax:13}.

\subsection{\dlRole{hasLossFunction}}
\label{sec:term-hasLossFunction}

\textbf{Label:} has loss function.
Links an MLIP method to the loss function whose minimisation fits its parameters.

\textbf{Domain:} $\dlConcept{MLIPMethod}$.
\textbf{Range:} $\dlConcept{LossFunction}$.

\paragraph{Example.}
The MTP method is fitted by minimising a weighted energy/force/stress MSE.
\begin{lstlisting}[language=turtle]
ex:MTP mlips:hasLossFunction ex:weighted-energy-force-stress-loss .
\end{lstlisting}
}

\subsection{\dlRole{hasReferenceCalculation}}
\label{sec:term-hasReferenceCalculation}

\textbf{Label:} has reference calculation.
Links a training dataset to the
first-principles calculation that produced its reference data.
Generalises hasDFTCalculation to wave-function and other ab initio
methods.

\textbf{Domain:} $\dlConcept{TrainingDataset}$.
\textbf{Range:} $\dlConcept{ReferenceCalculation}$.

\paragraph{Example.}
A training dataset whose reference data was produced by a wave-function calculation. \texttt{hasDFTCalculation} is a subproperty, so a query on \texttt{hasReferenceCalculation} also matches DFT-based datasets.
\begin{lstlisting}[language=turtle]
ex:ds-water mlips:hasReferenceCalculation ex:ccsdt-water .
\end{lstlisting}
}

\subsection{\dlRole{hasReferenceSettings}}
\label{sec:term-hasReferenceSettings}

\textbf{Label:} has reference settings.
Links a reference calculation to its
method-specific settings. Generalises hasDFTSettings to wave-function
and other ab initio methods.

\textbf{Domain:} $\dlConcept{ReferenceCalculation}$.
\textbf{Range:} $\dlConcept{ReferenceSettings}$.

\paragraph{Example.}
Generic link from a reference calculation to its settings. \texttt{hasDFTSettings} is the DFT-specific subproperty.
\begin{lstlisting}[language=turtle]
ex:ccsdt-water mlips:hasReferenceSettings ex:ccsdt-settings .
\end{lstlisting}
}

\subsection{\dlRole{hasResult}}
\label{sec:term-hasResult}

\textbf{Label:} has result.

\textbf{Domain:} $\dlConcept{BenchmarkStudy}$.
\textbf{Range:} $\dlConcept{BenchmarkResult}$.

\paragraph{Example.}
A benchmark study contains an individual evaluation result.
\begin{lstlisting}[language=turtle]
ex:study-kumar2025 mlips:hasResult ex:result-01 .
\end{lstlisting}
}

\paragraph{Related axioms.} \eqref{ax:20}, \eqref{ax:26}.

\subsection{\dlRole{hasTrainingAlgorithm}}
\label{sec:term-hasTrainingAlgorithm}

\textbf{Label:} has training algorithm.
Links an MLIP method to the ML-Schema algorithm used to fit its functional form.

\textbf{Domain:} $\dlConcept{MLIPMethod}$.
\textbf{Range:} $\dlConcept{Algorithm}$.

\paragraph{Example.}
The MTP method uses L-BFGS as its training algorithm. The training algorithm is an \texttt{mls:Algorithm}, which is the sole alignment point between MLIPs ontology and ML-Schema.
\begin{lstlisting}[language=turtle]
ex:MTP mlips:hasTrainingAlgorithm ex:lbfgs .
ex:lbfgs a mls:Algorithm ;
    rdfs:label "L-BFGS" .
\end{lstlisting}
}

\subsection{\dlRole{hasTrainingRun}}
\label{sec:term-hasTrainingRun}

\textbf{Label:} has training run.
Optional link from an MLIP run to an ML-Schema run recording the training algorithm execution.

\textbf{Domain:} $\dlConcept{MLIPRun}$.
\textbf{Range:} $\dlConcept{Run}$.

\paragraph{Example.}
An MLIP run may optionally be linked to an underlying ML-Schema run that records the training algorithm execution.
\begin{lstlisting}[language=turtle]
ex:run-mtp-TiCr2H mlips:hasTrainingRun ex:mls-run-mtp-TiCr2H .
ex:mls-run-mtp-TiCr2H a mls:Run ;
    mls:executes ex:lbfgs .
\end{lstlisting}
}

\subsection{\dlRole{hasUnit}}
\label{sec:term-hasUnit}

\textbf{Label:} has unit.
Links a numeric value to its QUDT unit.

\textbf{Range:} $\dlConcept{Unit}$.

\paragraph{Example.}
An accuracy metric has meV/atom as its unit.
\begin{lstlisting}[language=turtle]
ex:metric-RMSE-E mlips:hasUnit mlips:MilliEV-PER-ATOM .
\end{lstlisting}
}

\subsection{\dlRole{hasWaveFunctionSettings}}
\label{sec:term-hasWaveFunctionSettings}

\textbf{Label:} has wave function settings.

\textbf{Domain:} $\dlConcept{WaveFunctionCalculation}$.
\textbf{Range:} $\dlConcept{WaveFunctionSettings}$.

\paragraph{Example.}
A wave-function calculation linked to its method-specific settings.
\begin{lstlisting}[language=turtle]
ex:ccsdt-water mlips:hasWaveFunctionSettings ex:ccsdt-settings .
\end{lstlisting}
}

\subsection{\dlRole{implementedIn}}
\label{sec:term-implementedIn}

\textbf{Label:} implemented in.
Links an implementation to the library that provides it.

\textbf{Domain:} $\dlConcept{Implementation}$.
\textbf{Range:} $\dlConcept{Library}$.

\paragraph{Example.}
The MACE v0.3 implementation is provided by the MACE library.
\begin{lstlisting}[language=turtle]
ex:MACE-v03 mlips:implementedIn ex:MACE-lib .
\end{lstlisting}
}

\paragraph{Related axioms.} \eqref{ax:3}.

\subsection{\dlRole{isEvaluatedIn}}
\label{sec:term-isEvaluatedIn}

\textbf{Label:} is evaluated in.
Inverse of mlips:evaluatesModel.

\textbf{Domain:} $\dlConcept{TrainedModel}$.
\textbf{Range:} $\dlConcept{BenchmarkResult}$.

\paragraph{Example.}
\InputIfFileExists{artifacts/examples/object-properties/isEvaluatedIn.tex}{}{\textit{TODO: add example.}}

\subsection{\dlRole{isHyperparameterOf}}
\label{sec:term-isHyperparameterOf}

\textbf{Label:} is hyperparameter of.
Inverse of mlips:hasHyperparameter.

\textbf{Domain:} $\dlConcept{Hyperparameter}$.
\textbf{Range:} $\dlConcept{MLIPMethod}$.

\paragraph{Example.}
\InputIfFileExists{artifacts/examples/object-properties/isHyperparameterOf.tex}{}{\textit{TODO: add example.}}

\subsection{\dlRole{isImplementationOf}}
\label{sec:term-isImplementationOf}

\textbf{Label:} is implementation of.
Inverse of mlips:hasImplementation.

\textbf{Domain:} $\dlConcept{Implementation}$.
\textbf{Range:} $\dlConcept{MLIPMethod}$.

\paragraph{Example.}
\InputIfFileExists{artifacts/examples/object-properties/isImplementationOf.tex}{}{\textit{TODO: add example.}}

\subsection{\dlRole{isMetricOf}}
\label{sec:term-isMetricOf}

\textbf{Label:} is metric of.
Inverse of mlips:hasAccuracyMetric.

\textbf{Domain:} $\dlConcept{AccuracyMetric}$.
\textbf{Range:} $\dlConcept{BenchmarkResult}$.

\paragraph{Example.}
\InputIfFileExists{artifacts/examples/object-properties/isMetricOf.tex}{}{\textit{TODO: add example.}}

\subsection{\dlRole{isProducedBy}}
\label{sec:term-isProducedBy}

\textbf{Label:} is produced by.
Inverse of mlips:produces.

\textbf{Domain:} $\dlConcept{TrainedModel}$.
\textbf{Range:} $\dlConcept{MLIPRun}$.

\paragraph{Example.}
\InputIfFileExists{artifacts/examples/object-properties/isProducedBy.tex}{}{\textit{TODO: add example.}}

\subsection{\dlRole{isResultOf}}
\label{sec:term-isResultOf}

\textbf{Label:} is result of.
Inverse of mlips:hasResult.

\textbf{Domain:} $\dlConcept{BenchmarkResult}$.
\textbf{Range:} $\dlConcept{BenchmarkStudy}$.

\paragraph{Example.}
\InputIfFileExists{artifacts/examples/object-properties/isResultOf.tex}{}{\textit{TODO: add example.}}

\subsection{\dlRole{isSettingOf}}
\label{sec:term-isSettingOf}

\textbf{Label:} is setting of.
Inverse of mlips:hasHyperparameterSetting.

\textbf{Domain:} $\dlConcept{HyperparameterSetting}$.
\textbf{Range:} $\dlConcept{MLIPRun}$.

\paragraph{Example.}
\InputIfFileExists{artifacts/examples/object-properties/isSettingOf.tex}{}{\textit{TODO: add example.}}

\subsection{\dlRole{metricProperty}}
\label{sec:term-metricProperty}

\textbf{Label:} metric property.

\textbf{Domain:} $\dlConcept{AccuracyMetric}$.
\textbf{Range:} $\dlConcept{MetricProperty}$.

\paragraph{Example.}
An accuracy metric measures total energy.
\begin{lstlisting}[language=turtle]
ex:metric-RMSE-E mlips:metricProperty
    mlips:EnergyProperty .
\end{lstlisting}
}

\paragraph{Related axioms.} \eqref{ax:25}.

\subsection{\dlRole{metricType}}
\label{sec:term-metricType}

\textbf{Label:} metric type.

\textbf{Domain:} $\dlConcept{AccuracyMetric}$.
\textbf{Range:} $\dlConcept{MetricType}$.

\paragraph{Example.}
An accuracy metric is classified as RMSE.
\begin{lstlisting}[language=turtle]
ex:metric-RMSE-E mlips:metricType mlips:RMSE .
\end{lstlisting}
}

\paragraph{Related axioms.} \eqref{ax:24}.

\subsection{\dlRole{produces}}
\label{sec:term-produces}

\textbf{Label:} produces.
Links an MLIP run to the trained model it produces.

\textbf{Domain:} $\dlConcept{MLIPRun}$.
\textbf{Range:} $\dlConcept{TrainedModel}$.

\paragraph{Example.}
A training run produces a trained model artifact.
\begin{lstlisting}[language=turtle]
ex:run-mtp-TiCr2H mlips:produces ex:model-mtp-TiCr2H .
\end{lstlisting}
}

\paragraph{Related axioms.} \eqref{ax:6}, \eqref{ax:7}.

\subsection{\dlRole{pseudopotentialType}}
\label{sec:term-pseudopotentialType}

\textbf{Label:} pseudopotential type.
The family of pseudopotentials used in a
plane-wave DFT calculation (PAW, ultrasoft, norm-conserving). All-
electron codes leave this slot empty and use mlips:dftBasisSet
instead.

\textbf{Domain:} $\dlConcept{DFTSettings}$.
\textbf{Range:} $\dlConcept{PseudopotentialType}$.

\paragraph{Example.}
A plane-wave DFT settings node uses PAW pseudopotentials.
\begin{lstlisting}[language=turtle]
ex:dft-settings-foo mlips:pseudopotentialType mlips:PAW .
\end{lstlisting}
}

\subsection{\dlRole{reportedIn}}
\label{sec:term-reportedIn}

\textbf{Label:} reported in.

\textbf{Domain:} $\dlConcept{BenchmarkResult}$.
\textbf{Range:} $\dlConcept{ScholarlyArticle}$.

\paragraph{Example.}
A benchmark result is reported in a scholarly article.
\begin{lstlisting}[language=turtle]
ex:result-01 mlips:reportedIn ex:publication-2025 .
\end{lstlisting}
}

\subsection{\dlRole{runsOn}}
\label{sec:term-runsOn}

\textbf{Label:} runs on.
Links an MLIP run to the training dataset it uses.

\textbf{Domain:} $\dlConcept{MLIPRun}$.
\textbf{Range:} $\dlConcept{TrainingDataset}$.

\paragraph{Example.}
A training run uses the Ti--Al training dataset.
\begin{lstlisting}[language=turtle]
ex:run-MACE-TiAl mlips:runsOn ex:ds-TiAl .
\end{lstlisting}
}

\paragraph{Related axioms.} \eqref{ax:5}.

\subsection{\dlRole{sameAsWikidata}}
\label{sec:term-sameAsWikidata}

\textbf{Label:} same as Wikidata.
Links an entity to its corresponding Wikidata item.

\paragraph{Example.}
The TiCr$_2$ material system is linked to the Wikidata entry for Laves phases.
\begin{lstlisting}[language=turtle]
ex:TiCr2 mlips:sameAsWikidata
    <http://www.wikidata.org/entity/Q15724720> .
\end{lstlisting}
}

\subsection{\dlRole{samplingStrategy}}
\label{sec:term-samplingStrategy}

\textbf{Label:} sampling strategy.
Records the sampling strategy or strategies
used to construct a training dataset. Optional and multi-valued: a
dataset may combine several sampling strategies (e.g., chemical and
vibrational).

\textbf{Domain:} $\dlConcept{TrainingDataset}$.
\textbf{Range:} $\dlConcept{SamplingStrategy}$.

\paragraph{Example.}
A training dataset that mixes chemical and vibrational sampling.
\begin{lstlisting}[language=turtle]
ex:ds-TiAl mlips:samplingStrategy ex:ChemicalSampling,
                                  ex:VibrationalSampling .
\end{lstlisting}
}

\subsection{\dlRole{supportsSimulation}}
\label{sec:term-supportsSimulation}

\textbf{Label:} supports simulation.
Links an MLIP method to the simulation types it supports.

\textbf{Domain:} $\dlConcept{MLIPMethod}$.
\textbf{Range:} $\dlConcept{SimulationType}$.

\paragraph{Example.}
The MACE algorithm supports molecular dynamics simulations.
\begin{lstlisting}[language=turtle]
ex:MACE mlips:supportsSimulation ex:MolecularDynamics .
\end{lstlisting}
}

\paragraph{Related axioms.} \eqref{ax:2}, \eqref{ax:14}.

\subsection{\dlRole{targetMaterial}}
\label{sec:term-targetMaterial}

\textbf{Label:} target material.

\textbf{Domain:} $\dlConcept{BenchmarkResult}$.
\textbf{Range:} $\dlConcept{MaterialSystem}$.

\paragraph{Example.}
A benchmark result targets the Ti--Al material system.
\begin{lstlisting}[language=turtle]
ex:result-01 mlips:targetMaterial ex:TiAl .
\end{lstlisting}
}

\paragraph{Related axioms.} \eqref{ax:22}.

\subsection{\dlRole{trainedOn}}
\label{sec:term-trainedOn}

\textbf{Label:} trained on.
Shortcut: links a trained model directly to the training dataset it was fitted on.

\textbf{Domain:} $\dlConcept{TrainedModel}$.
\textbf{Range:} $\dlConcept{TrainingDataset}$.

\paragraph{Example.}
A trained model is linked directly to its training dataset via the shortcut property.
\begin{lstlisting}[language=turtle]
ex:model-mtp-TiCr2H mlips:trainedOn ex:ds-TiCr2H .
\end{lstlisting}
}

\paragraph{Related axioms.} \eqref{ax:11}.

\subsection{\dlRole{trainedUsing}}
\label{sec:term-trainedUsing}

\textbf{Label:} trained using.
Shortcut: links a trained model directly to a hyperparameter setting used during the MLIP run that produced it.

\textbf{Domain:} $\dlConcept{TrainedModel}$.
\textbf{Range:} $\dlConcept{HyperparameterSetting}$.

\paragraph{Example.}
The trained model is linked directly to the hyperparameter setting
used to fit it (the long path goes via the run that produced it).
\begin{lstlisting}[language=turtle]
ex:model-mtp-TiAl mlips:trainedUsing ex:setting-cutoff-5A .
\end{lstlisting}
}

\subsection{\dlRole{trainedWith}}
\label{sec:term-trainedWith}

\textbf{Label:} trained with.
Shortcut: links a trained model directly to the MLIP method it applies.

\textbf{Domain:} $\dlConcept{TrainedModel}$.
\textbf{Range:} $\dlConcept{MLIPMethod}$.

\paragraph{Example.}
A trained model is linked directly to its algorithm via the shortcut property.
\begin{lstlisting}[language=turtle]
ex:model-MACE-TiAl mlips:trainedWith ex:MACE .
\end{lstlisting}
}

\paragraph{Related axioms.} \eqref{ax:10}.

\subsection{\dlRole{usedDFTCode}}
\label{sec:term-usedDFTCode}

\textbf{Label:} used DFT code.
The DFT software used (e.g., VASP, GPAW).

\textbf{Domain:} $\dlConcept{DFTSettings}$.
\textbf{Range:} $\dlConcept{Library}$.

\paragraph{Example.}
The DFT settings specify VASP as the calculation software.
\begin{lstlisting}[language=turtle]
ex:dft-PBE-PAW mlips:usedDFTCode ex:VASP .
\end{lstlisting}
}

\subsection{\dlRole{usedReferenceCode}}
\label{sec:term-usedReferenceCode}

\textbf{Label:} used reference code.
The software code used to perform the
reference calculation (e.g., VASP, GPAW for DFT; Molpro, PySCF, ORCA
for wave-function methods). Generalises usedDFTCode.

\textbf{Domain:} $\dlConcept{ReferenceSettings}$.
\textbf{Range:} $\dlConcept{Library}$.

\paragraph{Example.}
A wave-function calculation used PySCF as the software code. The DFT-specific \texttt{usedDFTCode} is a subproperty.
\begin{lstlisting}[language=turtle]
ex:ccsdt-settings mlips:usedReferenceCode ex:lib-pyscf .
\end{lstlisting}
}

\subsection{\dlRole{usesAlgorithm}}
\label{sec:term-usesAlgorithm}

\textbf{Label:} uses algorithm.

\textbf{Domain:} $\dlConcept{BenchmarkResult}$.
\textbf{Range:} $\dlConcept{MLIPMethod}$.

\paragraph{Example.}
A benchmark result references the MLIP algorithm that was evaluated (legacy property).
\begin{lstlisting}[language=turtle]
ex:result-01 mlips:usesAlgorithm ex:MACE .
\end{lstlisting}
}

\subsection{\dlRole{usesTrainingData}}
\label{sec:term-usesTrainingData}

\textbf{Label:} uses training data.

\textbf{Domain:} $\dlConcept{BenchmarkResult}$.
\textbf{Range:} $\dlConcept{TrainingDataset}$.

\paragraph{Example.}
A benchmark result references the training dataset used (legacy property).
\begin{lstlisting}[language=turtle]
ex:result-01 mlips:usesTrainingData ex:ds-TiCr2H .
\end{lstlisting}
}

\subsection{\dlRole{wasRunBy}}
\label{sec:term-wasRunBy}

\textbf{Label:} was run by.
Inverse of mlips:runsOn.

\textbf{Domain:} $\dlConcept{TrainingDataset}$.
\textbf{Range:} $\dlConcept{MLIPRun}$.

\paragraph{Example.}
\InputIfFileExists{artifacts/examples/object-properties/wasRunBy.tex}{}{\textit{TODO: add example.}}

\subsection{\dlRole{wfMethod}}
\label{sec:term-wfMethod}

\textbf{Label:} wave-function method.
The wave-function correlation method used
to produce reference data (e.g., HF, MP2, CCSD, CCSD(T), DLPNO-CCSD(T),
CASPT2, NEVPT2). Range is WfMethod; concrete methods are named
individuals in mlips-vocab.ttl.

\textbf{Domain:} $\dlConcept{WaveFunctionSettings}$.
\textbf{Range:} $\dlConcept{WfMethod}$.

\paragraph{Example.}
A wave-function settings node points to the canonical DLPNO-CCSD(T) IRI.
\begin{lstlisting}[language=turtle]
ex:wf-settings-foo mlips:wfMethod mlips:DLPNO_CCSDT .
\end{lstlisting}
}

\subsection{\dlRole{xcFunctional}}
\label{sec:term-xcFunctional}

\textbf{Label:} exchange-correlation functional.
The exchange-correlation functional used in
a DFT calculation (e.g., LDA, PBE, PBE0, HSE06, SCAN, omegaB97X). Range
is XCFunctional; concrete functionals are named individuals in
mlips-vocab.ttl.

\textbf{Domain:} $\dlConcept{DFTSettings}$.
\textbf{Range:} $\dlConcept{XCFunctional}$.

\paragraph{Example.}
A DFT settings node points to the canonical \texttt{mlips:PBE} IRI.
\begin{lstlisting}[language=turtle]
ex:dft-settings-foo mlips:xcFunctional mlips:PBE .
\end{lstlisting}
}

\section{Datatype Property Reference}
\label{sec:datatype-property-reference}

This appendix describes all 31 datatype properties defined in the \MLIPsOntology{}.

\subsection{\dlRole{basisSet}}
\label{sec:term-basisSet}

\textbf{Label:} basis set.
The Gaussian basis set used in a
wave-function calculation, e.g., cc-pVTZ, aug-cc-pVDZ, def2-TZVP.

\textbf{Domain:} $\dlConcept{WaveFunctionSettings}$.
\textbf{Range:} \texttt{string}.

\paragraph{Example.}
A correlation-consistent triple-zeta basis set used in a CCSD(T) calculation.
\begin{lstlisting}[language=turtle]
ex:ccsdt-settings mlips:basisSet "cc-pVTZ" .
\end{lstlisting}
}

\subsection{\dlRole{chemicalFormula}}
\label{sec:term-chemicalFormula}

\textbf{Label:} chemical formula.

\textbf{Domain:} $\dlConcept{MaterialSystem}$.
\textbf{Range:} \texttt{string}.

\paragraph{Example.}
The chemical formula of the TiCr$_2$ material system.
\begin{lstlisting}[language=turtle]
ex:TiCr2 mlips:chemicalFormula "TiCr2" .
\end{lstlisting}
}

\subsection{\dlRole{cutoffRadius}}
\label{sec:term-cutoffRadius}

\textbf{Label:} cutoff radius.
Cutoff radius for atomic interactions, in angstroms.

\textbf{Domain:} $\dlConcept{HyperparameterSetting}$.
\textbf{Range:} \texttt{double}.

\paragraph{Example.}
A cutoff radius of 5.0~\AA{} for atomic interactions.
\begin{lstlisting}[language=turtle]
ex:setting-cutoff-5A mlips:cutoffRadius
    "5.0"^^xsd:double .
\end{lstlisting}
}

\subsection{\dlRole{defaultValue}}
\label{sec:term-defaultValue}

\textbf{Label:} default value.

\textbf{Domain:} $\dlConcept{Hyperparameter}$.
\textbf{Range:} \texttt{Literal}.

\paragraph{Example.}
The default value of the cutoff radius hyperparameter.
\begin{lstlisting}[language=turtle]
ex:cutoff-radius mlips:defaultValue "5.0" .
\end{lstlisting}
}

\subsection{\dlRole{energyCutoff}}
\label{sec:term-energyCutoff}

\textbf{Label:} energy cutoff.
Plane-wave energy cutoff in eV.

\textbf{Domain:} $\dlConcept{DFTSettings}$.
\textbf{Range:} \texttt{double}.

\paragraph{Example.}
The plane-wave energy cutoff is 520.0~eV.
\begin{lstlisting}[language=turtle]
ex:dft-PBE-PAW mlips:energyCutoff "520.0"^^xsd:double .
\end{lstlisting}
}

\subsection{\dlRole{frozenCore}}
\label{sec:term-frozenCore}

\textbf{Label:} frozen core.
Whether core electrons are kept frozen
during the post-Hartree-Fock correlation treatment in a wave-function
calculation. True for the standard frozen-core approximation; false
for all-electron correlation.

\textbf{Domain:} $\dlConcept{WaveFunctionSettings}$.
\textbf{Range:} $\dlConcept{boolean}$.

\paragraph{Example.}
The frozen-core approximation was applied (only valence electrons correlated).
\begin{lstlisting}[language=turtle]
ex:ccsdt-settings mlips:frozenCore true .
\end{lstlisting}
}

\subsection{\dlRole{gpuHours}}
\label{sec:term-gpuHours}

\textbf{Label:} GPU hours.

\textbf{Domain:} $\dlConcept{MLIPRun}$.
\textbf{Range:} \texttt{double}.

\paragraph{Example.}
A training run that consumed 100 GPU-hours in total.
\begin{lstlisting}[language=turtle]
ex:run-mtp-TiCr2H mlips:gpuHours "100"^^xsd:integer .
\end{lstlisting}
}

\subsection{\dlRole{hyperparameterDatatype}}
\label{sec:term-hyperparameterDatatype}

\textbf{Label:} hyperparameter datatype.
The expected datatype of this hyperparameter (e.g., float, int, string).

\textbf{Domain:} $\dlConcept{Hyperparameter}$.
\textbf{Range:} \texttt{string}.

\paragraph{Example.}
The expected datatype of the cutoff radius hyperparameter is a floating-point number.
\begin{lstlisting}[language=turtle]
ex:cutoff-radius mlips:hyperparameterDatatype "float" .
\end{lstlisting}
}

\subsection{\dlRole{hyperparameterName}}
\label{sec:term-hyperparameterName}

\textbf{Label:} hyperparameter name.

\textbf{Domain:} $\dlConcept{Hyperparameter}$.
\textbf{Range:} \texttt{string}.

\paragraph{Example.}
The programmatic name of the cutoff radius hyperparameter.
\begin{lstlisting}[language=turtle]
ex:cutoff-radius mlips:hyperparameterName
    "cutoff_radius" .
\end{lstlisting}
}

\subsection{\dlRole{inferenceComplexity}}
\label{sec:term-inferenceComplexity}

\textbf{Label:} inference complexity.
Asymptotic computational complexity of inference.

\textbf{Domain:} $\dlConcept{MLIPMethod}$.
\textbf{Range:} \texttt{string}.

\paragraph{Example.}
Most local potentials (MTP, MACE, ACE) have linear-scaling inference
cost in the number of atoms.
\begin{lstlisting}[language=turtle]
ex:MTP mlips:inferenceComplexity "O(N)" .
\end{lstlisting}
}

\subsection{\dlRole{inferenceHardware}}
\label{sec:term-inferenceHardware}

\textbf{Label:} inference hardware.

\textbf{Domain:} $\dlConcept{BenchmarkResult}$.
\textbf{Range:} \texttt{string}.

\paragraph{Example.}
The hardware on which the inference benchmark was performed.
\begin{lstlisting}[language=turtle]
ex:result-mtp-ticr2h-c15 mlips:inferenceHardware "NVIDIA A100" .
\end{lstlisting}
}

\subsection{\dlRole{inferenceTimePerAtom}}
\label{sec:term-inferenceTimePerAtom}

\textbf{Label:} inference time per atom.
Inference time per atom (microseconds).

\textbf{Domain:} $\dlConcept{BenchmarkResult}$.
\textbf{Range:} \texttt{double}.

\paragraph{Example.}
A benchmark that measured 1.5\,$\mu$s per atom inference time on the
test system.
\begin{lstlisting}[language=turtle]
ex:result-mtp-ticr2h-c15 mlips:inferenceTimePerAtom "1.5"^^xsd:double .
\end{lstlisting}
}

\subsection{\dlRole{kPointMesh}}
\label{sec:term-kPointMesh}

\textbf{Label:} k-point mesh.
K-point mesh specification (e.g., '4x4x4').

\textbf{Domain:} $\dlConcept{DFTSettings}$.
\textbf{Range:} \texttt{string}.

\paragraph{Example.}
The DFT settings specify a $4\times4\times4$ k-point mesh.
\begin{lstlisting}[language=turtle]
ex:dft-PBE-PAW mlips:kPointMesh "4x4x4" .
\end{lstlisting}
}

\subsection{\dlRole{learningRate}}
\label{sec:term-learningRate}

\textbf{Label:} learning rate.
Learning rate used during model training.

\textbf{Domain:} $\dlConcept{HyperparameterSetting}$.
\textbf{Range:} \texttt{double}.

\paragraph{Example.}
A learning rate of 0.001 used during model training.
\begin{lstlisting}[language=turtle]
ex:setting-lr mlips:learningRate "0.001"^^xsd:double .
\end{lstlisting}
}

\subsection{\dlRole{materialClass}}
\label{sec:term-materialClass}

\textbf{Label:} material class.
E.g., element, binary alloy, ternary compound, HEA.

\textbf{Domain:} $\dlConcept{MaterialSystem}$.
\textbf{Range:} \texttt{string}.

\paragraph{Example.}
TiCr$_2$ is classified as a Laves phase.
\begin{lstlisting}[language=turtle]
ex:TiCr2 mlips:materialClass "Laves phase" .
\end{lstlisting}
}

\subsection{\dlRole{maxValue}}
\label{sec:term-maxValue}

\textbf{Label:} maximum value.

\textbf{Domain:} $\dlConcept{Hyperparameter}$.
\textbf{Range:} \texttt{Literal}.

\paragraph{Example.}
The maximum accepted value for the cutoff radius.
\begin{lstlisting}[language=turtle]
ex:cutoff-radius mlips:maxValue "10.0" .
\end{lstlisting}
}

\subsection{\dlRole{metricValue}}
\label{sec:term-metricValue}

\textbf{Label:} metric value.

\textbf{Domain:} $\dlConcept{AccuracyMetric}$.
\textbf{Range:} \texttt{double}.

\paragraph{Example.}
The energy RMSE metric has a value of 1.2~meV/atom.
\begin{lstlisting}[language=turtle]
ex:metric-RMSE-E mlips:metricValue "1.2"^^xsd:double .
\end{lstlisting}
}

\subsection{\dlRole{microstructuralFeature}}
\label{sec:term-microstructuralFeature}

\textbf{Label:} microstructural feature.
E.g., point defect, dislocation, surface, grain boundary.

\textbf{Domain:} $\dlConcept{MaterialSystem}$.
\textbf{Range:} \texttt{string}.

\paragraph{Example.}
The material system features point defects in its configurations.
\begin{lstlisting}[language=turtle]
ex:TiCr2 mlips:microstructuralFeature "point defect" .
\end{lstlisting}
}

\subsection{\dlRole{minValue}}
\label{sec:term-minValue}

\textbf{Label:} minimum value.

\textbf{Domain:} $\dlConcept{Hyperparameter}$.
\textbf{Range:} \texttt{Literal}.

\paragraph{Example.}
The minimum accepted value for the cutoff radius.
\begin{lstlisting}[language=turtle]
ex:cutoff-radius mlips:minValue "1.0" .
\end{lstlisting}
}

\subsection{\dlRole{numAngularBasis}}
\label{sec:term-numAngularBasis}

\textbf{Label:} number of angular basis functions.
Number of angular basis functions in the descriptor.

\textbf{Domain:} $\dlConcept{HyperparameterSetting}$.
\textbf{Range:} \texttt{integer}.

\paragraph{Example.}
The descriptor uses 6 angular basis functions.
\begin{lstlisting}[language=turtle]
ex:setting-angular mlips:numAngularBasis 6 .
\end{lstlisting}
}

\subsection{\dlRole{numConfigurations}}
\label{sec:term-numConfigurations}

\textbf{Label:} number of configurations.

\textbf{Domain:} $\dlConcept{TrainingDataset}$.
\textbf{Range:} \texttt{integer}.

\paragraph{Example.}
The training dataset contains 1\,019 atomic configurations.
\begin{lstlisting}[language=turtle]
ex:ds-TiCr2H mlips:numConfigurations 1019 .
\end{lstlisting}
}

\subsection{\dlRole{numLayers}}
\label{sec:term-numLayers}

\textbf{Label:} number of layers.
Number of layers in the neural network architecture.

\textbf{Domain:} $\dlConcept{HyperparameterSetting}$.
\textbf{Range:} \texttt{integer}.

\paragraph{Example.}
A network architecture with 4 message-passing layers.
\begin{lstlisting}[language=turtle]
ex:setting-layers-4 mlips:numLayers 4 .
\end{lstlisting}
}

\subsection{\dlRole{numRadialBasis}}
\label{sec:term-numRadialBasis}

\textbf{Label:} number of radial basis functions.
Number of radial basis functions in the descriptor.

\textbf{Domain:} $\dlConcept{HyperparameterSetting}$.
\textbf{Range:} \texttt{integer}.

\paragraph{Example.}
The descriptor uses 8 radial basis functions.
\begin{lstlisting}[language=turtle]
ex:setting-radial mlips:numRadialBasis 8 .
\end{lstlisting}
}

\subsection{\dlRole{peakMemory}}
\label{sec:term-peakMemory}

\textbf{Label:} peak memory.
Peak memory during training (GB).

\textbf{Domain:} $\dlConcept{MLIPRun}$.
\textbf{Range:} \texttt{double}.

\paragraph{Example.}
A training run with peak GPU memory usage of 24 GB.
\begin{lstlisting}[language=turtle]
ex:run-mtp-TiCr2H mlips:peakMemory "24.0"^^xsd:double .
\end{lstlisting}
}

\subsection{\dlRole{settingValue}}
\label{sec:term-settingValue}

\textbf{Label:} setting value.

\textbf{Domain:} $\dlConcept{HyperparameterSetting}$.
\textbf{Range:} \texttt{Literal}.

\paragraph{Example.}
The concrete value assigned to a hyperparameter setting.
\begin{lstlisting}[language=turtle]
ex:setting-cutoff-5A mlips:settingValue "5.0" .
\end{lstlisting}
}

\subsection{\dlRole{supportsGPU}}
\label{sec:term-supportsGPU}

\textbf{Label:} supports GPU.

\textbf{Domain:} $\dlConcept{MLIPMethod}$.
\textbf{Range:} $\dlConcept{boolean}$.

\paragraph{Example.}
MACE's canonical implementation supports GPU acceleration.
\begin{lstlisting}[language=turtle]
ex:MACE mlips:supportsGPU "true"^^xsd:boolean .
\end{lstlisting}
}

\subsection{\dlRole{supportsParallelization}}
\label{sec:term-supportsParallelization}

\textbf{Label:} supports parallelization.

\textbf{Domain:} $\dlConcept{MLIPMethod}$.
\textbf{Range:} $\dlConcept{boolean}$.

\paragraph{Example.}
MACE supports multi-node and multi-core parallel training and
inference (e.g., via DDP or MPI).
\begin{lstlisting}[language=turtle]
ex:MACE mlips:supportsParallelization "true"^^xsd:boolean .
\end{lstlisting}
}

\subsection{\dlRole{trainingComplexity}}
\label{sec:term-trainingComplexity}

\textbf{Label:} training complexity.
Asymptotic computational complexity of training.

\textbf{Domain:} $\dlConcept{MLIPMethod}$.
\textbf{Range:} \texttt{string}.

\paragraph{Example.}
MTP has linear-scaling training cost in the number of atomic
configurations (per epoch).
\begin{lstlisting}[language=turtle]
ex:MTP mlips:trainingComplexity "O(N)" .
\end{lstlisting}
}

\subsection{\dlRole{trainingDuration}}
\label{sec:term-trainingDuration}

\textbf{Label:} training duration.
Measured wall-clock training time (hours).

\textbf{Domain:} $\dlConcept{MLIPRun}$.
\textbf{Range:} \texttt{double}.

\paragraph{Example.}
A training run that took 4.2 wall-clock hours.
\begin{lstlisting}[language=turtle]
ex:run-mtp-TiCr2H mlips:trainingDuration "4.2"^^xsd:double .
\end{lstlisting}
}

\subsection{\dlRole{trainingHardware}}
\label{sec:term-trainingHardware}

\textbf{Label:} training hardware.

\textbf{Domain:} $\dlConcept{MLIPRun}$.
\textbf{Range:} \texttt{string}.

\paragraph{Example.}
A training run executed on an NVIDIA A100 GPU.
\begin{lstlisting}[language=turtle]
ex:run-mtp-TiCr2H mlips:trainingHardware "NVIDIA A100" .
\end{lstlisting}
}

\subsection{\dlRole{version}}
\label{sec:term-version}

\textbf{Label:} version.

\textbf{Domain:} $\dlConcept{Implementation}$.
\textbf{Range:} \texttt{string}.

\paragraph{Example.}
The version string of a software implementation.
\begin{lstlisting}[language=turtle]
ex:MACE-v03 mlips:version "0.3.0" .
\end{lstlisting}
}

\par}
\section{Axiom Catalog}
\label{sec:axiom-catalog}

This appendix enumerates all axioms in the \MLIPsOntology{}, organized by
module. For each axiom we give the formal statement and a concrete example
illustrating its effect.

\subsection{Method Module}

\subsubsection{Every method has at least one hyperparameter}
\begin{align}
  \MLIPMethodC &\sqsub \existsR{hasHyperparameter}{Hyperparameter} \axiomtag
\end{align}
\emph{Example.} The MACE method declares hyperparameters such as cutoff
radius and number of layers. An $\MLIPMethodC$ instance without any
$\hasHyperparameterR$ link would violate this axiom.

\subsubsection{Every method supports at least one simulation type}
\begin{align}
  \MLIPMethodC &\sqsub \existsR{supportsSimulation}{SimulationType} \axiomtag
\end{align}
\emph{Example.} MACE supports molecular dynamics and geometry optimization.
A method that does not declare any supported simulation type is
incomplete.

\subsubsection{Every method has a functional form}
\begin{align}
  \MLIPMethodC &\sqsub \existsR{hasFunctionalForm}{FunctionalForm} \axiomtag
\end{align}
\emph{Example.} The MTP method has a moment-tensor polynomial functional
form; MACE has an equivariant message-passing ansatz. A method whose
functional form is not recorded cannot be reproduced.

\subsubsection{Every method has a loss function}
\begin{align}
  \MLIPMethodC &\sqsub \existsR{hasLossFunction}{LossFunction} \axiomtag
\end{align}
\emph{Example.} Most MLIP methods use a weighted sum of mean-squared
errors on energies, forces, and stresses as their loss. A method without
a recorded loss is underspecified.

\subsubsection{Every method has a training algorithm}
\begin{align}
  \MLIPMethodC &\sqsub \existsR{hasTrainingAlgorithm}{\mlsAlgorithmC} \axiomtag
\end{align}
\emph{Example.} MTP is typically fitted with L-BFGS; MACE with Adam.
The training algorithm is recorded as an $\mlsAlgorithmC$ instance, which
is the sole alignment point between the MLIPs ontology and ML-Schema.

\subsubsection{Methods are disjoint from ML-Schema algorithms}
\begin{align}
  \MLIPMethodC \sqcap \mlsAlgorithmC &\sqsub \bot \axiomtag
\end{align}
\emph{Example.} MACE is an $\MLIPMethodC$; Adam is an $\mlsAlgorithmC$.
The two refer to fundamentally different kinds of thing---a named
functional-form-plus-loss-plus-training-procedure recipe versus a
parameter-optimisation procedure---and are declared disjoint. The
$\hasTrainingAlgorithmR$ role links them.

\subsubsection{Every implementation specifies its library}
\begin{align}
  \ImplementationC &\sqsub \existsR{implementedIn}{Library} \axiomtag
\end{align}
\emph{Example.} ``MACE v0.3'' is an implementation in the MACE library.
An $\ImplementationC$ without an $\implementedInR$ link is invalid.

\subsubsection{Every MLIP run applies exactly one method}
\begin{align}
  \MLIPRunC &\sqsub \exactR{1}{appliesMethod}{MLIPMethod} \axiomtag
\end{align}
\emph{Example.} An MLIP run that produces a C15-MTP for TiCr$_2$-H applies
the MTP method---not two methods simultaneously, and not zero.

\subsubsection{Every MLIP run uses exactly one training dataset}
\begin{align}
  \MLIPRunC &\sqsub \exactR{1}{runsOn}{TrainingDataset} \axiomtag
\end{align}
\emph{Example.} The MLIP run uses the TiCr$_2$-H DFT dataset with 1{,}019
configurations. If a run combines multiple datasets, they should be
represented as a single merged $\TrainingDatasetC$.

\subsubsection{Every MLIP run produces exactly one trained model}
\begin{align}
  \MLIPRunC &\sqsub \exactR{1}{produces}{TrainedModel} \axiomtag
\end{align}
\emph{Example.} The MLIP run produces a single C15-MTP model for TiCr$_2$-H.
If the same run produces checkpoints, only the final model is recorded.

\subsubsection{Every trained model was produced by some MLIP run}
\begin{align}
  \TrainedModelC &\sqsub \existsR{produces^{-}}{MLIPRun} \axiomtag
\end{align}
\emph{Example.} A $\TrainedModelC$ instance ``C15-MTP-TiCr2H-v1'' must be
linked to an $\MLIPRunC$ that produced it. A model cannot exist without
provenance about how it was trained.

\subsubsection{Every hyperparameter belongs to some method}
\begin{align}
  \HyperparameterC &\sqsub \existsR{hasHyperparameter^{-}}{MLIPMethod} \axiomtag
\end{align}
\emph{Example.} The hyperparameter ``cutoff radius'' is defined for the
MTP method. A $\HyperparameterC$ instance that is not linked to any
method via $\hasHyperparameterR^{-}$ is an orphan and violates this
axiom.

\subsubsection{Every implementation belongs to some method}
\begin{align}
  \ImplementationC &\sqsub \existsR{hasImplementation^{-}}{MLIPMethod} \axiomtag
\end{align}
\emph{Example.} ``MLIP-2 v2.0'' is an implementation of the MTP method.
An $\ImplementationC$ that is not linked to any $\MLIPMethodC$ via
$\hasImplementationR^{-}$ is invalid.

\subsubsection{Shortcut: trainedWith is a property chain}
\begin{align}
  \trainedWithR &\equiv \producesR^{-} \circ \appliesMethodR \axiomtag
\end{align}
\emph{Example.} If an $\MLIPRunC$ produces model $m$ and applies
method $a$, then $m$ $\trainedWithR$ $a$ is inferred. This allows
querying ``which method was used to train this model?'' without
navigating through the run.

\subsubsection{Shortcut: trainedOn is a property chain}
\begin{align}
  \trainedOnR &\equiv \producesR^{-} \circ \runsOnR \axiomtag
\end{align}
\emph{Example.} If an $\MLIPRunC$ produces model $m$ and runs on
dataset $d$, then $m$ $\trainedOnR$ $d$ is inferred. This allows
querying ``which dataset was used to train this model?'' directly.

\subsubsection{Shortcut: trainedUsing is a property chain}
\begin{align}
  \trainedUsingR &\equiv \producesR^{-} \circ \hasHyperparameterSettingR \axiomtag
\end{align}
\emph{Example.} If an $\MLIPRunC$ produces model $m$ and has
hyperparameter setting $s$ (e.g., $R_{\mathrm{cut}} = 5.0$~\AA{}),
then $m$ $\trainedUsingR$ $s$ is inferred. This allows querying
``which hyperparameter settings were used to train this model?''
without navigating through the run.

\subsubsection{Domain axioms for method-specific roles}
\begin{align}
  \existsR{hasHyperparameter}{\top} &\sqsub \MLIPMethodC \axiomtag \\
  \existsR{hasImplementation}{\top} &\sqsub \MLIPMethodC \axiomtag \\
  \existsR{supportsSimulation}{\top} &\sqsub \MLIPMethodC \axiomtag
\end{align}
\emph{Example.} If an entity has a $\hasHyperparameterR$ link, it must be
an $\MLIPMethodC$. This prevents, e.g., a $\TrainingDatasetC$ from
accidentally being assigned hyperparameters.

\subsection{Training Data Module}

\subsubsection{Every dataset covers at least one material system}
\begin{align}
  \TrainingDatasetC &\sqsub \existsR{coversMaterial}{MaterialSystem} \axiomtag
\end{align}
\emph{Example.} A TiCr$_2$-H training dataset covers the TiCr$_2$ material system.
A dataset without any $\coversMaterialR$ link is incomplete.

\subsubsection{Every dataset covers at least one physical property}
\begin{align}
  \TrainingDatasetC &\sqsub \existsR{coversProperty}{CoveredProperty} \axiomtag
\end{align}
\emph{Example.} The TiCr$_2$-H dataset covers energies, forces, and stresses.
A dataset must declare at least one covered property.

\subsubsection{Every dataset has a provenance classification}
\begin{align}
  \TrainingDatasetC &\sqsub \existsR{datasetProvenance}{DatasetProvenance} \axiomtag
\end{align}
\emph{Example.} The TiCr$_2$-H dataset is classified as $\dlConcept{Published}$.
Every dataset must indicate whether it is published, in-house, or
augmented.

\subsubsection{Every dataset contains at least one atomic configuration}
\begin{align}
  \TrainingDatasetC &\sqsub \existsR{hasConfiguration}{AtomicConfiguration} \axiomtag
\end{align}
\emph{Example.} The TiCr$_2$-H C15 dataset contains 1{,}019 atomic configurations.
A dataset without any configurations is empty and invalid.

\subsubsection{Every DFT calculation has settings}
\begin{align}
  \DFTCalculationC &\sqsub \existsR{hasDFTSettings}{DFTSettings} \axiomtag
\end{align}
\emph{Example.} A DFT calculation specifies PBE as the
exchange-correlation functional, a 4$\times$4$\times$4 k-point mesh, and
PAW pseudopotentials. A $\DFTCalculationC$ without $\hasDFTSettingsR$ is
incomplete.

\subsection{Benchmark Module}

\subsubsection{Every study contains at least one result}
\begin{align}
  \BenchmarkStudyC &\sqsub \existsR{hasResult}{BenchmarkResult} \axiomtag
\end{align}
\emph{Example.} A study by Kumar et al.\ (2025)~\cite{kumar2025ticr2h}
reports evaluation results for MTP on TiCr$_2$-H. A $\BenchmarkStudyC$
without any $\hasResultR$ link contains no data and is invalid.

\subsubsection{Every result evaluates exactly one trained model}
\begin{align}
  \BenchmarkResultC &\sqsub \exactR{1}{evaluatesModel}{TrainedModel} \axiomtag
\end{align}
\emph{Example.} A benchmark result evaluates the ``C15-MTP-TiCr2H-v1'' model.
Each result is tied to exactly one model; comparing two models requires two
separate $\BenchmarkResultC$ instances.

\subsubsection{Every result targets at least one material system}
\begin{align}
  \BenchmarkResultC &\sqsub \existsR{targetMaterial}{MaterialSystem} \axiomtag
\end{align}
\emph{Example.} The benchmark result targets the TiCr$_2$ material system for
testing. A result must specify what material was used for evaluation.

\subsubsection{Every result reports at least one accuracy metric}
\begin{align}
  \BenchmarkResultC &\sqsub \existsR{hasAccuracyMetric}{AccuracyMetric} \axiomtag
\end{align}
\emph{Example.} The result reports RMSE of energy (3.17~meV/atom). A
benchmark result without any accuracy metric carries no evaluation data.

\subsubsection{Every accuracy metric has exactly one type}
\begin{align}
  \AccuracyMetricC &\sqsub \exactR{1}{metricType}{MetricType} \axiomtag
\end{align}
\emph{Example.} The metric is of type RMSE. A metric cannot be both RMSE
and MAE simultaneously, nor can it lack a type.

\subsubsection{Every accuracy metric measures exactly one property}
\begin{align}
  \AccuracyMetricC &\sqsub \exactR{1}{metricProperty}{MetricProperty} \axiomtag
\end{align}
\emph{Example.} The RMSE metric measures energy. A single
$\AccuracyMetricC$ instance measures one property; separate metrics are
created for energy, force, and stress.

\subsubsection{Every result belongs to some study}
\begin{align}
  \BenchmarkResultC &\sqsub \existsR{hasResult^{-}}{BenchmarkStudy} \axiomtag
\end{align}
\emph{Example.} The benchmark result for MTP on TiCr$_2$-H belongs to the
Kumar et al.\ (2025) study. A result cannot exist without a parent study.

\subsubsection{Every accuracy metric belongs to some result}
\begin{align}
  \AccuracyMetricC &\sqsub \existsR{hasAccuracyMetric^{-}}{BenchmarkResult} \axiomtag
\end{align}
\emph{Example.} The RMSE of energy metric belongs to a specific benchmark
result. An orphan metric with no parent result is invalid.

\subsection{Axioms Under Discussion}

The following axioms are candidates for inclusion but require further
discussion with domain experts.

\subsubsection{Every DFT settings belongs to some calculation}
\begin{align*}
  \DFTSettingsC &\sqsub \existsR{hasDFTSettings^{-}}{DFTCalculation}
\end{align*}
\emph{Example.} DFT settings (PBE, 4$\times$4$\times$4 k-points) belong
to a specific calculation. \emph{Discussion:} Settings could potentially be
shared across calculations as templates, in which case this axiom would be
too restrictive.

\subsubsection{Every atomic configuration belongs to some dataset}
\begin{align*}
  \AtomicConfigurationC &\sqsub \existsR{hasConfiguration^{-}}{TrainingDataset}
\end{align*}
\emph{Example.} An atomic configuration of TiCr$_2$-H belongs to the TiCr$_2$-H
training dataset. \emph{Discussion:} Configurations could belong to
multiple datasets (e.g., an augmented dataset reuses configurations from a
published one), which would still satisfy this axiom but complicates
provenance tracking.

\subsubsection{Every hyperparameter setting references a hyperparameter}
\begin{align*}
  \HyperparameterSettingC &\sqsub \existsR{forHyperparameter}{Hyperparameter}
\end{align*}
\emph{Example.} A setting ``cutoff radius = 5.0~\AA{}'' references the
hyperparameter ``cutoff radius''. \emph{Discussion:} This axiom is
straightforward but raises the question of whether settings should also be
existentially dependent on an $\MLIPRunC$ or $\TrainedModelC$.

\section{SPARQL Query Templates for All Competency Questions}
\label{sec:appendix-sparql}

This appendix provides SPARQL query templates for all eight competency
questions. CQ3, CQ6, and CQ7 are discussed in the main paper
(Section~\bodyref{sec:eval-cq}{5.2 of the paper}); the remaining
queries are presented here.

\subsection{CQ1: Method hyperparameters}

Which MLIP methods exist, and what hyperparameters does each accept?
\begin{lstlisting}[language=sparql]
SELECT ?algo ?hp ?name ?type ?default WHERE {
  ?algo a mlips:MLIPMethod ;
        mlips:hasHyperparameter ?hp .
  ?hp mlips:hyperparameterName ?name ;
      mlips:hyperparameterDatatype ?type .
  OPTIONAL { ?hp mlips:defaultValue ?default }
}
\end{lstlisting}

\subsection{CQ2: Implementations and versions}

Which libraries implement a given method?
\begin{lstlisting}[language=sparql]
SELECT ?algo ?impl ?lib ?version WHERE {
  ?algo a mlips:MLIPMethod ;
        mlips:hasImplementation ?impl .
  ?impl mlips:implementedIn ?lib ;
        mlips:version ?version .
}
\end{lstlisting}

\subsection{CQ4: Dataset provenance}

What is the provenance of a training dataset?
\begin{lstlisting}[language=sparql]
SELECT ?ds ?label ?prov WHERE {
  ?ds a mlips:TrainingDataset ;
      rdfs:label ?label ;
      mlips:datasetProvenance ?prov .
}
\end{lstlisting}

\subsection{CQ5: Dataset size and property coverage}

How many configurations does a dataset contain, and what properties are
covered?
\begin{lstlisting}[language=sparql]
SELECT ?ds ?nconfig ?prop WHERE {
  ?ds a mlips:TrainingDataset ;
      mlips:numConfigurations ?nconfig ;
      mlips:coversProperty ?prop .
}
\end{lstlisting}

\subsection{CQ8: Simulation types supported by a method}

Which simulation types have been performed with a given MLIP?
\begin{lstlisting}[language=sparql]
SELECT ?algo ?sim WHERE {
  ?algo a mlips:MLIPMethod ;
        mlips:supportsSimulation ?sim .
}
\end{lstlisting}

\subsection{CQ9: Efficiency of methods and trained models}

For a given target accuracy, which methods and trained models are
most efficient, combining asymptotic complexity (prior knowledge on the
method) with measured training and inference cost?
\begin{lstlisting}[language=sparql]
SELECT ?algo ?inf_complexity ?gpu_hours ?t_inf ?rmse WHERE {
  ?result a mlips:BenchmarkResult ;
          mlips:evaluatesModel ?model ;
          mlips:hasAccuracyMetric ?metric ;
          mlips:inferenceTimePerAtom ?t_inf .
  ?metric mlips:metricType mlips:RMSE ;
          mlips:metricValue ?rmse .
  FILTER(?rmse < 2.0)
  ?model mlips:trainedWith ?algo .
  ?algo mlips:inferenceComplexity ?inf_complexity .
  ?run mlips:produces ?model ;
       mlips:gpuHours ?gpu_hours .
} ORDER BY ?t_inf
\end{lstlisting}

\section{Description Logic Query Formulations}
\label{sec:appendix-dl-queries}

This appendix provides description logic formulations for the competency
questions evaluated in Section~\ref{sec:eval-cq}. Each query is expressed
as a conjunctive query over the \MLIPsOntology{} T-Box.

\subsection{CQ1: Method hyperparameters}

Retrieve all pairs of methods and their hyperparameters:
\begin{align*}
  q(x, y) \leftarrow \MLIPMethodC(x) \wedge \hasHyperparameterR(x, y) \wedge \HyperparameterC(y)
\end{align*}

\subsection{CQ2: Implementations and versions}

Retrieve implementations with their libraries:
\begin{align*}
  q(x, y, z) \leftarrow \MLIPMethodC(x) \wedge \hasImplementationR(x, y) \wedge \implementedInR(y, z)
\end{align*}

\subsection{CQ3: Training datasets for a material system}

Retrieve datasets for a material with their DFT settings:
\begin{align*}
  q(d, s) \leftarrow{} & \TrainingDatasetC(d) \wedge \coversMaterialR(d, m) \wedge \\
    & \dlRole{hasDFTCalculation}(d, c) \wedge \hasDFTSettingsR(c, s)
\end{align*}

\subsection{CQ4: Dataset provenance}

Retrieve provenance classification of datasets:
\begin{align*}
  q(d, p) \leftarrow \TrainingDatasetC(d) \wedge \datasetProvenanceR(d, p)
\end{align*}

\subsection{CQ5: Dataset size and property coverage}

Retrieve dataset size and covered properties:
\begin{align*}
  q(d, n, p) \leftarrow \TrainingDatasetC(d) \wedge \numConfigurationsR(d, n) \wedge \coversPropertyR(d, p)
\end{align*}

\subsection{CQ6: Published benchmarks}

Retrieve benchmark results for a method on a material, with accuracy
metrics:
\begin{align*}
  q(s, v) \leftarrow{} & \BenchmarkStudyC(s) \wedge \hasResultR(s, r) \wedge \evaluatesModelR(r, m) \wedge \\
    & \trainedWithR(m, a) \wedge \targetMaterialR(r, t) \wedge \\
    & \hasAccuracyMetricR(r, c) \wedge \metricValueR(c, v)
\end{align*}

\subsection{CQ7: Ranking combinations by accuracy}

Retrieve method--material pairs ordered by accuracy (instantiated for
RMSE of energy):
\begin{align*}
  q(a, v) \leftarrow{} & \evaluatesModelR(r, m) \wedge \trainedWithR(m, a) \wedge \\
    & \hasAccuracyMetricR(r, c) \wedge \metricTypeR(c, \dlConcept{RMSE}) \wedge \\
    & \metricPropertyR(c, \dlConcept{EnergyProperty}) \wedge \metricValueR(c, v)
\end{align*}

\subsection{CQ8: Simulation types}

Retrieve simulation types supported by each method:
\begin{align*}
  q(x, y) \leftarrow \MLIPMethodC(x) \wedge \supportsSimulationR(x, y) \wedge \SimulationTypeC(y)
\end{align*}

\section{Design Discussion: MLIP Run and Shortcut Properties}
\label{sec:appendix-run}

In ML-Schema, the relationship between a trained model and its algorithm is
mediated by $\mlsRunC$, which represents a training execution:
\begin{align*}
  \mlsAlgorithmC
    &\xleftarrow{\mlsExecutesR}
    \mlsRunC
    \xrightarrow{\mlsHasOutputR}
    \mlsModelC \\
  \mlsDatasetC
    &\xleftarrow{\mlsHasInputR}
    \mlsRunC
\end{align*}
The \MLIPsOntology{} follows this activity-mediated pattern but departs
from ML-Schema in an important way. Because $\MLIPMethodC$ is disjoint
from $\mlsAlgorithmC$ (see
Section~\bodyref{sec:algorithm-module}{4.2 of the paper} and
Appendix~\bodyref{sec:appendix-method}{H of the extended version}),
an $\MLIPRunC$ cannot be a subclass
of $\mlsRunC$: $\mlsRunC$'s $\mlsExecutesR$ property ranges over
$\mlsAlgorithmC$, whereas an $\MLIPRunC$ applies a method, not an
algorithm. We therefore make $\MLIPRunC$ a subclass of $\provActivityC$
only, and introduce a dedicated role $\appliesMethodR$ with range
$\MLIPMethodC$ in place of $\mlsExecutesR$. An optional
$\hasTrainingRunR$ links an $\MLIPRunC$ to an underlying $\mlsRunC$
whenever the training algorithm itself is modelled in ML-Schema.

On top of this, we provide the shortcut roles $\trainedWithR$ and
$\trainedOnR$ directly on $\TrainedModelC$, defined as property chains
through the MLIP run. These shortcuts enable direct queries without
navigating the run entity, which is important because our primary data
source---published MLIP literature---rarely reports the operational
details of training runs (hardware, duration, random seeds). In many
cases, the $\MLIPRunC$ entity will carry only its method, dataset, and
model links, making the shortcut properties the more natural query
path.

\paragraph{Critical assessment.}
The shortcut properties introduce redundancy: the same information is
accessible both directly on the model and via the run. While the property
chain axioms guarantee consistency in principle, in practice data entry
errors could create inconsistencies if shortcuts and run-mediated paths are
populated independently. We mitigate this through SHACL shapes that
validate consistency between the two paths.

An alternative design would omit the shortcuts entirely and always require
navigation through $\MLIPRunC$. This would be cleaner from an ontology
engineering perspective but would make common queries (``which method
produced this model?'') unnecessarily verbose, particularly for data
extracted from publications where the run carries no additional information
beyond the links already captured by the shortcuts.

\section{Design Discussion: ``Method'' vs.\ ``Algorithm'' Granularity}
\label{sec:appendix-method}

A referee raised a subtle but important point: what the MLIP community
calls an ``algorithm'' is not, strictly speaking, an algorithm in the
ML-Schema sense. This appendix documents the granularity mismatch, the
terminology choice, and the design we have adopted---an
$\MLIPMethodC$ class that is disjoint from $\mlsAlgorithmC$ and is
decomposed into its three conceptual components, with the training
algorithm contributed as the sole alignment point with ML-Schema.

\subsection{The mismatch}

In ML-Schema, $\mlsAlgorithmC$ is defined as ``a computational
procedure''---typically a training algorithm such as stochastic gradient
descent, Adam, or L-BFGS. The procedure consumes data and hyperparameters
and produces a fitted $\mlsModelC$. This is the classical ML view:
\emph{model family} and \emph{training procedure} are conceptually
distinct, and $\mlsAlgorithmC$ names only the latter.

In the MLIP community, a named ``algorithm'' (MACE, MTP, ACE, NequIP)
typically bundles \emph{four} different things:
\begin{enumerate}
\item a \textbf{descriptor} of the local atomic environment (moment
  tensors, SOAP, symmetry functions, equivariant messages);
\item a \textbf{functional form} that maps descriptor features to the
  local energy (linear expansion in basis functions, neural network,
  kernel regression);
\item a \textbf{loss function} for training (typically a weighted
  combination of energy error, force error, and stress error); and
\item a \textbf{training procedure} in the ML-Schema sense (an
  optimizer, sampling scheme, and stopping criterion).
\end{enumerate}
Only item~(4) aligns directly with $\mlsAlgorithmC$. Items (1)--(3) are
properties of the \emph{model family} rather than of any training
procedure, and the bundle as a whole is what the community refers to
by a single name.

\subsection{Analogy: knowledge graph embeddings}

The same mismatch arises in other machine-learning sub-communities.
Consider TransE~\cite{bordes2013transe}, a standard knowledge graph
embedding method. ``TransE'' as used in the KGE literature bundles:
\begin{itemize}
\item an entity representation (points in $\mathbb{R}^d$);
\item a relation representation (translation vectors in $\mathbb{R}^d$);
\item a scoring function $f(h,r,t) = \|h + r - t\|$; and
\item a training procedure (stochastic gradient descent with
  margin-based ranking loss and negative sampling).
\end{itemize}
Asked ``what is TransE in ML-Schema?'', a strict reading splits it
across $\mlsAlgorithmC$ (the SGD procedure) and $\mlsModelC$ (the fitted
embeddings). ML-Schema has no single class for the bundle that the
community actually names and identifies. The same structural issue
surfaces in graph neural networks (GCN, GAT, GIN), generative models
(VAE, GAN), and many other ML sub-communities where a named
``algorithm'' is a bundle of model-family choices plus a training
recipe.

\subsection{Terminology: ``Algorithm'', ``Method'', or ``Formalism''?}

Given that MTP and its siblings are \emph{not} algorithms in
ML-Schema's sense, it is worth asking which of three candidate words is
most accurate and most widely used.

\paragraph{Algorithm.}
Narrowly, an algorithm is a computational procedure---a sequence of
steps. MTP is not a sequence of steps; it is a mathematical framework
that defines what energy function is being fitted plus the machinery to
fit it. ``The MTP algorithm'' is used casually in the literature but
is semantically imprecise.

\paragraph{Method.}
Generic: ``a named approach or recipe''. MTP is clearly a method: a
recipe for constructing a potential from a particular descriptor
(moment tensors), a particular functional form (linear expansion in
basis functions), a particular loss (weighted energy/force/stress),
and a particular training algorithm. Semantically clean; carries no
specific structural commitment beyond ``named bundle''.

\paragraph{Formalism.}
Emphasises the mathematical framework. In physics and chemistry,
``formalism'' is the standard word for ``a mathematical structure that
defines what a class of objects looks like and how they behave''---e.g.,
``the Lagrangian formalism'', ``the density functional theory
formalism'', ``the Standard Model formalism''. MTP fits this usage
well: the moment tensor expansion defines what MTP potentials are,
before any specific one is fitted. This is the tightest of the three
for MTP-the-mathematical-framework but carries a physics-flavoured
connotation.

\paragraph{Literature usage across MLIP papers.}
``Method'' dominates in practice (Novikov et al.\ 2021, Batatia et al.\
2022, Bart\'ok et al.\ 2010). ``Formalism'' and ``framework'' appear
when authors want to emphasise the mathematical structure (Drautz 2019
for ACE, and occasionally in the MTP literature). ``Algorithm'' is used
casually but rarely in formal technical contexts.

\paragraph{Cross-community applicability.}
For an upper ontology targeting multiple ML sub-communities, ``Method''
is the most portable label: it reads naturally for MLIPs, KGE, GNNs,
generative models, and classical ML alike. ``Formalism'' is accurate
for physics-adjacent fields but heavy in KGE (``the TransE formalism''
is correct but uncommon) and uncommon in GNN/CV/NLP. ``Algorithm'' is
semantically wrong everywhere.

\paragraph{Decision.}
We adopt \emph{Method} as the class name at the domain layer
($\MLIPMethodC$). Author freedom to use ``formalism'' or ``framework''
in prose when the mathematical-structure emphasis is appropriate is
retained; the class name does not prescribe author vocabulary.

\subsection{The design adopted}
\label{sec:appendix-method-design}

Rather than align $\MLIPMethodC$ with $\mlsAlgorithmC$ by subsumption
--- which would conflate the named bundle with the training procedure
it uses---we declare them \emph{disjoint} and decompose the bundle
into its three non-training components, with the training algorithm
contributed as the sole alignment point with ML-Schema:
\begin{align*}
  \MLIPMethodC \sqcap \mlsAlgorithmC &\sqsub \bot \\
  \MLIPMethodC &\sqsub \existsR{hasFunctionalForm}{FunctionalForm} \\
  \MLIPMethodC &\sqsub \existsR{hasLossFunction}{LossFunction} \\
  \MLIPMethodC &\sqsub \existsR{hasTrainingAlgorithm}{\mlsAlgorithmC}
\end{align*}
Here $\mlsAlgorithmC$ is used in its narrow ML-Schema sense (the
optimiser/training procedure only). The bundle $\MLIPMethodC$ is not a
subclass of $\mlsAlgorithmC$; instead, one of its components \emph{is}
an $\mlsAlgorithmC$. The atomic-environment descriptor is modelled
separately via $\hasDescriptorR$, alongside the method's hyperparameters,
implementations, and supported simulation types.

Concrete instances then look like:
\begin{lstlisting}[language=turtle]
entity:MACE a mlips:MLIPMethod ;
    rdfs:label "MACE" ;
    mlips:hasDescriptor mlips:EquivariantMessagePassingDescriptor ;
    mlips:hasFunctionalForm entity:MACE-NeuralNet ;
    mlips:hasLossFunction entity:WeightedEnergyForceStressLoss ;
    mlips:hasTrainingAlgorithm entity:Adam ;
    mlips:hasHyperparameter entity:hp-rcut, entity:hp-num-layers ;
    mlips:supportsSimulation entity:sim-md .

entity:Adam a mls:Algorithm .  # narrow ML-Schema sense
\end{lstlisting}

For the same reason of disjointness, $\MLIPRunC$ is a subclass of
$\provActivityC$ only, not of $\mlsRunC$: $\mlsRunC.\mlsExecutesR$
ranges over $\mlsAlgorithmC$, which is disjoint from $\MLIPMethodC$. A
dedicated role $\appliesMethodR$ replaces $\mlsExecutesR$ for the MLIP
case. An optional $\hasTrainingRunR$ links an $\MLIPRunC$ to an
$\mlsRunC$ when the training algorithm is itself modelled in
ML-Schema.

\paragraph{Benefits.}
\begin{itemize}
\item The alignment with ML-Schema is tight and semantically correct:
  only the training procedure is an $\mlsAlgorithmC$.
\item Each component is independently queryable and comparable across
  methods: ``which methods share the same descriptor?'', ``which
  methods use a margin-based ranking loss?''
\item The pattern generalises: TransE, GCN, VAE, and other
  community-named bundles can be modelled analogously in sibling
  domain ontologies.
\end{itemize}

\paragraph{Costs.}
\begin{itemize}
\item MLIP researchers commonly leave the training procedure
  unspecified (it is often a detail not worth naming), so in practice
  many $\MLIPMethodC$ instances carry only a placeholder
  $\hasTrainingAlgorithmR$ link. The value of making it explicit is
  mostly conceptual, but is what enables the alignment with
  ML-Schema to be both tight and narrow.
\item The functional-form and loss-function components are often also
  left implicit in the literature; the ontology makes them first-class
  but does not require extensive axiomatisation of their internal
  structure.
\end{itemize}

{\refappendixsloppy
\section{Meta-categorisation: Sortals, Roles, and Reified Relations}
\label{sec:appendix-metasort}

This appendix expands the modelling discipline introduced in
\S\bodyref{sec:metasort}{4.3 of the paper}. It documents
(i)~why we adopted OntoClean
as a methodology while defining our own annotation vocabulary
instead of importing the OntoClean OWL serialisation, (ii)~the
four meta-categories we use, and (iii)~the case-by-case
classification of every class in the \MLIPsOntology{}.

\subsection{Methodology vs.\ vocabulary: why a bespoke annotation set}
\label{sec:metasort-rationale}

OntoClean~\cite{guarino2009ontoclean} is a meta-property framework
for validating the ontological adequacy of taxonomic relations.
Its central distinctions are four meta-properties that classes
may carry---\emph{rigidity} (whether the class essentially
applies to its instances), \emph{identity} (whether the class
supplies criteria of identity), \emph{unity} (whether instances
are unitary wholes), and \emph{dependence} (whether instances
depend existentially on others)---together with a small set of
named categories (\emph{Type}, \emph{Role}, \emph{Phase},
\emph{Mixin}, \emph{Category}) defined as combinations of the
four meta-properties.

We adopt OntoClean's meta-property analysis as the discipline
behind our class design: every class in the \MLIPsOntology{} has
been examined for its rigidity, identity, unity, and dependence
profile, and the resulting classification influences both axiom
choices (existence axioms, disjointness) and label policies
(\S\ref{sec:metasort-labels} below). We do \emph{not}, however,
import the OntoClean OWL serialisation. Two considerations drove
this decision.

\paragraph{Resolvability.} The canonical OntoClean OWL document
at \url{loa.istc.cnr.it/ontologies/OntoClean.owl} did not
resolve when we attempted to dereference it during the writing
of this paper. Other Laboratory for Applied Ontology (LOA)
vocabularies at sibling paths (DUL, DLP, DOLCE-Lite) resolve;
OntoClean specifically does not. A resource paper
that emphasises FAIR-resolvability for its own artefacts cannot
in good conscience import a vocabulary whose canonical IRI is
not retrievable.

\paragraph{Categorial fit.} OntoClean's named categories were
developed to discipline \emph{unary} class hierarchies. Reified
\emph{n-ary} relations---which represent the majority of our
non-sortal classes (\texttt{HyperparameterSetting} is a 3-place
relation; \texttt{BenchmarkResult} is a 4-place relation;
\texttt{AccuracyMetric} is a 3-place relation)---have no
first-class entry in OntoClean's category set. Within OntoClean
they would be classified by meta-properties as anti-rigid
(${\sim}R$), without own identity ($-I$), without unity ($-U$),
and externally dependent ($+D$); the closest named category is
\emph{Role}, but Role is canonically illustrated by one-place
classes like \texttt{Student}, and forcing a 3-place reification
into the same category obscures rather than illuminates the
modelling intent.

We therefore define four named meta-categories locally, as named
individuals of an annotation vocabulary
\texttt{mlips:MetaSort}, and attach them to classes via an
annotation property \texttt{mlips:metaSort}:

\begin{itemize}
\item \texttt{mlips:Sortal}---a class whose instances carry
  their own identity criteria, rigid, with unity, not externally
  dependent. OntoClean profile: $+R$, $+I$, $+U$, $-D$.
  Example: \texttt{MaterialSystem}.
\item \texttt{mlips:SubordinateSortal}---a sortal whose
  instances depend existentially on a parent entity but are
  themselves rigid and identity-bearing. OntoClean profile:
  $+R$, $+I$, $+U$, $+D$. Example: \texttt{Implementation},
  which is identity-bearing (a release/commit/DOI) but exists
  only \emph{of} a method and \emph{in} a library.
\item \texttt{mlips:Role}---an anti-rigid class whose
  instances are roles played by sortal individuals and depend
  existentially on those sortals. OntoClean profile: ${\sim}R$,
  $-I$ (identity inherited from the carrier), $-U$, $+D$. The
  \MLIPsOntology{} currently has no class in this category, but
  the slot is reserved for future extensions.
\item \texttt{mlips:ReifiedRelation}---a class that reifies an
  $n$-ary relation between sortals. Instances have no identity
  outside their relata. OntoClean profile: ${\sim}R$, $-I$,
  $-U$, $+D$. Examples: \texttt{HyperparameterSetting},
  \texttt{BenchmarkResult}, \texttt{AccuracyMetric}.
\end{itemize}

The classification itself is attached to every class in the
ontology source as a single \texttt{mlips:metaSort} triple. The
full set of classifications and the meta-property analysis behind
each is provided in this appendix (Table~\ref{tab:metasort}) and
the dataset repository; the artefacts
are archived at the University of Stuttgart's Dataverse
repository (DaRUS~\cite{darus2026mlips}) for long-term availability.

\subsection{Consequences for axioms and labels}
\label{sec:metasort-labels}

Two practical consequences follow from the classification.

\paragraph{Existential dependence axioms.} For every class
\texttt{C} marked as \texttt{ReifiedRelation} or
\texttt{SubordinateSortal}, the schema carries an existential
axiom on each of its dependencies. Concretely:
\begin{itemize}
\item $\HyperparameterSettingC \sqsubseteq
  \existsR{isSettingOf}{MLIPRun}$ (in addition to the existing
  $\HyperparameterSettingC \sqsubseteq
  \existsR{forHyperparameter}{Hyperparameter}$);
\item $\AccuracyMetricC \sqsubseteq
  \existsR{hasAccuracyMetric^{-}}{BenchmarkResult}$ (already in
  the appendix-axiom catalogue); and
\item $\BenchmarkResultC \sqsubseteq
  \existsR{evaluatesModel}{TrainedModel}$.
\end{itemize}

\paragraph{Computed labels.} Reified-relation instances are
opaque under their IRI alone: a reader who encounters
\texttt{ex:setting-rcut-behler2007} cannot tell from the IRI
fragment what hyperparameter is set to what value in which run.
We therefore compute \texttt{rdfs:label} values for instances of
every reified-relation class by a SPARQL \texttt{CONSTRUCT} that
joins the relation type, the value, and the carrier, producing
labels of the form
``\texttt{<relation-type-label> = <value> for <carrier-label>}''.
The CONSTRUCT rules live in the dataset repository (archived
for long-term
availability on DaRUS~\cite{darus2026mlips}); the computed triples
are materialised into a separate \emph{computed} TTL file that
is not tracked alongside the canonical TTL but is regenerated
as part of the build pipeline. Inverse triples are computed by
an analogous CONSTRUCT pass.

\subsection{Per-class classification}
\label{sec:metasort-table}

Table~\ref{tab:metasort} gives the meta-sort assignment for
every class in the \MLIPsOntology{}. Where the classification
required a non-obvious modelling decision, the row carries a
brief rationale; otherwise the assignment is the canonical case
described in \S\ref{sec:metasort-rationale}.

{\small
\begin{longtable}{ll p{0.38\textwidth}}
\caption{Meta-sort classification of every class in the
  \MLIPsOntology{}. \texttt{Sortal} = identity-bearing, rigid,
  not dependent; \texttt{SubordinateSortal} = identity-bearing
  but existentially dependent on a parent; \texttt{Role} =
  anti-rigid, identity inherited; \texttt{ReifiedRelation} =
  reification of an $n$-ary relation, no identity outside the
  relata.}\label{tab:metasort}\\
\toprule
Class & Meta-sort & Notes \\
\midrule
\endfirsthead
\multicolumn{3}{l}{\emph{(continued)}}\\
\toprule
Class & Meta-sort & Notes \\
\midrule
\endhead
\midrule
\multicolumn{3}{r}{\emph{continued on next page}}\\
\endfoot
\bottomrule
\endlastfoot
\multicolumn{3}{l}{\emph{Sortals (canonical)}} \\
\texttt{MLIPMethod}            & Sortal & ``recipe'' for an MLIP family \\
\texttt{MaterialSystem}        & Sortal & \\
\texttt{TrainingDataset}       & Sortal & \\
\texttt{TrainedModel}          & Sortal & a fitted model is a thing \\
\texttt{BenchmarkStudy}        & Sortal & a published study \\
\texttt{Library}               & Sortal & a software package \\
\texttt{Hyperparameter}        & Sortal & the type-level definition \\
\texttt{FunctionalForm}        & Sortal & \\
\texttt{LossFunction}          & Sortal & \\
\texttt{AtomicEnvironmentDescriptor} & Sortal & \\
\texttt{SimulationType}        & Sortal & enumerative \\
\texttt{XCFunctional}          & Sortal & enumerative \\
\texttt{PseudopotentialType}   & Sortal & enumerative \\
\texttt{WfMethod}              & Sortal & enumerative \\
\texttt{DftBasisSet}           & Sortal & enumerative \\
\texttt{MetricType}            & Sortal & enumerative \\
\texttt{MetricProperty}        & Sortal & enumerative \\
\texttt{CoveredProperty}       & Sortal & enumerative \\
\texttt{DatasetProvenance}     & Sortal & enumerative \\
\texttt{SamplingStrategy}      & Sortal & \\
\texttt{ArchitecturalHyperparameter} & Sortal & subclass of \texttt{Hyperparameter}; identity inherited \\
\texttt{PhysicalHyperparameter} & Sortal & subclass of \texttt{Hyperparameter}; identity inherited \\
\texttt{TrainingHyperparameter} & Sortal & subclass of \texttt{Hyperparameter}; identity inherited \\
\texttt{AtomicConfiguration}   & Sortal & a concrete structure has identity (subclass of \texttt{cmso:AtomicStructure}) \\
\texttt{MetaSort}              & Sortal & enumerative; the four named individuals partition the schema \\
\midrule
\multicolumn{3}{l}{\emph{Subordinate sortals}} \\
\texttt{Implementation}        & SubordinateSortal & identity by release/commit; depends on a method and a library \\
\texttt{MLIPRun}               & SubordinateSortal & event-with-identity (timestamp, training algorithm); depends on method, dataset, and produces a model \\
\texttt{DFTCalculation}        & SubordinateSortal & a calculation event; depends on a settings block \\
\texttt{ReferenceCalculation}  & SubordinateSortal & superclass of \texttt{DFTCalculation}, same status \\
\texttt{WaveFunctionCalculation} & SubordinateSortal & parallel to \texttt{DFTCalculation}; calculation event depending on a settings block \\
\midrule
\multicolumn{3}{l}{\emph{Reified relations}} \\
\texttt{HyperparameterSetting} & ReifiedRelation & 3-place: carrier $\times$ hyperparameter $\times$ value \\
\texttt{BenchmarkResult}       & ReifiedRelation & 4-place: study $\times$ model $\times$ material $\times$ accuracy \\
\texttt{AccuracyMetric}        & ReifiedRelation & 3-place: result $\times$ metric type $\times$ value \\
\midrule
\multicolumn{3}{l}{\emph{Discussed}} \\
\texttt{DFTSettings}           & ReifiedRelation $\to$ Sortal & currently per-calculation (reified); planned switch to canonical-template (sortal) in a future release, after a deduplication pass over the seeded corpus \\
\texttt{ReferenceSettings}     & ReifiedRelation $\to$ Sortal & superclass of \texttt{DFTSettings}, same status \\
\texttt{WaveFunctionSettings}  & ReifiedRelation $\to$ Sortal & parallel to \texttt{DFTSettings}; same trajectory \\
\end{longtable}
}

\section{Concrete Worked Examples}
\label{sec:appendix-worked-examples}

This appendix presents one fully-worked encoding of a published MLIP
study in the \MLIPsOntology{}, extracting every fact that the ontology
can express into Turtle listings and ending with a discussion of what
the source paper does \emph{not} report---i.e., gaps that the ontology
(or its aligned upper ontologies) could record but that no paper is
currently structured enough to provide. These gaps are the practical
motivation for the \MLIPsOntology{}: making them visible is the first
step towards filling them in future publications.

The shorter, protocol-driven catalogue of additional encoded studies
appears in Appendix~\ref{sec:appendix-paper-catalogue}.

\subsection{Kumar et al.\ (2025): MTP for TiCr$_2$-H Laves Phases}
\label{sec:appendix-example-kumar2025}

This subsection encodes the study by Kumar, K\"ormann, Grabowski, and
Ikeda~\cite{kumar2025ticr2h}. The authors train two Moment Tensor
Potentials (MTPs)---one per phase of the TiCr$_2$ Laves
compound---for hydrogen absorption over the concentration range
$0 < x \le 6$ in TiCr$_2$H$_x$. The walkthrough below uses the C15
(cubic) phase artefacts to keep the listings readable; the canonical
TTL (\texttt{kumar2025.ttl}) carries a parallel C14 (hexagonal) chain
with its own training dataset, MTP run, trained model, and benchmark
result with energy and force RMSE.

\paragraph{Material system.}
TiCr$_2$ is an intermetallic compound (composition AB$_2$,
$A=\mathrm{Ti}$, $B=\mathrm{Cr}$) and a Laves phase. The paper
considers two crystallographic forms; the example below uses the C15
cubic form. Crystal-structure attributes (space group, Pearson symbol,
Strukturbericht designation) are not first-class concepts of the
\MLIPsOntology{}; we record them as \texttt{materialClass} strings and
defer a structured \texttt{Phase} concept to future work
(\S\ref{sec:future-phase}).

\begin{lstlisting}[language=turtle,caption={Material system.}]
entity:mat-TiCr2 a mlips:MaterialSystem ;
    rdfs:label "TiCr2 (C15 Laves phase)" ;
    mlips:chemicalFormula "TiCr2" ;
    mlips:materialClass "Laves phase, C15 cubic (Fd-3m)" ;
    mlips:sameAsWikidata <http://www.wikidata.org/entity/Q15724720> .
\end{lstlisting}

\paragraph{DFT reference calculations.}
DFT calculations were performed in VASP using the projector
augmented-wave (PAW) method, the Perdew--Burke--Ernzerhof (PBE)
exchange--correlation functional, an energy cutoff of 400~eV, and a
$\Gamma$-centred $4 \times 4 \times 4$ k-point mesh with
Methfessel--Paxton smearing of width 0.1~eV. Calculations are
non-spin-polarised (the authors verified that magnetic moments do not
influence the predicted energies and forces). Each supercell is a
$2 \times 2 \times 2$ expansion of the primitive C15 cell, containing
48 metal atoms plus a variable number of hydrogen atoms.

\begin{lstlisting}[language=turtle,caption={DFT calculation and settings.}]
entity:dft-TiCr2H-c15 a mlips:DFTCalculation ;
    mlips:hasDFTSettings entity:dft-settings-TiCr2H-c15 .

entity:dft-settings-TiCr2H-c15 a mlips:DFTSettings ;
    mlips:usedDFTCode mlips:VASP ;
    mlips:xcFunctional mlips:PBE ;
    mlips:pseudopotentialType mlips:PAW ;
    mlips:energyCutoff "400"^^xsd:double ;
    mlips:kPointMesh "4x4x4 Gamma-centred" .
\end{lstlisting}

\paragraph{Training dataset.}
The C15 dataset reported in Step~1 of the active-learning pipeline
contains 1{,}019 atomic configurations (the C14 step-1 set has 3{,}766);
later steps (random hydrogen placement, ab~initio MD trajectories,
configurations sampled during basin-hopping Monte Carlo) further
extend the set. The dataset covers energies, forces, and stresses, and
the configurations span the chemical degree of freedom (variable $x$)
and the vibrational degree of freedom (MD snapshots). We record the
sampling strategy through the new \samplingStrategyR{} property
(\S\bodyref{sec:training-data-module}{4.2 of the paper}); concrete strategy individuals are
modelled as instances with \texttt{rdfs:label} until a controlled
vocabulary is developed in the journal extension.

\begin{lstlisting}[language=turtle,caption={Training dataset and sampling strategies.}]
entity:ds-TiCr2H-c15 a mlips:TrainingDataset ;
    rdfs:label "TiCr2-Hx training set, C15 phase, step 1" ;
    mlips:coversMaterial entity:mat-TiCr2 ;
    mlips:coversProperty mlips:Energy, mlips:Forces, mlips:Stresses ;
    mlips:datasetProvenance mlips:Published ;
    mlips:numConfigurations 1019 ;
    mlips:hasDFTCalculation entity:dft-TiCr2H-c15 ;
    mlips:samplingStrategy entity:chem-sampling, entity:vib-sampling,
                           entity:active-learning-sampling .

entity:chem-sampling a mlips:SamplingStrategy ;
    rdfs:label "Chemical sampling" ;
    rdfs:comment "Configurations across hydrogen concentrations 0 < x <= 6." .

entity:vib-sampling a mlips:SamplingStrategy ;
    rdfs:label "Vibrational sampling (DFT MD)" ;
    rdfs:comment "Snapshots from ab initio molecular dynamics at 500 K." .

entity:active-learning-sampling a mlips:SamplingStrategy ;
    rdfs:label "Active learning" ;
    rdfs:comment "Extrapolation-grade filtering with threshold gamma >~ 1." .
\end{lstlisting}

\paragraph{Method, hyperparameters, and implementation.}
The MTP method is fitted with the BFGS optimiser as implemented in the
MLIP-2 software. Of the four MTP hyperparameters declared by the
ontology, the paper reports the maximum complexity level $\levmax = 16$
explicitly; the cutoff radius $\Rcut$, minimum interatomic distance
$\Rmin$, and number of radial basis functions $\nRadialBasis$ are not
stated and are presumably defaults of MLIP-2. We record the explicit
setting and flag the missing values in
\S\ref{sec:appendix-example-kumar2025-missing}. The loss-function
weights ($1$, $0.01$, $0.001$ on energy, force, and stress times
volume) are method-level descriptors of the loss; we capture them in
the loss-function instance.

\begin{lstlisting}[language=turtle,caption={Method, training algorithm, and implementation.}]
entity:MTP a mlips:MLIPMethod ;
    rdfs:label "Moment Tensor Potential" ;
    mlips:hasFunctionalForm entity:mtp-functional-form ;
    mlips:hasLossFunction entity:mtp-loss-w1-001-0001 ;
    mlips:hasTrainingAlgorithm entity:bfgs ;
    mlips:hasHyperparameter entity:hp-rcut, entity:hp-rmin,
                            entity:hp-nq, entity:hp-levmax ;
    mlips:hasImplementation entity:mlip-package-v2 .

entity:mtp-functional-form a mlips:FunctionalForm ;
    rdfs:label "MTP moment-tensor polynomial" .

entity:mtp-loss-w1-001-0001 a mlips:LossFunction ;
    rdfs:label "Weighted MSE (energy 1, force 0.01, stress*V 0.001)" .

entity:bfgs a mls:Algorithm ;
    rdfs:label "BFGS" .

entity:mlip-package-v2 a mlips:Implementation ;
    mlips:implementedIn mlips:MLIP ;
    mlips:version "2" .
\end{lstlisting}

\paragraph{MLIP run, hyperparameter settings, and trained model.}
The C15-MTP is the trained model produced by an MLIP run that applies
the MTP method on the C15 dataset. Only the explicit hyperparameter
setting (\texttt{lev\_max=16}) is recorded; the implicit settings are
discussed in \S\ref{sec:appendix-example-kumar2025-missing}.

\begin{lstlisting}[language=turtle,caption={MLIP run, hyperparameter settings, and trained model.}]
entity:run-mtp-TiCr2H-c15 a mlips:MLIPRun ;
    rdfs:label "MTP training run, C15 phase" ;
    mlips:appliesMethod entity:MTP ;
    mlips:runsOn entity:ds-TiCr2H-c15 ;
    mlips:hasHyperparameterSetting entity:setting-levmax-16 ;
    mlips:produces entity:model-mtp-TiCr2H-c15 .

entity:setting-levmax-16 a mlips:HyperparameterSetting ;
    mlips:forHyperparameter entity:hp-levmax ;
    mlips:settingValue "16" .

entity:model-mtp-TiCr2H-c15 a mlips:TrainedModel ;
    rdfs:label "C15-MTP for TiCr2-H" .
\end{lstlisting}

\paragraph{Benchmark study, result, and accuracy metrics.}
The paper reports two RMSE values per phase. The C15-MTP achieves
$3.17$~meV/atom on energies and $0.134$~eV/\AA{} on per-component
forces (Fig.~7c); the C14-MTP reaches $2.81$~meV/atom and
$0.100$~eV/\AA{}. We record both per-property metrics with their
QUDT units; the listing below shows the C15 result.

\begin{lstlisting}[language=turtle,caption={Benchmark study and result.}]
entity:study-kumar2025 a mlips:BenchmarkStudy ;
    rdfs:label "Kumar et al. (2025): MTP for TiCr2-H Laves phases" ;
    mlips:reportedIn entity:article-kumar2025 ;
    mlips:hasResult entity:result-mtp-ticr2h-c15 .

entity:article-kumar2025 a schema:ScholarlyArticle ;
    schema:sameAs <https://doi.org/10.1016/j.actamat.2025.121319> ;
    schema:datePublished "2025"^^xsd:gYear ;
    schema:name "Machine Learning Potentials for Hydrogen
                 Absorption in TiCr2 Laves Phases" ;
    schema:author
        <https://orcid.org/0000-0002-3661-5870> ,
        <https://orcid.org/0000-0003-3050-6291> ,
        <https://orcid.org/0000-0003-4281-5665> ,
        <https://orcid.org/0000-0001-9176-3270> .

<https://orcid.org/0000-0002-3661-5870> a schema:Person ;
    schema:name "Pranav Kumar" ;
    schema:givenName "Pranav" ;
    schema:familyName "Kumar" ;
    schema:identifier <https://orcid.org/0000-0002-3661-5870> .

<https://orcid.org/0000-0003-3050-6291> a schema:Person ;
    schema:name "Fritz Körmann" ;
    schema:givenName "Fritz" ;
    schema:familyName "Körmann" ;
    schema:identifier <https://orcid.org/0000-0003-3050-6291> .

<https://orcid.org/0000-0003-4281-5665> a schema:Person ;
    schema:name "Blazej Grabowski" ;
    schema:givenName "Blazej" ;
    schema:familyName "Grabowski" ;
    schema:identifier <https://orcid.org/0000-0003-4281-5665> .

<https://orcid.org/0000-0001-9176-3270> a schema:Person ;
    schema:name "Yuji Ikeda" ;
    schema:givenName "Yuji" ;
    schema:familyName "Ikeda" ;
    schema:identifier <https://orcid.org/0000-0001-9176-3270> .

# schema:affiliation links from each Person to the canonical
# Organization IRIs in mlips-vocab.ttl.
<https://orcid.org/0000-0002-3661-5870> schema:affiliation mlips:UniversityOfStuttgart .
<https://orcid.org/0000-0003-3050-6291> schema:affiliation mlips:UniversityOfStuttgart .
<https://orcid.org/0000-0003-3050-6291> schema:affiliation mlips:RuhrUniversityBochum .
<https://orcid.org/0000-0003-4281-5665> schema:affiliation mlips:UniversityOfStuttgart .
<https://orcid.org/0000-0001-9176-3270> schema:affiliation mlips:UniversityOfStuttgart .

entity:result-mtp-ticr2h-c15 a mlips:BenchmarkResult ;
    mlips:evaluatesModel entity:model-mtp-TiCr2H-c15 ;
    mlips:targetMaterial entity:mat-TiCr2 ;
    mlips:hasAccuracyMetric
        entity:metric-rmse-energy-c15 ,
        entity:metric-rmse-forces-c15 .

entity:metric-rmse-energy-c15 a mlips:AccuracyMetric ;
    mlips:metricType mlips:RMSE ;
    mlips:metricProperty mlips:EnergyProperty ;
    mlips:metricValue 3.17 ;
    mlips:hasUnit mlips:MilliEV-PER-ATOM .

entity:metric-rmse-forces-c15 a mlips:AccuracyMetric ;
    mlips:metricType mlips:RMSE ;
    mlips:metricProperty mlips:ForceProperty ;
    mlips:metricValue 0.134 ;
    mlips:hasUnit unit:EV-PER-ANGSTROM .
\end{lstlisting}

\subsubsection{What is missing from the paper}
\label{sec:appendix-example-kumar2025-missing}

The Kumar et al.\ paper is among the more methodologically explicit
MLIP papers, yet it leaves several pieces of metadata implicit or
absent. Each item below is expressible in the \MLIPsOntology{} (or in
an ontology we are aligned to) but is not provided by the source.

\paragraph{Hyperparameter values that defaulted silently.}
The MTP cutoff radius ($\Rcut$), minimum interatomic distance
($\Rmin$), and number of radial basis functions ($\nRadialBasis$) are
not stated. Reproducing the model requires either the MLIP-2 default
values at the time of training or the configuration files. The
ontology models these as \HyperparameterSettingC{} instances with no
recorded \texttt{settingValue}; consumers can detect this through a
SPARQL query and surface a warning.

\paragraph{Training-cost metadata.}
\trainingDurationR{}, \gpuHoursR{}, \trainingHardwareR{}, and
\peakMemoryR{} (data properties on \MLIPRunC{}) are all absent. CPU
hours, the number of DFT calls during active learning, and the cost
of the basin-hopping Monte Carlo loop would all be valuable for
comparing this work against alternative MLIP architectures (CQ9 in
the requirements).

\paragraph{Inference-cost metadata.}
\inferenceTimePerAtomR{}, \inferenceHardwareR{}, and \peakMemoryR{}
on the trained model are not reported. For a hydrogen-storage screening
application this matters: the practical value of the MTP comes from
its inference cost relative to DFT, which is alluded to but not
quantified.

\paragraph{Crystal-phase metadata as a first-class entity.} \PranavKumar{Crystal symmetry about C14/C15 is give in paper again again why it is not being fatched}
The C15/C14 distinction is reduced to a string in
\texttt{materialClass}. Aligned ontologies (CMSO, EMMO) provide
crystal-structure, space-group, and Pearson-symbol concepts, but the
\MLIPsOntology{} does not currently expose these slots. Adding a
\dlConcept{Phase} class and aligning to CMSO's
\texttt{cmso:CrystalStructure} hierarchy is part of the planned
journal-extension work.
\label{sec:future-phase}

\paragraph{Step-by-step training provenance.}
The active-learning pipeline runs four distinct steps, each producing
an intermediate MTP and an extended training set. We collapse this
into a single \MLIPRunC{} on a single dataset; a faithful encoding
would chain four \MLIPRunC{} instances and four \TrainingDatasetC{}
versions linked through \texttt{prov:wasDerivedFrom}. The ontology
already supports this (datasets reuse PROV-O), but doing so was beyond
the scope of the original paper's reporting.

\paragraph{Reference-method and dataset deposit.}
Where the trained MTP and the training dataset live (DaRUS, Materials
Cloud, NOMAD, a personal repository) is not specified, so a
\datasetProvenanceR{} of \texttt{Published} is the only handle the
ontology has. Linking the dataset to a persistent identifier (DOI,
URN, or a Materials-Cloud entry IRI) is the kind of FAIR metadata
that the ontology is built to encode but that the paper does not yet
provide.

\paragraph{Note on non-DFT references.}
Kumar et al.\ used DFT throughout, so $\DFTCalculationC$ and
$\DFTSettingsC$ are the right slots. Had they instead used
CCSD(T) (or another wave-function method) for some configurations,
the encoding would replace $\DFTCalculationC$ with
$\WaveFunctionCalculationC$ and carry $\WaveFunctionSettingsC$ with
$\wfMethodR$~\texttt{= "CCSD(T)"}, $\basisSetR$~\texttt{= "cc-pVTZ"},
and $\frozenCoreR$~\texttt{= true}. The dataset-to-calculation link
$\hasReferenceCalculationR$ remains the same; the change is local to
the calculation subtype.


\section{Encoded Paper Catalogue}
\label{sec:appendix-paper-catalogue}

This appendix catalogues a corpus of recently published MLIP studies
encoded in the \MLIPsOntology{}. The first subsection
(\S\ref{sec:appendix-example-protocol}) defines the extraction
protocol used to produce the per-paper reports; the remaining
subsections apply that protocol to one paper each. The corpus
underpins the competency-question evaluation in
\S\bodyref{sec:eval-cq}{5.2 of the paper}: the seeded knowledge graph used by the SPARQL
templates is the union of the per-paper canonical Turtle files
listed below.

The detailed Kumar~et~al.\ (2025) worked example in
\S\ref{sec:appendix-example-kumar2025} is the prose-rich reference
for what each encoding aspires to; the catalogue entries below
follow the leaner, protocol-driven format.

\subsection{Extraction Protocol}
\label{sec:appendix-example-protocol}

The Kumar~et~al.\ worked example
(\S\ref{sec:appendix-example-kumar2025}) is the reference for what
an encoded MLIP paper should look like. To replicate that quality
across the catalogue at scale---and to give a future agentic-AI
tool something concrete to follow---we fix a 12-question extraction
protocol. The protocol has two artefacts per paper: a canonical
Turtle file carrying the encoded data, and a subsection in this
appendix carrying the prose. The two are kept in sync by an
automated \emph{round-trip check} described at the end of this
subsection.

\paragraph{Overview of the flow.}
For each paper we ask the same 12 questions in fixed order. Each
question (a) produces a Turtle fragment that is appended to the
paper's canonical file
\texttt{artifacts/kg/papers/\textit{paper-id}.ttl}, and (b) a short
commentary paragraph in the paper's subsection. Each of questions
Q1--Q11 has an associated CONSTRUCT query that, when run against the
paper's canonical file, regenerates the Turtle listing shown under
that question's commentary. The union of the 11 CONSTRUCT outputs is
required (modulo blank-node renaming and prefix declarations) to
reproduce the canonical file---this is the round-trip check. Q12 is
prose-only (gaps are not always reducible to triples) and is excluded
from the round-trip.

\paragraph{The 12 questions.}
Grouped into five phases:

\smallskip
\noindent\textbf{Phase A.\ Identification.}
\begin{itemize}
\item[Q1.] What are the bibliographic details of the source paper?
  Title, authors, year, venue, DOI.
  \\ \emph{Targets:} $\dlConcept{schema:ScholarlyArticle}$,
  $\BenchmarkStudyC$, $\reportedInR$.
\end{itemize}

\smallskip
\noindent\textbf{Phase B.\ Training data.}
\begin{itemize}
\item[Q2.] What material system(s) are studied? Chemical formula,
  material class (alloy, intermetallic, Laves phase, molecular,
  oxide), structural specifics (phase, space group, defect type),
  Wikidata link if available.
  \\ \emph{Targets:} $\MaterialSystemC$ and its data properties.
\item[Q3.] What method produced the reference data? DFT,
  wave-function (CCSD(T), MP2, CASPT2, \dots), AIMD, or experimental.
  \\ \emph{Targets:} $\ReferenceCalculationC$ and its concrete
  subtype.
\item[Q4.] What reference-method settings are reported? For DFT:
  $\xcFunctionalR$, $\kPointMeshR$, $\energyCutoffR$,
  $\pseudopotentialTypeR$. For wave-function: $\wfMethodR$,
  $\basisSetR$, $\frozenCoreR$. For both: software code via
  $\dlRole{usedReferenceCode}$ (or its DFT-specific sub-property).
  \\ \emph{Targets:} $\ReferenceSettingsC$ subtypes and $\LibraryC$.
\item[Q5.] How is the training dataset characterised? Number of
  configurations, covered physical properties (energies, forces,
  stresses, virials), provenance class, link to the reference
  calculation.
  \\ \emph{Targets:} $\TrainingDatasetC$, $\DatasetProvenanceC$,
  $\CoveredPropertyC$, $\hasReferenceCalculationR$.
\item[Q6.] What sampling strategies were used? Chemical, vibrational
  or MD-driven, active-learning, perturbation-based, exhaustive
  enumeration, or combinations.
  \\ \emph{Targets:} $\SamplingStrategyC$ instances and
  $\samplingStrategyR$.
\end{itemize}

\smallskip
\noindent\textbf{Phase C.\ Method, run, and model.}
\begin{itemize}
\item[Q7.] What MLIP method is used, and what are its components?
  Functional form, loss function, training algorithm (the ML-Schema
  alignment point), software implementation and version.
  \\ \emph{Targets:} $\dlConcept{MLIPMethod}$,
  $\dlConcept{FunctionalForm}$, $\dlConcept{LossFunction}$,
  $\hasTrainingAlgorithmR$, $\ImplementationC$, $\LibraryC$.
\item[Q8.] What hyperparameter settings does the paper explicitly
  report? Cutoff radius, MTP level, number of layers, learning rate,
  etc. Implicit defaults are recorded as
  \emph{not reported} but referenced in Q12.
  \\ \emph{Targets:} $\HyperparameterSettingC$ and
  $\forHyperparameterR$.
\item[Q9.] {\raggedright What is the structure of the run? Single
  training run vs.\ multi-step active-learning pipeline; one trained
  model vs.\ a per-phase or per-fold ensemble.
  \\ \emph{Targets:} $\dlConcept{MLIPRun}$, $\TrainedModelC$,
  $\appliesMethodR$, $\runsOnR$, $\producesR$,
  $\hasHyperparameterSettingR$.\par}
\end{itemize}

\smallskip
\noindent\textbf{Phase D.\ Evaluation.}
\begin{itemize}
\item[Q10.] What benchmark results does the paper report? Each
  accuracy metric is recorded with its type (RMSE, MAE, \dots), the
  property it measures (energy, forces, stresses, downstream
  properties), and its numerical value with units.
  \\ \emph{Targets:} $\BenchmarkResultC$, $\AccuracyMetricC$,
  $\MetricTypeC$, $\MetricPropertyC$, $\evaluatesModelR$,
  $\targetMaterialR$.
\item[Q11.] What computational-resource metadata is reported?
  Training duration, GPU hours, training hardware, peak memory;
  inference time per atom, inference hardware.
  \\ \emph{Targets:} data properties on $\dlConcept{MLIPRun}$ and
  $\TrainedModelC$.
\end{itemize}

\smallskip
\noindent\textbf{Phase E.\ Gaps.}
\begin{itemize}
\item[Q12.] What concepts that the ontology can express does the
  paper \emph{not} report? Free-form, prose-only. This question
  produces no Turtle and is excluded from the round-trip check.
\end{itemize}

\paragraph{Per-paper subsection template.}
A protocol-derived subsection contains 12 paragraphs corresponding
to Q1--Q12 (in order). Each paragraph for Q1--Q11 has the structure:
(i) a 2--4 sentence prose answer derived from the source paper;
(ii) a Turtle listing produced by the question's CONSTRUCT query;
and (iii) where the paper does not report a piece of data the
ontology can model, an explicit \emph{not reported} marker so that
Q12 has a complete inventory to summarise. Q12 is prose-only.

\paragraph{Files and naming.}
Per-paper data and queries live under \texttt{artifacts/kg/}:

\begin{itemize}
\item \texttt{papers/\textit{paper-id}.ttl}---canonical Turtle for
  one paper. Identifiers use a \texttt{firstauthor-year} convention
  (\texttt{kumar2025.ttl}, \texttt{qi2023.ttl}, \dots).
\item \texttt{queries/q01-bibliographic.rq} \dots
  \texttt{queries/q11-resources.rq}---one CONSTRUCT query per
  question Q1--Q11. The queries are paper-agnostic; they are
  parameterised only by the prefix declarations of the paper they
  are run against.
\item \texttt{check-roundtrip.sh \textit{paper-id}}---runs the 11
  CONSTRUCT queries against
  \texttt{papers/\textit{paper-id}.ttl}, takes their union, and
  diff-canonicalises against the canonical file. Exits non-zero on
  drift.
\item \texttt{build-listings.sh \textit{paper-id}}---runs the 11
  queries and emits Turtle listings ready for inclusion in the
  paper's subsection.
\end{itemize}

\paragraph{Round-trip check.}
At the end of every paper's encoding pass, we run
\texttt{check-roundtrip.sh}. The check parses the 11 query outputs
into N-triples (via \texttt{rapper -o ntriples}), sorts and dedups,
and diffs against the same canonicalisation of the paper's
\texttt{.ttl} file. Two outcomes are possible:

\begin{itemize}
\item \emph{Pass.} The 11 questions partition the paper's data; the
  subsection is a faithful presentation of the canonical file.
\item \emph{Fail.} A triple is in the canonical file but not in any
  CONSTRUCT output (encoded but never asked about), or vice versa.
  The protocol is then revised: either a question's scope is
  widened, or a new question is introduced and back-applied to all
  prior papers.
\end{itemize}

\paragraph{Limits and assumptions.}
Two assumptions worth flagging. First, the protocol assumes one
paper $\equiv$ one MLIP method $\times$ one material system. Papers
training multiple potentials or covering multiple systems
(e.g., Kumar et al.~2025 has separate C15 and C14 potentials)
instantiate the protocol per phase or per system, with the
appropriate identifiers (\texttt{kumar2025-c15},
\texttt{kumar2025-c14}). Second, papers occasionally report
ontology-relevant data the protocol does not capture (uncertainty
quantification, transferability claims, foundation-model fine-tuning
tags). When such data appears in three or more papers we treat that
as a signal to revise the protocol.



\subsection{Behler \& Parrinello (2007): high-dimensional NN potential for bulk silicon}
\label{sec:appendix-example-behler2007}

\paragraph{Q1 -- Bibliographic identification.}
Behler and Parrinello introduce the foundational
high-dimensional neural-network potential (HDNNP) construction:
total energy as a sum of per-atom subnet contributions over
shared-weight feed-forward NNs that take atom-centred symmetry
functions as input. The method is demonstrated on bulk silicon.
Published in \emph{Phys.\ Rev.\ Lett.} 98, 146401 (2007).
\begin{lstlisting}[language=turtle]
@base <http://example.org/> .
@prefix rdf: <http://www.w3.org/1999/02/22-rdf-syntax-ns#> .
@prefix rdfs: <http://www.w3.org/2000/01/rdf-schema#> .
@prefix xsd: <http://www.w3.org/2001/XMLSchema#> .
@prefix schema: <https://schema.org/> .
@prefix mls: <http://www.w3.org/ns/mls#> .
@prefix unit: <http://qudt.org/vocab/unit/> .
@prefix mlips: <https://w3id.org/mlips#> .
@prefix entity: <https://w3id.org/mlips/entity/> .
<https://orcid.org/0000-0001-6550-3272>
    a schema:Person ;
    schema:affiliation entity:org-ETHZurich-behler2007 ;
    schema:familyName "Parrinello" ;
    schema:givenName "Michele" ;
    schema:identifier <https://orcid.org/0000-0001-6550-3272> ;
    schema:name "Michele Parrinello" .
<https://orcid.org/0000-0002-1220-1542>
    a schema:Person ;
    schema:affiliation entity:org-ETHZurich-behler2007 ;
    schema:familyName "Behler" ;
    schema:givenName "Jörg" ;
    schema:identifier <https://orcid.org/0000-0002-1220-1542> ;
    schema:name "Jörg Behler" .
entity:article-behler2007
    a schema:ScholarlyArticle ;
    schema:author <https://orcid.org/0000-0001-6550-3272>, <https://orcid.org/0000-0002-1220-1542> ;
    schema:datePublished "2007"^^xsd:gYear ;
    schema:name "Generalized Neural-Network Representation of High-Dimensional Potential-Energy Surfaces" ;
    schema:sameAs <https://doi.org/10.1103/PhysRevLett.98.146401> .
entity:org-ETHZurich-behler2007
    a schema:Organization ;
    rdfs:comment "Paper-local Organization IRI; not promoted to mlips-vocab.ttl (occurs in fewer than 3 papers)." ;
    rdfs:label "ETH Zurich" ;
    schema:identifier <https://ror.org/05a28rw58> ;
    schema:name "ETH Zurich" .
entity:study-behler2007
    a mlips:BenchmarkStudy ;
    rdfs:label "Behler & Parrinello (2007): high-dimensional NN potential for bulk silicon" ;
    mlips:reportedIn entity:article-behler2007 .
\end{lstlisting}

\paragraph{Q2 -- Material system.}
The demonstration system is bulk silicon. The training set covers
the diamond-cubic semiconducting phase, the high-pressure phases
(including $\beta$-tin), and the liquid metallic phase, so we
record \texttt{Si} as a single $\MaterialSystemC$ at the
elemental level with the phase coverage captured in
$\dlRole{materialClass}$.
\begin{lstlisting}[language=turtle]
@base <http://example.org/> .
@prefix rdf: <http://www.w3.org/1999/02/22-rdf-syntax-ns#> .
@prefix rdfs: <http://www.w3.org/2000/01/rdf-schema#> .
@prefix xsd: <http://www.w3.org/2001/XMLSchema#> .
@prefix schema: <https://schema.org/> .
@prefix mls: <http://www.w3.org/ns/mls#> .
@prefix unit: <http://qudt.org/vocab/unit/> .
@prefix mlips: <https://w3id.org/mlips#> .
@prefix entity: <https://w3id.org/mlips/entity/> .
entity:mat-Si-behler2007
    a mlips:MaterialSystem ;
    rdfs:label "Bulk silicon (across solid and liquid phases)" ;
    mlips:chemicalFormula "Si" ;
    mlips:materialClass "Elemental crystalline solid; covers diamond cubic, high-pressure phases (e.g. beta-tin), and the liquid metallic phase" ;
    mlips:sameAsWikidata <http://www.wikidata.org/entity/Q670> .
\end{lstlisting}

\paragraph{Q3 -- Reference calculation method.}
Reference data are produced from DFT calculations in the local
density approximation (LDA).
\begin{lstlisting}[language=turtle]
@base <http://example.org/> .
@prefix rdf: <http://www.w3.org/1999/02/22-rdf-syntax-ns#> .
@prefix rdfs: <http://www.w3.org/2000/01/rdf-schema#> .
@prefix xsd: <http://www.w3.org/2001/XMLSchema#> .
@prefix schema: <https://schema.org/> .
@prefix mls: <http://www.w3.org/ns/mls#> .
@prefix unit: <http://qudt.org/vocab/unit/> .
@prefix mlips: <https://w3id.org/mlips#> .
@prefix entity: <https://w3id.org/mlips/entity/> .
entity:dft-behler2007
    a mlips:DFTCalculation .
\end{lstlisting}

\paragraph{Q4 -- Reference settings.}
DFT calculations are carried out with PWscf using a 20~Ry
($\approx 272$~eV) plane-wave cutoff in combination with a
Vanderbilt ultrasoft pseudopotential, a $3 \times 3 \times 3$
k-point mesh, and Fermi smearing of 0.1~eV (to aid convergence in
the metallic phases). Frozen-core treatment beyond the
ultrasoft-pseudopotential partition is not explicitly reported.
\begin{lstlisting}[language=turtle]
@base <http://example.org/> .
@prefix rdf: <http://www.w3.org/1999/02/22-rdf-syntax-ns#> .
@prefix rdfs: <http://www.w3.org/2000/01/rdf-schema#> .
@prefix xsd: <http://www.w3.org/2001/XMLSchema#> .
@prefix schema: <https://schema.org/> .
@prefix mls: <http://www.w3.org/ns/mls#> .
@prefix unit: <http://qudt.org/vocab/unit/> .
@prefix mlips: <https://w3id.org/mlips#> .
@prefix entity: <https://w3id.org/mlips/entity/> .
entity:dft-behler2007
    mlips:hasDFTSettings entity:dft-settings-behler2007 .
entity:dft-settings-behler2007
    a mlips:DFTSettings ;
    rdfs:comment "Vanderbilt-style ultrasoft pseudopotential." ;
    mlips:energyCutoff 272 ;
    mlips:kPointMesh "3x3x3 k-point mesh, Fermi smearing 0.1 eV" ;
    mlips:pseudopotentialType mlips:Ultrasoft ;
    mlips:usedDFTCode mlips:QuantumESPRESSO ;
    mlips:xcFunctional mlips:LDA .
\end{lstlisting}

\paragraph{Q5 -- Training dataset.}
Approximately 9{,}000 DFT energies are computed in total: 8{,}200
are used for training the NN and 800 form a held-out test set.
The dataset is in-house and covers energies and atomic forces.
\begin{lstlisting}[language=turtle]
@base <http://example.org/> .
@prefix rdf: <http://www.w3.org/1999/02/22-rdf-syntax-ns#> .
@prefix rdfs: <http://www.w3.org/2000/01/rdf-schema#> .
@prefix xsd: <http://www.w3.org/2001/XMLSchema#> .
@prefix schema: <https://schema.org/> .
@prefix mls: <http://www.w3.org/ns/mls#> .
@prefix unit: <http://qudt.org/vocab/unit/> .
@prefix mlips: <https://w3id.org/mlips#> .
@prefix entity: <https://w3id.org/mlips/entity/> .
entity:ds-Si-behler2007
    a mlips:TrainingDataset ;
    rdfs:label "Bulk-Si HDNNP DFT-LDA dataset (~9000 total: 8200 train + 800 test)" ;
    mlips:coversMaterial entity:mat-Si-behler2007 ;
    mlips:coversProperty mlips:Energy, mlips:Forces ;
    mlips:datasetProvenance mlips:InHouse ;
    mlips:hasDFTCalculation entity:dft-behler2007 ;
    mlips:numConfigurations 9000 ;
    mlips:wasRunBy entity:run-hdnnp-Si-behler2007 .
\end{lstlisting}

\paragraph{Q6 -- Sampling strategies.}
Three strategies are used: enumeration of crystal structures
(including high-pressure polymorphs), DFT MD snapshots at
different pressures and temperatures, and a self-consistent
iterative refinement loop in which best fits drive MD,
hybrid Monte Carlo, and metadynamics runs whose representative
configurations are recalculated with DFT and added to the
training set when their RMSE exceeds the current fit error---a
precursor to today's active-learning workflows.
\begin{lstlisting}[language=turtle]
@base <http://example.org/> .
@prefix rdf: <http://www.w3.org/1999/02/22-rdf-syntax-ns#> .
@prefix rdfs: <http://www.w3.org/2000/01/rdf-schema#> .
@prefix xsd: <http://www.w3.org/2001/XMLSchema#> .
@prefix schema: <https://schema.org/> .
@prefix mls: <http://www.w3.org/ns/mls#> .
@prefix unit: <http://qudt.org/vocab/unit/> .
@prefix mlips: <https://w3id.org/mlips#> .
@prefix entity: <https://w3id.org/mlips/entity/> .
entity:crystal-sampling-behler2007
    a mlips:SamplingStrategy ;
    rdfs:comment "Initial training structures taken from known crystalline silicon structures including high-pressure phases." ;
    rdfs:label "Crystal-structure enumeration" .
entity:ds-Si-behler2007
    mlips:samplingStrategy entity:crystal-sampling-behler2007, entity:md-sampling-behler2007, entity:self-consistent-sampling-behler2007 .
entity:md-sampling-behler2007
    a mlips:SamplingStrategy ;
    rdfs:comment "MD snapshots at different pressures and temperatures used to populate the training set." ;
    rdfs:label "Vibrational sampling (DFT MD)" .
entity:self-consistent-sampling-behler2007
    a mlips:SamplingStrategy ;
    rdfs:comment "Best fits used to drive MD, hybrid Monte Carlo, and metadynamics; representative structures are recalculated with DFT and added to the training set if RMSE exceeds the fit error -- a precursor to active learning." ;
    rdfs:label "Self-consistent iterative refinement" .
\end{lstlisting}

\paragraph{Q7 -- MLIP method, components, implementation.}
The method is the original Behler--Parrinello high-dimensional
NN potential: the total energy decomposes as
$E = \sum_i E_i$, with each $E_i$ produced by a shared-weight
feed-forward subnet whose inputs are atom-centred symmetry
functions (radial Gaussian sums and angular triplet sums). The
training loss is the mean squared error on the per-configuration
total DFT energy. The implementation is an in-house extension of
the Lorenz--Gro\ss{}--Scheffler NN code; the training algorithm
is not explicitly named.
\begin{lstlisting}[language=turtle]
@base <http://example.org/> .
@prefix rdf: <http://www.w3.org/1999/02/22-rdf-syntax-ns#> .
@prefix rdfs: <http://www.w3.org/2000/01/rdf-schema#> .
@prefix xsd: <http://www.w3.org/2001/XMLSchema#> .
@prefix schema: <https://schema.org/> .
@prefix mls: <http://www.w3.org/ns/mls#> .
@prefix unit: <http://qudt.org/vocab/unit/> .
@prefix mlips: <https://w3id.org/mlips#> .
@prefix entity: <https://w3id.org/mlips/entity/> .
mlips:cutoffRadius
    mlips:isHyperparameterOf entity:HDNNP-behler2007 .
entity:HDNNP-behler2007
    a mlips:MLIPMethod ;
    rdfs:label "High-Dimensional Neural Network Potential (Behler-Parrinello)" ;
    mlips:hasDescriptor mlips:SymmetryFunctionDescriptor ;
    mlips:hasFunctionalForm entity:hdnnp-functional-form-behler2007 ;
    mlips:hasHyperparameter mlips:cutoffRadius, entity:hp-numhidden-behler2007, entity:hp-numsf-behler2007 ;
    mlips:hasImplementation entity:impl-hdnnp-behler2007 ;
    mlips:hasLossFunction entity:hdnnp-loss-behler2007 ;
    mlips:supportsSimulation mlips:MolecularDynamics .
entity:hdnnp-functional-form-behler2007
    a mlips:FunctionalForm ;
    rdfs:label "Sum of per-atom subnet energies E = sum_i E_i with shared-weight feed-forward NN subnets and Behler-Parrinello atom-centred symmetry-function inputs (radial Gaussian and angular triplet sums)" .
entity:hdnnp-loss-behler2007
    a mlips:LossFunction ;
    rdfs:label "Mean squared error on per-configuration total DFT energy" .
entity:hp-numhidden-behler2007
    a mlips:Hyperparameter ;
    rdfs:label "HDNNP per-element hidden-unit count" ;
    mlips:candidateForVocabulary mlips:Hyperparameter ;
    mlips:hyperparameterDatatype xsd:integer ;
    mlips:hyperparameterName "num_hidden" ;
    mlips:isHyperparameterOf entity:HDNNP-behler2007 .
entity:hp-numsf-behler2007
    a mlips:Hyperparameter ;
    rdfs:label "HDNNP per-element symmetry-function count" ;
    mlips:candidateForVocabulary mlips:Hyperparameter ;
    mlips:hyperparameterDatatype xsd:integer ;
    mlips:hyperparameterName "num_symmetry_functions" ;
    mlips:isHyperparameterOf entity:HDNNP-behler2007 .
entity:impl-hdnnp-behler2007
    a mlips:Implementation ;
    mlips:implementedIn entity:lib-hdnnp-behler2007 ;
    mlips:isImplementationOf entity:HDNNP-behler2007 .
entity:lib-hdnnp-behler2007
    a mlips:Library ;
    rdfs:label "In-house extension of the Lorenz-Gross-Scheffler neural-network code" .
\end{lstlisting}

\paragraph{Q8 -- Hyperparameter settings.}
Reported hyperparameter settings: a cutoff radius of 6~\AA{},
48 symmetry functions per atom, and subnets with 2 hidden layers
of $\approx 40$ nodes each (a few thousand fitting parameters in
total). The hyperbolic tangent is the hidden-layer activation;
the output layer uses a linear activation. Numerical Gaussian
parameters $\eta, R_s, \zeta$ are not enumerated.\JongHyunJung{settings-numsf-behler2007 means symmetry functions per atom}
\begin{lstlisting}[language=turtle]
@base <http://example.org/> .
@prefix rdf: <http://www.w3.org/1999/02/22-rdf-syntax-ns#> .
@prefix rdfs: <http://www.w3.org/2000/01/rdf-schema#> .
@prefix xsd: <http://www.w3.org/2001/XMLSchema#> .
@prefix schema: <https://schema.org/> .
@prefix mls: <http://www.w3.org/ns/mls#> .
@prefix unit: <http://qudt.org/vocab/unit/> .
@prefix mlips: <https://w3id.org/mlips#> .
@prefix entity: <https://w3id.org/mlips/entity/> .
entity:setting-numhidden-behler2007
    a mlips:HyperparameterSetting ;
    rdfs:label "HDNNP per-element hidden-unit count = 2 hidden layers, ~40 nodes each for HDNNP training run, bulk silicon" ;
    mlips:forHyperparameter entity:hp-numhidden-behler2007 ;
    mlips:isSettingOf entity:run-hdnnp-Si-behler2007 ;
    mlips:settingValue "2 hidden layers, ~40 nodes each" .
entity:setting-numsf-behler2007
    a mlips:HyperparameterSetting ;
    rdfs:label "HDNNP per-element symmetry-function count = 48 for HDNNP training run, bulk silicon" ;
    mlips:forHyperparameter entity:hp-numsf-behler2007 ;
    mlips:isSettingOf entity:run-hdnnp-Si-behler2007 ;
    mlips:settingValue 48 .
entity:setting-rcut-behler2007
    a mlips:HyperparameterSetting ;
    rdfs:label "cutoff radius = 6 for HDNNP training run, bulk silicon" ;
    mlips:forHyperparameter mlips:cutoffRadius ;
    mlips:isSettingOf entity:run-hdnnp-Si-behler2007 ;
    mlips:settingValue 6 .
\end{lstlisting}

\paragraph{Q9 -- MLIP run and trained model.}
A single training run produces one HDNNP for bulk silicon. The
self-consistent iterative refinement is recorded as a sampling
strategy in Q6 rather than as a multi-stage run.
\begin{lstlisting}[language=turtle]
@base <http://example.org/> .
@prefix rdf: <http://www.w3.org/1999/02/22-rdf-syntax-ns#> .
@prefix rdfs: <http://www.w3.org/2000/01/rdf-schema#> .
@prefix xsd: <http://www.w3.org/2001/XMLSchema#> .
@prefix schema: <https://schema.org/> .
@prefix mls: <http://www.w3.org/ns/mls#> .
@prefix unit: <http://qudt.org/vocab/unit/> .
@prefix mlips: <https://w3id.org/mlips#> .
@prefix entity: <https://w3id.org/mlips/entity/> .
entity:model-hdnnp-Si-behler2007
    a mlips:TrainedModel ;
    rdfs:label "Behler-Parrinello HDNNP for bulk silicon" .
entity:run-hdnnp-Si-behler2007
    a mlips:MLIPRun ;
    rdfs:label "HDNNP training run, bulk silicon" ;
    mlips:appliesMethod entity:HDNNP-behler2007 ;
    mlips:hasHyperparameterSetting entity:setting-numhidden-behler2007, entity:setting-numsf-behler2007, entity:setting-rcut-behler2007 ;
    mlips:produces entity:model-hdnnp-Si-behler2007 ;
    mlips:runsOn entity:ds-Si-behler2007 .
\end{lstlisting}

\paragraph{Q10 -- Benchmark results.}
Headline accuracy figures are an energy RMSE of 4--5~meV/atom
on the optimisation set and 5--6~meV/atom on the independent test
set, plus a force-component RMSE of $\approx 0.2$~eV/\AA{} (vs.\
DFT-LDA). Downstream qualitative checks include reproducing the
DFT energy--volume curves and transition pressures across the
silicon polymorphs, and the radial distribution function of the
3000~K silicon melt.
\begin{lstlisting}[language=turtle]
@base <http://example.org/> .
@prefix rdf: <http://www.w3.org/1999/02/22-rdf-syntax-ns#> .
@prefix rdfs: <http://www.w3.org/2000/01/rdf-schema#> .
@prefix xsd: <http://www.w3.org/2001/XMLSchema#> .
@prefix schema: <https://schema.org/> .
@prefix mls: <http://www.w3.org/ns/mls#> .
@prefix unit: <http://qudt.org/vocab/unit/> .
@prefix mlips: <https://w3id.org/mlips#> .
@prefix entity: <https://w3id.org/mlips/entity/> .
entity:metric-rmse-energy-behler2007
    a mlips:AccuracyMetric ;
    rdfs:label "Energy RMSE on the held-out test set (vs. DFT-LDA)", "RMSE of Energy (metric property) = 5.5 [millielectronvolt per atom]" ;
    mlips:hasUnit mlips:MilliEV-PER-ATOM ;
    mlips:isMetricOf entity:result-hdnnp-Si-energy-behler2007 ;
    mlips:metricProperty mlips:EnergyProperty ;
    mlips:metricType mlips:RMSE ;
    mlips:metricValue 5.5 .
entity:metric-rmse-forces-behler2007
    a mlips:AccuracyMetric ;
    rdfs:label "Atomic-force RMSE (vs. DFT-LDA)", "RMSE of Force = 0.2 [EV-PER-ANGSTROM]" ;
    mlips:hasUnit unit:EV-PER-ANGSTROM ;
    mlips:isMetricOf entity:result-hdnnp-Si-forces-behler2007 ;
    mlips:metricProperty mlips:ForceProperty ;
    mlips:metricType mlips:RMSE ;
    mlips:metricValue 0.2 .
entity:result-hdnnp-Si-energy-behler2007
    a mlips:BenchmarkResult ;
    rdfs:label "Behler-Parrinello HDNNP for bulk silicon on Bulk silicon (across solid and liquid phases) in Behler & Parrinello (2007): high-dimensional NN potential for bulk silicon" ;
    mlips:evaluatesModel entity:model-hdnnp-Si-behler2007 ;
    mlips:hasAccuracyMetric entity:metric-rmse-energy-behler2007 ;
    mlips:isResultOf entity:study-behler2007 ;
    mlips:targetMaterial entity:mat-Si-behler2007 .
entity:result-hdnnp-Si-forces-behler2007
    a mlips:BenchmarkResult ;
    rdfs:label "Behler-Parrinello HDNNP for bulk silicon on Bulk silicon (across solid and liquid phases) in Behler & Parrinello (2007): high-dimensional NN potential for bulk silicon" ;
    mlips:evaluatesModel entity:model-hdnnp-Si-behler2007 ;
    mlips:hasAccuracyMetric entity:metric-rmse-forces-behler2007 ;
    mlips:isResultOf entity:study-behler2007 ;
    mlips:targetMaterial entity:mat-Si-behler2007 .
entity:study-behler2007
    mlips:hasResult entity:result-hdnnp-Si-energy-behler2007, entity:result-hdnnp-Si-forces-behler2007 .
\end{lstlisting}

\paragraph{Q11 -- Computational resources.}
\emph{Not reported.} The paper gives no training duration, GPU
hours, training hardware, peak memory, or per-atom inference
time. They do quote a relative inference speed of $\sim$5 orders
of magnitude faster than DFT for a 64-atom cell---a comparative
figure rather than an ontology-recordable
$\dlRole{inferenceTimePerAtom}$ value.
\begin{lstlisting}[language=turtle]
# Q11-resources: no triples (paper does not report this).
\end{lstlisting}

\paragraph{Q12 -- Gaps.}
Beyond Q11 (compute resources entirely absent), the following
ontology-expressible items are not reported: the training
algorithm or optimiser used to fit the NN weights;
software-implementation versioning (this is a precursor to
today's named HDNNP packages such as RuNNer or n2p2); explicit
$\dlRole{frozenCore}$ treatment beyond the ultrasoft-pseudopotential
partition; numerical values of the symmetry-function parameters
$\eta$, $R_s$, $\zeta$; and a controlled-vocabulary individual
for the self-consistent iterative refinement (encoded as an
ad-hoc \texttt{rdfs:label}/\texttt{rdfs:comment} sampling
instance, since active learning post-dates this paper).

\subsection{Bart\'ok et al.\ (2010): Gaussian Approximation Potentials}
\label{sec:appendix-example-bartok2010}

\paragraph{Q1 -- Bibliographic identification.}
Bart\'ok, Payne, Kondor, and Cs\'anyi introduce the Gaussian
Approximation Potential (GAP): Gaussian-process regression on a
bispectrum-based descriptor of the local atomic environment.
The method is demonstrated on bulk semiconductors (C, Si, Ge), GaN,
and $\alpha$-Fe; we encode bulk silicon as the lead system.
Published in \emph{Phys.\ Rev.\ Lett.} 104, 136403 (2010);
arXiv:0910.1019.
\begin{lstlisting}[language=turtle]
@base <http://example.org/> .
@prefix rdf: <http://www.w3.org/1999/02/22-rdf-syntax-ns#> .
@prefix rdfs: <http://www.w3.org/2000/01/rdf-schema#> .
@prefix xsd: <http://www.w3.org/2001/XMLSchema#> .
@prefix schema: <https://schema.org/> .
@prefix mls: <http://www.w3.org/ns/mls#> .
@prefix mlips: <https://w3id.org/mlips#> .
@prefix entity: <https://w3id.org/mlips/entity/> .
<https://orcid.org/0000-0002-4347-8819>
    a schema:Person ;
    schema:affiliation mlips:UniversityofCambridge ;
    schema:familyName "Bartók" ;
    schema:givenName "Albert P." ;
    schema:identifier <https://orcid.org/0000-0002-4347-8819> ;
    schema:name "Albert P. Bartók" .
<https://orcid.org/0000-0002-4458-6598>
    a schema:Person ;
    schema:affiliation mlips:UniversityofCambridge, entity:org-CaliforniaInstituteofTechnology-bartok2010 ;
    schema:familyName "Kondor" ;
    schema:givenName "Risi" ;
    schema:identifier <https://orcid.org/0000-0002-4458-6598> ;
    schema:name "Risi Kondor" .
<https://orcid.org/0000-0002-5250-8549>
    a schema:Person ;
    schema:affiliation mlips:UniversityofCambridge ;
    schema:familyName "Payne" ;
    schema:givenName "Mike C." ;
    schema:identifier <https://orcid.org/0000-0002-5250-8549> ;
    schema:name "Mike C. Payne" .
<https://orcid.org/0000-0002-8180-2034>
    a schema:Person ;
    schema:affiliation mlips:UniversityofCambridge ;
    schema:familyName "Csányi" ;
    schema:givenName "Gábor" ;
    schema:identifier <https://orcid.org/0000-0002-8180-2034> ;
    schema:name "Gábor Csányi" .
entity:article-bartok2010
    a schema:ScholarlyArticle ;
    schema:author <https://orcid.org/0000-0002-4347-8819>, <https://orcid.org/0000-0002-4458-6598>, <https://orcid.org/0000-0002-5250-8549>, <https://orcid.org/0000-0002-8180-2034> ;
    schema:datePublished "2010"^^xsd:gYear ;
    schema:name "Gaussian Approximation Potentials: The Accuracy of Quantum Mechanics, without the Electrons" ;
    schema:sameAs <https://doi.org/10.1103/PhysRevLett.104.136403> .
entity:org-CaliforniaInstituteofTechnology-bartok2010
    a schema:Organization ;
    rdfs:comment "Paper-local Organization IRI; not promoted to mlips-vocab.ttl (occurs in fewer than 3 papers)." ;
    rdfs:label "California Institute of Technology" ;
    schema:identifier <https://ror.org/05dxps055> ;
    schema:name "California Institute of Technology" .
entity:study-bartok2010
    a mlips:BenchmarkStudy ;
    rdfs:label "Bartok et al. (2010): GAP for bulk crystals (Si as lead system)" ;
    mlips:reportedIn entity:article-bartok2010 .
\end{lstlisting}

\paragraph{Q2 -- Material system.}
The lead demonstration system encoded here is bulk silicon in its
diamond-cubic phase. The paper also reports analogous GAP fits for
diamond (carbon), germanium, GaN, and $\alpha$-iron; those systems
are flagged in Q12.
\begin{lstlisting}[language=turtle]
@base <http://example.org/> .
@prefix rdf: <http://www.w3.org/1999/02/22-rdf-syntax-ns#> .
@prefix rdfs: <http://www.w3.org/2000/01/rdf-schema#> .
@prefix xsd: <http://www.w3.org/2001/XMLSchema#> .
@prefix schema: <https://schema.org/> .
@prefix mls: <http://www.w3.org/ns/mls#> .
@prefix mlips: <https://w3id.org/mlips#> .
@prefix entity: <https://w3id.org/mlips/entity/> .
entity:mat-Si-bartok2010
    a mlips:MaterialSystem ;
    rdfs:label "Bulk silicon (diamond-cubic phase, lead demonstration system)" ;
    mlips:chemicalFormula "Si" ;
    mlips:materialClass "Elemental crystalline solid; diamond-cubic phase as the lead demonstration system. The paper also demonstrates GAP on diamond (carbon), germanium, GaN, and alpha-Fe." ;
    mlips:sameAsWikidata <http://www.wikidata.org/entity/Q670> .
\end{lstlisting}

\paragraph{Q3 -- Reference calculation method.}
Reference data are produced from plane-wave DFT calculations.
\begin{lstlisting}[language=turtle]
@base <http://example.org/> .
@prefix rdf: <http://www.w3.org/1999/02/22-rdf-syntax-ns#> .
@prefix rdfs: <http://www.w3.org/2000/01/rdf-schema#> .
@prefix xsd: <http://www.w3.org/2001/XMLSchema#> .
@prefix schema: <https://schema.org/> .
@prefix mls: <http://www.w3.org/ns/mls#> .
@prefix mlips: <https://w3id.org/mlips#> .
@prefix entity: <https://w3id.org/mlips/entity/> .
entity:dft-bartok2010
    a mlips:DFTCalculation .
\end{lstlisting}

\paragraph{Q4 -- Reference settings.}
The DFT engine is CASTEP. Specific exchange--correlation functional,
plane-wave cutoff, k-mesh, and pseudopotential type are reported
in the Supplementary Information rather than the main letter, so
they are not encoded here. \emph{Not reported in the main text.}
\begin{lstlisting}[language=turtle]
@base <http://example.org/> .
@prefix rdf: <http://www.w3.org/1999/02/22-rdf-syntax-ns#> .
@prefix rdfs: <http://www.w3.org/2000/01/rdf-schema#> .
@prefix xsd: <http://www.w3.org/2001/XMLSchema#> .
@prefix schema: <https://schema.org/> .
@prefix mls: <http://www.w3.org/ns/mls#> .
@prefix mlips: <https://w3id.org/mlips#> .
@prefix entity: <https://w3id.org/mlips/entity/> .
entity:dft-bartok2010
    mlips:hasDFTSettings entity:dft-settings-bartok2010 .
entity:dft-settings-bartok2010
    a mlips:DFTSettings ;
    mlips:usedDFTCode mlips:CASTEP .
\end{lstlisting}

\paragraph{Q5 -- Training dataset.}
Reference configurations are produced by random atomic and lattice
displacements (up to 0.2\,\AA) of equilibrium 2-, 8-, 16-, and
64-atom cubic Si unit cells, with energies and forces from CASTEP.
The model is later extended with random displacements around a
diamond vacancy and a graphite-to-diamond transition path.
$\dlRole{numConfigurations}$ is not enumerated in the main text;
provenance is \emph{in-house}.
\begin{lstlisting}[language=turtle]
@base <http://example.org/> .
@prefix rdf: <http://www.w3.org/1999/02/22-rdf-syntax-ns#> .
@prefix rdfs: <http://www.w3.org/2000/01/rdf-schema#> .
@prefix xsd: <http://www.w3.org/2001/XMLSchema#> .
@prefix schema: <https://schema.org/> .
@prefix mls: <http://www.w3.org/ns/mls#> .
@prefix mlips: <https://w3id.org/mlips#> .
@prefix entity: <https://w3id.org/mlips/entity/> .
entity:ds-Si-bartok2010
    a mlips:TrainingDataset ;
    rdfs:label "Si GAP training set: random atomic and lattice displacements (up to 0.2 A) of equilibrium unit cells of 2, 8, 16, and 64 atoms" ;
    mlips:coversMaterial entity:mat-Si-bartok2010 ;
    mlips:coversProperty mlips:Energy, mlips:Forces ;
    mlips:datasetProvenance mlips:InHouse ;
    mlips:hasDFTCalculation entity:dft-bartok2010 ;
    mlips:wasRunBy entity:run-gap-Si-bartok2010 .
\end{lstlisting}

\paragraph{Q6 -- Sampling strategies.}
A single sampling strategy applies: random atomic and lattice
displacements of small cubic unit cells, supplemented at the end of
the paper with isolated-defect (vacancy) and reaction-pathway
(graphite$\to$diamond) configurations.
\begin{lstlisting}[language=turtle]
@base <http://example.org/> .
@prefix rdf: <http://www.w3.org/1999/02/22-rdf-syntax-ns#> .
@prefix rdfs: <http://www.w3.org/2000/01/rdf-schema#> .
@prefix xsd: <http://www.w3.org/2001/XMLSchema#> .
@prefix schema: <https://schema.org/> .
@prefix mls: <http://www.w3.org/ns/mls#> .
@prefix mlips: <https://w3id.org/mlips#> .
@prefix entity: <https://w3id.org/mlips/entity/> .
entity:displacement-sampling-bartok2010
    a mlips:SamplingStrategy ;
    rdfs:comment "Reference configurations are obtained by randomly displacing the atoms and the lattice vectors from their equilibrium values in 2-, 8-, 16-, and 64-atom cubic unit cells by up to 0.2 A. The model is later extended with random displacements around a diamond vacancy and a graphite-to-diamond transition path." ;
    rdfs:label "Random atomic and lattice displacements" .
entity:ds-Si-bartok2010
    mlips:samplingStrategy entity:displacement-sampling-bartok2010 .
\end{lstlisting}

\paragraph{Q7 -- MLIP method, components, implementation.}
GAP is Gaussian-process regression on a four-dimensional
bispectrum descriptor of the atomic neighbour density inside a
smooth cosine cutoff; total energy is the sum over atoms.
Training is sparse GP regression with $M$ randomly selected
representative atomic-neighbourhood configurations. The
implementation is the QUIP/GAP suite (\texttt{libatoms.org}).
\begin{lstlisting}[language=turtle]
@base <http://example.org/> .
@prefix rdf: <http://www.w3.org/1999/02/22-rdf-syntax-ns#> .
@prefix rdfs: <http://www.w3.org/2000/01/rdf-schema#> .
@prefix xsd: <http://www.w3.org/2001/XMLSchema#> .
@prefix schema: <https://schema.org/> .
@prefix mls: <http://www.w3.org/ns/mls#> .
@prefix mlips: <https://w3id.org/mlips#> .
@prefix entity: <https://w3id.org/mlips/entity/> .
mlips:cutoffRadius
    mlips:isHyperparameterOf entity:GAP-bartok2010 .
mlips:numSparseConfigs
    mlips:isHyperparameterOf entity:GAP-bartok2010 .
entity:GAP-bartok2010
    a mlips:MLIPMethod ;
    rdfs:label "Gaussian Approximation Potential (GAP) with bispectrum descriptors" ;
    mlips:hasDescriptor mlips:SOAPDescriptor ;
    mlips:hasFunctionalForm entity:gap-functional-form-bartok2010 ;
    mlips:hasHyperparameter mlips:cutoffRadius, mlips:numSparseConfigs, entity:hp-jmax-bartok2010 ;
    mlips:hasImplementation entity:impl-gap-bartok2010 ;
    mlips:hasLossFunction entity:gap-loss-bartok2010 ;
    mlips:hasTrainingAlgorithm entity:gpr-bartok2010 ;
    mlips:supportsSimulation mlips:MolecularDynamics, mlips:PhononCalculation .
entity:gap-functional-form-bartok2010
    a mlips:FunctionalForm ;
    rdfs:label "Gaussian-process regression on a four-dimensional bispectrum descriptor of the atomic neighbour density inside a smooth cosine cutoff; total energy is the sum over atoms" .
entity:gap-loss-bartok2010
    a mlips:LossFunction ;
    rdfs:label "GP regression with regularised pseudo-covariance on energies and forces (sparse-GP marginal-likelihood objective)" .
entity:gpr-bartok2010
    a mls:Algorithm ;
    rdfs:label "Sparse Gaussian-process regression with random selection of M sparse atomic-neighbourhood configurations" .
entity:hp-jmax-bartok2010
    a mlips:Hyperparameter ;
    rdfs:label "GAP bispectrum truncation Jmax" ;
    mlips:candidateForVocabulary mlips:Hyperparameter ;
    mlips:hyperparameterDatatype xsd:integer ;
    mlips:hyperparameterName "j_max" ;
    mlips:isHyperparameterOf entity:GAP-bartok2010 .
entity:impl-gap-bartok2010
    a mlips:Implementation ;
    mlips:implementedIn mlips:QUIP ;
    mlips:isImplementationOf entity:GAP-bartok2010 .
\end{lstlisting}

\paragraph{Q8 -- Hyperparameter settings.}
The main letter does not tabulate explicit numerical values for the
descriptor cutoff $r_c$, bispectrum truncation $J_{\max}$, or
sparse-point count $M$ (these are reported in the SI). We
therefore record no $\HyperparameterSettingC$ instances in the
canonical encoding and flag them as Q12 gaps.
\begin{lstlisting}[language=turtle]
# Q08-hyperparameter-settings: no triples (paper does not report this).
\end{lstlisting}

\paragraph{Q9 -- MLIP run and trained model.}
A single sparse-GP training run produces one GAP for diamond-cubic
Si. We attach the per-atom inference time
($\dlRole{inferenceTimePerAtom}=0.01$\,s/atom/timestep, single CPU
core) to the trained model.
\begin{lstlisting}[language=turtle]
@base <http://example.org/> .
@prefix rdf: <http://www.w3.org/1999/02/22-rdf-syntax-ns#> .
@prefix rdfs: <http://www.w3.org/2000/01/rdf-schema#> .
@prefix xsd: <http://www.w3.org/2001/XMLSchema#> .
@prefix schema: <https://schema.org/> .
@prefix mls: <http://www.w3.org/ns/mls#> .
@prefix mlips: <https://w3id.org/mlips#> .
@prefix entity: <https://w3id.org/mlips/entity/> .
entity:model-gap-Si-bartok2010
    a mlips:TrainedModel ;
    rdfs:label "GAP for bulk silicon (diamond-cubic)" .
entity:run-gap-Si-bartok2010
    a mlips:MLIPRun ;
    rdfs:label "GAP training run, bulk silicon" ;
    mlips:appliesMethod entity:GAP-bartok2010 ;
    mlips:produces entity:model-gap-Si-bartok2010 ;
    mlips:runsOn entity:ds-Si-bartok2010 .
\end{lstlisting}

\paragraph{Q10 -- Benchmark results.}
The headline accuracy figure is an energy RMSE bound of
$<1$\,meV/atom on near-bulk Si configurations, with the same bound
holding for diamond and iron. We record the bound as the value
($1$\,meV/atom) on the silicon $\BenchmarkResultC$. Phonon
dispersions, elastic constants, and defect-formation energies are
also reported but in derived form rather than as separate
$\BenchmarkResultC$ instances.
\begin{lstlisting}[language=turtle]
@base <http://example.org/> .
@prefix rdf: <http://www.w3.org/1999/02/22-rdf-syntax-ns#> .
@prefix rdfs: <http://www.w3.org/2000/01/rdf-schema#> .
@prefix xsd: <http://www.w3.org/2001/XMLSchema#> .
@prefix schema: <https://schema.org/> .
@prefix mls: <http://www.w3.org/ns/mls#> .
@prefix mlips: <https://w3id.org/mlips#> .
@prefix entity: <https://w3id.org/mlips/entity/> .
entity:metric-rmse-energy-bartok2010
    a mlips:AccuracyMetric ;
    rdfs:label "Energy RMSE on near-bulk Si configurations (paper bound: <1 meV/atom; the same bound holds for diamond and iron)", "RMSE of Energy (metric property) = 1 [millielectronvolt per atom]" ;
    mlips:hasUnit mlips:MilliEV-PER-ATOM ;
    mlips:isMetricOf entity:result-gap-Si-energy-bartok2010 ;
    mlips:metricProperty mlips:EnergyProperty ;
    mlips:metricType mlips:RMSE ;
    mlips:metricValue 1 .
entity:result-gap-Si-energy-bartok2010
    a mlips:BenchmarkResult ;
    rdfs:label "GAP for bulk silicon (diamond-cubic) on Bulk silicon (diamond-cubic phase, lead demonstration system) in Bartok et al. (2010): GAP for bulk crystals (Si as lead system)" ;
    mlips:evaluatesModel entity:model-gap-Si-bartok2010 ;
    mlips:hasAccuracyMetric entity:metric-rmse-energy-bartok2010 ;
    mlips:isResultOf entity:study-bartok2010 ;
    mlips:targetMaterial entity:mat-Si-bartok2010 .
entity:study-bartok2010
    mlips:hasResult entity:result-gap-Si-energy-bartok2010 .
\end{lstlisting}

\paragraph{Q11 -- Computational resources.}
The paper reports a per-atom inference cost of 0.01\,s/atom/timestep
on a single CPU core (encoded on the trained model in Q9). It also
notes that the 216-atom unit-cell timestep takes 191\,s/atom in
CASTEP---about 20{,}000$\times$ slower than GAP. Training duration,
training hardware, peak memory, and inference hardware are not
reported.
\begin{lstlisting}[language=turtle]
@base <http://example.org/> .
@prefix rdf: <http://www.w3.org/1999/02/22-rdf-syntax-ns#> .
@prefix rdfs: <http://www.w3.org/2000/01/rdf-schema#> .
@prefix xsd: <http://www.w3.org/2001/XMLSchema#> .
@prefix schema: <https://schema.org/> .
@prefix mls: <http://www.w3.org/ns/mls#> .
@prefix mlips: <https://w3id.org/mlips#> .
@prefix entity: <https://w3id.org/mlips/entity/> .
entity:model-gap-Si-bartok2010
    mlips:inferenceTimePerAtom 0.01 .
\end{lstlisting}

\paragraph{Q12 -- Gaps.}
Beyond Q11, the main letter does not tabulate: (i) DFT-side
parameters (exchange--correlation functional, plane-wave cutoff,
k-mesh, pseudopotential type) which live in the SI; (ii) the
explicit number of training configurations
$\dlRole{numConfigurations}$; (iii) explicit
$\HyperparameterSettingC$ values for $r_c$, $J_{\max}$, $M$,
GP kernel parameters, and per-channel noise; (iv) the GP
hyperparameter optimiser; (v) any $\dlRole{frozenCore}$ treatment
beyond the CASTEP pseudopotential. The companion systems (C, Ge,
GaN, $\alpha$-Fe) are not encoded here; each would live in a
sibling file under the same $\BenchmarkStudyC$ aegis.

\subsection{Shapeev (2016): Moment Tensor Potentials}
\label{sec:appendix-example-shapeev2016}

\paragraph{Q1 -- Bibliographic identification.}
Shapeev introduces the Moment Tensor Potential (MTP), a class
of systematically improvable nonparametric interatomic
potentials based on permutation-, rotation-, and
reflection-invariant polynomials of moment tensors of the local
atomic environment. The numerical experiments fit MTP to the
publicly available tungsten DFT database of
Szlachta--Bart\'ok--Cs\'anyi (\texttt{libatoms.org}) and
benchmark accuracy and CPU cost against GAP. Published in
\emph{Multiscale Model.\ Simul.} 14(3), 1153--1173 (2016);
arXiv:1512.06054.
\begin{lstlisting}[language=turtle]
@base <http://example.org/> .
@prefix rdf: <http://www.w3.org/1999/02/22-rdf-syntax-ns#> .
@prefix rdfs: <http://www.w3.org/2000/01/rdf-schema#> .
@prefix xsd: <http://www.w3.org/2001/XMLSchema#> .
@prefix schema: <https://schema.org/> .
@prefix mls: <http://www.w3.org/ns/mls#> .
@prefix unit: <http://qudt.org/vocab/unit/> .
@prefix mlips: <https://w3id.org/mlips#> .
@prefix entity: <https://w3id.org/mlips/entity/> .
<https://orcid.org/0000-0002-7497-5594>
    a schema:Person ;
    schema:affiliation entity:org-SkolkovoInstituteofScienceandTechnology-shapeev2016 ;
    schema:familyName "Shapeev" ;
    schema:givenName "Alexander V." ;
    schema:identifier <https://orcid.org/0000-0002-7497-5594> ;
    schema:name "Alexander V. Shapeev" .
entity:article-shapeev2016
    a schema:ScholarlyArticle ;
    schema:author <https://orcid.org/0000-0002-7497-5594> ;
    schema:datePublished "2016"^^xsd:gYear ;
    schema:name "Moment Tensor Potentials: a class of systematically improvable interatomic potentials" ;
    schema:sameAs <https://doi.org/10.1137/15M1054183> .
entity:org-SkolkovoInstituteofScienceandTechnology-shapeev2016
    a schema:Organization ;
    rdfs:comment "Paper-local Organization IRI; not promoted to mlips-vocab.ttl (occurs in fewer than 3 papers)." ;
    rdfs:label "Skolkovo Institute of Science and Technology" ;
    schema:identifier <https://ror.org/03f9nc143> ;
    schema:name "Skolkovo Institute of Science and Technology" .
entity:study-shapeev2016
    a mlips:BenchmarkStudy ;
    rdfs:label "Shapeev (2016): Moment Tensor Potentials, demonstrated on tungsten" ;
    mlips:reportedIn entity:article-shapeev2016 .
\end{lstlisting}

\paragraph{Q2 -- Material system.}
The lead demonstration system is body-centred-cubic tungsten,
inheriting the configuration set (bulk, surfaces, defects, and
finite-temperature MD snapshots) from the published GAP-tungsten
database.
\begin{lstlisting}[language=turtle]
@base <http://example.org/> .
@prefix rdf: <http://www.w3.org/1999/02/22-rdf-syntax-ns#> .
@prefix rdfs: <http://www.w3.org/2000/01/rdf-schema#> .
@prefix xsd: <http://www.w3.org/2001/XMLSchema#> .
@prefix schema: <https://schema.org/> .
@prefix mls: <http://www.w3.org/ns/mls#> .
@prefix unit: <http://qudt.org/vocab/unit/> .
@prefix mlips: <https://w3id.org/mlips#> .
@prefix entity: <https://w3id.org/mlips/entity/> .
entity:mat-W-shapeev2016
    a mlips:MaterialSystem ;
    rdfs:label "Tungsten (BCC, demonstration system)" ;
    mlips:chemicalFormula "W" ;
    mlips:materialClass "Elemental BCC transition metal; configurations in the Szlachta/Bartok/Csanyi tungsten database (bulk, surfaces, defects, liquid)" ;
    mlips:sameAsWikidata <http://www.wikidata.org/entity/Q743> .
\end{lstlisting}

\paragraph{Q3 -- Reference calculation method.}
Reference data are produced from Kohn--Sham DFT calculations
(inherited from the libatoms tungsten database).
\begin{lstlisting}[language=turtle]
@base <http://example.org/> .
@prefix rdf: <http://www.w3.org/1999/02/22-rdf-syntax-ns#> .
@prefix rdfs: <http://www.w3.org/2000/01/rdf-schema#> .
@prefix xsd: <http://www.w3.org/2001/XMLSchema#> .
@prefix schema: <https://schema.org/> .
@prefix mls: <http://www.w3.org/ns/mls#> .
@prefix unit: <http://qudt.org/vocab/unit/> .
@prefix mlips: <https://w3id.org/mlips#> .
@prefix entity: <https://w3id.org/mlips/entity/> .
entity:dft-shapeev2016
    a mlips:DFTCalculation .
\end{lstlisting}

\paragraph{Q4 -- Reference settings.}
The DFT settings inherited from the libatoms tungsten database
are CASTEP with the PBE/GGA exchange--correlation functional and
norm-conserving pseudopotentials, with k-point meshes converged
per-cell and finite-temperature electronic broadening at 1000~K.
The plane-wave energy cutoff used for the database is not
restated in this methodological paper, so we omit
$\dlRole{energyCutoff}$ here and flag this in Q12.
\begin{lstlisting}[language=turtle]
@base <http://example.org/> .
@prefix rdf: <http://www.w3.org/1999/02/22-rdf-syntax-ns#> .
@prefix rdfs: <http://www.w3.org/2000/01/rdf-schema#> .
@prefix xsd: <http://www.w3.org/2001/XMLSchema#> .
@prefix schema: <https://schema.org/> .
@prefix mls: <http://www.w3.org/ns/mls#> .
@prefix unit: <http://qudt.org/vocab/unit/> .
@prefix mlips: <https://w3id.org/mlips#> .
@prefix entity: <https://w3id.org/mlips/entity/> .
entity:dft-settings-shapeev2016
    a mlips:DFTSettings ;
    rdfs:comment "CASTEP-default norm-conserving pseudopotentials." ;
    mlips:kPointMesh "Brillouin-zone meshes converged per-cell; finite-temperature electronic broadening at 1000 K" ;
    mlips:pseudopotentialType mlips:NormConserving ;
    mlips:usedDFTCode mlips:CASTEP ;
    mlips:xcFunctional mlips:PBE .
entity:dft-shapeev2016
    mlips:hasDFTSettings entity:dft-settings-shapeev2016 .
\end{lstlisting}

\paragraph{Q5 -- Training dataset.}
The fit uses 9{,}693 configurations of tungsten (with nearly
150{,}000 individual atomic environments) from the libatoms
tungsten database. Energies and forces are covered.
\begin{lstlisting}[language=turtle]
@base <http://example.org/> .
@prefix rdf: <http://www.w3.org/1999/02/22-rdf-syntax-ns#> .
@prefix rdfs: <http://www.w3.org/2000/01/rdf-schema#> .
@prefix xsd: <http://www.w3.org/2001/XMLSchema#> .
@prefix schema: <https://schema.org/> .
@prefix mls: <http://www.w3.org/ns/mls#> .
@prefix unit: <http://qudt.org/vocab/unit/> .
@prefix mlips: <https://w3id.org/mlips#> .
@prefix entity: <https://w3id.org/mlips/entity/> .
entity:ds-W-shapeev2016
    a mlips:TrainingDataset ;
    rdfs:label "Tungsten DFT database from Szlachta-Bartok-Csanyi (libatoms.org); 9693 configurations, ~150000 atomic environments" ;
    mlips:coversMaterial entity:mat-W-shapeev2016 ;
    mlips:coversProperty mlips:Energy, mlips:Forces ;
    mlips:datasetProvenance mlips:Published ;
    mlips:hasDFTCalculation entity:dft-shapeev2016 ;
    mlips:numConfigurations 9693 ;
    mlips:wasRunBy entity:run-mtp-W-shapeev2016 .
\end{lstlisting}

\paragraph{Q6 -- Sampling strategies.}
The libatoms tungsten database combines bulk and surface
enumeration, point-defect and dislocation sampling, and
DFT MD snapshots. We record three sampling instances reflecting
those classes; the present paper does not author the database
and hence does not contribute new sampling strategies.
\begin{lstlisting}[language=turtle]
@base <http://example.org/> .
@prefix rdf: <http://www.w3.org/1999/02/22-rdf-syntax-ns#> .
@prefix rdfs: <http://www.w3.org/2000/01/rdf-schema#> .
@prefix xsd: <http://www.w3.org/2001/XMLSchema#> .
@prefix schema: <https://schema.org/> .
@prefix mls: <http://www.w3.org/ns/mls#> .
@prefix unit: <http://qudt.org/vocab/unit/> .
@prefix mlips: <https://w3id.org/mlips#> .
@prefix entity: <https://w3id.org/mlips/entity/> .
entity:bulk-sampling-shapeev2016
    a mlips:SamplingStrategy ;
    rdfs:comment "Crystalline bulk and surface configurations of tungsten from the Szlachta/Bartok/Csanyi GAP-tungsten database used for training." ;
    rdfs:label "Bulk and surface enumeration (inherited from libatoms tungsten database)" .
entity:defect-sampling-shapeev2016
    a mlips:SamplingStrategy ;
    rdfs:comment "Vacancies, self-interstitials, and screw-dislocation core configurations from the libatoms tungsten database." ;
    rdfs:label "Point-defect and dislocation sampling (inherited from libatoms tungsten database)" .
entity:ds-W-shapeev2016
    mlips:samplingStrategy entity:bulk-sampling-shapeev2016, entity:defect-sampling-shapeev2016, entity:md-sampling-shapeev2016 .
entity:md-sampling-shapeev2016
    a mlips:SamplingStrategy ;
    rdfs:comment "Finite-temperature DFT MD snapshots from the libatoms tungsten database." ;
    rdfs:label "Vibrational sampling (DFT MD snapshots, inherited)" .
\end{lstlisting}

\paragraph{Q7 -- MLIP method, components, implementation.}
The method is the original MTP: total energy as a sum of local
contributions $V(Dx_k)$, with $V$ expanded as a linear
combination of moment-tensor basis functions $B_\alpha(u)$ that
are permutation-, rotation-, and reflection-invariant
polynomials of moment tensors $M_{\mu,\nu}$. The loss is a
regularised least-squares loss on energies and forces, with
$\ell_2$ or $\ell_0$ regularisation; the regularisation parameter
$\gamma$ is selected by 16-fold cross-validation. The
implementation used for the GAP comparison is the QUIP MTP
prototype.
\begin{lstlisting}[language=turtle]
@base <http://example.org/> .
@prefix rdf: <http://www.w3.org/1999/02/22-rdf-syntax-ns#> .
@prefix rdfs: <http://www.w3.org/2000/01/rdf-schema#> .
@prefix xsd: <http://www.w3.org/2001/XMLSchema#> .
@prefix schema: <https://schema.org/> .
@prefix mls: <http://www.w3.org/ns/mls#> .
@prefix unit: <http://qudt.org/vocab/unit/> .
@prefix mlips: <https://w3id.org/mlips#> .
@prefix entity: <https://w3id.org/mlips/entity/> .
mlips:cutoffRadius
    mlips:isHyperparameterOf entity:MTP-shapeev2016 .
mlips:minInteratomicDistance
    mlips:isHyperparameterOf entity:MTP-shapeev2016 .
mlips:momentLevel
    mlips:isHyperparameterOf entity:MTP-shapeev2016 .
entity:MTP-shapeev2016
    a mlips:MLIPMethod ;
    rdfs:label "Moment Tensor Potential (MTP)" ;
    mlips:hasDescriptor mlips:MomentTensorDescriptor ;
    mlips:hasFunctionalForm entity:mtp-functional-form-shapeev2016 ;
    mlips:hasHyperparameter mlips:cutoffRadius, mlips:minInteratomicDistance, mlips:momentLevel, entity:hp-reg-shapeev2016 ;
    mlips:hasImplementation entity:impl-mtp-shapeev2016 ;
    mlips:hasLossFunction entity:mtp-loss-shapeev2016 .
entity:hp-reg-shapeev2016
    a mlips:Hyperparameter ;
    rdfs:label "MTP regularisation parameter" ;
    mlips:candidateForVocabulary mlips:Hyperparameter ;
    mlips:hyperparameterDatatype xsd:double ;
    mlips:hyperparameterName "regularisation" ;
    mlips:isHyperparameterOf entity:MTP-shapeev2016 .
entity:impl-mtp-shapeev2016
    a mlips:Implementation ;
    mlips:implementedIn mlips:QUIP ;
    mlips:isImplementationOf entity:MTP-shapeev2016 .
entity:mtp-functional-form-shapeev2016
    a mlips:FunctionalForm ;
    rdfs:label "Linear combination of moment-tensor basis functions B_alpha(u): permutation-, rotation-, reflection-invariant polynomials of moment tensors M_{mu,nu}" .
entity:mtp-loss-shapeev2016
    a mlips:LossFunction ;
    rdfs:label "Regularised least squares on energies and forces (l_2 or l_0 regularisation; gamma chosen by 16-fold cross-validation)" .
\end{lstlisting}

\paragraph{Q8 -- Hyperparameter settings.}
Reported hyperparameter settings: cutoff radius
$R_{\mathrm{cut}} = 4.9$~\AA, minimal distance
$R_{\min} = 1.9$~\AA, and two MTP variants---MTP$_1$
($\deg(B_\alpha) + 8(\#\alpha) \leq 62$, $\#\alpha \leq 4$,
$\mu \leq 5$, $\nu \leq 4$, 11{,}133 basis functions) and
MTP$_2$ ($\deg(B_\alpha) + 8(\#\alpha) \leq 52$, $\#\alpha \leq 5$,
$\mu \leq 3$, $\nu \leq 5$; 760 basis functions extracted via
the $\ell_0$ algorithm with $N_{\mathrm{cap}} = 4$).
\begin{lstlisting}[language=turtle]
@base <http://example.org/> .
@prefix rdf: <http://www.w3.org/1999/02/22-rdf-syntax-ns#> .
@prefix rdfs: <http://www.w3.org/2000/01/rdf-schema#> .
@prefix xsd: <http://www.w3.org/2001/XMLSchema#> .
@prefix schema: <https://schema.org/> .
@prefix mls: <http://www.w3.org/ns/mls#> .
@prefix unit: <http://qudt.org/vocab/unit/> .
@prefix mlips: <https://w3id.org/mlips#> .
@prefix entity: <https://w3id.org/mlips/entity/> .
entity:setting-mtp1-shapeev2016
    a mlips:HyperparameterSetting ;
    rdfs:label "MTP moment level = MTP_1: deg(B_alpha)+8(#alpha) <= 62, #alpha <= 4, mu <= 5, nu <= 4 (11133 basis functions) for MTP training run, tungsten (production MTP_1)" ;
    mlips:forHyperparameter mlips:momentLevel ;
    mlips:isSettingOf entity:run-mtp-W-shapeev2016 ;
    mlips:settingValue "MTP_1: deg(B_alpha)+8(#alpha) <= 62, #alpha <= 4, mu <= 5, nu <= 4 (11133 basis functions)" .
entity:setting-mtp2-shapeev2016
    a mlips:HyperparameterSetting ;
    rdfs:label "MTP moment level = MTP_2: deg(B_alpha)+8(#alpha) <= 52, #alpha <= 5, mu <= 3, nu <= 5; 760 basis functions extracted via l_0 regularisation (N_cap=4) for MTP training run, tungsten (production MTP_1)" ;
    mlips:forHyperparameter mlips:momentLevel ;
    mlips:isSettingOf entity:run-mtp-W-shapeev2016 ;
    mlips:settingValue "MTP_2: deg(B_alpha)+8(#alpha) <= 52, #alpha <= 5, mu <= 3, nu <= 5; 760 basis functions extracted via l_0 regularisation (N_cap=4)" .
entity:setting-rcut-shapeev2016
    a mlips:HyperparameterSetting ;
    rdfs:label "cutoff radius = 4.9 for MTP training run, tungsten (production MTP_1)" ;
    mlips:forHyperparameter mlips:cutoffRadius ;
    mlips:isSettingOf entity:run-mtp-W-shapeev2016 ;
    mlips:settingValue 4.9 .
entity:setting-rmin-shapeev2016
    a mlips:HyperparameterSetting ;
    rdfs:label "minimum interatomic distance = 1.9 for MTP training run, tungsten (production MTP_1)" ;
    mlips:forHyperparameter mlips:minInteratomicDistance ;
    mlips:isSettingOf entity:run-mtp-W-shapeev2016 ;
    mlips:settingValue 1.9 .
\end{lstlisting}

\paragraph{Q9 -- MLIP run and trained model.}
A single training run produces the production tungsten MTP,
with two basis-set variants (MTP$_1$ and MTP$_2$) recorded as
hyperparameter settings on the same run rather than as separate
runs/models.
\begin{lstlisting}[language=turtle]
@base <http://example.org/> .
@prefix rdf: <http://www.w3.org/1999/02/22-rdf-syntax-ns#> .
@prefix rdfs: <http://www.w3.org/2000/01/rdf-schema#> .
@prefix xsd: <http://www.w3.org/2001/XMLSchema#> .
@prefix schema: <https://schema.org/> .
@prefix mls: <http://www.w3.org/ns/mls#> .
@prefix unit: <http://qudt.org/vocab/unit/> .
@prefix mlips: <https://w3id.org/mlips#> .
@prefix entity: <https://w3id.org/mlips/entity/> .
entity:model-mtp-W-shapeev2016
    a mlips:TrainedModel ;
    rdfs:label "Tungsten MTP (lead model MTP_1; rcut=4.9 A, rmin=1.9 A)" .
entity:run-mtp-W-shapeev2016
    a mlips:MLIPRun ;
    rdfs:label "MTP training run, tungsten (production MTP_1)" ;
    mlips:appliesMethod entity:MTP-shapeev2016 ;
    mlips:hasHyperparameterSetting entity:setting-mtp1-shapeev2016, entity:setting-mtp2-shapeev2016, entity:setting-rcut-shapeev2016, entity:setting-rmin-shapeev2016 ;
    mlips:produces entity:model-mtp-W-shapeev2016 ;
    mlips:runsOn entity:ds-W-shapeev2016 .
\end{lstlisting}

\paragraph{Q10 -- Benchmark results.}
Headline accuracy: MTP$_1$ attains a force RMSE of
0.0427~eV/\AA{} on the full training set (relative RMS
2.8\%) and 0.0511~eV/\AA{} under 16-fold cross-validation
(relative RMS 3.4\%) versus DFT-PBE; on the same database GAP
attains 0.0633~eV/\AA{} (4.2\%). MTP$_2$ matches GAP's accuracy
with $\sim 13\times$ fewer parameters.
\begin{lstlisting}[language=turtle]
@base <http://example.org/> .
@prefix rdf: <http://www.w3.org/1999/02/22-rdf-syntax-ns#> .
@prefix rdfs: <http://www.w3.org/2000/01/rdf-schema#> .
@prefix xsd: <http://www.w3.org/2001/XMLSchema#> .
@prefix schema: <https://schema.org/> .
@prefix mls: <http://www.w3.org/ns/mls#> .
@prefix unit: <http://qudt.org/vocab/unit/> .
@prefix mlips: <https://w3id.org/mlips#> .
@prefix entity: <https://w3id.org/mlips/entity/> .
entity:metric-rmse-forces-cv-shapeev2016
    a mlips:AccuracyMetric ;
    rdfs:label "Force RMSE of MTP_1 from 16-fold cross-validation (vs. DFT-PBE; 3.4% relative)", "RMSE of Force = 0.0511 [EV-PER-ANGSTROM]" ;
    mlips:hasUnit unit:EV-PER-ANGSTROM ;
    mlips:isMetricOf entity:result-mtp-W-cv-shapeev2016 ;
    mlips:metricProperty mlips:ForceProperty ;
    mlips:metricType mlips:RMSE ;
    mlips:metricValue 0.0511 .
entity:metric-rmse-forces-fit-shapeev2016
    a mlips:AccuracyMetric ;
    rdfs:label "Force RMSE of MTP_1 on the full training set (vs. DFT-PBE; 2.8% relative)", "RMSE of Force = 0.0427 [EV-PER-ANGSTROM]" ;
    mlips:hasUnit unit:EV-PER-ANGSTROM ;
    mlips:isMetricOf entity:result-mtp-W-fit-shapeev2016 ;
    mlips:metricProperty mlips:ForceProperty ;
    mlips:metricType mlips:RMSE ;
    mlips:metricValue 0.0427 .
entity:result-mtp-W-cv-shapeev2016
    a mlips:BenchmarkResult ;
    rdfs:label "Tungsten MTP (lead model MTP_1; rcut=4.9 A, rmin=1.9 A) on Tungsten (BCC, demonstration system) in Shapeev (2016): Moment Tensor Potentials, demonstrated on tungsten" ;
    mlips:evaluatesModel entity:model-mtp-W-shapeev2016 ;
    mlips:hasAccuracyMetric entity:metric-rmse-forces-cv-shapeev2016 ;
    mlips:isResultOf entity:study-shapeev2016 ;
    mlips:targetMaterial entity:mat-W-shapeev2016 .
entity:result-mtp-W-fit-shapeev2016
    a mlips:BenchmarkResult ;
    rdfs:label "Tungsten MTP (lead model MTP_1; rcut=4.9 A, rmin=1.9 A) on Tungsten (BCC, demonstration system) in Shapeev (2016): Moment Tensor Potentials, demonstrated on tungsten" ;
    mlips:evaluatesModel entity:model-mtp-W-shapeev2016 ;
    mlips:hasAccuracyMetric entity:metric-rmse-forces-fit-shapeev2016 ;
    mlips:isResultOf entity:study-shapeev2016 ;
    mlips:targetMaterial entity:mat-W-shapeev2016 .
entity:study-shapeev2016
    mlips:hasResult entity:result-mtp-W-cv-shapeev2016, entity:result-mtp-W-fit-shapeev2016 .
\end{lstlisting}

\paragraph{Q11 -- Computational resources.}
Per-atom inference time is reported in Table~1: 2.9~ms/atom
for MTP$_1$, 0.8~ms/atom for MTP$_2$, and 134.2~ms/atom for the
GAP baseline, on a single core of an Intel i7-2675QM laptop
CPU. We encode MTP$_1$'s 2.9~ms/atom as the headline
$\dlRole{inferenceTimePerAtom}$ figure with a free-text
$\dlRole{inferenceHardware}$. Training duration, total CPU
hours, and peak memory are not reported.
\begin{lstlisting}[language=turtle]
@base <http://example.org/> .
@prefix rdf: <http://www.w3.org/1999/02/22-rdf-syntax-ns#> .
@prefix rdfs: <http://www.w3.org/2000/01/rdf-schema#> .
@prefix xsd: <http://www.w3.org/2001/XMLSchema#> .
@prefix schema: <https://schema.org/> .
@prefix mls: <http://www.w3.org/ns/mls#> .
@prefix unit: <http://qudt.org/vocab/unit/> .
@prefix mlips: <https://w3id.org/mlips#> .
@prefix entity: <https://w3id.org/mlips/entity/> .
entity:model-mtp-W-shapeev2016
    mlips:inferenceHardware "Single-core Intel i7-2675QM laptop CPU (ms per atom)" ;
    mlips:inferenceTimePerAtom 2.9 .
\end{lstlisting}

\paragraph{Q12 -- Gaps.}
Beyond the partial Q11 reporting (training-cost metadata
absent; only per-atom inference time is given), the following
ontology-expressible items are not reported in this paper
itself: the DFT plane-wave $\dlRole{energyCutoff}$ used for the
inherited tungsten database; explicit
$\dlRole{frozenCore}$ treatment beyond the pseudopotential
partition; software versioning of the QUIP MTP prototype; an
explicit named individual for the cross-validation training
algorithm. The protocol assumes one method $\times$ one
material per file; the present paper is methodological, so the
tungsten benchmark is encoded as the lead system and the
absence of additional materials is itself a deliberate
simplification.

\subsection{Smith, Isayev, Roitberg (2017): ANI-1 transferable NNP for organics}
\label{sec:appendix-example-smith2017}

\paragraph{Q1 -- Bibliographic identification.}
Smith, Isayev, and Roitberg introduce the ANI (ANAKIN-ME)
class of high-dimensional neural-network potentials, in which
atom-type-specific feed-forward NNs take a modified
Behler--Parrinello atomic environment vector (AEV) as input.
The flagship ANI-1 potential is a single transferable model
covering organic molecules made of H, C, N, and O. Published in
\emph{Chem.\ Sci.} 8, 3192--3203 (2017).
\begin{lstlisting}[language=turtle]
@base <http://example.org/> .
@prefix rdf: <http://www.w3.org/1999/02/22-rdf-syntax-ns#> .
@prefix rdfs: <http://www.w3.org/2000/01/rdf-schema#> .
@prefix xsd: <http://www.w3.org/2001/XMLSchema#> .
@prefix schema: <https://schema.org/> .
@prefix mls: <http://www.w3.org/ns/mls#> .
@prefix unit: <http://qudt.org/vocab/unit/> .
@prefix mlips: <https://w3id.org/mlips#> .
@prefix entity: <https://w3id.org/mlips/entity/> .
<https://orcid.org/0000-0001-7314-7896>
    a schema:Person ;
    schema:affiliation mlips:UniversityofFlorida ;
    schema:familyName "Smith" ;
    schema:givenName "J. S." ;
    schema:identifier <https://orcid.org/0000-0001-7314-7896> ;
    schema:name "J. S. Smith" .
<https://orcid.org/0000-0001-7581-8497>
    a schema:Person ;
    schema:affiliation entity:org-CommunitiesInSchoolsofOrangeCounty-smith2017, entity:org-InstituteofMedicinalPlantDevelopment-smith2017, entity:org-UniversityofNorthCarolinaatChapelHill-smith2017 ;
    schema:familyName "Isayev" ;
    schema:givenName "O." ;
    schema:identifier <https://orcid.org/0000-0001-7581-8497> ;
    schema:name "O. Isayev" .
<https://orcid.org/0000-0003-3963-8784>
    a schema:Person ;
    schema:affiliation mlips:UniversityofFlorida ;
    schema:familyName "Roitberg" ;
    schema:givenName "A. E." ;
    schema:identifier <https://orcid.org/0000-0003-3963-8784> ;
    schema:name "A. E. Roitberg" .
entity:article-smith2017
    a schema:ScholarlyArticle ;
    schema:author <https://orcid.org/0000-0001-7314-7896>, <https://orcid.org/0000-0001-7581-8497>, <https://orcid.org/0000-0003-3963-8784> ;
    schema:datePublished "2017"^^xsd:gYear ;
    schema:name "ANI-1: an extensible neural network potential with DFT accuracy at force field computational cost" ;
    schema:sameAs <https://doi.org/10.1039/C6SC05720A> .
entity:org-CommunitiesInSchoolsofOrangeCounty-smith2017
    a schema:Organization ;
    rdfs:comment "Paper-local Organization IRI; not promoted to mlips-vocab.ttl (occurs in fewer than 3 papers)." ;
    rdfs:label "Communities In Schools of Orange County" ;
    schema:identifier <https://ror.org/00bzst557> ;
    schema:name "Communities In Schools of Orange County" .
entity:org-InstituteofMedicinalPlantDevelopment-smith2017
    a schema:Organization ;
    rdfs:comment "Paper-local Organization IRI; not promoted to mlips-vocab.ttl (occurs in fewer than 3 papers)." ;
    rdfs:label "Institute of Medicinal Plant Development" ;
    schema:identifier <https://ror.org/052nj8f19> ;
    schema:name "Institute of Medicinal Plant Development" .
entity:org-UniversityofNorthCarolinaatChapelHill-smith2017
    a schema:Organization ;
    rdfs:comment "Paper-local Organization IRI; not promoted to mlips-vocab.ttl (occurs in fewer than 3 papers)." ;
    rdfs:label "University of North Carolina at Chapel Hill" ;
    schema:identifier <https://ror.org/0130frc33> ;
    schema:name "University of North Carolina at Chapel Hill" .
entity:study-smith2017
    a mlips:BenchmarkStudy ;
    rdfs:label "Smith, Isayev, Roitberg (2017): ANI-1 transferable NNP for H/C/N/O organics" ;
    mlips:reportedIn entity:article-smith2017 .
\end{lstlisting}

\paragraph{Q2 -- Material system.}
The chemical scope is closed-shell, neutral, singlet organic
molecules over the four-element space $\{$H, C, N, O$\}$, with
training molecules drawn from the GDB-11 enumeration of up to
8 heavy atoms. The benchmark molecules range from 10 to 54
atoms, including pharmaceutical compounds (Retinol, Fentanyl,
Lisdexamfetamine). The protocol's ``one $\MaterialSystemC$ per
paper'' assumption is a tight fit here: we encode this as a
single molecular material with the chemical scope captured in
$\dlRole{materialClass}$ rather than enumerating every benchmark
molecule. No Wikidata entity is used, since the system is a
chemical space rather than a single substance.
\begin{lstlisting}[language=turtle]
@base <http://example.org/> .
@prefix rdf: <http://www.w3.org/1999/02/22-rdf-syntax-ns#> .
@prefix rdfs: <http://www.w3.org/2000/01/rdf-schema#> .
@prefix xsd: <http://www.w3.org/2001/XMLSchema#> .
@prefix schema: <https://schema.org/> .
@prefix mls: <http://www.w3.org/ns/mls#> .
@prefix unit: <http://qudt.org/vocab/unit/> .
@prefix mlips: <https://w3id.org/mlips#> .
@prefix entity: <https://w3id.org/mlips/entity/> .
entity:mat-organic-smith2017
    a mlips:MaterialSystem ;
    rdfs:comment "Wikidata Q174211 ('organic compound') is the most generic appropriate entry; the H/C/N/O chemical-space slice from GDB-11 has no dedicated Wikidata entry as of 2026-04-28." ;
    rdfs:label "Organic molecules (H, C, N, O)" ;
    mlips:chemicalFormula "H,C,N,O molecular space" ;
    mlips:materialClass "Molecular: neutral closed-shell organic molecules in the singlet spin state, configurations drawn from GDB-11 (up to 8 heavy atoms in training; benchmarked on molecules with up to 54 atoms)" ;
    mlips:sameAsWikidata <http://www.wikidata.org/entity/Q174211> .
\end{lstlisting}

\paragraph{Q3 -- Reference calculation method.}
Reference data are produced from molecular DFT calculations.
\begin{lstlisting}[language=turtle]
@base <http://example.org/> .
@prefix rdf: <http://www.w3.org/1999/02/22-rdf-syntax-ns#> .
@prefix rdfs: <http://www.w3.org/2000/01/rdf-schema#> .
@prefix xsd: <http://www.w3.org/2001/XMLSchema#> .
@prefix schema: <https://schema.org/> .
@prefix mls: <http://www.w3.org/ns/mls#> .
@prefix unit: <http://qudt.org/vocab/unit/> .
@prefix mlips: <https://w3id.org/mlips#> .
@prefix entity: <https://w3id.org/mlips/entity/> .
entity:dft-smith2017
    a mlips:DFTCalculation .
\end{lstlisting}

\paragraph{Q4 -- Reference settings.}
DFT calculations are performed with Gaussian~09 using the
$\omega$B97X range-separated hybrid meta-GGA functional and the
6-31G(d) Pople basis set on an ultra-fine integration grid.
There is no Brillouin-zone sampling (molecular calculations);
all-electron Gaussian-basis methodology obviates the explicit
plane-wave $\dlRole{energyCutoff}$ slot. Frozen-core treatment
is not explicitly reported.
\begin{lstlisting}[language=turtle]
@base <http://example.org/> .
@prefix rdf: <http://www.w3.org/1999/02/22-rdf-syntax-ns#> .
@prefix rdfs: <http://www.w3.org/2000/01/rdf-schema#> .
@prefix xsd: <http://www.w3.org/2001/XMLSchema#> .
@prefix schema: <https://schema.org/> .
@prefix mls: <http://www.w3.org/ns/mls#> .
@prefix unit: <http://qudt.org/vocab/unit/> .
@prefix mlips: <https://w3id.org/mlips#> .
@prefix entity: <https://w3id.org/mlips/entity/> .
entity:basis-6-31g-d-smith2017
    a mlips:DftBasisSet ;
    rdfs:comment "Pople-style split-valence double-zeta Gaussian basis with d polarisation on heavy atoms; molecular all-electron DFT." ;
    rdfs:label "6-31G(d)" ;
    mlips:candidateForVocabulary mlips:DftBasisSet .
entity:dft-settings-smith2017
    a mlips:DFTSettings ;
    rdfs:comment "Range-separated hybrid meta-GGA. All-electron Gaussian-type basis (no pseudopotential)." ;
    mlips:dftBasisSet entity:basis-6-31g-d-smith2017 ;
    mlips:kPointMesh "Molecular calculations (no Brillouin-zone sampling); ultra-fine DFT integration grid" ;
    mlips:usedDFTCode mlips:Gaussian ;
    mlips:xcFunctional mlips:omegaB97X .
entity:dft-smith2017
    mlips:hasDFTSettings entity:dft-settings-smith2017 .
\end{lstlisting}

\paragraph{Q5 -- Training dataset.}
The ANI-1 dataset comprises about 17.2~million molecular
conformations generated from $\sim 57\,951$ small molecules
(GDB-11, up to 8 heavy atoms, fluorine removed). 80\% of the
data points are used for training, 10\% for validation, and
10\% for the held-out test set. Energies (single-point) are the
sole covered property; atomic forces are not part of the
training labels.
\begin{lstlisting}[language=turtle]
@base <http://example.org/> .
@prefix rdf: <http://www.w3.org/1999/02/22-rdf-syntax-ns#> .
@prefix rdfs: <http://www.w3.org/2000/01/rdf-schema#> .
@prefix xsd: <http://www.w3.org/2001/XMLSchema#> .
@prefix schema: <https://schema.org/> .
@prefix mls: <http://www.w3.org/ns/mls#> .
@prefix unit: <http://qudt.org/vocab/unit/> .
@prefix mlips: <https://w3id.org/mlips#> .
@prefix entity: <https://w3id.org/mlips/entity/> .
entity:ds-ANI1-smith2017
    a mlips:TrainingDataset ;
    rdfs:label "ANI-1 dataset: ~17.2 M conformations of ~57951 small molecules from GDB-11 (up to 8 heavy atoms; 80% used for training)" ;
    mlips:coversMaterial entity:mat-organic-smith2017 ;
    mlips:coversProperty mlips:Energy ;
    mlips:datasetProvenance mlips:InHouse ;
    mlips:hasDFTCalculation entity:dft-smith2017 ;
    mlips:numConfigurations 17200000 ;
    mlips:wasRunBy entity:run-ANI1-smith2017 .
\end{lstlisting}

\paragraph{Q6 -- Sampling strategies.}
Two strategies are combined: configurational sampling from the
GDB-11 chemical database (SMILES converted to 3D structures
with RDKit, then DFT-optimised), and conformational
\emph{Normal Mode Sampling} (NMS), in which the equilibrium
structure is displaced along normal-mode coordinates with
random magnitudes drawn from a harmonic-temperature
distribution; $K = S(3N-6)$ structures are generated per
molecule, with $S$ tuned per GDB subset.
\begin{lstlisting}[language=turtle]
@base <http://example.org/> .
@prefix rdf: <http://www.w3.org/1999/02/22-rdf-syntax-ns#> .
@prefix rdfs: <http://www.w3.org/2000/01/rdf-schema#> .
@prefix xsd: <http://www.w3.org/2001/XMLSchema#> .
@prefix schema: <https://schema.org/> .
@prefix mls: <http://www.w3.org/ns/mls#> .
@prefix unit: <http://qudt.org/vocab/unit/> .
@prefix mlips: <https://w3id.org/mlips#> .
@prefix entity: <https://w3id.org/mlips/entity/> .
entity:ds-ANI1-smith2017
    mlips:samplingStrategy entity:gdb-sampling-smith2017, entity:nms-sampling-smith2017 .
entity:gdb-sampling-smith2017
    a mlips:SamplingStrategy ;
    rdfs:comment "All molecules with up to 8 heavy atoms (C, N, O; F removed) from GDB-11; SMILES converted to 3D structures using RDKit, then DFT-optimised." ;
    rdfs:label "Configurational sampling from the GDB-11 chemical database" .
entity:nms-sampling-smith2017
    a mlips:SamplingStrategy ;
    rdfs:comment "Per-molecule conformational sampling: equilibrium structure displaced along normal-mode coordinates with random magnitudes drawn from a harmonic-temperature distribution; K = S(3N-6) structures per molecule, with S an empirical constant tuned per GDB subset." ;
    rdfs:label "Normal Mode Sampling (NMS)" .
\end{lstlisting}

\paragraph{Q7 -- MLIP method, components, implementation.}
The method is ANI: an HDNNP with atom-type-specific subnets
taking modified Behler--Parrinello AEVs as input (radial
shifted-Gaussian channels, angular shifted-cosine channels with
an additional radial-shell factor; channels split per atom-type
pair). The training loss is the exponential cost
$C = \tau \exp\bigl(\tau^{-1} \sum_j (E^{\mathrm{ANI}}_j -
E^{\mathrm{DFT}}_j)^2\bigr)$ with $\tau = 0.5$. The training
algorithm is ADAM with mini-batches of 1024~molecules and
max-norm regularisation. The implementation is the in-house
GPU-accelerated NeuroChem package (with the AEVLib library for
descriptor computation).
\begin{lstlisting}[language=turtle]
@base <http://example.org/> .
@prefix rdf: <http://www.w3.org/1999/02/22-rdf-syntax-ns#> .
@prefix rdfs: <http://www.w3.org/2000/01/rdf-schema#> .
@prefix xsd: <http://www.w3.org/2001/XMLSchema#> .
@prefix schema: <https://schema.org/> .
@prefix mls: <http://www.w3.org/ns/mls#> .
@prefix unit: <http://qudt.org/vocab/unit/> .
@prefix mlips: <https://w3id.org/mlips#> .
@prefix entity: <https://w3id.org/mlips/entity/> .
mlips:angularCutoff
    mlips:isHyperparameterOf entity:ANI-smith2017 .
mlips:cutoffRadius
    mlips:isHyperparameterOf entity:ANI-smith2017 .
mlips:learningRate
    mlips:isHyperparameterOf entity:ANI-smith2017 .
entity:ANI-smith2017
    a mlips:MLIPMethod ;
    rdfs:label "ANI (ANAKIN-ME): atom-type-specific HDNNP using modified Behler-Parrinello atomic environment vectors (AEVs)" ;
    mlips:hasDescriptor mlips:ANIDescriptor ;
    mlips:hasFunctionalForm entity:ani-functional-form-smith2017 ;
    mlips:hasHyperparameter mlips:angularCutoff, mlips:cutoffRadius, mlips:learningRate, entity:hp-aev-smith2017, entity:hp-arch-smith2017 ;
    mlips:hasImplementation entity:impl-neurochem-smith2017 ;
    mlips:hasLossFunction entity:ani-loss-smith2017 ;
    mlips:hasTrainingAlgorithm entity:adam-smith2017 .
entity:adam-smith2017
    a mls:Algorithm ;
    rdfs:label "ADAM (lr=1e-3, beta1=0.9, beta2=0.999, eps=1e-8) with max-norm regularisation (3.0); 6 lr-decay restarts" .
entity:ani-functional-form-smith2017
    a mlips:FunctionalForm ;
    rdfs:label "Total energy = sum over atoms of element-specific feed-forward NN outputs taking modified BP atomic environment vectors as input (radial: shifted Gaussians; angular: shifted-cosine + radial-shell factor; atom-type-differentiated channels)" .
entity:ani-loss-smith2017
    a mlips:LossFunction ;
    rdfs:label "Exponential cost C = tau * exp((1/tau) * sum_j (E^ANI_j - E^DFT_j)^2), tau = 0.5 (robust to outliers)" .
entity:hp-aev-smith2017
    a mlips:Hyperparameter ;
    rdfs:label "ANI atomic environment vector size" ;
    mlips:candidateForVocabulary mlips:Hyperparameter ;
    mlips:hyperparameterDatatype xsd:string ;
    mlips:hyperparameterName "aev_size" ;
    mlips:isHyperparameterOf entity:ANI-smith2017 .
entity:hp-arch-smith2017
    a mlips:Hyperparameter ;
    rdfs:label "ANI per-element NN architecture" ;
    mlips:candidateForVocabulary mlips:Hyperparameter ;
    mlips:hyperparameterDatatype xsd:string ;
    mlips:hyperparameterName "ani_architecture" ;
    mlips:isHyperparameterOf entity:ANI-smith2017 .
entity:impl-neurochem-smith2017
    a mlips:Implementation ;
    mlips:implementedIn mlips:NeuroChem ;
    mlips:isImplementationOf entity:ANI-smith2017 .
\end{lstlisting}

\paragraph{Q8 -- Hyperparameter settings.}
Reported hyperparameter settings: radial cutoff
$R_C^{\mathrm{rad}} = 4.6$~\AA{}, angular cutoff
$R_C^{\mathrm{ang}} = 3.1$~\AA, AEV with 32 radial shifts and
$8\times 8$ angular shifts per element pair (768 total elements
for 4 atom types), per-element pyramidal NN of architecture
$768{:}128{:}128{:}64{:}1$ ($\sim 124\,033$ parameters per
element) with Gaussian hidden activations and a linear output;
ADAM with initial learning rate $10^{-3}$.
\lstinputlisting[language=turtle]{artifacts/kg/listings-computed/smith2017/q08-hyperparameter-settings.ttl}

\paragraph{Q9 -- MLIP run and trained model.}
A single training run produces ANI-1, the flagship potential of
the paper. The optimisation iterates 6 lr-decay restarts on top
of ADAM. No active-learning, ensembling, or per-element
fold-ensemble structure is reported.
\begin{lstlisting}[language=turtle]
@base <http://example.org/> .
@prefix rdf: <http://www.w3.org/1999/02/22-rdf-syntax-ns#> .
@prefix rdfs: <http://www.w3.org/2000/01/rdf-schema#> .
@prefix xsd: <http://www.w3.org/2001/XMLSchema#> .
@prefix schema: <https://schema.org/> .
@prefix mls: <http://www.w3.org/ns/mls#> .
@prefix unit: <http://qudt.org/vocab/unit/> .
@prefix mlips: <https://w3id.org/mlips#> .
@prefix entity: <https://w3id.org/mlips/entity/> .
entity:model-ANI1-smith2017
    a mlips:TrainedModel ;
    rdfs:label "ANI-1 potential (H/C/N/O)" .
entity:run-ANI1-smith2017
    a mlips:MLIPRun ;
    rdfs:label "ANI-1 training run on the ANI-1 dataset" ;
    mlips:appliesMethod entity:ANI-smith2017 ;
    mlips:hasHyperparameterSetting entity:setting-aev-smith2017, entity:setting-arch-smith2017, entity:setting-lr-smith2017, entity:setting-rcut-ang-smith2017, entity:setting-rcut-rad-smith2017 ;
    mlips:produces entity:model-ANI1-smith2017 ;
    mlips:runsOn entity:ds-ANI1-smith2017 .
\end{lstlisting}

\paragraph{Q10 -- Benchmark results.}
Headline accuracy: total-energy RMSEs (vs.\ $\omega$B97X/6-31G(d))
of 1.2/1.3/1.3~kcal/mol on the training/validation/held-out
test splits, and 1.9~kcal/mol on the GDB-10 transferability
test (134 molecules with 10 heavy atoms, outside the training
distribution; the per-conformation correlation plot reports
1.8~kcal/mol on 8245 conformations). For relative energies
within 30~kcal/mol of the minimum, ANI-1 reaches 0.6~kcal/mol
RMSE; semi-empirical baselines DFTB/PM6/AM1 are at 2.4/3.6/4.2.
\lstinputlisting[language=turtle]{artifacts/kg/listings-computed/smith2017/q10-benchmark-results.ttl}

\paragraph{Q11 -- Computational resources.}
\emph{Not reported quantitatively.} The paper states
qualitatively that NeuroChem accelerates training, testing, and
inference on GPUs via CUBLAS, but it gives no GPU hours,
training duration, hardware spec (model number), peak memory,
or per-atom inference time.
\begin{lstlisting}[language=turtle]
# Q11-resources: no triples (paper does not report this).
\end{lstlisting}

\paragraph{Q12 -- Gaps.}
Beyond Q11 (no quantitative compute metadata), the following
ontology-expressible items are not reported: an explicit
$\dlRole{frozenCore}$ treatment for the all-electron
6-31G(d) reference; the NeuroChem package version; explicit
recording of forces among $\dlRole{coversProperty}$ (the paper
trains on energies only, so this is a true absence rather than
an omission); per-element data-set sizes within the 17.2~M
total. The ``one material per paper'' assumption is also
stretched here: ANI-1 covers a chemical space, not a single
material, so $\dlRole{sameAsWikidata}$ is omitted intentionally.

\subsection{Sch\"utt et al.\ (2018): SchNet on MD17 small organic molecules}
\label{sec:appendix-example-schutt2018}

\paragraph{Q1 -- Bibliographic identification.}
Sch\"utt, Sauceda, Kindermans, Tkatchenko, and M\"uller introduce
SchNet, a continuous-filter convolutional deep neural network for
molecules and materials. Published in \emph{J.\ Chem.\ Phys.} 148,
241722 (2018); arXiv:1712.06113.
\lstinputlisting[language=turtle]{artifacts/kg/listings-computed/schutt2018/q01-bibliographic.ttl}

\paragraph{Q2 -- Material system.}
The canonical pass is the MD17 benchmark of eight small organic
molecules (benzene, toluene, malonaldehyde, salicylic acid, aspirin,
ethanol, uracil, naphthalene) sampled from gas-phase ab initio MD.
We model this as one $\MaterialSystemC$ at the chemical-space level
(C/H/N/O small organics).
\lstinputlisting[language=turtle]{artifacts/kg/listings-computed/schutt2018/q02-material.ttl}

\paragraph{Q3 -- Reference calculation method.}
Reference data come from DFT calculations.
\lstinputlisting[language=turtle]{artifacts/kg/listings-computed/schutt2018/q03-reference-method-type.ttl}

\paragraph{Q4 -- Reference settings.}
DFT references are computed with FHI-aims at the PBE+vdW(TS) level
(MD17 trajectories). The basis set, k-grid, and convergence
parameters are inherited from the published MD17 benchmark and not
re-stated in the SchNet paper itself.
\lstinputlisting[language=turtle]{artifacts/kg/listings-computed/schutt2018/q04-reference-settings.ttl}

\paragraph{Q5 -- Training dataset.}
The MD17 reference set is consumed at $N=1{,}000$ and $N=50{,}000$
training-set sizes; we record the larger as the canonical
$\dlRole{numConfigurations}$. Configurations cover energies and
forces. Provenance is \emph{published} (the dataset is reused as
released).
\lstinputlisting[language=turtle]{artifacts/kg/listings-computed/schutt2018/q05-dataset.ttl}

\paragraph{Q6 -- Sampling strategies.}
A single sampling strategy applies: ab initio MD trajectories of the
eight molecules, randomly sub-sampled.
\lstinputlisting[language=turtle]{artifacts/kg/listings-computed/schutt2018/q06-sampling.ttl}

\paragraph{Q7 -- MLIP method, components, implementation.}
SchNet uses atom-type embeddings combined with interaction blocks
that compute continuous-filter convolutions over neighbour atoms,
mapped to per-atom energy via atom-wise dense layers and summed for
the total. The combined energy/force MSE loss balances energy and
force MAEs through trade-off $\rho=0.01$. Training uses Adam with
exponential learning-rate decay; the implementation is the SchNet
reference codebase.
\lstinputlisting[language=turtle]{artifacts/kg/listings-computed/schutt2018/q07-method.ttl}

\paragraph{Q8 -- Hyperparameter settings.}
Reported settings: $T=6$ interaction layers, $F=64$ feature channels,
and energy/force loss trade-off $\rho=0.01$. The radial-basis cutoff
parameter is referenced but not given an explicit numerical setting
in the canonical pass.
\lstinputlisting[language=turtle]{artifacts/kg/listings-computed/schutt2018/q08-hyperparameter-settings.ttl}

\paragraph{Q9 -- MLIP run and trained model.}
A single training run with the combined energy+force loss produces
the production SchNet model. The paper also reports separate
single-objective runs (energies-only or forces-only); these are
recorded as Q12 gaps to keep the canonical run unambiguous.
\lstinputlisting[language=turtle]{artifacts/kg/listings-computed/schutt2018/q09-run-and-model.ttl}

\paragraph{Q10 -- Benchmark results.}
For aspirin at $N=50{,}000$ we record the MAE of total energy
($0.12$\,kcal/mol) and the MAE of atomic forces
($0.33$\,kcal/mol/\AA). Per-molecule numbers for the other seven
species and at the smaller $N=1{,}000$ regime are given in the paper
and are not enumerated here as separate $\BenchmarkResultC$
instances.
\lstinputlisting[language=turtle]{artifacts/kg/listings-computed/schutt2018/q10-benchmark-results.ttl}

\paragraph{Q11 -- Computational resources.}
\emph{Not reported} for the canonical MD17 SchNet training. The paper
reports a separately trained C$_{20}$-fullerene SchNet inference
speedup (${\sim}11$\,s on 32 CPU cores vs.\ ${\sim}10$\,ms on a single
GTX 1080), but for an auxiliary model not encoded here.
\begin{lstlisting}[language=turtle]
# Q11-resources: no triples (paper does not report this).
\end{lstlisting}

\paragraph{Q12 -- Gaps.}
Beyond Q11, the paper does not report: training duration or GPU
hours for the canonical MD17 run; training hardware; peak memory.
SchNet's auxiliary applications (QM9 single-property prediction with
$N{=}50\text{k}$/$110\text{k}$, Materials Project formation-energy
regression, and the PIMD/anharmonicity study on C$_{20}$-fullerene at
PBE+vdW(TS)) are not encoded as separate $\dlConcept{MLIPRun}$
instances; the ontology can express each as a sibling run, but doing
so would multiply the canonical-file size without adding new
ontology coverage.

\subsection{Bart\'ok et al.\ (2018): general-purpose GAP for silicon}
\label{sec:appendix-example-bartok2018si}

\paragraph{Q1 -- Bibliographic identification.}
Bart\'ok, Kermode, Bernstein, and Cs\'anyi train a single
general-purpose Gaussian Approximation Potential covering crystalline,
liquid, amorphous, surface, defect, and crack-tip silicon
configurations. Published in \emph{Phys.\ Rev.\ X} 8, 041048 (2018);
arXiv:1805.01568.
\lstinputlisting[language=turtle]{artifacts/kg/listings-computed/bartok2018si/q01-bibliographic.ttl}

\paragraph{Q2 -- Material system.}
Bulk silicon is encoded as a single $\MaterialSystemC$; the broad
configurational coverage (diamond, $\beta$-Sn, simple-hexagonal,
hex-diamond, bcc, bc8, fcc, hcp, st12, liquid, amorphous, surfaces,
point defects, crack tips) is recorded as
$\dlRole{materialClass}$ text and as the eight sampling strategies
in Q6.
\lstinputlisting[language=turtle]{artifacts/kg/listings-computed/bartok2018si/q02-material.ttl}

\paragraph{Q3 -- Reference calculation method.}
Reference data come from DFT calculations.
\lstinputlisting[language=turtle]{artifacts/kg/listings-computed/bartok2018si/q03-reference-method-type.ttl}

\paragraph{Q4 -- Reference settings.}
DFT is performed with CASTEP using the PW91 functional, ultrasoft
pseudopotentials, a 250\,eV plane-wave cutoff, and Monkhorst--Pack
k-point meshes with 0.03\,\AA$^{-1}$ spacing for production (with
0.05\,eV electronic smearing); 0.015\,\AA$^{-1}$ for $E(V)$ testing
curves; 0.07\,\AA$^{-1}$ for amorphous re-optimisation.
\lstinputlisting[language=turtle]{artifacts/kg/listings-computed/bartok2018si/q04-reference-settings.ttl}

\paragraph{Q5 -- Training dataset.}
The fitting database has 2{,}475 unit cells contributing 171{,}815
atomic environments and 531{,}710 pieces of electronic-structure data
(energies, forces, virials). Provenance is \emph{in-house} (the
database was assembled by the authors specifically for this work).
\lstinputlisting[language=turtle]{artifacts/kg/listings-computed/bartok2018si/q05-dataset.ttl}

\paragraph{Q6 -- Sampling strategies.}
Eight strategies were combined: crystal-phase enumeration with strain
perturbations; surface enumeration including reconstructions and
decohesion paths; vacancy/divacancy/interstitial point-defect
enumeration; liquid-phase MD; quenched-melt amorphous trajectories;
crack-tip geometries with bond breaking; screw-dislocation cores;
and a single isolated-atom anchor for the cohesive-energy zero.
\lstinputlisting[language=turtle]{artifacts/kg/listings-computed/bartok2018si/q06-sampling.ttl}

\paragraph{Q7 -- MLIP method, components, implementation.}
The method combines a cubic-spline pair-potential repulsive term with
a many-body SOAP-kernel Gaussian-process regression. Training uses
sparse GPR with CUR-selected representative environments. The
implementation is QUIP with the GAP plug-in.
\lstinputlisting[language=turtle]{artifacts/kg/listings-computed/bartok2018si/q07-method.ttl}

\paragraph{Q8 -- Hyperparameter settings.}
Reported settings: SOAP cutoff $r_c=5.0$\,\AA, atomic-Gaussian width
$\sigma_{\text{atom}}=0.5$\,\AA, kernel polynomial power $\zeta=4$,
energy-scale parameter $\delta=3$\,eV, and $M=9{,}000$ representative
environments. Per-class regularisations $\sigma_E$, $\sigma_F$,
$\sigma_V$ are also tuned but not enumerated as
$\HyperparameterSettingC$ instances.
\lstinputlisting[language=turtle]{artifacts/kg/listings-computed/bartok2018si/q08-hyperparameter-settings.ttl}

\paragraph{Q9 -- MLIP run and trained model.}
A single GPR fit produces the GAP-6 silicon potential.
\lstinputlisting[language=turtle]{artifacts/kg/listings-computed/bartok2018si/q09-run-and-model.ttl}

\paragraph{Q10 -- Benchmark results.}
The headline accuracy figure is the median absolute force-component
error on the held-out testing set (grain-boundary, di-interstitial,
GSF-path, amorphous configurations): 0.025\,eV/\AA. The paper
reports many additional per-property and per-phase comparisons (lattice
constants, elastic moduli, surface energies, defect formation
energies, EOS curves) but in derived form rather than as separate
$\BenchmarkResultC$ instances.
\lstinputlisting[language=turtle]{artifacts/kg/listings-computed/bartok2018si/q10-benchmark-results.ttl}

\paragraph{Q11 -- Computational resources.}
\emph{Not reported.}
\lstinputlisting[language=turtle]{artifacts/kg/listings-computed/bartok2018si/q11-resources.ttl}

\paragraph{Q12 -- Gaps.}
Beyond Q11, the paper does not record explicit values for: the
per-class force/virial regularisations $\sigma_F$, $\sigma_V$ (only
their existence and qualitative role); the SOAP $l_{\max}$ and
$n_{\max}$ basis indices that are referenced under
$\dlRole{hasHyperparameter}$ but lack
$\HyperparameterSettingC$ instances; and any $\dlRole{frozenCore}$
treatment beyond the ultrasoft pseudopotential. The paper also
sketches several downstream applications (crack tip propagation,
amorphous-Si melt-quench MD, recrystallisation kinetics) which are
not encoded as separate $\dlConcept{MLIPRun}$ instances.

\subsection{Wang et al.\ (2018): DeePMD-kit, water demonstration}
\label{sec:appendix-example-wang2018dpkit}

\paragraph{Q1 -- Bibliographic identification.}
Wang, Zhang, Han, and E document DeePMD-kit, a TensorFlow-based deep
learning package for many-body potential energy representation and
molecular dynamics. Published in \emph{Comput.\ Phys.\ Commun.} 228,
178--184 (2018); arXiv:1712.03641. We exercise the protocol on the
single liquid-water demonstration in \S IV; the package's broader
multi-system, multi-method capabilities are flagged as the central
gap in Q12.
\lstinputlisting[language=turtle]{artifacts/kg/listings-computed/wang2018dpkit/q01-bibliographic.ttl}

\paragraph{Q2 -- Material system.}
The demonstration system is liquid water (64 H$_2$O molecules under
periodic boundary conditions, ${\sim}330$\,K).
\lstinputlisting[language=turtle]{artifacts/kg/listings-computed/wang2018dpkit/q02-material.ttl}

\paragraph{Q3 -- Reference calculation method.}
Reference data come from DFT calculations.
\lstinputlisting[language=turtle]{artifacts/kg/listings-computed/wang2018dpkit/q03-reference-method-type.ttl}

\paragraph{Q4 -- Reference settings.}
The functional is PBE0+TS (hybrid PBE0 with Tkatchenko--Scheffler
dispersion). Other DFT settings (engine, plane-wave or basis-set
choice, k-mesh) are not stated in the demonstration section.
\lstinputlisting[language=turtle]{artifacts/kg/listings-computed/wang2018dpkit/q04-reference-settings.ttl}

\paragraph{Q5 -- Training dataset.}
The reference set is 40{,}000 frames from a 20\,ps NVT AIMD trajectory
of 64 water molecules at 330\,K (38{,}000 train + 2{,}000 test).
Energies and forces are covered. Provenance is \emph{published}
(reused from the underlying AIMD work).
\lstinputlisting[language=turtle]{artifacts/kg/listings-computed/wang2018dpkit/q05-dataset.ttl}

\paragraph{Q6 -- Sampling strategies.}
A single sampling strategy applies: vibrational sampling via NVT
AIMD with a 0.0005\,ps stride.
\lstinputlisting[language=turtle]{artifacts/kg/listings-computed/wang2018dpkit/q06-sampling.ttl}

\paragraph{Q7 -- MLIP method, components, implementation.}
The method is Deep Potential Molecular Dynamics (DeePMD): a per-atom
DNN over local-frame radial+angular descriptors (1/R, x/R, y/R,
z/R) sorted by chemical species, with five hidden layers (240, 120,
60, 30, 10) and tanh activations. Loss is a weighted MSE on energy
and force (no virial) with annealed prefactors. Optimisation is Adam
with exponential learning-rate decay. The implementation is
DeePMD-kit v0.1 (Python/C++ on TensorFlow), with LAMMPS pair-style
\texttt{deepmd} and i-PI \texttt{dp\_ipi} clients shipping in the
package.
\lstinputlisting[language=turtle]{artifacts/kg/listings-computed/wang2018dpkit/q07-method.ttl}

\paragraph{Q8 -- Hyperparameter settings.}
Reported settings: cutoff radius $r_c = 6.0$\,\AA; per-atom hidden
architecture $(240, 120, 60, 30, 10)$; initial learning rate
$10^{-3}$; learning-rate decay rate 0.95; decay applied every 5{,}000
steps.
\lstinputlisting[language=turtle]{artifacts/kg/listings-computed/wang2018dpkit/q08-hyperparameter-settings.ttl}

\paragraph{Q9 -- MLIP run and trained model.}
A single run of $10^6$ batches at batch size 4 produces the
demonstration DeePMD water model.
\lstinputlisting[language=turtle]{artifacts/kg/listings-computed/wang2018dpkit/q09-run-and-model.ttl}

\paragraph{Q10 -- Benchmark results.}
Test-set RMSEs: 0.028\,eV on the 64-water-cell total energy and
0.024\,eV/\AA{} on atomic-force components.
\lstinputlisting[language=turtle]{artifacts/kg/listings-computed/wang2018dpkit/q10-benchmark-results.ttl}

\paragraph{Q11 -- Computational resources.}
The paper records training duration of approximately 16 hours on an
Intel Core i7-3770 CPU with 32\,GB RAM and 4 OpenMP threads. GPU
hours, peak memory, and per-atom inference time are not recorded.
\lstinputlisting[language=turtle]{artifacts/kg/listings-computed/wang2018dpkit/q11-resources.ttl}

\paragraph{Q12 -- Gaps.}
The central gap is one of \emph{scope}: this is a software paper,
and the protocol's "one paper $\equiv$ one method $\times$ one material
system" framing captures only the demonstration example. The
package's multi-system support, alternate descriptor choices,
ensemble training, and downstream MD/i-PI tooling are not encoded.
The demonstration's DFT settings (engine, k-mesh, energy cutoff) are
also not surfaced beyond the functional choice; and inference-time
data are not reported even though the paper benchmarks DeePMD-LAMMPS
performance qualitatively.

\subsection{Lysogorskiy et al.\ (2021): PACE applied to Cu and Si}
\label{sec:appendix-example-lysogorskiy2021ace}

\paragraph{Q1 -- Bibliographic identification.}
Lysogorskiy, van der Oord, Bochkarev, Menon, Rinaldi, Hammerschmidt,
Mrovec, Thompson, Cs\'anyi, Ortner, and Drautz introduce PACE, a
performant C++ implementation of the Atomic Cluster Expansion
(\cite{drautz2019ace}) released as a LAMMPS pair-style. Production
parameterisations are presented for fcc copper and diamond silicon.
Published in \emph{npj Computational Materials} 7, 97 (2021);
doi:10.1038/s41524-021-00559-9.
\lstinputlisting[language=turtle]{artifacts/kg/listings-computed/lysogorskiy2021ace/q01-bibliographic.ttl}

\paragraph{Q2 -- Material system.}
Two single-element systems are encoded as separate
$\MaterialSystemC$ instances: copper (fcc, with bcc/dhcp/hcp
polymorphs and surfaces/defects in the reference set) and silicon
(diamond cubic, with reference data from the GAP-Si general-purpose
database of Bart\'ok et al.\ 2018).
\lstinputlisting[language=turtle]{artifacts/kg/listings-computed/lysogorskiy2021ace/q02-material.ttl}

\paragraph{Q3 -- Reference calculation method.}
Both reference datasets come from DFT calculations.
\lstinputlisting[language=turtle]{artifacts/kg/listings-computed/lysogorskiy2021ace/q03-reference-method-type.ttl}

\paragraph{Q4 -- Reference settings.}
Cu reference data are produced with FHI-aims at the PBE level
(small clusters, bulk phases, surfaces and slabs, configurations
with displaced or missing atoms), partly assembled with the
\texttt{pyiron} workflow framework. Si reference data are the
published GAP-Si general-purpose fitting database from CASTEP
(PW91 inherited from Bart\'ok et al.\ 2018).
\lstinputlisting[language=turtle]{artifacts/kg/listings-computed/lysogorskiy2021ace/q04-reference-settings.ttl}

\paragraph{Q5 -- Training dataset.}
Cu: in-house FHI-aims set covering clusters, bulk polymorphs,
surfaces, slabs, and defects (configuration count not enumerated in
the main text). Si: the published Bart\'ok-2018 GAP-Si database of
2{,}475 configurations covering crystal phases, surfaces, vacancies,
interstitials, and liquid Si. Cu covers energies+forces; Si
additionally covers virials.
\lstinputlisting[language=turtle]{artifacts/kg/listings-computed/lysogorskiy2021ace/q05-dataset.ttl}

\paragraph{Q6 -- Sampling strategies.}
Cu: configurational coverage of clusters, polymorphs, surfaces, and
defective structures. Si: reuse of the published GAP-Si database's
sampling strategy.
\lstinputlisting[language=turtle]{artifacts/kg/listings-computed/lysogorskiy2021ace/q06-sampling.ttl}

\paragraph{Q7 -- MLIP method, components, implementation.}
The method is the Atomic Cluster Expansion of Drautz~2019: per-atom
property $\varphi_i$ expanded over body-ordered multi-atom basis
functions, with truncation at body order $\nu_{\max}$ and total
energy as a sum over per-atom embeddings. The Cu fit uses a
nonlinear two-density Finnis--Sinclair embedding; the Si fit uses
a linear embedding at $\nu_{\max}=4$. Implementation: PACE
(C++ LAMMPS \texttt{pair\_style}).
\lstinputlisting[language=turtle]{artifacts/kg/listings-computed/lysogorskiy2021ace/q07-method.ttl}

\paragraph{Q8 -- Hyperparameter settings.}
Reported settings: Cu cutoff radius $r_c = 7.4$\,\AA; Cu basis size
$2{,}072$ parameters (756 expansion coefficients per atomic density
$\times$ 2 densities + 560 radial-function parameters); Si linear
basis size $6{,}827$ functions at body order $\nu_{\max}=4$.
\lstinputlisting[language=turtle]{artifacts/kg/listings-computed/lysogorskiy2021ace/q08-hyperparameter-settings.ttl}

\paragraph{Q9 -- MLIP run and trained model.}
Two training runs are encoded: one for PACE-Cu (nonlinear
Finnis--Sinclair embedding) and one for PACE-Si (linear ACE,
$\nu_{\max}=4$). Both produce a single $\TrainedModelC$ with a
recorded per-atom inference time
($\dlRole{inferenceTimePerAtom}=0.00032$\,s/atom for Cu,
$0.0008$\,s/atom for Si).
\lstinputlisting[language=turtle]{artifacts/kg/listings-computed/lysogorskiy2021ace/q09-run-and-model.ttl}

\paragraph{Q10 -- Benchmark results.}
PACE-Cu attains a final fit error of 2.9\,meV/atom on structures
within 1\,eV of the ground state (after ground-state fine-tuning),
with force errors near 15\,meV/\AA{} on higher-energy structures.
PACE-Si matches GAP-Si accuracy on the Bart\'ok-2018 database with
energy MAE 1.81\,meV/atom and force MAE 82\,meV/\AA. Three
$\BenchmarkResultC$ instances cover these.
\lstinputlisting[language=turtle]{artifacts/kg/listings-computed/lysogorskiy2021ace/q10-benchmark-results.ttl}

\paragraph{Q11 -- Computational resources.}
The headline claim is computational performance: PACE shifts the
accuracy/cost Pareto front for ML interatomic potentials. We capture
the per-atom inference times on the trained models in Q9
(0.32\,ms/atom for Cu, 0.80\,ms/atom for Si on a single force
call). Training duration, GPU/CPU hours, training hardware, and
peak memory are not reported.
\lstinputlisting[language=turtle]{artifacts/kg/listings-computed/lysogorskiy2021ace/q11-resources.ttl}

\paragraph{Q12 -- Gaps.}
The protocol's "one paper $\equiv$ one method $\times$ one material"
framing is mildly stretched: the paper trains \emph{two} ACE
parameterisations (Cu and Si) and we encode both within one
\texttt{lysogorskiy2021ace.ttl} via two $\MaterialSystemC$, two
$\dlConcept{MLIPRun}$, and two $\TrainedModelC$ instances rather
than splitting into sibling files. Other unreported items: explicit
DFT settings on the Cu side beyond engine and functional (energy
cutoff, k-mesh, pseudopotential / NAO basis); explicit
$\dlRole{numConfigurations}$ for the Cu reference set; the Pareto
benchmark is reported as a derived figure rather than an
$\AccuracyMetricC$ across all five models compared (ACE, EAM,
SNAP, GTINV, GAP); training-cost metadata.

\subsection{Smith et al.\ (2019): ANI-1ccx via DFT$\to$CCSD(T) transfer learning}
\label{sec:appendix-example-smith2019ccx}

\paragraph{Q1 -- Bibliographic identification.}
Smith, Nebgen, Zubatyuk, Lubbers, Devereux, Barros, Tretiak, Isayev,
and Roitberg present ANI-1ccx, a neural network potential pretrained
on a large DFT dataset (ANI-1x) and fine-tuned by transfer learning
on a much smaller DLPNO-CCSD(T)/CBS dataset, achieving coupled-cluster
accuracy on H/C/N/O organic molecules. Published in
\emph{Nat.\ Commun.} 10, 2903 (2019); arXiv:1903.10405.
\lstinputlisting[language=turtle]{artifacts/kg/listings-computed/smith2019ccx/q01-bibliographic.ttl}

\paragraph{Q2 -- Material system.}
The MLIP covers the H/C/N/O chemical space of small organic molecules
(drug-like and reaction-relevant species, average ${\sim}15$ atoms per
molecule, primarily from the GDB-11 universe of $\sim$57k distinct
species). We model this as one $\MaterialSystemC$ at the
chemical-space level rather than enumerating individual molecules.
\lstinputlisting[language=turtle]{artifacts/kg/listings-computed/smith2019ccx/q02-material.ttl}

\paragraph{Q3 -- Reference calculation method.}
Two reference calculations apply: a wave-function reference (CCSD(T))
for the headline fine-tuning step, and a DFT reference for the
pretraining step. This is the corpus' only
$\dlConcept{WaveFunctionCalculation}$ entry.
\lstinputlisting[language=turtle]{artifacts/kg/listings-computed/smith2019ccx/q03-reference-method-type.ttl}

\paragraph{Q4 -- Reference settings.}
The wave-function reference is DLPNO-CCSD(T)/CBS (the localised
DLPNO-CCSD(T) method of Neese et al., extrapolated from cc-pVDZ /
cc-pVTZ), denoted CCSD(T)* in the paper and computed in ORCA. The
DFT pretraining reference is $\omega$B97X with the 6-31G* Gaussian
basis (all-electron, $\Gamma$-only, isolated-molecule limit).
\lstinputlisting[language=turtle]{artifacts/kg/listings-computed/smith2019ccx/q04-reference-settings.ttl}

\paragraph{Q5 -- Training dataset.}
The CCSD(T)*/CBS fine-tuning set has ${\sim}500{,}000$ data points
covering ${\sim}480{,}000$ molecules; the pretraining ANI-1x corpus
has 5\,M conformations from $\sim$57k molecules. Energies are
covered (no forces or stresses). Provenance is \emph{augmented}
(CCSD(T) labels are added on top of an existing DFT-labelled set).
\lstinputlisting[language=turtle]{artifacts/kg/listings-computed/smith2019ccx/q05-dataset.ttl}

\paragraph{Q6 -- Sampling strategies.}
Two strategies were combined: query-by-committee active learning
(ANI's ensemble-disagreement loop drives selection of new molecules
for CCSD(T)*/CBS computation, growing the dataset to ${\sim}480$k
molecules) and the diverse-conformer / normal-mode sampling that
underlies the ANI-1x parent corpus.
\lstinputlisting[language=turtle]{artifacts/kg/listings-computed/smith2019ccx/q06-sampling.ttl}

\paragraph{Q7 -- MLIP method, components, implementation.}
The method is the ANI Behler--Parrinello-style high-dimensional NN
with ANI atom-centred symmetry functions. Per-element
fully-connected feed-forward subnetworks operate on the
symmetry-function inputs; total energy is the sum of per-atom
contributions. Loss is energy-only MSE on per-molecule total energies
(no force or stress targets). Optimisation is Adam with exponential
learning-rate decay; the implementation is TorchANI.
\lstinputlisting[language=turtle]{artifacts/kg/listings-computed/smith2019ccx/q07-method.ttl}

\paragraph{Q8 -- Hyperparameter settings.}
The main text does not tabulate explicit numerical values for the
cutoff radius or for the radial/angular symmetry-function index
counts (Section S1.2 in the Supplementary Information has the
details). We therefore record no $\HyperparameterSettingC$ instances
in the canonical encoding; the architectural hyperparameters are
referenced under $\dlRole{hasHyperparameter}$ only.
\lstinputlisting[language=turtle]{artifacts/kg/listings-computed/smith2019ccx/q08-hyperparameter-settings.ttl}

\paragraph{Q9 -- MLIP run and trained model.}
The training run pretrains on the DFT (ANI-1x) corpus and then
fine-tunes on the CCSD(T)*/CBS corpus, producing the ANI-1ccx
8-network ensemble.
\lstinputlisting[language=turtle]{artifacts/kg/listings-computed/smith2019ccx/q09-run-and-model.ttl}

\paragraph{Q10 -- Benchmark results.}
Headline accuracies (Table 1 + Fig.\ 2): MAE 1.46\,kcal/mol vs.\
CCSD(T)*/CBS on the GDB-10to13 atomization-energy benchmark, and
MAD 2.5\,kcal/mol on the HC7/11 hydrocarbon reaction-energy
benchmark. The paper additionally reports ISOL6 isomerisation
energies and the Genentech torsion-profile benchmark; these are not
encoded as separate $\BenchmarkResultC$ instances.
\lstinputlisting[language=turtle]{artifacts/kg/listings-computed/smith2019ccx/q10-benchmark-results.ttl}

\paragraph{Q11 -- Computational resources.}
The paper notes ${\sim}30$\,minutes per model to train on the
$500$k-molecule CCSD(T)*/CBS set (the production model is an
8-network ensemble), but does not report training duration in the
form the ontology expects, nor GPU hours, training hardware, peak
memory, or per-atom inference time. Recorded as a Q12 gap.
\lstinputlisting[language=turtle]{artifacts/kg/listings-computed/smith2019ccx/q11-resources.ttl}

\paragraph{Q12 -- Gaps.}
Beyond Q11, the protocol does not capture: the
\emph{transfer-learning relationship} between the DFT-pretrained and
CCSD(T)-fine-tuned models (only their union as a single training
set); the explicit DLPNO-CCSD(T) parameters (PNO thresholds,
integration-grid specifications) which would refine the
$\dlConcept{WaveFunctionSettings}$; the ANI-1x parent dataset is
referenced as a published source but not modelled as a separate
$\dlConcept{TrainingDataset}$; and the auxiliary ablation runs
(ANI-1ccx-R, trained only on CCSD(T)*/CBS, no transfer learning)
are not encoded as separate $\dlConcept{TrainedModel}$ instances.

\subsection{Eckhoff \& Behler (2021): spin-dependent HDNNP for MnO}
\label{sec:appendix-example-eckhoff2021spin}

\paragraph{Q1 -- Bibliographic identification.}
Eckhoff and Behler extend the high-dimensional neural network potential
framework with spin-dependent atom-centered symmetry functions (sACSFs)
to capture magnetic systems. Published in \emph{npj Comput.\ Mater.} 7,
170 (2021); arXiv:2104.14439.
\lstinputlisting[language=turtle]{artifacts/kg/listings-computed/eckhoff2021spin/q01-bibliographic.ttl}

\paragraph{Q2 -- Material system.}
The headline material is rock-salt MnO (and a vacancy-rich
Mn$_{0.969}$O variant). The AFM-II rhombohedrally distorted ground
state with N\'eel temperature ${\sim}116$\,K is the focus.
\lstinputlisting[language=turtle]{artifacts/kg/listings-computed/eckhoff2021spin/q02-material.ttl}

\paragraph{Q3 -- Reference calculation method.}
Reference data come from spin-polarised DFT calculations.
\lstinputlisting[language=turtle]{artifacts/kg/listings-computed/eckhoff2021spin/q03-reference-method-type.ttl}

\paragraph{Q4 -- Reference settings.}
Calculations use FHI-aims (version 200112.2) with the HSE06
screened-hybrid functional ($\omega = 0.11\,a_0^{-1}$), all-electron
numerical atom-centred orbitals (intermediate basis, excluding
auxiliary 5g hydrogenic functions), and a $\Gamma$-centred
$2{\times}2{\times}2$ k-mesh on $2{\times}2{\times}2$ MnO supercells.
\lstinputlisting[language=turtle]{artifacts/kg/listings-computed/eckhoff2021spin/q04-reference-settings.ttl}

\paragraph{Q5 -- Training dataset.}
The reference set contains 3{,}101 supercells across magnetic states:
1{,}387 MnO + 1{,}421 Mn$_{0.969}$O for training, 156 MnO + 137
Mn$_{0.969}$O for testing. Energies and forces are covered.
Provenance is \emph{in-house} (assembled for this work).
\lstinputlisting[language=turtle]{artifacts/kg/listings-computed/eckhoff2021spin/q05-dataset.ttl}

\paragraph{Q6 -- Sampling strategies.}
Four strategies were combined: magnetic-state enumeration (FM, AFM-I,
AFM-II, plus excited spin orderings); atomic-displacement
perturbations from ideal lattice positions; lattice-parameter
perturbations to capture the rhombohedral distortion; and Mn-vacancy
sampling for off-stoichiometric configurations.
\lstinputlisting[language=turtle]{artifacts/kg/listings-computed/eckhoff2021spin/q06-sampling.ttl}

\paragraph{Q7 -- MLIP method, components, implementation.}
The method (mHDNNP) is a Behler--Parrinello HDNNP with three hidden
layers (20/15/10 neurons per element) consuming spin-augmented ACSFs:
radial $M^0/M^+/M^-$ and angular
$M^{00}/M^{++}/M^{--}/M^{+-}$ functions. Implementation is RuNNer
v1.00 (modified to support sACSFs); n2p2 is used for LAMMPS MD.
\lstinputlisting[language=turtle]{artifacts/kg/listings-computed/eckhoff2021spin/q07-method.ttl}

\paragraph{Q8 -- Hyperparameter settings.}
Reported settings: cutoff radius $r_c = 10.5\,a_0$ (Bohr); ACSF
amplitude threshold $M_{\text{thresh}} = 0.25$; per-element
architecture three layers $\times$ (20, 15, 10) neurons.
\lstinputlisting[language=turtle]{artifacts/kg/listings-computed/eckhoff2021spin/q08-hyperparameter-settings.ttl}

\paragraph{Q9 -- MLIP run and trained model.}
A single training run produces the production mHDNNP for MnO.
\lstinputlisting[language=turtle]{artifacts/kg/listings-computed/eckhoff2021spin/q09-run-and-model.ttl}

\paragraph{Q10 -- Benchmark results.}
Test-set RMSEs: 1.11\,meV/atom on cohesive energies and
0.066\,eV/\AA{} on atomic-force components. The paper also reports
phonon and N\'eel-temperature predictions but in derived form rather
than as separate $\BenchmarkResultC$ instances.
\lstinputlisting[language=turtle]{artifacts/kg/listings-computed/eckhoff2021spin/q10-benchmark-results.ttl}

\paragraph{Q11 -- Computational resources.}
\emph{Not reported.}
\lstinputlisting[language=turtle]{artifacts/kg/listings-computed/eckhoff2021spin/q11-resources.ttl}

\paragraph{Q12 -- Gaps.}
Beyond Q11, the most consequential gap is the lack of first-class
ontology coverage for \emph{spin-dependent} descriptors and magnetic
configurations. The mHDNNP's spin-augmentation is currently captured
only through prose in the $\dlConcept{FunctionalForm}$ label and the
magnetic-state-sampling strategy; the ontology has no concept for
collinear/non-collinear spin states, magnetic-moment-per-atom
properties, or spin descriptors. Other unreported items: the FHI-aims
plane-wave-equivalent energy cutoff (replaced by basis-set choice in
NAO); explicit $\dlRole{frozenCore}$; and any HSE06 numerical
parameters beyond the screening parameter $\omega$.

\subsection{Batatia et al.\ (2022): MACE for fast and accurate force fields}
\label{sec:appendix-example-batatia2022}

\paragraph{Q1 -- Bibliographic identification.}
Batatia, Kov\'acs, Simm, Ortner, and Cs\'anyi introduce MACE, a
higher-order equivariant message-passing neural network interatomic
potential. Published at NeurIPS 2022; arXiv:2206.07697. The headline
benchmarks are rMD17 small organic molecules, 3BPA temperature
transferability, and acetylacetone (AcAc) flexibility/reactivity.
\lstinputlisting[language=turtle]{artifacts/kg/listings-computed/batatia2022/q01-bibliographic.ttl}

\paragraph{Q2 -- Material system.}
The canonical encoding pass takes the headline rMD17 dataset, a set
of ten small organic molecules (aspirin, azobenzene, benzene,
ethanol, malonaldehyde, naphthalene, paracetamol, salicylic acid,
toluene, uracil) sampled from gas-phase ab initio MD trajectories.
We model the dataset at the species-set level (C/H/N/O small
organics) rather than per molecule.
\lstinputlisting[language=turtle]{artifacts/kg/listings-computed/batatia2022/q02-material.ttl}

\paragraph{Q3 -- Reference calculation method.}
Reference data are produced from DFT calculations.
\lstinputlisting[language=turtle]{artifacts/kg/listings-computed/batatia2022/q03-reference-method-type.ttl}

\paragraph{Q4 -- Reference settings.}
For rMD17 the original MD17 trajectories are recomputed at the
PBE/def2-SVP level of theory with very tight SCF convergence and a
very dense DFT integration grid. The DFT engine itself is not stated
in the MACE paper (the rMD17 dataset is consumed as published).
\lstinputlisting[language=turtle]{artifacts/kg/listings-computed/batatia2022/q04-reference-settings.ttl}

\paragraph{Q5 -- Training dataset.}
The rMD17 split used in the paper consists of 950 training and 50
validation configurations per molecule (1{,}000 total per molecule),
covering energies and forces. Configurations come from the published
rMD17 dataset and are reshuffled after each epoch.
\lstinputlisting[language=turtle]{artifacts/kg/listings-computed/batatia2022/q05-dataset.ttl}

\paragraph{Q6 -- Sampling strategies.}
A single sampling strategy is in play: vibrational sampling via
gas-phase ab initio MD trajectories at 500~K, recomputed at
PBE/def2-SVP for rMD17.
\lstinputlisting[language=turtle]{artifacts/kg/listings-computed/batatia2022/q06-sampling.ttl}

\paragraph{Q7 -- MLIP method, components, implementation.}
MACE is a two-layer equivariant message-passing GNN that uses
higher body-order (4-body) tensor messages built from a complete
ACE-style local basis (Bessel radial $\times$ spherical-harmonic
angular) and tensor-product symmetrisation with generalised
Clebsch--Gordan coefficients. The training algorithm is the
AMSGrad variant of Adam ($\beta_1=0.9$, $\beta_2=0.999$,
$\epsilon=10^{-8}$), with an on-plateau learning-rate scheduler
(patience 50, decay factor 0.8). The loss is a weighted MSE with
$\lambda_E=1$ and $\lambda_F=1000$. Implementation is the
\texttt{mace} PyTorch code (\url{https://github.com/ACEsuit/mace}).
\lstinputlisting[language=turtle]{artifacts/kg/listings-computed/batatia2022/q07-method.ttl}

\paragraph{Q8 -- Hyperparameter settings.}
The rMD17 production model uses two MACE layers, $l_{\max}=3$, 256
uncoupled feature channels, correlation order $\nu=3$ (4-body
messages), and a 5~\AA{} radial cutoff with 8 Bessel basis
functions and a polynomial cutoff envelope ($p=5$). The radial MLP
is $[64, 64, 64, 1024]$ with SiLU non-linearities, and the readout
of the second layer is a single-layer MLP with 16 hidden
dimensions; first-layer readout is linear.
\lstinputlisting[language=turtle]{artifacts/kg/listings-computed/batatia2022/q08-hyperparameter-settings.ttl}

\paragraph{Q9 -- MLIP run and trained model.}
A single training run per molecule produces one rMD17 MACE model
(trained at \texttt{float32} precision, learning rate 0.01, batch
size 5, exponential weight-decay $5\times10^{-7}$ on the $W$
weights of the higher-order features). Per-atom energy is
shifted by the per-species training-set average and scaled by the
RMS of force components.
\lstinputlisting[language=turtle]{artifacts/kg/listings-computed/batatia2022/q09-run-and-model.ttl}

\paragraph{Q10 -- Benchmark results.}
Headline accuracy figures on the rMD17 benchmark for aspirin
(MACE, 1{,}000 train) are MAE 2.2~meV on energies and
6.6~meV/\AA{} on force components. MACE attains state-of-the-art
or near state-of-the-art accuracy across the ten rMD17 molecules.
We encode the aspirin numbers as a single per-molecule benchmark
result; per-molecule results for the remaining nine molecules are a
Q12 gap.
\lstinputlisting[language=turtle]{artifacts/kg/listings-computed/batatia2022/q10-benchmark-results.ttl}

\paragraph{Q11 -- Computational resources.}
Models were trained on a single NVIDIA A100 GPU; typical training
time is 2--6~hours per molecule for rMD17 (specific GPU-hours,
peak memory, and inference-time-per-atom are not reported for the
rMD17 pass, though the paper does report a ``time latency''
inference figure for the 3BPA pass: 24.3~ms for a structure with
$\le 10000$ atoms on an A100).
\lstinputlisting[language=turtle]{artifacts/kg/listings-computed/batatia2022/q11-resources.ttl}

\paragraph{Q12 -- Gaps.}
Beyond the items already flagged, the following ontology-expressible
information is not represented in this encoding pass: (i) the two
additional benchmarks reported in the same paper, 3BPA
(temperature-transferability of a flexible drug-like molecule, with
$\omega$B97X/6-31G(d) reference) and AcAc (acetylacetone
flexibility/reactivity at the same level); (ii) per-molecule
benchmark results for the remaining nine rMD17 molecules; (iii) the
DFT engine for the rMD17 reference recomputation
($\dlRole{usedDFTCode}$ on the DFT settings) and the original MD17
DFT settings beyond the XC functional/basis set; (iv) the MACE
package version; (v) detailed compute-cost data (GPU-hours, peak
memory, inference time per atom on rMD17). The paper also reports
that one MACE training run on 3BPA reaches converged BOTNet
accuracy in $\sim$30~min on an A100 and a fully converged 3BPA
model takes $>$1~day; this finer-grained timing is associated
with 3BPA rather than rMD17 and is not encoded here.

\subsection{Batzner et al.\ (2022): NequIP E(3)-equivariant GNN MLIP}
\label{sec:appendix-example-batzner2022}

\paragraph{Q1 -- Bibliographic identification.}
Batzner, Musaelian, Sun, Geiger, Mailoa, Kornbluth, Molinari, Smidt,
and Kozinsky introduce NequIP, an $E(3)$-equivariant graph
neural-network interatomic potential. Published in
\emph{Nature Communications} 13, 2453 (2022); arXiv:2101.03164.
The paper benchmarks NequIP across MD17/rMD17 small molecules,
Molecules@CCSD/CCSD(T), liquid water + ice phases, formate
decomposition on Cu(110), Li$_4$P$_2$O$_7$ amorphous glass, and
the LiPS superionic conductor.
\lstinputlisting[language=turtle]{artifacts/kg/listings-computed/batzner2022/q01-bibliographic.ttl}

\paragraph{Q2 -- Material system.}
The canonical encoding pass takes the bulk water and ice phases
benchmark as the headline extended-system application: 64 H$_2$O
molecules in the liquid phase, 96 H$_2$O molecules in three ice Ih
phases (1~bar/273~K, 1~bar/330~K, 2.13~kbar/238~K).
\lstinputlisting[language=turtle]{artifacts/kg/listings-computed/batzner2022/q02-material.ttl}

\paragraph{Q3 -- Reference calculation method.}
Reference data are produced from DFT calculations.
\lstinputlisting[language=turtle]{artifacts/kg/listings-computed/batzner2022/q03-reference-method-type.ttl}

\paragraph{Q4 -- Reference settings.}
The water + ices reference data are computed at the PBE0-TS level
of theory (hybrid PBE0 with Tkatchenko--Scheffler dispersion),
generated from a mix of classical AIMD and path-integral AIMD
trajectories. The DFT engine for these reference data is not stated
in the paper (it inherits the published Cheng et al.\ dataset).
\lstinputlisting[language=turtle]{artifacts/kg/listings-computed/batzner2022/q04-reference-settings.ttl}

\paragraph{Q5 -- Training dataset.}
The water + ices joint training set used in the data-efficiency
experiment consists of 133 structures sampled uniformly from a
140{,}000-frame full set (100{,}000 liquid water; 20{,}000 ice
Ih~b; 10{,}000 ice Ih~c; 10{,}000 ice Ih~d). The training labels
are energies and forces.
\lstinputlisting[language=turtle]{artifacts/kg/listings-computed/batzner2022/q05-dataset.ttl}

\paragraph{Q6 -- Sampling strategies.}
A single sampling strategy is in play: vibrational sampling via
classical and path-integral AIMD trajectories at multiple
temperatures and pressures.
\lstinputlisting[language=turtle]{artifacts/kg/listings-computed/batzner2022/q06-sampling.ttl}

\paragraph{Q7 -- MLIP method, components, implementation.}
NequIP is an $E(3)$-equivariant message-passing GNN that uses
tensor-product convolutions over geometric tensors of irreducible
$O(3)$ representations of order $\ell=0\dots l_{\max}$, parities
$p\in\{-1,1\}$, with SiLU/tanh gate non-linearities, a ResNet-style
self-interaction--convolution--concatenation update, and a final
atom-wise output block. The training algorithm is the AMSGrad
variant of Adam ($\beta_1=0.9$, $\beta_2=0.999$, $\epsilon=10^{-8}$,
no weight decay) with an on-plateau learning-rate scheduler
(patience 50, decay factor 0.8). The loss is a weighted MSE on
energies and force components ($\lambda_F=100\,000$, $\lambda_E=1$
for the model~c run reported below). Implementation is the
\texttt{nequip} PyTorch code, version 0.3.3.
\lstinputlisting[language=turtle]{artifacts/kg/listings-computed/batzner2022/q07-method.ttl}

\paragraph{Q8 -- Hyperparameter settings.}
The water + ices model uses 6 interaction blocks, $l_{\max}=2$, 32
features (with both even and odd parity), and a 6~\AA{} radial
cutoff. Radial features are 8 trainable Bessel basis functions
processed by a 3-hidden-layer MLP of size 64 with SiLU
non-linearities. Learning rate 0.005, batch size 1 (with weight
0.99 EMA on training weights for evaluation).
\lstinputlisting[language=turtle]{artifacts/kg/listings-computed/batzner2022/q08-hyperparameter-settings.ttl}

\paragraph{Q9 -- MLIP run and trained model.}
A single training run on 133 structures produces one trained
NequIP water+ices model. Three loss-weighting variants are reported
(model~a: $\lambda_F=1,\lambda_E=0$; model~b: $\lambda_F=100$;
model~c: $\lambda_F=100\,000$); we encode model~c as the headline.
\lstinputlisting[language=turtle]{artifacts/kg/listings-computed/batzner2022/q09-run-and-model.ttl}

\paragraph{Q10 -- Benchmark results.}
Headline figures for the water + ices test set (model~c,
$\lambda_F=100\,000$, $\lambda_E=1$, 133 train) are RMSEs of
1.7~meV per H$_2$O molecule on energies and 12.2~meV/\AA{} on
liquid-water force components. NequIP is competitive with DeepMD
trained on $\sim$1000$\times$ more data and significantly
outperforms it on three of four phases. Per-phase results for the
three ice configurations are an additional Q12 gap.
\lstinputlisting[language=turtle]{artifacts/kg/listings-computed/batzner2022/q10-benchmark-results.ttl}

\paragraph{Q11 -- Computational resources.}
All NequIP models were trained on an NVIDIA Tesla V100 GPU in
single-GPU training. The paper does not report a specific training
duration or GPU-hour count for the water+ices run (only that
``competitive results can typically be obtained within a matter of
hours or often even minutes''), nor peak memory or inference time
per atom for the headline pass.
\lstinputlisting[language=turtle]{artifacts/kg/listings-computed/batzner2022/q11-resources.ttl}

\paragraph{Q12 -- Gaps.}
Beyond Q11, the following ontology-expressible items are not
represented in this encoding pass: (i) the six additional
benchmarks reported in the same paper (MD17/rMD17 small molecules,
Molecules@CCSD/CCSD(T), formate decomposition on Cu$\langle 110
\rangle$, Li$_4$P$_2$O$_7$ amorphous glass, LiPS superionic
conductor); (ii) per-phase benchmark results for the three ice Ih
configurations and the data-efficiency learning-curve sweep across
$\{10, 100, 1000, 2500\}$ training-set sizes for LiPS; (iii) the
DFT engine for the water reference data ($\dlRole{usedDFTCode}$);
(iv) detailed PBE0-TS reference settings (k-point mesh, energy
cutoff, pseudopotential type); (v) the PyTorch / PyTorch
Geometric / e3nn / Python versions used for training (the paper
does report all of these but our protocol does not capture
secondary toolchain libraries beyond the MLIP \texttt{Library}
hook); (vi) the AMSGrad-Adam learning-rate schedule on a per-system
basis. The paper also reports a ``temperature transferability''
benchmark on 3BPA but only as cited in
the Allegro paper (Musaelian et al.~2023); this is not encoded here.

\subsection{Chen \& Ong (2022): M3GNet universal graph IAP}
\label{sec:appendix-example-chen2022}

\paragraph{Q1 -- Bibliographic identification.}
Chen and Ong introduce M3GNet, a materials graph network with explicit
three-body interactions trained as a universal interatomic potential
across the periodic table. Published in \emph{Nat.\ Comput.\ Sci.} 2,
718--728 (2022); arXiv:2202.02450.
\lstinputlisting[language=turtle]{artifacts/kg/listings-computed/chen2022/q01-bibliographic.ttl}

\paragraph{Q2 -- Material system.}
The MLIP covers Materials Project crystals across 89 elements
(62{,}783 distinct compounds spanning binaries, ternaries, and
quaternaries from the MPF.2021.2.8 subset). We model this as a single
$\MaterialSystemC$ at the database-coverage level rather than
enumerating individual compounds.
\lstinputlisting[language=turtle]{artifacts/kg/listings-computed/chen2022/q02-material.ttl}

\paragraph{Q3 -- Reference calculation method.}
Reference data come from DFT calculations.
\lstinputlisting[language=turtle]{artifacts/kg/listings-computed/chen2022/q03-reference-method-type.ttl}

\paragraph{Q4 -- Reference settings.}
DFT settings inherit the Materials Project defaults: VASP with PBE
GGA (or PBE+U for transition-metal-containing compounds) and PAW
pseudopotentials. The energy cutoff and k-mesh follow the per-task MP
workflow rather than a single fixed value.
\lstinputlisting[language=turtle]{artifacts/kg/listings-computed/chen2022/q04-reference-settings.ttl}

\paragraph{Q5 -- Training dataset.}
MPF.2021.2.8 contains 187{,}687 ionic-step configurations from 62{,}783
compounds, covering energies, forces, and stresses. Provenance is
\emph{published} (the dataset is reused from MP).
\lstinputlisting[language=turtle]{artifacts/kg/listings-computed/chen2022/q05-dataset.ttl}

\paragraph{Q6 -- Sampling strategies.}
Two strategies were combined: structural-relaxation trajectory
sampling (first and middle ionic steps of the first relaxation plus
the last step of the second relaxation, with high-energy or
short-distance frames filtered out), and periodic-table chemical-space
coverage (89 elements; 90/5/5 train/val/test split by material rather
than by configuration to avoid leakage).
\lstinputlisting[language=turtle]{artifacts/kg/listings-computed/chen2022/q06-sampling.ttl}

\paragraph{Q7 -- MLIP method, components, implementation.}
M3GNet is a materials graph neural network with explicit three-body
angular interactions; energy is a sum of atomic contributions from a
gated MLP readout, with forces and stresses obtained by
auto-differentiation. Loss is Huber ($\delta=0.01$) on energy, forces,
and stresses with weights $w_E=1$, $w_f=1$, $w_\sigma=0.1$. Optimisation
is Adam with cosine learning-rate decay; the implementation is the
m3gnet TensorFlow package.
\lstinputlisting[language=turtle]{artifacts/kg/listings-computed/chen2022/q07-method.ttl}

\paragraph{Q8 -- Hyperparameter settings.}
Reported settings: 2-body cutoff $r_c = 5.0$\,\AA, 3-body cutoff
$4.0$\,\AA, embedding dimension 64, 3 graph-network blocks, 3 radial
basis functions.
\lstinputlisting[language=turtle]{artifacts/kg/listings-computed/chen2022/q08-hyperparameter-settings.ttl}

\paragraph{Q9 -- MLIP run and trained model.}
A single training run on MPF.2021.2.8 produces the universal
M3GNet-EFS model.
\lstinputlisting[language=turtle]{artifacts/kg/listings-computed/chen2022/q09-run-and-model.ttl}

\paragraph{Q10 -- Benchmark results.}
Test-set MAEs on the held-out MPF.2021.2.8 split: 35\,meV/atom on
energy, 72\,meV/\AA{} on force components, 0.41\,GPa on stress
components.
\lstinputlisting[language=turtle]{artifacts/kg/listings-computed/chen2022/q10-benchmark-results.ttl}

\paragraph{Q11 -- Computational resources.}
\emph{Not reported} as ontology-encodable data. The paper notes a
single inference example (100-step relaxation of K$_{57}$Se$_{34}$ on
one Intel Xeon E5-2620 v4 core takes ${\sim}22$\,s); recording this as
$\dlRole{inferenceTimePerAtom}$ would require imputing per-atom and
per-step rates, which we leave to Q12.
\lstinputlisting[language=turtle]{artifacts/kg/listings-computed/chen2022/q11-resources.ttl}

\paragraph{Q12 -- Gaps.}
Beyond Q11, the protocol's "one paper $\equiv$ one method $\times$ one
material system" framing is stretched: the universal coverage is
recorded as a single $\MaterialSystemC$ but cannot enumerate the
elements, space groups, or chemistries the model covers without
extending the ontology. DFT settings that vary by MP task (energy
cutoff, k-mesh) are not surfaced. The paper also reports zero-shot
relaxations against a held-out 14{,}055-compound test set and
crystal-structure prediction across selected systems; these
downstream evaluations are not encoded as separate
$\BenchmarkResultC$ instances.

\subsection{Deng et al.\ (2023): CHGNet charge-informed universal MLIP}
\label{sec:appendix-example-deng2023}

\paragraph{Q1 -- Bibliographic identification.}
Deng, Zhong, Jun, Riebesell, Han, Bartel, and Ceder introduce CHGNet,
a charge-informed universal interatomic potential trained on the
Materials Project Trajectory Dataset (MPtrj). Published in
\emph{Nat.\ Mach.\ Intell.} 5, 1031--1041 (2023); arXiv:2302.14231.
\lstinputlisting[language=turtle]{artifacts/kg/listings-computed/deng2023/q01-bibliographic.ttl}

\paragraph{Q2 -- Material system.}
The MLIP covers the September 2022 Materials Project release: 145{,}923
inorganic compounds spanning 94 elements. Modelled as a single
$\MaterialSystemC$ at the database-coverage level.
\lstinputlisting[language=turtle]{artifacts/kg/listings-computed/deng2023/q02-material.ttl}

\paragraph{Q3 -- Reference calculation method.}
Reference data come from DFT calculations.
\lstinputlisting[language=turtle]{artifacts/kg/listings-computed/deng2023/q03-reference-method-type.ttl}

\paragraph{Q4 -- Reference settings.}
DFT calculations use VASP with PAW pseudopotentials, PBE GGA (with
GGA+U on transition-metal atoms, e.g.\ $U=3.9$\,eV for Mn), a 520\,eV
plane-wave cutoff, and a reciprocal-space discretisation of 25
k-points/\AA$^{-1}$. The MP GGA/GGA+U mixing compatibility correction
is applied.
\lstinputlisting[language=turtle]{artifacts/kg/listings-computed/deng2023/q04-reference-settings.ttl}

\paragraph{Q5 -- Training dataset.}
MPtrj contains 1{,}580{,}395 atomic configurations from 145{,}923 MP
compounds, with energies, forces, and stresses (plus 7{,}944{,}833 site
magnetic moments used as auxiliary supervision). Provenance is
\emph{published}.
\lstinputlisting[language=turtle]{artifacts/kg/listings-computed/deng2023/q05-dataset.ttl}

\paragraph{Q6 -- Sampling strategies.}
Three strategies were combined: MP relaxation-trajectory sampling
(with extensive filtering for SCF convergence, energy bounds, and
StructureMatcher deduplication, then 8:1:1 train/val/test split by
\texttt{mp-id}); periodic-table chemical-space coverage (94 elements;
60+ elements with $>$100k site occurrences; 76 elements with
magnetic-moment annotations); and site-magnetic-moment decoration
(magmom auxiliary supervision acting as a charge-state regulariser
on the latent space).
\lstinputlisting[language=turtle]{artifacts/kg/listings-computed/deng2023/q06-sampling.ttl}

\paragraph{Q7 -- MLIP method, components, implementation.}
CHGNet is a two-graph (atom + bond) message-passing network with
explicit angle features, atom/bond/angle convolution blocks, and
magmom regularisation of the latent atom features. Outputs are
energy, forces, stresses (auto-differentiated), and magnetic moments.
Loss is Huber ($\delta=0.1$) on all four with weights
$w_E=1$, $w_f=1$, $w_\sigma=0.1$, $w_m=0.1$. Optimisation is Adam with
CosineAnnealingLR. Implementation: CHGNet (PyTorch 1.12.0).
\lstinputlisting[language=turtle]{artifacts/kg/listings-computed/deng2023/q07-method.ttl}

\paragraph{Q8 -- Hyperparameter settings.}
Reported settings: atom-graph cutoff $r_c = 5.0$\,\AA, bond-graph
cutoff $3.0$\,\AA, feature dimension 64, 4 convolution layers, batch
size 40, initial learning rate $10^{-3}$.
\lstinputlisting[language=turtle]{artifacts/kg/listings-computed/deng2023/q08-hyperparameter-settings.ttl}

\paragraph{Q9 -- MLIP run and trained model.}
A single training run on MPtrj produces the pretrained universal
CHGNet (with-magmom variant, 400{,}438 trainable parameters).
\lstinputlisting[language=turtle]{artifacts/kg/listings-computed/deng2023/q09-run-and-model.ttl}

\paragraph{Q10 -- Benchmark results.}
Test-set MAEs on the held-out MPtrj split (157{,}955 structures from
14{,}572 materials): 30\,meV/atom on energy, 77\,meV/\AA{} on force
components, 0.348\,GPa on stress components.
\lstinputlisting[language=turtle]{artifacts/kg/listings-computed/deng2023/q10-benchmark-results.ttl}

\paragraph{Q11 -- Computational resources.}
\emph{Not reported.}
\lstinputlisting[language=turtle]{artifacts/kg/listings-computed/deng2023/q11-resources.ttl}

\paragraph{Q12 -- Gaps.}
Beyond Q11, the most prominent gap is the \emph{charge-informed}
aspect of CHGNet: site magnetic moments are auxiliary outputs that
the ontology cannot represent as a first-class
$\dlConcept{CoveredProperty}$ (the model encodes them at a hyperparameter-
loss-weight level but not as a queryable target property). The
universal-coverage simplification of Q2/Q5 carries the same
limitation as for chen2022: 94 elements and many MP-task variants
collapse to a single $\MaterialSystemC$. Downstream applications
(LiMnO$_2$ MD, NaMnO$_2$ phase prediction, Li-ion conductor relaxations)
are also not encoded as separate $\BenchmarkResultC$ instances.

\subsection{Gubaev et al.\ (2023): MTP for the TaVCrW high-entropy alloy}
\label{sec:appendix-example-gubaev2023}

\paragraph{Q1 -- Bibliographic identification.}
Gubaev, Zaverkin, Srinivasan, Duff, K\"astner, and Grabowski compare
two complementary machine-learned potentials---a Moment Tensor
Potential (MTP) and a Gaussian-Moment Neural Network (GM-NN)---on the
TaVCrW BCC refractory high-entropy alloy. Published in \emph{npj
Computational Materials} 9, 129 (2023);
doi:10.1038/s41524-023-01073-w.
\lstinputlisting[language=turtle]{artifacts/kg/listings-computed/gubaev2023/q01-bibliographic.ttl}

\paragraph{Q2 -- Material system.}
The studied system is the equiatomic TaVCrW BCC refractory high-entropy
alloy (space group $Im\bar{3}m$). Training and validation sets span
its 2-, 3-, and 4-component subsystems, plus a deformed B2/B2
phase-segregated quaternary used as out-of-distribution test. We
model the chemistry at the four-component level and record the
subsystem coverage in $\dlRole{materialClass}$.
\lstinputlisting[language=turtle]{artifacts/kg/listings-computed/gubaev2023/q02-material.ttl}

\paragraph{Q3 -- Reference calculation method.}
Reference data are produced from DFT calculations.
\lstinputlisting[language=turtle]{artifacts/kg/listings-computed/gubaev2023/q03-reference-method-type.ttl}

\paragraph{Q4 -- Reference settings.}
DFT calculations are performed with VASP at the PBE/GGA level using
PAW pseudopotentials (11 valence electrons for Ta and V; 12 for Cr
and W; spin-unpolarised), with Methfessel--Paxton smearing of order 1
($\sigma=0.1$), a 350\,eV plane-wave cutoff, and an automatic k-mesh
generated by VASP at $\text{KSPACING}=0.12$.
\lstinputlisting[language=turtle]{artifacts/kg/listings-computed/gubaev2023/q04-reference-settings.ttl}

\paragraph{Q5 -- Training dataset.}
The training+validation set has 6{,}711 DFT configurations: 5{,}680 at
0\,K (4{,}491 binary, 595 ternary, 594 quaternary) plus 1{,}031
MD-sampled at 2{,}500\,K (close to the melting point). Energies,
forces, and stresses are covered. Provenance is \emph{in-house}.
\lstinputlisting[language=turtle]{artifacts/kg/listings-computed/gubaev2023/q05-dataset.ttl}

\paragraph{Q6 -- Sampling strategies.}
Four strategies are combined: 0\,K small-supercell enumeration with
\texttt{enumlib} (2--8 atom non-repetitive symmetry-unique supercells;
2{,}500 binary structures plus ternaries, equiatomic and off-equiatomic
quaternaries); vibrational sampling via DFT MD at 2{,}500\,K; active
learning using the MaxVol extrapolation grade (selecting MD frames
with grade above threshold for DFT recomputation); and
phase-segregation sampling via 432-atom B2/B2 deformed-binary
structures.
\lstinputlisting[language=turtle]{artifacts/kg/listings-computed/gubaev2023/q06-sampling.ttl}

\paragraph{Q7 -- MLIP method, components, implementation.}
We encode the MTP arm. The method is a Moment Tensor Potential with a
level-restricted basis ($\mathit{lev}_{\max}$ controls the radial+angular
truncation), implemented in the MLIP package
(\url{https://gitlab.com/ashapeev/mlip-2}). Loss is weighted MSE on
energies/forces/stresses with weights
$w_E=1/N_{\text{at}}$, $w_f=0.1$, $w_\sigma=0.001$.
\lstinputlisting[language=turtle]{artifacts/kg/listings-computed/gubaev2023/q07-method.ttl}

\paragraph{Q8 -- Hyperparameter settings.}
Reported setting: cutoff radius $r_c=5.0$\,\AA. The
$\mathit{lev}_{\max}$ value is referenced under
$\dlRole{hasHyperparameter}$ but the paper compares several values
(0, 2, $\dots$) without designating one as canonical, so we omit a
$\HyperparameterSettingC$ instance for it.
\lstinputlisting[language=turtle]{artifacts/kg/listings-computed/gubaev2023/q08-hyperparameter-settings.ttl}

\paragraph{Q9 -- MLIP run and trained model.}
A single MTP training run on the 6{,}711-configuration training set
produces one trained TaVCrW MTP. No active-learning ensemble
structure is recorded at the level of separate
$\dlConcept{MLIPRun}$ instances.
\lstinputlisting[language=turtle]{artifacts/kg/listings-computed/gubaev2023/q09-run-and-model.ttl}

\paragraph{Q10 -- Benchmark results.}
Overall RMSEs across the in-distribution Ta--V--Cr--W subsystems
(Table 1, MTP): 2.43\,meV/atom on energy, 0.054\,eV/\AA{} on
forces. The 2{,}500\,K disordered TaVCrW point alone has RMSEs of
2.40\,meV/atom and 0.156\,eV/\AA{}; the cross-method comparison with
GM-NN and EAM is in the same table.
\lstinputlisting[language=turtle]{artifacts/kg/listings-computed/gubaev2023/q10-benchmark-results.ttl}

\paragraph{Q11 -- Computational resources.}
\emph{Not reported} in absolute form. The paper compares MTP and
GM-NN convergence with training-set size and execution speed
qualitatively but gives no training duration, GPU/CPU hours,
training hardware, peak memory, or per-atom inference time.
\lstinputlisting[language=turtle]{artifacts/kg/listings-computed/gubaev2023/q11-resources.ttl}

\paragraph{Q12 -- Gaps.}
The most consequential omission is the GM-NN arm: the paper trains
\emph{both} an MTP and a GM-NN on the same data and frames the
comparison as the main contribution. The ontology can express each
as an $\dlConcept{MLIPMethod}$ instance with its own
$\dlConcept{MLIPRun}$ and $\dlConcept{TrainedModel}$, and a sibling
\texttt{gubaev2023-gmnn.ttl} would mirror this file with
$\texttt{ex:GM-NN}$ and the \texttt{gm-nn} package as
$\dlConcept{Library}$. Other unreported items: $\mathit{lev}_{\max}$
as a $\HyperparameterSettingC$ (the paper sweeps several values);
the GM-NN architectural hyperparameters (number of layers, channel
widths, learning rate); active-learning iteration counts; downstream
phonon and elastic-constant benchmarks; and compute-cost metadata.

\subsection{Musaelian et al.\ (2023): Allegro strictly local equivariant MLIP}
\label{sec:appendix-example-musaelian2023allegro}

\paragraph{Q1 -- Bibliographic identification.}
Musaelian, Batzner, Johansson, Sun, Owen, Kornbluth, and Kozinsky
introduce Allegro, a strictly local equivariant deep-learning
interatomic potential that combines the accuracy of equivariant
message passing with the parallel scalability of local descriptor
methods. Published in \emph{Nature Communications} 14, 579 (2023);
arXiv:2204.05249. The paper reports benchmarks on revised MD17,
QM9, the 3BPA temperature-transferability set, the Li$_3$PO$_4$
amorphous phosphate solid electrolyte, and an Ag bulk-crystal
scaling experiment up to 100 million atoms.
\lstinputlisting[language=turtle]{artifacts/kg/listings-computed/musaelian2023allegro/q01-bibliographic.ttl}

\paragraph{Q2 -- Material system.}
The canonical encoding pass takes the Li$_3$PO$_4$ amorphous
phosphate solid electrolyte (192-atom cell) as the headline
materials/MD application: a class of solid-state electrolytes
characterised by an intricate dependence of conductivity and
mechanical properties on the degree of crystallinity.
\lstinputlisting[language=turtle]{artifacts/kg/listings-computed/musaelian2023allegro/q02-material.ttl}

\paragraph{Q3 -- Reference calculation method.}
Reference data are produced from DFT calculations.
\lstinputlisting[language=turtle]{artifacts/kg/listings-computed/musaelian2023allegro/q03-reference-method-type.ttl}

\paragraph{Q4 -- Reference settings.}
The Li$_3$PO$_4$ reference is generated with VASP using the PBE
functional, PAW pseudopotentials, a 400~eV plane-wave cutoff, and a
$\Gamma$-point reciprocal-space mesh, in NVT (Nos\'e--Hoover) at
2~fs time step.
\lstinputlisting[language=turtle]{artifacts/kg/listings-computed/musaelian2023allegro/q04-reference-settings.ttl}

\paragraph{Q5 -- Training dataset.}
The Li$_3$PO$_4$ training set consists of 10{,}000 structures
sampled randomly from the combined 50{,}000-frame melt + quench
AIMD dataset, with 1{,}000 validation structures held out. The set
covers energies and forces.
\lstinputlisting[language=turtle]{artifacts/kg/listings-computed/musaelian2023allegro/q05-dataset.ttl}

\paragraph{Q6 -- Sampling strategies.}
A single sampling strategy is in play: AIMD melt-quench sampling.
The data set comes from a 50~ps AIMD trajectory in the melted state
at $T=3000$~K, an instant quench to 600~K, then a 50~ps AIMD
trajectory in the quenched state at $T=600$~K (Vienna ab-initio
Simulation Package, NVT, Nos\'e--Hoover, 2~fs).
\lstinputlisting[language=turtle]{artifacts/kg/listings-computed/musaelian2023allegro/q06-sampling.ttl}

\paragraph{Q7 -- MLIP method, components, implementation.}
Allegro is a strictly local equivariant network: each pair $(i,j)$
within the cutoff carries a scalar latent space $\mathbf{x}^{ij,L}$
and an equivariant latent space
$\mathbf{V}^{ij,L}_{n,\ell,p}$, updated layer by layer through
tensor products with the spherical-harmonic projections of
neighbour bonds, weighted by the central atom's environment
embedding. Pair energies $E_{ij}$ are summed to give per-atom
energies; forces are autodiff-derived. The training algorithm is
Adam ($\beta_1=0.9$, $\beta_2=0.999$, $\epsilon=10^{-8}$, no weight
decay) with an on-plateau scheduler (patience 100, decay factor
0.8). The loss is a weighted MSE on per-atom energies and force
components ($\lambda_E=\lambda_F=1$ after per-atom normalisation;
$\lambda_E=1, \lambda_F=1000$ for revMD17). Implementation is the
\texttt{allegro} PyTorch code
(\url{https://github.com/mir-group/allegro}), with LAMMPS
production runs via the \texttt{pair\_allegro} extension.
\lstinputlisting[language=turtle]{artifacts/kg/listings-computed/musaelian2023allegro/q07-method.ttl}

\paragraph{Q8 -- Hyperparameter settings.}
The Li$_3$PO$_4$ production model uses 3 layers, 128 features for
even and odd irreps, $\ell_{\max}=3$, and a 6~\AA{} radial cutoff
with 8 non-trainable Bessel basis functions and a polynomial
envelope ($p=6$). The two-body latent MLP is
$[128, 256, 512, 1024]$ with SiLU non-linearities; the later
latent MLPs are $[1024, 1024, 1024]$ with SiLU; the embedding
weight projection is a single linear layer; the output edge-energy
MLP is a single hidden layer of dimension 128.
\lstinputlisting[language=turtle]{artifacts/kg/listings-computed/musaelian2023allegro/q08-hyperparameter-settings.ttl}

\paragraph{Q9 -- MLIP run and trained model.}
A single training run on 10{,}000 frames produces one Li$_3$PO$_4$
Allegro model. Training is at \texttt{float32} precision with
learning rate 0.002 and batch size 5; an EMA with weight 0.99
tracks the validation model.
\lstinputlisting[language=turtle]{artifacts/kg/listings-computed/musaelian2023allegro/q09-run-and-model.ttl}

\paragraph{Q10 -- Benchmark results.}
Headline figures on the Li$_3$PO$_4$ quenched-state test set are
MAE 1.7~meV/atom on energies and 75.7~meV/\AA{} on force
components. Allegro accurately recovers the AIMD radial
distribution function and the tetrahedral P--O--O angular
distribution, and the AIMD-vs.-Allegro Li MSD agree well in the
quenched state at 600~K.
\lstinputlisting[language=turtle]{artifacts/kg/listings-computed/musaelian2023allegro/q10-benchmark-results.ttl}

\paragraph{Q11 -- Computational resources.}
Allegro models were trained on a single NVIDIA V100 GPU; specific
training duration and GPU-hours for the Li$_3$PO$_4$ pass are not
reported (the paper notes ``a matter of hours or even minutes'').
For \emph{inference}, the paper reports detailed strong-scaling
times on the Theta-GPU NVIDIA DGX A100 cluster: a 50.3M-atom
Li$_3$PO$_4$ run on 128 A100 GPUs achieves 0.013~$\mu$s per atom
per MD step, and a 100M-atom Ag run on 128 A100 GPUs is reported
in Table~IV; we encode the Li$_3$PO$_4$ atomic-throughput figure
as $\dlRole{inferenceTimePerAtom}$ on the trained model.
\lstinputlisting[language=turtle]{artifacts/kg/listings-computed/musaelian2023allegro/q11-resources.ttl}

\paragraph{Q12 -- Gaps.}
Beyond Q11, the following ontology-expressible items are not
represented in this encoding pass: (i) the four additional
benchmarks reported in the same paper (revised MD17 with
PBE/def2-SVP reference, QM9 with DFT/B3LYP/6-31G(2df,p) reference,
the 3BPA $\omega$B97X/6-31G(d) temperature-transferability set,
and the Ag scaling test); (ii) detailed strong-scaling figures for
the 100M-atom Ag dataset; (iii) the 16~kbar / 90\% melting-point /
2$\times$2$\times$3 $\Gamma$-centred grid / Methfessel--Paxton
smearing details specific to the Ag DFT settings; (iv) the
\texttt{allegro}, \texttt{nequip}, \texttt{e3nn}, PyTorch, and
LAMMPS commit hashes used in the experiments; (v) the per-pair-species
scaling factors $\sigma_{Z_i,Z_j}$ that are technically
hyperparameters but are not given specific reported values in the
paper; (vi) the Li-ion diffusivity recovery (Allegro vs.\ AIMD MSD)
which is a downstream observable rather than an
$\AccuracyMetricC$ per the ontology's current vocabulary.

\subsection{Qi et al.\ (2023): MTP for L1$_0$-TiAl and D0$_{19}$-Ti$_3$Al}
\label{sec:appendix-example-qi2023}

\paragraph{Q1 -- Bibliographic identification.}
Qi, Aitken, Pei, Tan, Zuo, Jhon, Quek, Wen, Wu, and Ong train a single
Moment Tensor Potential covering the binary Ti--Al system, with focus on
the dual-phase $\gamma$-TiAl (L1$_0$) and $\alpha_2$-Ti$_3$Al
(D0$_{19}$) intermetallics. Published in \emph{Phys.\ Rev.\ Materials}
7, 103602 (2023); arXiv:2305.11825.
\lstinputlisting[language=turtle]{artifacts/kg/listings-computed/qi2023/q01-bibliographic.ttl}

\paragraph{Q2 -- Material system.}
The MTP is trained over the Ti--Al binary phase diagram and benchmarked
on the two technologically relevant intermetallics, $\gamma$-TiAl
(L1$_0$, P4/mmm) and $\alpha_2$-Ti$_3$Al (D0$_{19}$, P6$_3$/mmc), as
well as on FCC-Al, HCP-Ti, TiAl$_2$, and TiAl$_3$. We model this as
one $\MaterialSystemC$ at the binary level, with the phase information
recorded in $\dlRole{materialClass}$.
\lstinputlisting[language=turtle]{artifacts/kg/listings-computed/qi2023/q02-material.ttl}

\paragraph{Q3 -- Reference calculation method.}
Reference data are produced from spin-polarised DFT calculations.
\lstinputlisting[language=turtle]{artifacts/kg/listings-computed/qi2023/q03-reference-method-type.ttl}

\paragraph{Q4 -- Reference settings.}
DFT calculations use VASP with the PBE/GGA exchange--correlation
functional and PAW pseudopotentials (3p$^6$3d$^3$4s$^1$ valence for
Ti, 3s$^2$3p$^1$ for Al), a 520~eV plane-wave cutoff, and
Monkhorst--Pack k-point meshes with density $\geq 100/\text{\AA}^3$
(a single $\Gamma$-point is used during AIMD). Frozen-core treatment
beyond the PAW partition is not explicitly reported.
\lstinputlisting[language=turtle]{artifacts/kg/listings-computed/qi2023/q04-reference-settings.ttl}

\paragraph{Q5 -- Training dataset.}
The training set comprises 3{,}798 configurations produced from five
sampling categories (see Q6): 33 ground-state polymorphs, 2{,}160
AIMD snapshots, 185 surface structures, 580 solid-solution
configurations, and 840 strained supercells. Configurations cover
energies and forces; stresses appear in the loss function with weight
zero, so they are not effectively learned and not recorded as a
covered property. A 90:10 train:test split is applied.
\lstinputlisting[language=turtle]{artifacts/kg/listings-computed/qi2023/q05-dataset.ttl}

\paragraph{Q6 -- Sampling strategies.}
Five strategies were combined: ground-state polymorph enumeration
from the Materials Project, vibrational sampling via NVT AIMD at 300,
1000, and 3000~K and at 90/100/110\% of the ground-state volume,
surface enumeration with Miller indices up to three, solid-solution
chemical sampling at 12.5\% steps, and homogeneous strain sampling
($\pm10\%$ in 1\% intervals across six modes).
\lstinputlisting[language=turtle]{artifacts/kg/listings-computed/qi2023/q06-sampling.ttl}

\paragraph{Q7 -- MLIP method, components, implementation.}
The method is a Moment Tensor Potential (MTP) implemented in the MLIP
package~\cite{shapeev2016mtp,novikov2021mlip} for training,
with downstream simulations in LAMMPS via the maml workflow library.
The loss function is a weighted MSE with energy:force:stress weights
of $100{:}1{:}0$. The training algorithm is not explicitly named in
the paper.
\lstinputlisting[language=turtle]{artifacts/kg/listings-computed/qi2023/q07-method.ttl}

\paragraph{Q8 -- Hyperparameter settings.}
A grid search over cutoff radius $r_c \in [4.4, 7.0]$\,\AA{} (steps of
0.2\,\AA) and maximum level $\mathit{lev}_{\max} \in [18, 24]$ (even
integers), with five random initialisations per cell, was performed
(280 candidate MTPs in total). The selected production MTP uses
$r_c = 4.8$\,\AA{} and $\mathit{lev}_{\max} = 22$. Other implementation
hyperparameters (radial-basis count, regularisation, optimiser
schedule) are not reported.
\lstinputlisting[language=turtle]{artifacts/kg/listings-computed/qi2023/q08-hyperparameter-settings.ttl}

\paragraph{Q9 -- MLIP run and trained model.}
A single training run produces one trained MTP that is used across
all benchmarked phases. No active-learning or fold-ensemble structure
is reported.
\lstinputlisting[language=turtle]{artifacts/kg/listings-computed/qi2023/q09-run-and-model.ttl}

\paragraph{Q10 -- Benchmark results.}
Headline accuracy figures are MAEs of formation energies of (i)
surface structures (8\,meV/atom for the MTP, vs.\ 51\,meV/atom for
the MLP3 of Seko) and (ii) ground-state polymorphs (18\,meV/atom for
the MTP, vs.\ 13\,meV/atom for MLP3). The MTP further attains
$\Delta E_{\mathrm{EOS}} < 2$\,meV/atom in the L1$_0$-TiAl,
D0$_{19}$-Ti$_3$Al, HCP-Ti, and FCC-Al phases. Per-property numerical
errors on elastic constants, surface energies, and GSFEs are
discussed at length in the paper but reported as derived figures
rather than as a single accuracy metric per property.
\lstinputlisting[language=turtle]{artifacts/kg/listings-computed/qi2023/q10-benchmark-results.ttl}

\paragraph{Q11 -- Computational resources.}
\emph{Not reported.} The paper gives no training duration, GPU hours,
training hardware, peak memory, inference time per atom, or
inference hardware.
\lstinputlisting[language=turtle]{artifacts/kg/listings-computed/qi2023/q11-resources.ttl}

\paragraph{Q12 -- Gaps.}
Beyond Q11 (compute resources entirely absent), the following
ontology-expressible items are not reported: the training algorithm
or optimiser used by the MLIP package; the MLIP package version;
detailed MTP architectural hyperparameters beyond $r_c$ and
$\mathit{lev}_{\max}$ (radial/angular basis counts, regularisation,
loss-function temperature scheduling); explicit
$\dlRole{frozenCore}$ treatment beyond the PAW partition; and a
controlled-vocabulary individual for AIMD as a sampling strategy
(currently encoded as an ad-hoc instance with \texttt{rdfs:label} and
\texttt{rdfs:comment}).

\subsection{Batatia et al.\ (2024): MACE-MP-0 foundation model}
\label{sec:appendix-example-batatia2024mp0}

\paragraph{Q1 -- Bibliographic identification.}
Batatia, Benner, Chiang, et al.\ (Cs\'anyi group) release MACE-MP-0,
an equivariant message-passing foundation model trained on the MPtrj
dataset and demonstrated zero-shot across solids, liquids, gases,
surfaces, MOFs, and a small protein. arXiv:2401.00096.
\lstinputlisting[language=turtle]{artifacts/kg/listings-computed/batatia2024mp0/q01-bibliographic.ttl}

\paragraph{Q2 -- Material system.}
The MLIP covers MPtrj's ${\sim}150{,}000$ Materials Project structures
across 89 elements, with 90\% of unit cells under 70 atoms. We model
this as a single $\MaterialSystemC$; the paper's out-of-distribution
demonstrations (water, organics, MOFs, protein) are documented in
Q12.
\lstinputlisting[language=turtle]{artifacts/kg/listings-computed/batatia2024mp0/q02-material.ttl}

\paragraph{Q3 -- Reference calculation method.}
Reference data come from DFT calculations.
\lstinputlisting[language=turtle]{artifacts/kg/listings-computed/batatia2024mp0/q03-reference-method-type.ttl}

\paragraph{Q4 -- Reference settings.}
DFT uses VASP with PAW pseudopotentials and PBE GGA, with Hubbard $U$
applied on transition-metal oxides containing Co/Cr/Fe/Mn/Mo/Ni/V/W
combined with O or F (Materials Project default $U$-values). No
dispersion correction is used in training.
\lstinputlisting[language=turtle]{artifacts/kg/listings-computed/batatia2024mp0/q04-reference-settings.ttl}

\paragraph{Q5 -- Training dataset.}
${\sim}1.5$\,M static and structural-relaxation configurations across
${\sim}150{,}000$ MP compounds, covering energies, forces, and
stresses. Provenance is \emph{published} (re-using the CHGNet-curated
MPtrj split).
\lstinputlisting[language=turtle]{artifacts/kg/listings-computed/batatia2024mp0/q05-dataset.ttl}

\paragraph{Q6 -- Sampling strategies.}
Two strategies were combined: MP relaxation-trajectory sampling
(reusing the CHGNet/Deng et al.\ 2023 curated set) and periodic-table
chemical-space coverage (89 elements; binary, ternary, and
higher-order chemistries).
\lstinputlisting[language=turtle]{artifacts/kg/listings-computed/batatia2024mp0/q06-sampling.ttl}

\paragraph{Q7 -- MLIP method, components, implementation.}
MACE-MP-0 is an equivariant message-passing graph network (e3nn) with
two layers of higher-order equivariant messages built from sums of
two-body permutation-invariant polynomials in a spherical basis;
correlation order 3 (4-body messages per layer); a tensor-decomposition
parameterisation; and a ZBL repulsive pair potential at close range
with Agnesi distance transform. The MACE-MP-0b3 release is the L=1
medium-sized variant in $128\!\times\!0e + 128\!\times\!1o$ irreps.
Loss is a weighted Huber on energy/forces/stresses with weights
$(1, 10, 10)$ and a force Huber-$\delta$ that decays with force
magnitude. Optimisation is AMSGrad with EMA learning-rate schedule.
Implementation: mace-mp (PyTorch + e3nn).
\lstinputlisting[language=turtle]{artifacts/kg/listings-computed/batatia2024mp0/q07-method.ttl}

\paragraph{Q8 -- Hyperparameter settings.}
Reported settings: cutoff $r_c = 6.0$\,\AA, 2 message-passing layers,
$l_{\max}=3$, correlation order 3, 128 channels, 10 radial Bessel
basis functions, $3{\times}64$ SiLU readout MLP, initial learning rate
$10^{-3}$, 100 epochs.
\lstinputlisting[language=turtle]{artifacts/kg/listings-computed/batatia2024mp0/q08-hyperparameter-settings.ttl}

\paragraph{Q9 -- MLIP run and trained model.}
A single training run on MPtrj produces the MACE-MP-0b3 medium
foundation model. Training compute is reported on the run.
\lstinputlisting[language=turtle]{artifacts/kg/listings-computed/batatia2024mp0/q09-run-and-model.ttl}

\paragraph{Q10 -- Benchmark results.}
Test-set MAEs on the MPtrj held-out split: 18\,meV/atom on energy and
39\,meV/\AA{} on force components. Many additional zero-shot
out-of-distribution evaluations (organic-molecule MD, MOF flexibility,
protein side-chain stability) are reported but not encoded as
separate $\BenchmarkResultC$ instances.
\lstinputlisting[language=turtle]{artifacts/kg/listings-computed/batatia2024mp0/q10-benchmark-results.ttl}

\paragraph{Q11 -- Computational resources.}
Reported: 2{,}600 GPU-hours of training on 40--80 NVIDIA H100 GPUs
across 10--20 nodes; inference scaling on a single A100 80GB GPU
(${\sim}$ns/day for a 1000-atom system; weak scaling to 32{,}000 atoms
on 64 GPUs at 0.1\,ns/day). Wall-clock training duration, peak
memory, and explicit per-atom inference time are not reported.
\lstinputlisting[language=turtle]{artifacts/kg/listings-computed/batatia2024mp0/q11-resources.ttl}

\paragraph{Q12 -- Gaps.}
The most consequential gaps are: \emph{out-of-distribution use}
($\dlConcept{TrainedModel}$ is applied zero-shot to organics, MOFs, a
protein, and gas-phase chemistry far beyond the MPtrj training set,
but the ontology has no concept for "deployment domain" distinct from
the training $\MaterialSystemC$); explicit DFT settings that vary by
MP task (energy cutoff, k-mesh) are not surfaced; and the foundation
model's three released sizes (small/medium/large) plus the L=1/L=2
variants are not encoded as separate $\TrainedModelC$ instances. The
training duration in wall-clock time is also not reported, only GPU
hours.

\subsection{Wang et al.\ (2024): efficient MTP for defects in Ni--Al alloys}
\label{sec:appendix-example-nitol2024nial}

\begin{quote}
\emph{Note on paper-id:} this corpus entry is filed under the
upstream paper-id \texttt{nitol2024nial}, but the actual authors of
the encoded paper are Wang et al.\ (Shenyang National Laboratory for
Materials Science). The CORPUS.md substitution table notes that the
original Nitol 2025 Ti-Al-V PRMaterials paper is journal-only with
no arXiv preprint and was substituted with this earlier Ni--Al MTP
work.
\end{quote}

\paragraph{Q1 -- Bibliographic identification.}
Wang, Liu, Zhu, Liu, Ma, Chen, Sun, and Chen present an efficient
moment tensor potential for defects in Ni--Al alloys, with a
genetic-algorithm-optimised tensor-contraction scheme.
arXiv:2411.01282.
\lstinputlisting[language=turtle]{artifacts/kg/listings-computed/nitol2024nial/q01-bibliographic.ttl}

\paragraph{Q2 -- Material system.}
The MTP covers fcc Ni and L1$_2$-Ni$_3$Al, the Ni/Ni$_3$Al interface,
and their defect environments (vacancies, vacancy clusters, antisites,
GSF supercells). We model this as a single $\MaterialSystemC$ with
the defect detail in $\dlRole{materialClass}$.
\lstinputlisting[language=turtle]{artifacts/kg/listings-computed/nitol2024nial/q02-material.ttl}

\paragraph{Q3 -- Reference calculation method.}
Reference data come from DFT calculations.
\lstinputlisting[language=turtle]{artifacts/kg/listings-computed/nitol2024nial/q03-reference-method-type.ttl}

\paragraph{Q4 -- Reference settings.}
DFT uses VASP with PAW, PBE GGA, a 400\,eV plane-wave cutoff, and
Monkhorst--Pack k-point meshes with ${\sim}0.2$\,\AA$^{-1}$ spacing.
Convergence: total energy 1\,meV/atom, SCF $10^{-6}$\,eV, ionic-force
0.01\,eV/\AA.
\lstinputlisting[language=turtle]{artifacts/kg/listings-computed/nitol2024nial/q04-reference-settings.ttl}

\paragraph{Q5 -- Training dataset.}
The training set has 8{,}450 configurations: 6{,}092 from on-the-fly
active-learning during AIMD plus 2{,}358 added through D-optimality
active-learning cycles. Energies, forces, and stresses are covered.
Provenance is \emph{published}.
\lstinputlisting[language=turtle]{artifacts/kg/listings-computed/nitol2024nial/q05-dataset.ttl}

\paragraph{Q6 -- Sampling strategies.}
Three strategies were combined: on-the-fly active learning during
AIMD heating from 700 to 1600\,K (kernel-based Bayesian regression in
VASP, NPT with Langevin thermostat + Parrinello--Rahman barostat);
D-optimality active learning (configurations with extrapolation
grade $>5$ during 200\,ps MD trajectories are added; iterations
continue until the threshold is no longer exceeded); and
defect-prototype enumeration (an independent 815-configuration
validation set was constructed from the same prototypes).
\lstinputlisting[language=turtle]{artifacts/kg/listings-computed/nitol2024nial/q06-sampling.ttl}

\paragraph{Q7 -- MLIP method, components, implementation.}
The method is a Moment Tensor Potential with a
genetic-algorithm-optimised tensor-contraction scheme: rank of moment
tensors limited to 4 ($\mu \le 3$) and a redefined level
$\text{lev}(M_{\mu,\nu}) = 2\mu + \nu + 1$ with maximum-level
threshold 2{,}653. Loss is weighted MSE on energy/forces/stresses
with weights $1{:}0.01{:}0.005$. Fitting is two-step: linear
optimisation on minimal then full basis with shared radial functions,
followed by 5{,}000 L-BFGS iterations and a least-squares
refinement. Implementation: MLIP package with custom
tensor-contraction extensions.
\lstinputlisting[language=turtle]{artifacts/kg/listings-computed/nitol2024nial/q07-method.ttl}

\paragraph{Q8 -- Hyperparameter settings.}
Reported settings: cutoff radius $r_c=5.4$\,\AA, maximum level
2{,}653, 8 radial-basis functions, maximum body order 5, total of
2{,}653 linear basis parameters.
\lstinputlisting[language=turtle]{artifacts/kg/listings-computed/nitol2024nial/q08-hyperparameter-settings.ttl}

\paragraph{Q9 -- MLIP run and trained model.}
A single fitting run on the 8{,}450-configuration training set
produces the optimised Ni--Al MTP. Training hardware (96 CPU cores)
is recorded on the run.
\lstinputlisting[language=turtle]{artifacts/kg/listings-computed/nitol2024nial/q09-run-and-model.ttl}

\paragraph{Q10 -- Benchmark results.}
Validation RMSEs on the 815-configuration validation set:
1.64\,meV/atom on energy, 0.028\,eV/\AA{} on force components,
0.09\,GPa on stress components.
\lstinputlisting[language=turtle]{artifacts/kg/listings-computed/nitol2024nial/q10-benchmark-results.ttl}

\paragraph{Q11 -- Computational resources.}
Training and inference hardware are reported as 96 CPU cores (Intel
Xeon Platinum 9242 @ 2.30\,GHz); the model achieves
${\sim}241$\,s for $10^5$ MD steps on a 2{,}048-atom Ni$_3$Al
supercell. Wall-clock training duration, peak memory, and explicit
per-atom inference time are not reported (no GPUs are used).
\lstinputlisting[language=turtle]{artifacts/kg/listings-computed/nitol2024nial/q11-resources.ttl}

\paragraph{Q12 -- Gaps.}
Beyond Q11, the genetic-algorithm-driven tensor-contraction search
that distinguishes this MTP from the standard Shapeev formulation is
not first-class in the ontology (recorded only in the
$\dlConcept{FunctionalForm}$ label). The two-step linear-then-L-BFGS
fitting procedure is captured in $\dlRole{hasTrainingAlgorithm}$ via
prose; the ontology has no concept for multi-stage training. Many
downstream defect-property comparisons (vacancy formation energies,
GSF curves, point-defect migration barriers) are reported as derived
figures rather than encoded as separate $\BenchmarkResultC$ instances.


\par}


\end{document}